\documentclass[lettersize,journal]{IEEEtran} %,onecolumn gln  ,runningheads='123456'
\usepackage{soul, color} %\hl
\usepackage{amsmath,amsfonts}
\usepackage{algorithmic}
\usepackage{algorithm}
\usepackage{array}
\usepackage[caption=true,font=normalsize,labelfont=sf,textfont=sf]{subfig} %原来的模板中caption=false!!!!!!!!!gln
\definecolor{myGreen}{rgb}{0.73,0.88,0.73} %gln
\definecolor{myOrange}{rgb}{1,0.5,0} %gln
\definecolor{myPurple}{rgb}{0.5,0,0.5} %gln
\usepackage{caption} %gln 表格标题问题
\usepackage{textcomp}
\usepackage{stfloats}
\usepackage{url}
\usepackage{verbatim}
\usepackage{graphicx}
\usepackage{cite}
\usepackage{extarrows}
\usepackage[switch]{lineno} %[left]代表行号在左侧  %gln switch表示左栏的左侧+右栏的右侧

\usepackage{bm} %equation
\usepackage{bbm}
\usepackage{amssymb} %equation
\usepackage{cases} %equation
\usepackage{amssymb} %equation
\usepackage{amsthm} %equation
\usepackage{booktabs}
\soulregister{\ref}7 % 注册\ref命令 %为了对cite部分进行高亮
\soulregister{\cite}7 % 注册\cite命令 %为了对ref部分进行高亮

\begin{document}
% \linenumbers %gln

\title{Self-Supervised Multi-Scale Network for Blind Image Deblurring via Alternating Optimization}

\author{Lening Guo, Jing Yu, Ning Zhang and Chuangbai Xiao
        % <-this % stops a space
%\thanks{}
        % <-this % stops a space
%This paper was produced by the IEEE Publication Technology Group. They are in Piscataway, NJ.  gln
%\thanks{}%Manuscript received April 19, 2021; revised August 16, 2021.  gln

\thanks{Corresponding author: Chuangbai Xiao (\textit{E-mail}: cbxiao@bjut.edu.cn).}
\thanks{Lening Guo, Jing Yu, Ning Zhang and Chuangbai Xiao are with the College of Computer Science, Beijing University of Technology, Beijing 100124, China.}
}

%\markboth{IEEE Transactions on Image Processing,~Vol.~xx, No.~xx, June~2024}%Journal of \LaTeX\ Class Files
%{Guo \MakeLowercase{\textit{et al.}}: Self-Supervised Multi-Scale Network for Blind Image Deblurring via Alternating Optimization}

%\IEEEpubid{0000--0000/00\$00.00~\copyright~2024 IEEE}

\maketitle

\begin{abstract}
Blind image deblurring is a challenging low-level vision task that involves estimating the unblurred image when the blur kernel is unknown. In this paper, we present a self-supervised multi-scale blind image deblurring method to jointly estimate the latent image and the blur kernel via alternating optimization. In the image estimation step, we construct a multi-scale generator network with multiple inputs and multiple outputs to collaboratively estimate latent images at various scales, supervised by an image pyramid constructed from only the blurred image. This generator places architectural constraints on the network and avoids the need for mathematical expression of image priors. In the blur kernel estimation step, the blur kernel at each scale is independently estimated with a direct solution to a quadratic regularized least-squares model for its flexible adaptation to the proposed multi-scale generator for image estimation. Thanks to the collaborative estimation across multiple scales, our method avoids the computationally intensive coarse-to-fine propagation and additional image deblurring processes used in traditional mathematical optimization-based methods. Quantitative and qualitative experimental results on synthetic and realistic datasets demonstrate the superior performance of our method, especially for handling large and real-world blurs.
\end{abstract}

\begin{IEEEkeywords}
Self-supervised learning, multi-scale, alternating minimization, image deblurring.
\end{IEEEkeywords}

\section{Introduction}
\IEEEPARstart{M}{OTION} blur is a common image degradation issue caused by camera shaking during exposure. Image deblurring is the process of recovering the latent sharp image from its blurry observation, and many methods have been developed in this field. The degradation process of the uniformly blurred image can be mathematically formulated as,
\begin{equation}
\bm{y} = \bm{h} \, {\ast} \, \bm{x} + \bm{n}
\label{eq:eq1}
\end{equation}
where $\bm{y}$ is the blurred image, $\bm{h}$ is the blur kernel, $\bm{x}$ is the latent sharp image, ${\ast}$ is the two-dimensional convolution operator, and $\bm{n}$ is the additive noise. Depending on whether the blur kernel is known or not, image deblurring can be classified into two categories: non-blind image deblurring and blind image deblurring. In blind image deblurring, the goal is to estimate the sharp image when the blur kernel is unknown, which poses a severely ill-posed inverse problem. More often, this work involves simultaneously estimating both the sharp image and the blur kernel.

Nowadays, deep convolutional neural networks (CNNs) have shown great success in the field of image deblurring where supervised learning-based methods are more widespread than unsupervised learning-based methods. To handle large and complex blurs, traditional mathematical optimization-based methods typically deconvolve a blurred image in a coarse-to-fine manner. Inspired by the coarse-to-fine scheme in mathematical optimization-based methods, supervised learning-based methods are moving from single-scale design\cite{schuler2015learning,kupyn2018deblurgan} to multi-scale design\cite{kupyn2019deblurganv2,cho2021rethinking,nah2017deep,zhang2017learning}. Single-scale design often fails to deal with large blurs, and multi-scale design avoids direct blur kernel estimation from a largely blurred image. Nah et al.\cite{nah2017deep} present a multi-scale convolutional neural network that cascades multiple sub-networks to mimic the image pyramid. Cho et al.\cite{cho2021rethinking} instead propose a multi-input multi-output U-Net (MIMO-UNet) which performs multi-scale deblurring in a single U-Net. The multi-scale architecture demonstrates its effectiveness in handling largely blurred images. In most cases, since pairs of real blurred images and ground truth sharp images are not available for supervised learning, the network typically trained on only simulation data does not perform well on real data. Self-supervised learning-based methods can be generalized better to real-world applications. Ulyanov et al.\cite{ulyanov2018deep} suggest a deep image prior network (DIP-Net) based on U-Net\cite{UNet2015}, using the inherent architecture of the deep network to implicitly regularize the smoothness prior of the latent image. SelfDeblur\cite{ren2020neural} applies the DIP-Net to estimate the image and the fully-connected network (FCN) to estimate the blur kernel, with alternating updates of the parameters of both networks during the training.

Mathematical optimization-based methods are flexible, adaptable and interpretable, and conveniently generalized to handle various inverse problems, but they heavily rely on handcrafted priors and inevitably suffer from high numerical computational costs, especially for non-convex optimization problems. On the contrary, deep learning-based methods increase the speed of image restoration through the parallelization on a GPU. However, their performance heavily depends on well-designed network architectures which should be tailor-made for certain image inverse problems, leading to a lack of flexibility. As a result, it is fascinating to integrate their respective strengths of these two types of methods. With the aid of variable splitting techniques, IRCNN decouples the denoising subproblem from the general optimization-based model and solves the denoising subproblem through the trained CNN denoiser\cite{zhang2017learning}. Fast-SelfDeblur\cite{bai2023fastselfdeblur} splits the joint optimization problem over the latent image and the blur kernel into two subproblems, where the image estimation subproblem is a parameter learning process of the DIP-Net and the blur kernel estimation subproblem is a regularized least-squares model.

Inspired by MIMO-UNet \cite{cho2021rethinking} and Fast-SelfDeblur \cite{bai2023fastselfdeblur}, we propose a self-supervised multi-scale blind image deblurring method (Self-MSNet), which integrates the multi-scale scheme into the self-supervised learning framework. We formulate blind deconvolution as a joint optimization problem on both the latent image and the blur kernel, and split it into the blur kernel estimation subproblem and the image estimation subproblem. These two subproblems are respectively optimized using deep learning and mathematical optimization. We develop a self-supervised multi-scale generator network to simultaneously generate unblurred images of multiple scales from random inputs supervised by an image pyramid constructed from the only blurred image. The blur kernel at each scale is independently estimated by solving a regularized least-squares optimization problem. The latent image and the blur kernel are jointly estimated through the iterative and alternating optimization of these two subproblems. Extensive experiments on synthetic and realistic datasets demonstrate the superior performance of our method to state-of-the-art mathematical optimization-based methods, supervised learning-based methods and self-supervised learning-based methods, particularly in handling large real-world blurs.

Our contributions are summarized as follows:
\begin{enumerate}
\item{Self-MSNet is a \textbf{self-supervised} method that learns network parameters by minimizing the objective function of a total variation (TV) regularized least-squares model supervised by the only blurred image itself. The proposed method does not require the training on ground truth sharp images or/and blur kernels that are required by supervised learning-based methods.}

\item{Self-MSNet involves the \textbf{integration} of the learning capabilities of deep learning-based methods and the adaptability of mathematical optimization-based methods, with the help of alternating optimization. We use a deep network to learn the image prior and generate the unblurred image from randomly distributed data through gradient descent on the objective function, requiring no intricate handcrafted priors commonly used in traditional mathematical optimization-based methods. Together with the powerful network learning, the direct solution to a regularized least-squares problem is evaluated for its simple generation to the kernel estimation of the motion blur.}

\item{The \textbf{multi-scale} scheme is incorporated into a single U-Net to form the generator network with multiple inputs and multiple outputs for image estimation. The U-Net architecture facilitates the propagation of feature representations across various scales. The design of multiple inputs and outputs enables the collaborative restoration of sharp images at different scales. This design achieves faster convergence by avoiding the computationally intensive coarse-to-fine propagation process in the traditional mathematical optimization-based methods.}

\item{Self-MSNet performs \textbf{joint estimation} of the latent image and the blur kernel via alternating optimization. Different from mathematical optimization-based methods or kernel estimation methods, the deblurred image is directly estimated after the alternating and iterative optimization process between updating the image and the blur kernel, requiring no additional image deblurring processes.}
\end{enumerate}
%-------------------------------------------------------------------------
\section{Related Work}
\label{sec:relate}
To resolve the challenging blind image deblurring problem, we propose a self-supervised multi-scale framework to generate sharp images and blur kernels across multiple scales, supervised by the only blurred image. The joint estimation of the image and the blur kernel integrates the learning capabilities of deep networks with the adaptability of mathematical optimization. In this section, we present a comprehensive overview related to this work.

\subsection{Mathematical Optimization vs. Deep Learning}
Since blind image deblurring is an ill-posed inverse problem, it is necessary to explicitly or implicitly incorporate prior information of images/kernels.

Mathematical optimization-based methods impose statistical priors or regularizers over images or/and kernels under the domination of the blur formation model. Some methods incorporate various image priors to restrict the feasible set of the latent sharp image. Based on the assumption that natural images typically follow a heavy-tailed distribution on their gradients, early pioneering works mainly elaborate image gradient-based sparsity priors to approximate this distribution, such as TV regularization\cite{chan1998total,whyte2014IJCVshaken,perrone2014total}, mixture-of-Gaussians prior\cite{fergus2006removing,Levin2011CVPRefficient}, $\ell_1$-norm prior\cite{shan2008highquality}, $\ell_2$-norm prior\cite{levin2009understanding,Yang2019CVPRAdaptiveEdgeSelection}, hyper-Laplacian prior\cite{Krishnan2009hyperLaplacian}, and $\ell_1 / \ell_2$-norm prior\cite{Krishnan2011CVPRnormalized}. Since the high-contrast and step-like structures can facilitate the estimation of blur kernel, unnatural priors are proposed including $\ell_0$-norm prior\cite{xu2013unnatural,pan2014CVPRtextL0,Anger2019L0,pan2016PAMIL0text} and heuristic edge selection\cite{cho2009fast,Joshi2008CVPRsharpedge,xu2010two,money2008TVshock,sun2013ICCPedgeBased,lai2015CVPRnormalized,Pan2013SPICsalient,gong2016CVPRautomatic}. Instead of using an explicitly expressed regularizer in the objective function, the heuristic edges promote the estimation of the blur kernel by enhancing salient edges with image filters\cite{cho2009fast,Joshi2008CVPRsharpedge,xu2010two,money2008TVshock,Pan2013SPICsalient} or locating informative edge pixels with edge masks\cite{sun2013ICCPedgeBased,lai2015CVPRnormalized,gong2016CVPRautomatic}. The image gradient operator operates on the neighbourhood of pixels, while the patch-based priors capture larger structures in the image, e.g. sparse representation prior\cite{zhang2011close, czc2017sparse,peng2021ksvd}, nonlocal self-similarity (NSS) prior\cite{michaeli2014blind, czc2017sparse}, color-line prior\cite{lai2015CVPRnormalized}, low-rank prior\cite{wang2013ACCVnsp,ren2016image, peng2022lowrank}, dark channel prior (DCP)\cite{pan2016CVPRdarkchannel, pan2018PAMIdarkchannel}, extreme channel prior (ECP)\cite{yan2017extremechannel}, local maximum gradient prior\cite{chen2019blind}, and Gaussian mixture model of image patches\cite{Zoran2011ICCVnaturalimagepatch}. Another methods incorporate kernel regularizers about prior knowledge of the physics of motion blur, such as $\ell_2$-norm prior \cite{xu2013unnatural,cai2012TIPframelet,gong2016CVPRautomatic,bai2023fastselfdeblur}, $\ell_1$-norm prior \cite{shan2008highquality,Kotera2013CAIPL1norm,Krishnan2011CVPRnormalized,yue2014ECCVhybrid}, bi-$\ell_0$-$\ell_2$-norm prior \cite{shao2015BiL0L2norm}, TV regularizer \cite{Almeida2010TIPsemi,chan1998total,you1999TIPanisotropicTV}, maximum convolution eigenvalue-based kernel regularizer\cite{liu2014spectral}, and continuity prior of the blur kernel \cite{cai2012TIPframelet}. In this work, we adopt the common $\Vert {\bm h} \Vert^2_2$ to enforce the sparsity of the blur kernel.

While it is hard to mathematically formulate the sophisticated distribution or prior of images using traditional optimization-based methods above, deep learning-based methods place architectural constraints on the network. In \cite{ulyanov2018deep}, DIP-Net pioneers the implicit representation of image priors using deep generative models. It is based on the assumption that high-dimensional visual data (e.g., images) usually lie on low-dimensional manifolds embedded in the high-dimensional space. Many deep learning-based methods\cite{ren2020neural,liang2021CVPRFKP,fang2023CVPRflow,li2023deepEM,bai2023fastselfdeblur} are developed based on the framework of DIP-Net for image estimation. These methods employ diverse strategies for kernel estimation. These work of \cite{ren2020neural,liang2021CVPRFKP,fang2023CVPRflow,li2023deepEM} adopts deep networks to implicitly model priors of blur kernels. In \cite{ren2020neural}, a FCN is used to generate the blur kernel from random vectors where the blur kernel is flattened into a vector form and its inherent spatial correlation is disregarded. The methods of \cite{liang2021CVPRFKP,fang2023CVPRflow} train normalizing flow models on blur kernel samples to learn the manifold of blur kernels as a prior. Inspired by Double-DIP\cite{gandelsman2019double}, the U-Net is used in \cite{li2023deepEM} to generate the kernel set for the non-uniform motion deblurring.

\subsection{Single-scale vs. Multi-scale}
To deblur the image with large blur kernel, conventional mathematical optimization-based methods utilize a pyramid framework to estimate images and blur kernels in a coarse-to-fine manner. Motivated by this idea, deep learning-based methods mimic the coarse-to-fine strategy to merge multi-scale information for improving the deblurring performance. In contrast to multi-scale methods, these methods that directly estimate the blur kernel or the latent sharp image at the original image resolution can be referred to as single-scale methods. Some early multi-scale deblurring methods cascade multiple sub-networks in a pyramid structure, progressively restoring the sharp image from coarse to fine scales\cite{schuler2015learning,nah2017deep,tao2018scale,gao2019CVPRnested}. However, this network design often leads to increased computational burden and a large number of parameters. Kaufman et al.\cite{kaufman2020deblurring} propose a single network where the blur kernel is estimated from coarse to fine. In MIMO-UNet \cite{cho2021rethinking}, multiple inputs and multiple outputs are incorporated into a single U-Net, resulting in a faster and lighter multi-scale deblurring network. Compared to single-scale deblurring networks, multi-scale methods perform better to handle diverse large and complex blur kernels in blind image deblurring.

\subsection{Supervised vs. Unsupervised}
According to whether depending on the training dataset, deep learning-based methods can be broadly classified into two categories: supervised learning- and unsupervised learning-based methods.

Supervised learning-based blind image deblurring methods typically require a large set of ground truth sharp images or/and blur kernels for training. These methods of \cite{schuler2015learning,kupyn2018deblurgan,nah2017deep,tao2018scale,kupyn2019deblurganv2,zhang2019deep,cho2021rethinking,Zamir2021CVPRMPRNet,fang2023CVPRflow,li2022ECCVdegradation,xu2018TIPmotion,Ramakrishnan2017ICCVWGFMD,gao2019CVPRnested,yuan2020efficient,Purohit2020regionadaptive} learn the mapping from blurry images to their sharp counterparts by training on pairs of blurry and sharp images, and some approaches\cite{sun2015CVPRlearning,yan2016blind,kaufman2020deblurring,gong2017CVPRfromMBMF} take blurry images as inputs and their corresponding blur kernels as outputs by training on blurry images with known ground truth blur kernels. Furthermore, some approaches require access to triplets of blurry images, sharp images and corresponding ground truth blur kernels for training\cite{zuo2016TIPtriple,li2019ICASSPunrolling,kaufman2020deblurring,Pan2017ICCVdataFitting}. Supervised learning-based methods, due to their reliance on paired training data, tend to perform poorly in real-world scenarios.

To mitigate the shortcomings of supervised learning-based methods, unsupervised learning-based methods have been developed for image restoration, mainly including unpaired transfer learning and self-supervised learning. Unpaired transfer learning-based methods focus on disentangling domain-invariant and domain-specific feature representations, and then transfer the learned domain-invariant feature representation from the source domain to an unlabeled target domain via generative adversarial networks (GANs) \cite{Lu2019CVPRgan,xia2019NIPStraining,zhang2020CVPRrealistic}. Motivated by the practical challenge of data collection, there is a growing interest in dataset-free deep learning-based methods for image deblurring. Self-supervised methods directly learn the prior information from the blurred image itself rather than external datasets. DIP\cite{ulyanov2018deep}, SelfDeblur\cite{ren2020neural}, Fast-SelfDeblur\cite{bai2023fastselfdeblur}, Li et al.'s method\cite{li2023deepEM}, and DEBID\cite{Chen2023TCSVTensemble} are representative examples of self-supervised learning-based image deblurring methods by minimizing the data fidelity term derived from the blur formation model, supervised by the blurred image.

%-------------------------------------------------------------------------
\section{Methodology}
\label{sec:proposed}
The proposed Self-MSNet, a self-supervised multi-scale blind image deblurring method, jointly models the latent image and the blur kernel. And the joint model can be split into two subproblems about image and blur kernel, allowing for alternating optimization by utilizing the deep network learning and the mathematical optimization method respectively.

\subsection{Model Architecture}
By introducing the multi-scale deep network into the TV regularized blind deblurring model, the optimization problem is formulated as follows:
\begin{equation}
    \begin{aligned}
        \min_{{\bm \theta},{{\bm{h}}^{s}}} \,\, \Big\{ L \left({\bm \theta},{\bm{h}}^{s} \right)
        &= \sum_{s}\left \Vert {\bm y}^s - {{\bm h}^s}* \bm{x}^s \right \Vert^2_{\mathrm F} + \lambda_h \Vert {\bm{h}^s} \Vert^2_{\mathrm F} \\ &+\lambda_{x} \Vert \nabla \bm{x}^{s} \Vert_{1} \Big\}, \, \{ \bm{x}^{s} \} = {f}( \{ \bm{z}^{s} \};{\bm{\theta}})
    \end{aligned}
    \label{eq:eq2}
\end{equation}
where $s$ indicates the scale index with $s=0$ denoting the original scale, $\bm{y}^s$, $\bm{h}^s$ and ${\bm{x}}^s$ are the blurred image, the blur kernel and the sharp image at the $s$-th scale, respectively, $\lambda_{h}$ and $\lambda_{x}$ represent the regularization parameters, $\lVert \cdot \rVert_{\mathrm F}$ denotes the Frobenius norm of a matrix, $\lVert \cdot \rVert_{1}$ denotes the entrywise $\ell_{1}$ norm of a matrix\footnote{We use the definition of a matrix norm rather than that of a vector norm for the variables, as all the variables inside the operator norms are matrices. Given a matrix ${\bm A}$, the Frobenius norm of the matrix is defined as $\lVert {\bm A} \rVert_{\mathrm F} \xlongequal{\mathrm{def}} \left( \sum\limits_{i=1}^{m}\sum\limits_{j=1}^{n} \lvert a_{ij} \rvert ^2 \right)^{1/2}$ and the entrywise $\ell_{1}$ norm of the matrix is defined as $\lVert {\bm A} \rVert_{1} \xlongequal{\mathrm{def}} \sum\limits_{i=1}^{m} \sum\limits_{j=1}^{n} \lvert a_{ij} \rvert$.}, and ${f}( \{ \bm{z}^{s} \};{\bm{\theta}})$ represents the multi-scale generator network with parameters ${\bm{\theta}}$ to generate sharp images $\{ \bm{x}^{s} \}$ from multiple matrices of random values $\{ {\bm{z}}^s \}$ with different scales. In Eq.(\ref{eq:eq2}), the first term is the data fidelity term, ensuring the solution in accordance with the blur formation model. The second term introduces the Frobenius norm to impose squared $\ell_{2}$-norm regularization on blur kernel $\bm{h}$, while the third term uses TV regularization to enforce the smoothness of the estimated image.

Eq.(\ref{eq:eq2}) can be split into two subproblems, \textit{i.e.} the blur kernel estimation subproblem as shown in Eq.(\ref{eq:eq3a}) and the image estimation subproblem as shown in Eq.(\ref{eq:eq3b}). This unfolding allows for the alternating optimization over the blur kernel and the unblurred image.
%\begin{footnotesize}
\begin{subequations}
    \begin{numcases}{}
        %\begin{aligned}
	{{\bm{h}}_k^{s}} = \mathop{\arg \min_{\bm{h}^{s}}} \Vert \bm{y}^{s} \, - \, \bm{h}^{s} \ast {\bm{x}_{k-1}^{s}} \Vert^2_{\mathrm F} \, + \, \lambda_h \Vert \bm{h}^{s} \Vert^2_{\mathrm F} \label{eq:eq3a} \\
        %\end{aligned}\\
        \begin{aligned}
	{\bm{\theta}}_{k} = &\mathop{\arg \min_{\bm{\theta}}} \sum\limits_{s} (\Vert \bm{y}^{s} \! - \! {\bm{h}_k^{s}} \ast {\bm{x}^{s}} \Vert^2_{\mathrm F} \! + \! \lambda_{x} \Vert \nabla {\bm{x}^{s}} \Vert_{1}), \\
        &{\bm{x}^{s}} = {f}^{s} \left( \{ \bm{z}^{s} \};{\bm{\theta}} \right) \label{eq:eq3b}
        \end{aligned}
    \end{numcases}
    \label{eq:eq3}
\end{subequations}
\noindent $\!\!\!$where $k$ stands for the iteration, and ${f}^{s} \left( \{ {\bm{z}}^{s} \};{\bm{\theta}} \right)$ denotes the output at the $s$-th scale of the multi-scale generator network ${f} \left( \{ {\bm{z}}^{s} \};{\bm{\theta}} \right)$, with the superscript $s$ serving as the scale index. At each iteration $k$, we solve Eq.(\ref{eq:eq3a}) for the blur kernels $\{ {\bm{h}}_{k}^s \}$ according to the unblurred images $\{ {\bm{x}}_{k-1}^s \}$, and then update network parameters ${\bm{\theta}}_{k}$ from ${\bm{\theta}}_{k-1}$ through Eq.(\ref{eq:eq3b}) according to the estimated blur kernels $\{ {\bm{h}}_{k}^s \}$; and subsequently, the unblurred images $\{ {\bm{x}}_{k}^s \}$ are derived from the network parameter ${\bm{\theta}}_{k}$.

Eq.(\ref{eq:eq3a}) is a quadratic regularized least-squares problem about $\bm h$, and we directly solve it for the blur kernel ${\bm h}^s$ at each scale according to ${\bm x}^{s}$. Eq.(\ref{eq:eq3b}) is a TV regularized least-squares problem about ${\bm{x}}^s$, and the multi-scale generator network is used to generate images from random inputs. In this way, Eq.(\ref{eq:eq3b}) becomes an optimization problem about network parameters ${\bm \theta}$, and then solving it is a network learning process using gradient descend updates. By minimizing the total loss across all scales as defined in the objective function of Eq.(\ref{eq:eq3b}), the generator network outputs unblurred images $\{ {\bm x}^{s} \} = {f}( \{ \bm{z}^{s} \};{\bm{\theta}})$ at all different scales.

\begin{figure}[hbtp]
    \centering
    \includegraphics[width=0.95\linewidth]{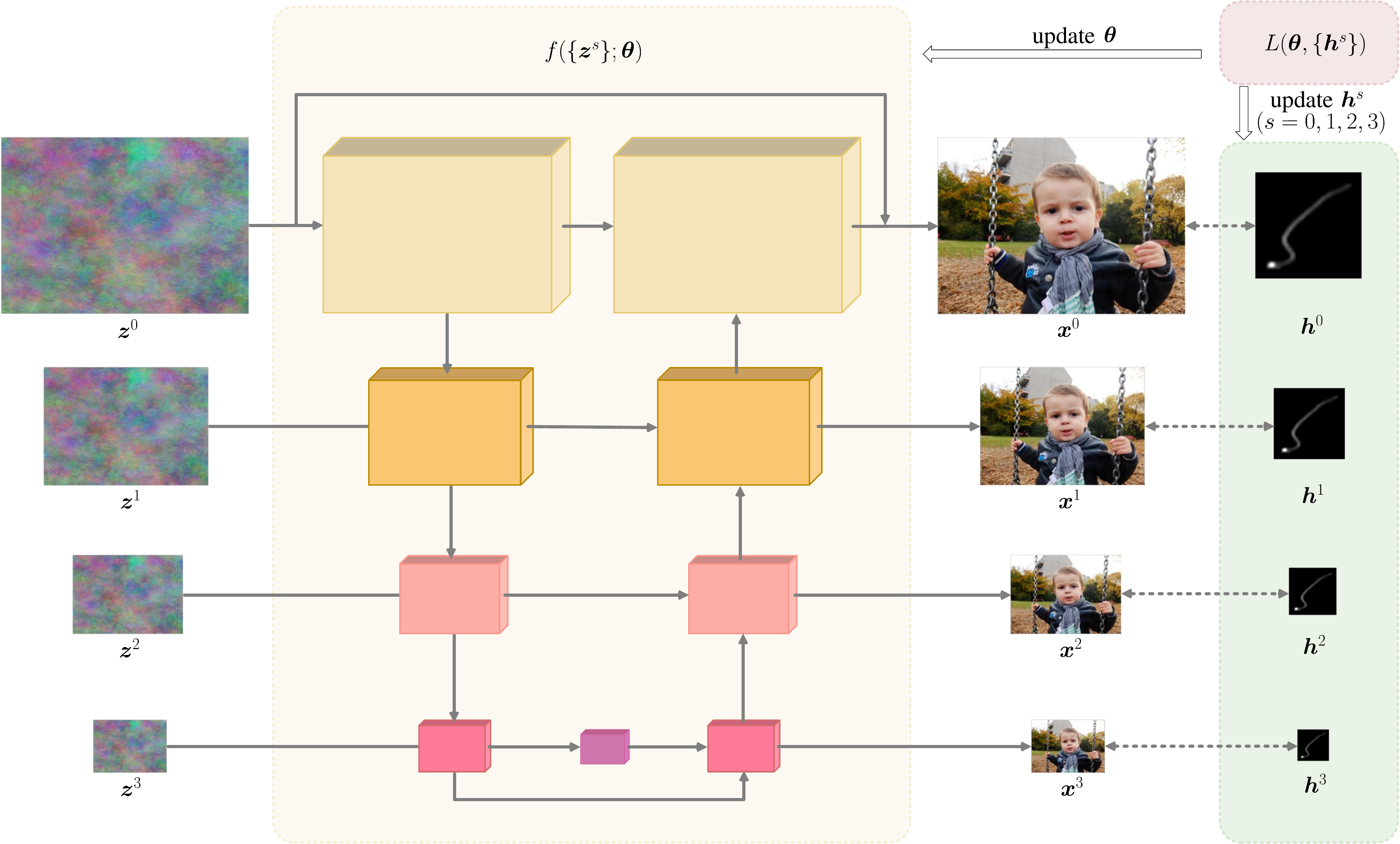}
    \caption{Overview of our Self-MSNet.}
    \label{fig:framework-overall}
\end{figure}

Figure \ref{fig:framework-overall} illustrates the overview architecture of our Self-MSNet. We construct a multi-input and multi-output deep network similar to U-Net to simultaneously generate images $\{ {\bm x}^s \} = {f}( \{ \bm{z}^{s} \};{\bm{\theta}})$ of different scales, and solve for the blur kernel ${\bm h}^s$ according to ${\bm x}^s$ at each scale independently. The image estimation and the blur kernel estimation are iteratively alternated to minimize the objective function $L \left( {\bm \theta}, \bm{h}^{s} \right)$ in Eq.(\ref{eq:eq2}). In Section \ref{ablationstudy}, we report an ablation study exploring the benefit of the multi-scale network architecture.

\subsection{Blur Kernel Estimation Subproblem}
\label{sec:blur_kernel_estimation_subproblem}
In the blur kernel estimation step, Eq.(\ref{eq:eq3a}) is solved and the blur kernels $\{ {\bm h}_{k}^s \}$ are updated when ${\bm x}_{k-1}$ is fixed. We omit the superscript $s$ for brevity, and the optimization problem in Eq.(\ref{eq:eq3a}) can be reformulated as
\begin{equation}
\begin{aligned}
    {\bm{h}_k} =  &\mathop{\arg \min_{\bm{h}}} \Vert \bm{y} - \bm{h} \ast \bm{x}_{k-1} \Vert^2_{\mathrm F} + \lambda_h \Vert \bm{h} \Vert^2_{\mathrm F}, \,\, \\ &\{ \bm{x}_{k-1} \} = {f} \left( \{ \bm{z} \};{\bm{\theta}}_{k-1} \right)
    \label{eq:eq4}
\end{aligned}
\end{equation}

\noindent We solve Eq.(\ref{eq:eq4}) for the respective blur kernel at each scale. Additionally, a sparse kernel refinement process \cite{xu2010two} is then performed to preserve the sparsity of the blur kernel.

Fourier transform can be used to perform fast convolution. The discrete Fourier transform (DFT) assumes an image is extended periodically. This assumption causes boundary related ringing in the deblurred images. Since pixels are highly correlated in most of images, the first-order derivative of the image is zero or close to zero almost everywhere except for details such as edge or texture. To avoid ringing artifacts, we define the data fidelity term in the gradient domain. In order to encourage the center of mass of the blur kernel to be at its coordinate center, Eq.(\ref{eq:eq4}) can be reformulated as follows:
\begin{equation}
    \begin{aligned}
        {{\bm{h}}_k} =  &\mathop{\arg}\min_{\bm{h}} \,\, \Vert  \nabla \bm{y} - \bm{h} \ast \nabla \bm{x}_{k-1} \Vert^2_{\mathrm F} + \lambda_h \Vert \bm{h} \Vert^2_{\mathrm F} \\ &+ \gamma \bigg{\lVert} {\left( x_0, y_0 \right)^{\rm{T}}} - {\left( {\bar x}, {\bar y} \right)^{\rm{T}}} \bigg{\rVert}^2_2
    \end{aligned}
    \label{eq:eq5}
\end{equation}
where $\nabla = [\partial_r, \partial_c]^{\rm T} $ denotes the gradient operator, $\partial_r $ and $\partial_c $ are the partial derivatives of the image along the row and column directions respectively, $\gamma$ is the regularization parameter, $\left( x_0, y_0 \right)$ refers to the coordinate center of the blur kernel, and $\left( {\bar x}, {\bar y} \right)$ represents the center of mass of the blur kernel. Assuming that the blur kernel is of size $m \times n$, we have $\left( x_0, y_0 \right) = \left( \frac{m}{2}, \frac{n}{2} \right)$ and $\left( {\bar x}, {\bar y} \right) = {\left( \frac{ {\mathbbm{1}}_{n}^{\rm{T}} {\bm{h}}^{\rm{T}} {{\bm t}_m} }{ {\mathbbm{1}}_{m}^{\rm{T}} {\bm{h}} {{\mathbbm{1}}_{n}} }, \frac{ {\mathbbm{1}}_{m}^{\rm{T}} {\bm{h}} {{\bm t}_n} }{ {\mathbbm{1}}_{m}^{\rm{T}} {\bm{h}} {{\mathbbm{1}}_{n}} } \right)}$ where ${{\mathbbm{1}}_{m}} = ( \underbrace{1,1,\cdots,1}_{m} )^{\rm{T}}$, ${{\mathbbm{1}}_{n}} = ( \underbrace{1,1,\cdots,1}_{n} )^{\rm{T}}$, ${\bm t}_m = \left( 1,2,\cdots,m \right)^{\rm{T}}$ and ${\bm t}_n = \left( 1,2,\cdots,n \right)^{\rm{T}}$. Equation (\ref{eq:eq5}) is a quadratic regularized least-squares problem with respect to $\bm{h}$, which admits a global minimum solution. By setting the derivative of the objective function in Eq.(\ref{eq:eq5}) to zero, a linear system of equations is derived as follows
\begin{equation}
    {\bm A} {\rm{vec}} \left( {\bm{h}_{k}}  \right) = {\rm{vec}} \left( {\bm Z} \right)
    \label{eq:eq6}
\end{equation}
where ${\rm{vec}} \left( \cdot \right)$ represents the operation of stacking the elements of a matrix column-wise,
\begin{footnotesize}
\begin{equation}
    \bm{Z} = {\mathcal F}^{-1} \left( \frac{ \overline{\mathcal{F}(\partial_r \bm{x}_{k-1})}\mathcal{F}( \partial_r \bm{y})+\overline{\mathcal{F}(\partial_c \bm{x}_{k-1})}\mathcal{F}( \partial_c \bm{y}) } { \overline{\mathcal{F} (\partial_r \bm{x}_{k-1})}\mathcal{F} (\partial_r \bm{x}_{k-1})+\overline{\mathcal{F}(\partial_c \bm{x}_{k-1})} \mathcal{F} (\partial_c \bm{x}_{k-1}) + \lambda_{h} } \right) \,\, , \\
    \notag
\end{equation}
\end{footnotesize}

\noindent and the coefficient matrix is
\begin{equation}
    {\bm A} = \gamma \left[ {\rm{vec}} \left( {\bm F} \right) {\rm{vec}} \left( {\bm U} \right)^{\rm T}
    + {\rm{vec}} \left( {\bm K} \right) {\rm{vec}} \left( {\bm V} \right)^{\rm T} \right] + {\bm I},
    \notag
\end{equation}
with ${\bm I}$ denoting an identity matrix of size $mn \times mn$. These matrices ${\bm U},{\bm V},{\bm F},{\bm K}$ are denoted as follows, respectively.

\begin{footnotesize} %scriptsize 
\begin{equation}
    \begin{aligned}
        &{\bm U} = \left({\bm t}_m - {x_0} \right){\mathbbm 1}_{n}^{\rm T}, \,\,\,\,
        {\bm V} = {\mathbbm 1}_m \left( {\bm t}_n - {y_0}\right)^{\rm T}, \\
        &{\bm{F}} = {\mathcal F}^{-1} \left( \frac{ \mathcal{F} \left({{\left( {\bm t}_m - {x_0} \right)}  {\mathbbm 1}_{n}^{\rm{T}}} \right)  }  { \overline{\mathcal{F} \left(\partial_r \bm{x}_{k-1}\right)}\mathcal{F} \left(\partial_r \bm{x}_{k-1}\right)+\overline{\mathcal{F}\left(\partial_c \bm{x}_{k-1}\right)} \mathcal{F} \left( \partial_c \bm{x}_{k-1} \right) + \lambda_{h} }  \right), \\
        &{\bm{K}} = {\mathcal F}^{-1} \left( \frac{\mathcal{F} \left( {{\mathbbm 1}_m {\left( {\bm t}_n - {y_0} \right)} ^{\rm{T}}} \right)} { \overline{\mathcal{F} \left( \partial_r \bm{x}_{k-1} \right)}\mathcal{F} \left( \partial_r \bm{x}_{k-1} \right)+\overline{\mathcal{F}\left(\partial_c \bm{x}_{k-1}\right)} \mathcal{F} \left(\partial_c \bm{x}_{k-1}\right) + \lambda_{h} }  \right).
    \end{aligned}
    \notag
\end{equation}
\end{footnotesize}

\noindent Here, $\mathcal{F}(\cdot)$ is Fourier transform, $\mathcal{F}^{-1}(\cdot)$ is inverse Fourier transform, and $ \overline{\mathcal{F}(\cdot)} $ is complex conjugate of Fourier transform. We can obtain a direct, exact and analytic solution for blur kernel ${\bm{h}_{k}}$ by solving Eq.(\ref{eq:eq6}). The detailed derivation is included in the Appendix \ref{sec:appendix_blur_kernel_derivation}.

The flat or textural regions in the image could not contribute to blur kernel estimation. Only the significant edges whose width is larger than that of the blur kernel, can provide reliable information for blur kernel estimation. We restrict the fidelity term contributions to salient edge regions to guide the estimation of the blur kernel. The simple Sobel filters are employed to calculate the gradient magnitude and direction of pixels in four orientations: horizontal, vertical, and $\pm 45^\circ$. Subsequently, we find a threshold to keep at least top $10\%$ of pixels with the largest filter responses in each direction. This edge selection process significantly improves the accuracy of blur kernel estimation.

\subsection{Image Estimation Subproblem}
In the image estimation step, Eq.(\ref{eq:eq3b}) is solved through a network learning process supervised by the image pyramid constructed from the only blurred image $\{ {\bm y}^{s} \}$ and the network parameters ${\bm \theta}_{k}$ are updated according to $\bm{\theta}_{k-1}$ when the blur kernels $\{ \bm{h}_{k}^s \}$ are fixed; and then the latent images $\{ {\bm x}^s_k \} = {f}( \{ \bm{z}^{s} \};{\bm{\theta}}_k)$ are estimated according to $\bm{\theta}_{k}$.

The objective function in Eq.(\ref{eq:eq3b}) is a function of network parameters $\bm{\theta}$, denoted as
\begin{equation}
    \begin{aligned}
    L \left( {\bm \theta} \right) = \sum\limits_{s} ( &\Vert \bm{y}^{s} \! - \! {\bm{h}_k^{s}} \ast {f}^{s} \left( \{ \bm{z}^{s} \};{\bm{\theta}} \right) \Vert^2_{\mathrm F} \,\! \\
    &+ \,\! \lambda_{x} \Vert \nabla {f}^{s} \left( \{ \bm{z}^{s} \};{\bm{\theta}} \right) \Vert_{1} )\\
    \end{aligned}
    \label{eq:eq6_2}
\end{equation}

\noindent To minimize $L \left( {\bm \theta} \right)$, a gradient descent algorithm is applied where the gradient of $L \left( {\bm \theta} \right)$ is evaluated with respect to the network parameters $\bm{\theta}$. The learnable parameters are iteratively updated by taking small steps in the direction that reduces the loss at each iteration.
\begin{equation}
    {\bm \theta}_{k} = {\bm \theta}_{k-1} - {\eta} \frac{\partial L \left( {\bm \theta} \right)}{\partial {\bm \theta}} \bigg|_{{\bm \theta} = {\bm \theta}_{k-1}}
    \label{eq:eq7}
\end{equation}
where ${\eta}$ is the learning rate. Based on the parameters ${\bm{\theta}}_{k}$, we obtain the unblurred images $\{ \bm{x}_{k}^s \} = {f}( \{ \bm{z}^s \};{\bm{\theta}_{k}})$ for all the scales at the iteration $k$.

\begin{figure*}[htbp]
    \centering
    \includegraphics[width=0.98\linewidth]{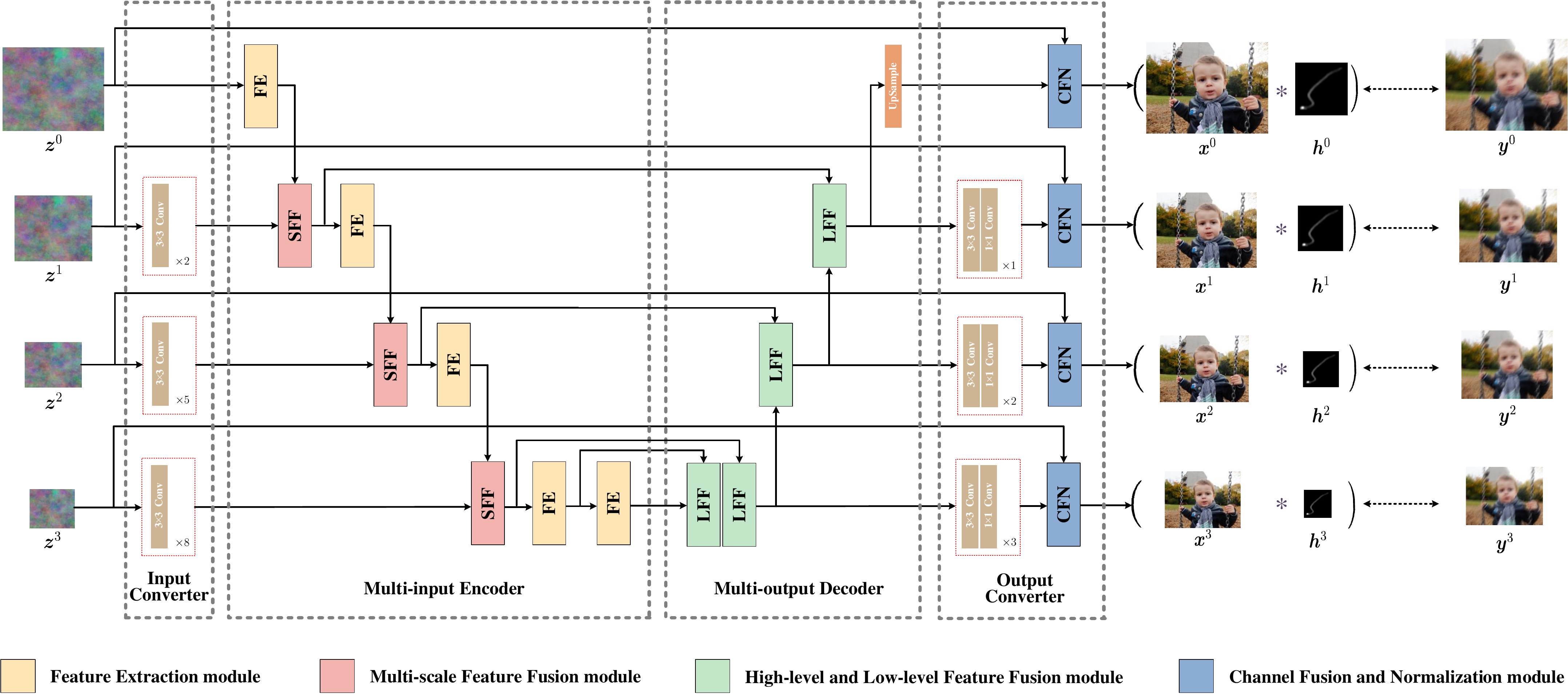}
        \vskip 5pt
    \caption{Detailed architecture of the multi-scale generator network for image estimation in Self-MSNet.}
    \label{fig:framework-detailed}
\end{figure*}

\begin{figure}[!t]
\centering
\subfloat[FE]{ \includegraphics[width=1.5in]{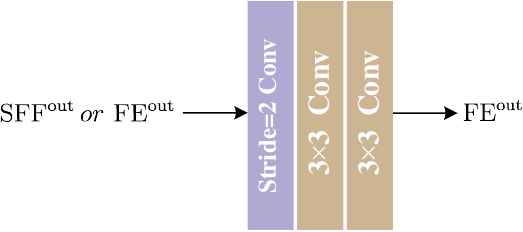} \label{fig:FE}}
%\hfill
\subfloat[SFF]{\includegraphics[width=1.7in]{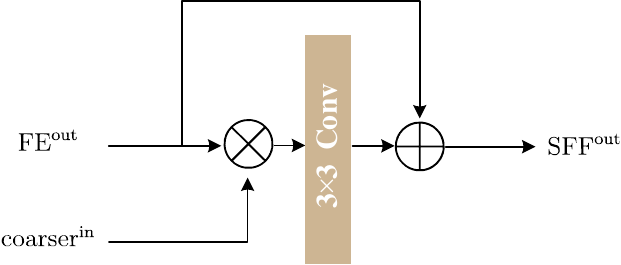}
\label{fig:SFF}}

\subfloat[LFF]{\includegraphics[width=1.7in]{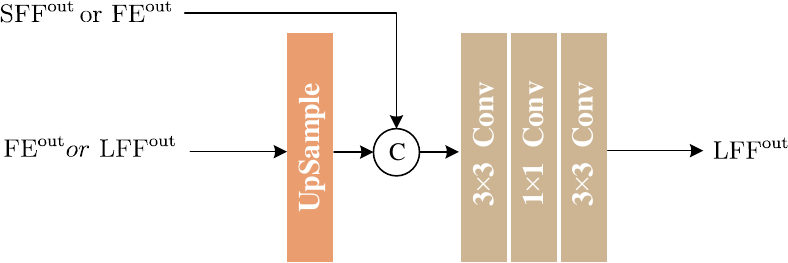}
\label{fig:LFF}}
%\hfill
\subfloat[CFN]{\includegraphics[width=1.6in]{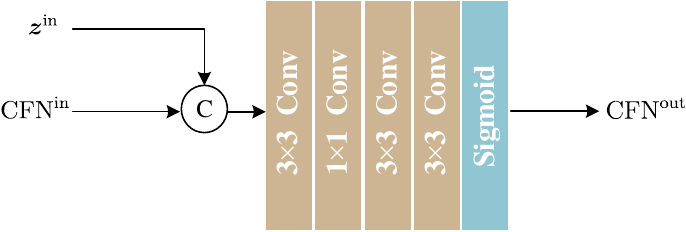}
\label{fig:CFN}}
\caption{The structures of modules.}
\label{fig:framework-modules}
\end{figure}

We incorporate the multi-scale scheme of MIMO-UNet \cite{cho2021rethinking} into the network design of Fast-SelfDeblur \cite{bai2023fastselfdeblur} to propose a self-supervised multi-scale deblurring network with multiple inputs and multiple outputs. As illustrated in Fig.\ref{fig:framework-detailed}, the multi-scale generator network for image estimation has an encoder-decoder structure similar to U-Net \cite{UNet2015}, coarsely divided into four parts: Input Converter, Multi-input Encoder, Multi-output Decoder, and Output Converter. \textbf{(i)} Input Converter takes the matrices of random values at various scales as inputs and produces feature maps as outputs. It uses additional convolutional layers at the coarser scales to ensure consistent semantic levels of features obtained from the multi-scale random input. \textbf{(ii)} Multi-input Encoder performs an image reduction process by repeatedly extracting features and downsampling. The feature extraction (FE) module (as shown in Fig.\ref{fig:FE}) is composed of several convolutional layers in which the first convolution layer with a stride of 2 downsamples the spatial dimensions by a factor of two. The multi-scale feature fusion (SFF) module (as shown in Fig.\ref{fig:SFF}) integrates the features obtained from FE with the coarser-scale features $\rm{{coarser}^{in}}$ through an element-wise product operation and adopts the residual connection to learn an additive change to the current fused feature. \textbf{(iii)} Multi-output Decoder achieves an image expansion process through repeated feature extractions and upsampling operations. The high- and low-level feature fusion (LFF) module (as shown in Fig.\ref{fig:LFF}) merges the spatial resolution and high-level semantic features, where skip connections are made by directly concatenating outputs of corresponding layers. \textbf{(iv)} Output Converter generates the unblurred images from the output feature maps. Similarly to the Input Converter part, there are more stacked convolutional layers at the coarser scale to ensure the consistency of semantic levels across various scales. The channel fusion and normalization (CFN) module (as shown in Fig.\ref{fig:CFN}) concatenates the feature maps from the last layer with the matrix of random values input ${\bm{z}}^{\rm in}$ using skip connections to aid convergence. A \textit{Sigmoid} nonlinearity function is applied to convert the output in the range $\left[ 0,1 \right]$. The loss function is defined on all three channels RGB rather than a single luminance channel\cite{ren2020neural} to enhance color fidelity and reduce color distortion of the deblurred images. Note that these modules are not independent, and we package repetitive operations for simplifying the illustration of our architecture.

Our multi-scale generator network incorporates four-scale inputs and outputs ($s=0,1,2,3$) to concurrently deblur images at different scales. The inputs of the generator network are matrices of random values. The random input ${\bm{z}}^3$ is of the same size as the coarsest scale $s=3$ and each element is independent and identically distributed with a uniform distribution on $\left( 0,1 \right)$. The input of the finer scale $s-1$ is a 2-fold upsampled version of the random input at the coarser scale $s$ using nearest neighbor interpolation. These inputs are consistently configured with 16 channels. The number of scales is chosen such that the size of the coarsest scale remains large enough to ensure robust and reliable estimation.

% \begin{figure}[!tbp]
% 	\centering
% 	\label{fig:framework-modules}
% 	\begin{minipage}{0.47\linewidth}
% 		\centering
% 		\includegraphics[width=1\linewidth]{images/FE_revise2_crop.pdf}
% 		\caption*{(a)FE\centering}
% 		\label{fig:FE}
% 	\end{minipage}
% 	\begin{minipage}{0.47\linewidth}
% 		\centering
% 		\includegraphics[width=1\linewidth]{images/SFF_revise2_crop.pdf}
% 		\caption*{(b)SFF\centering}
% 		\label{fig:SFF}
% 	\end{minipage}
%         \vskip 7pt
%         \begin{minipage}{0.47\linewidth}
% 		\centering
% 		\includegraphics[width=1\linewidth]{images/LFF_revise2_crop.pdf}
% 		\caption*{(c)LFF\centering}
% 		\label{fig:LFF}
% 	\end{minipage}
% 	\begin{minipage}{0.47\linewidth}
% 		\centering
% 		\includegraphics[width=1\linewidth]{images/CFN_revise2_crop.pdf}
% 		\caption*{(d)CFN\centering}
% 		\label{fig:CFN}
% 	\end{minipage}
%         \vskip 5pt
% 	\caption{The structures of modules.}
% \end{figure}

\subsection{Implementation}
The pseudo-code of our method is presented in Alg.\ref{alg:algorithm_1}. Given a blurred image $\bm{y}$, Self-MSNet outputs a deblurred image $\bm{x}$ and its associated blur kernel $\bm{h}$. For the preparation stage, We randomly initialize the network parameter ${\bm{\theta}_{0}}$ with He initialization \cite{kaiming2015initialization}, draw the network inputs $ \{ {\bm{z}}^s \}$ from a uniform distribution and pass this through the network to form the initial latent images so that $\{ \bm{x}_{0}^s \} = {f}( \{ \bm{z}^s \};{\bm{\theta}_{0}})$ for all the scales. For the alternating optimization stage, the blur kernel $\bm{h}_{k}^s$ for each scale $s$ is firstly updated by directly solving the regularized problem shown in Eq.(\ref{eq:eq5}); and then, this is followed by an iteration step of the Adam gradient descent method \cite{kingma2014adam} to update parameters ${\bm{\theta}}_{k}$ of the multi-scale generator network for image estimation $ \{ \bm{x}_{k}^s \} = {f}( \{ \bm{z}^s \};{\bm{\theta}_{k}})$. Self-MSNet performs an iterative process that alternates between optimizing the parameter ${\bm{\theta}}$ and the blur kernels $\{ {\bm{h}}^s \}$ until the loss function converges to its minimum value or reaching the maximum number of iterations $K$. The results ${\bm x}_{K}^{0}$ and ${\bm h}_{K}^{0}$ at the original image scale $s=0$ are regarded as the final estimates of deblurred image and the blur kernel, respectively.

\begin{algorithm}[!t]
    \caption{Self-MSNet.}
    {\bf Input}: Blurred image $\bm {y}$ \\
    {\bf Output}: Deblurred images $\{ \bm{x}_{K}^{s} \}_{s=0,1,2,3}$; blur kernels $\{ \bm{h}_{K}^{s} \}_{s=0,1,2,3}$
    \begin{algorithmic}[1] %[1] enables line numbers
        \STATE Initialize network parameters ${\bm{\theta}_{0}}$ and generate random inputs $\{ {\bm{z}}^{s} \}$.
        \STATE Generate the initial unblurred images $\{ \bm{x}_{0}^s \} = {f}( \{ \bm{z}^s \};{\bm{\theta}_{0}}), s=0,1,2,3$.	
	\FOR{$k=1$ \TO $K$} 
        \STATE Given $\{ \bm{x}_{k-1}^s \}$, solve for blur kernels $\{ \bm{h}_{k}^s \}, s=0,1,2,3$ using Eq.(\ref{eq:eq6});
        \STATE Given $\{ \bm{h}_{k}^s \}$ and ${\bm{\theta}}_{k-1}$, update parameters ${\bm{\theta}}_{k}$ of the multi-scale generator network using the Adam optimizer\cite{kingma2014adam};
        \STATE Given ${\bm{\theta}}_{k}$, update unblurred images $\{ \bm{x}_{k}^s \} = {f}(\{ \bm{z}^s \};{\bm{\theta}_{k}}), s=0,1,2,3$;
        \STATE $k = k+1$;
        \ENDFOR
    \end{algorithmic}
    The final deblurred image is ${\bm x} = \bm{x}_{K}^{0}$ and the estimated blur kernel is ${\bm h} = \bm{h}_{K}^{0}$. 
    \label{alg:algorithm_1}
\end{algorithm}

\section{Experiments}
\label{sec:experiments}
In this section, we conduct several experiments on Lai et al.'s dataset \cite{Lai2016} and Kohler et al.'s dataset \cite{Kohler2012} for quantitative and qualitative evaluations. Peak Signal-to-Noise Ratio (PSNR) and Structural Similarity Index Measure (SSIM) are measured quantitatively for assessing the quality of the deblurred image. The proposed Self-MSNet is compared with state-of-the-art mathematical optimization-based and deep learning-based methods including supervised learning- and unsupervised learning-based methods. Our Self-MSNet is implemented using the Pytorch platform. The experiments are run on a PC with a 2.5GHz Xeon E5-2678 CPU (128GB) and an NVIDIA Geforce RTX 3090 GPU (24GB).

\subsection{Experimental Results on Lai's Dataset}
Lai et al.'s dataset \cite{Lai2016} comprises 100 uniformly blurred images, which are generated by convolving 25 sharp images with 4 ground truth kernels while adding 1\% Gaussian noise. These blur kernels are of sizes ranging from $31 \times 31$ to $75 \times 75$. Experiments are conducted on this dataset using the following settings. The maximum allowed number of iterations is set to 2000, the initial learning rate is 0.001 which is halved every 500 iterations, and the regularization parameters for the blur kernel are set to $\lambda_{h}=10$ and $\gamma=10$. Considering the negligible noise in the blurred images of Lai's dataset, we set the TV regularization parameter of the image as $\lambda_{x} = 0$ such that only the data fidelity term remains in the loss function.

Our Self-MSNet is compared with the mathematical optimization-based methods \cite{fergus2006removing, cho2009fast, xu2010two, whyte2010nonUniform, krishnan2011normalized, levin2011understanding, xu2013unnatural, zhang2013multiImage, sun2013ICCPedgeBased, Zhong2013handling, whyte2014IJCVshaken, pan2014CVPRtextL0, michaeli2014blind, perrone2014total, RDS2014, pan2016PAMIL0text, yan2017extremechannel, pan2018PAMIdarkchannel, bai2019TIPgraph, Yang2019CVPRAdaptiveEdgeSelection, chen2019blind, chen2020enhanced, chen2020OID, shao2020gradient, wen2021TCSVT, shao2023Revisiting}, the supervised learning-based methods \cite{sun2015CVPRlearning, Pan2017ICCVdataFitting, nah2017deep, kupyn2018deblurgan, tao2018scale, kupyn2019deblurganv2, kaufman2020deblurring, cho2021rethinking}, and the self-supervised learning-based methods \cite{ren2020neural,bai2023fastselfdeblur,shao2023Revisiting,li2023deepEM,Chen2023TCSVTensemble}. Table \ref{tab:Lai-ave-PSNRssim} presents the average PSNR and SSIM results of these methods on Lai's dataset. The PSNR between the ground truth image and the restored image is the ratio of the maximum possible pixel value of the image and the mean squared error between images, while SSIM measures the weighted sum of mean value, standard deviation, and correlation coefficient between images. A larger value is associated with higher image quality. The results of \cite{RDS2014,wen2021TCSVT,cho2021rethinking} are generated using their released codes, those of \cite{shao2020gradient,shao2023Revisiting,kaufman2020deblurring,ren2020neural,bai2023fastselfdeblur,li2023deepEM,Chen2023TCSVTensemble} are quoted from their original papers, and the remaining results \cite{fergus2006removing,whyte2010nonUniform, levin2011understanding,zhang2013multiImage, sun2013ICCPedgeBased, Zhong2013handling, whyte2014IJCVshaken, pan2014CVPRtextL0,yan2017extremechannel, pan2018PAMIdarkchannel, bai2019TIPgraph, Yang2019CVPRAdaptiveEdgeSelection, chen2019blind, chen2020enhanced, chen2020OID,sun2015CVPRlearning, Pan2017ICCVdataFitting, nah2017deep, kupyn2018deblurgan, tao2018scale, kupyn2019deblurganv2} are quoted from other literature.

As listed in Tab.\ref{tab:Lai-ave-PSNRssim}, ${}^\triangle$ indicates the mathematical optimization-based models that obtain the final deblurred images using non-blind deblurring methods, ${}^{\dag}$ identifies the supervised learning-based methods that require additional datasets for network training, and \textbf{bolded} refers to the optimal result. One can see that our Self-MSNet achieves the highest PSNR and SSIM values among all the compared methods. Since the mathematical optimization-based image deblurring methods typically estimate blur kernels in a coarse-to-fine manner, and require non-blind deblurring methods to generate the final deblurred images, deep learning-based methods tend to fall behind mathematical optimization-based methods in the field of image deblurring. Nevertheless, our method achieves superior quantitative evaluation results against all the compared mathematical optimization-based methods. Even without ground truth images or blur kernels for network training, the proposed method gains superior results compared to all the supervised learning-based methods metioned above. Our Self-MSNet, as well as \cite{ren2020neural,bai2023fastselfdeblur,shao2023Revisiting,li2023deepEM,Chen2023TCSVTensemble} are all self-supervised learning-based methods with the joint estimation of the latent image and the blur kernel. In terms of PSNR results, the proposed method outperforms SelfDeblur \cite{ren2020neural} by 4.45dB, Fast-SelfDeblur \cite{bai2023fastselfdeblur} by 2.35dB, the method of \cite{li2023deepEM} by 1.8dB, DEBID \cite{Chen2023TCSVTensemble} by 1.22dB, and RDP-SelfDeblur \cite{shao2023Revisiting} by 3.64dB.

\begin{table}[h!] %表示和前文一致
    \caption{Average PSNR/SSIM Comparisons with State-of-the-art Methods on Lai's Dataset.}
    \label{tab:Lai-ave-PSNRssim}
    \centering
    \begin{tabular}{p{3.3cm}p{1.3cm}<{\centering}p{1.3cm}<{\centering}}
        \hline%\toprule%\hline
        Method  &PSNR(dB)   &SSIM \\
        \hline%\midrule
    Fergus et al.${}^\triangle$ \cite{fergus2006removing}    &15.60    &0.5080 \\           
    Cho and Lee${}^\triangle$ \cite{cho2009fast} &17.06	&0.4801 \\                             
    Xu and Jia${}^\triangle$ \cite{xu2010two} &20.18	&0.7080  \\                            
    Whyte et al.${}^\triangle$ \cite{whyte2010nonUniform} &17.44	&0.5700  \\                
    Krishnan et al.${}^\triangle$ \cite{krishnan2011normalized} &17.72	&0.4875 \\             
    Levin et al.${}^\triangle$ \cite{levin2011understanding} &16.57	&0.5690 \\                             
    Xu et al.${}^\triangle$ \cite{xu2013unnatural} &19.23	&0.6593 \\                         
    Zhang et al.${}^\triangle$ \cite{zhang2013multiImage} &16.70	&0.5660 \\                 
    Sun et al.${}^\triangle$ \cite{sun2013ICCPedgeBased} &20.47	&0.7180  \\                
    Zhong et al.${}^\triangle$ \cite{Zhong2013handling} &18.95	&0.7190  \\                    
    Whyte et al.${}^\triangle$ \cite{whyte2014IJCVshaken} &24.41	&0.7312  \\                
    Pan et al.${}^\triangle$ \cite{pan2014CVPRtextL0} &19.33	&0.7550 \\                 
    Michaeli and Irani${}^\triangle$ \cite{michaeli2014blind} &18.37	&0.5181  \\            
    Perrone and Favaro${}^\triangle$ \cite{perrone2014total} &18.48	&0.6130  \\                
    RDS${}^\triangle$ \cite{RDS2014} &20.39	&0.6569 \\             
    Pan-$\ell_0{}^\triangle$ \cite{pan2016PAMIL0text} &18.54	&0.6248  \\                                        
    ECP${}^\triangle$ \cite{yan2017extremechannel}    &21.02    &0.6750  \\                    
    DCP${}^\triangle$ \cite{pan2018PAMIdarkchannel} &19.89	&0.6656  \\                        
    Bai et al.${}^\triangle$ \cite{bai2019TIPgraph} &20.15	&0.7271  \\                        
    Yang et al.${}^\triangle$ \cite{Yang2019CVPRAdaptiveEdgeSelection} &21.79	&0.7040  \\
    Chen et al.${}^\triangle$ \cite{chen2019blind} &21.82	&0.8135  \\                        
    ESM${}^\triangle$ \cite{chen2020enhanced} &22.47	&0.8517\\
    OID${}^\triangle$ \cite{chen2020OID} &21.17	&0.7750\\
    Shao et al.${}^\triangle$ \cite{shao2020gradient} &21.60	&0.8148\\
    Wen et al.${}^\triangle$ \cite{wen2021TCSVT} &22.04	&0.7458\\
    RDP${}^\triangle$ \cite{shao2023Revisiting} &22.27	&0.8613\\
    Sun et al.${}^{\dag}$ \cite{sun2015CVPRlearning}    &24.58    &0.7379\\
    Pan et al.${}^\dag$ \cite{Pan2017ICCVdataFitting} &17.80	&0.5590\\
    DeepDeblur${}^{\dag}$ \cite{nah2017deep} &17.42	&0.4935\\
    DeblurGAN${}^{\dag}$ \cite{kupyn2018deblurgan} &17.49	&0.4731\\
    SRN-DeblurNet${}^{\dag}$ \cite{tao2018scale} &17.65	&0.5360\\
    DeblurGAN-v2${}^{\dag}$\cite{kupyn2019deblurganv2}    &17.98    &0.5950\\
    Kaufman et al.${}^{\dag}$ \cite{kaufman2020deblurring}    &20.89    &0.8190\\
    MIMO-UNet${}^{\dag}$\cite{cho2021rethinking}    &18.06    &0.4984\\
    SelfDeblur \cite{ren2020neural} &20.97	&0.7524\\
    Fast-SelfDeblur \cite{bai2023fastselfdeblur} &23.07	&0.8124\\
    Li et al.\cite{li2023deepEM}    &23.62    &0.7660\\
    DEBID \cite{Chen2023TCSVTensemble} &24.20	&0.8310\\
    RDP-SelfDeblur \cite{shao2023Revisiting} &21.78	&0.7903\\
    Self-MSNet-S &23.15	&0.7346\\
    Self-MSNet ({\textbf{Ours}}) &\textbf{25.42}	&\textbf{0.8874}\\
    %\specialrule{0em}{1.5pt}{1.5pt}
    %\specialrule{0em}{2pt}{2pt}
		\hline%\bottomrule
    \end{tabular}
\end{table}

\begin{figure*}[!tbp]
	\centering
	\begin{minipage}{0.22\linewidth}
		\centering
		\includegraphics[width=1\linewidth]{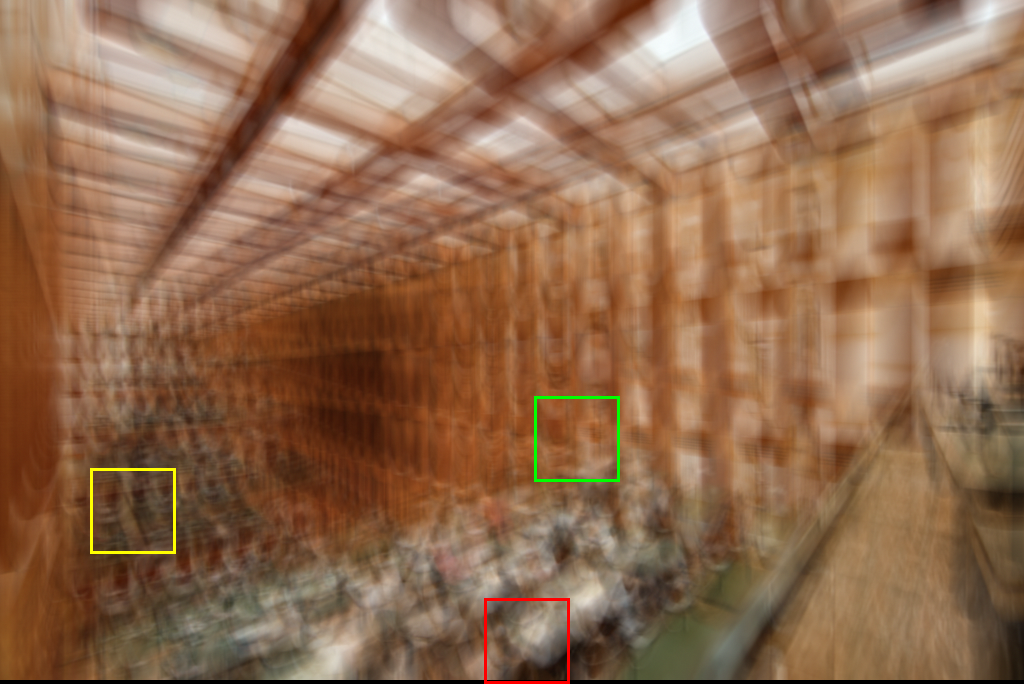}
		\vskip 4pt
		\includegraphics[width=1\linewidth]{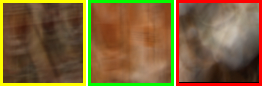}
		\caption*{(a)Blurred image \protect\\ {\textcolor{white}{***}}\centering}
		\label{lai_manmade_01_kernel_04_Blurry image}%文中引用该图片代号
	\end{minipage}
	\begin{minipage}{0.22\linewidth}
		\centering
		\includegraphics[width=1\linewidth]{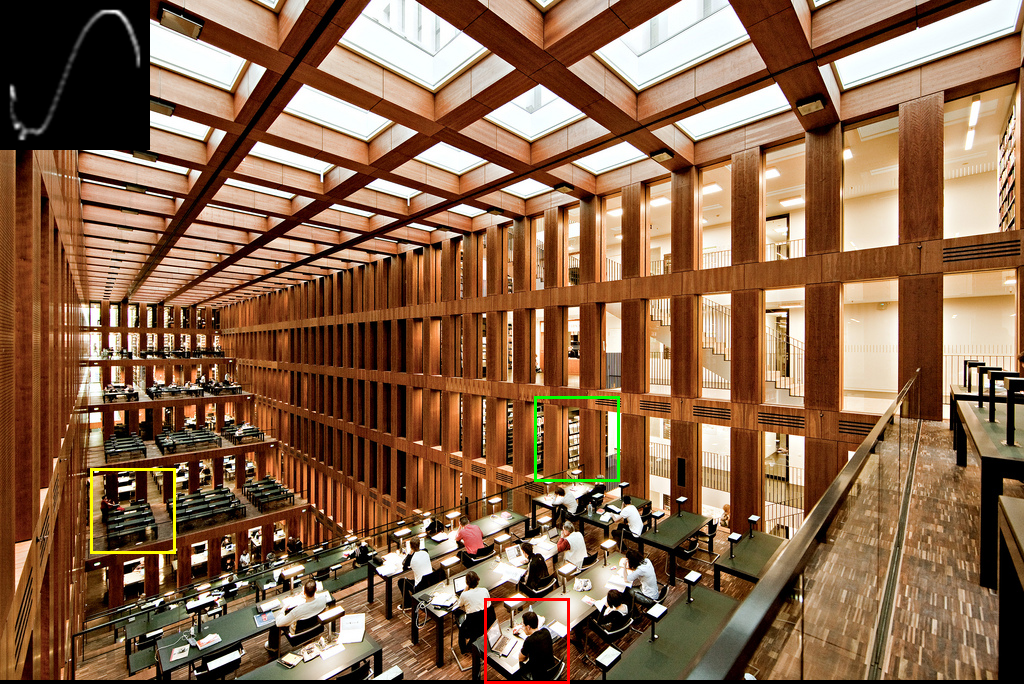}
		\vskip 4pt
		\includegraphics[width=1\linewidth]{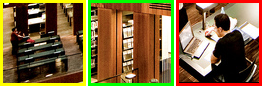}
		\caption*{(b)Ground-truth \protect\\ {\textcolor{white}{***}}\centering}
		\label{lai_manmade_01_kernel_04_Ground-truth}%文中引用该图片代号
	\end{minipage}
	\begin{minipage}{0.22\linewidth}
		\centering
		\includegraphics[width=1\linewidth]{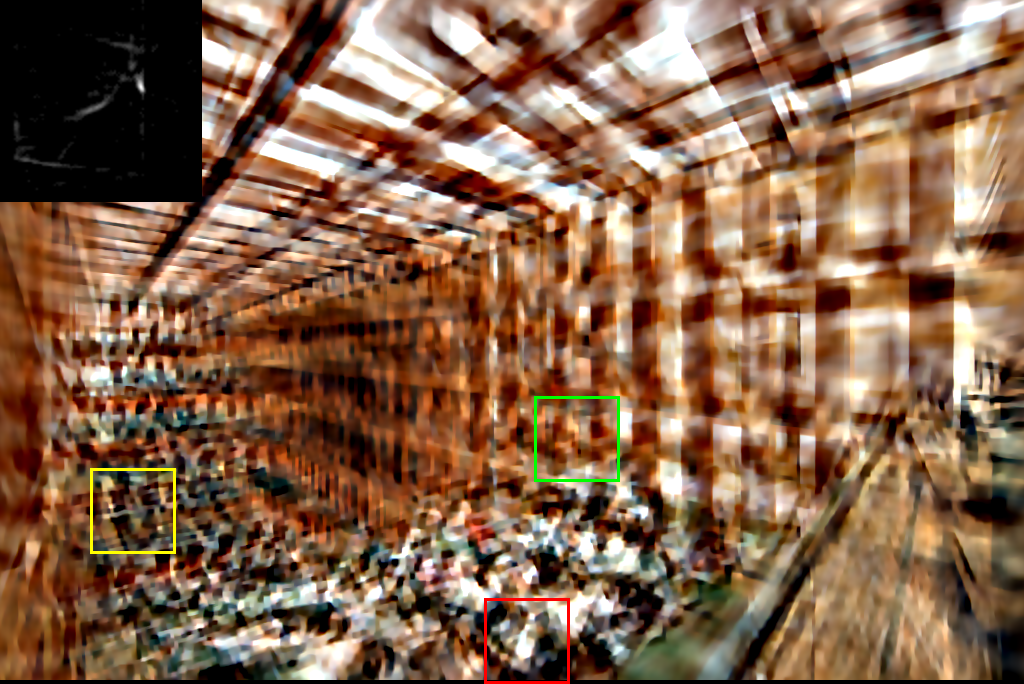}
		\vskip 4pt
		\includegraphics[width=1\linewidth]{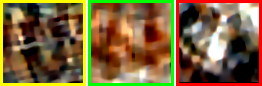}
		\caption*{(c)Cho and Lee\cite{cho2009fast} \protect\\ {PSNR: 11.77}\centering}
		\label{lai_manmade_01_kernel_04_cho}%文中引用该图片代号
	\end{minipage}
        \begin{minipage}{0.22\linewidth}
		\centering
		\includegraphics[width=1\linewidth]{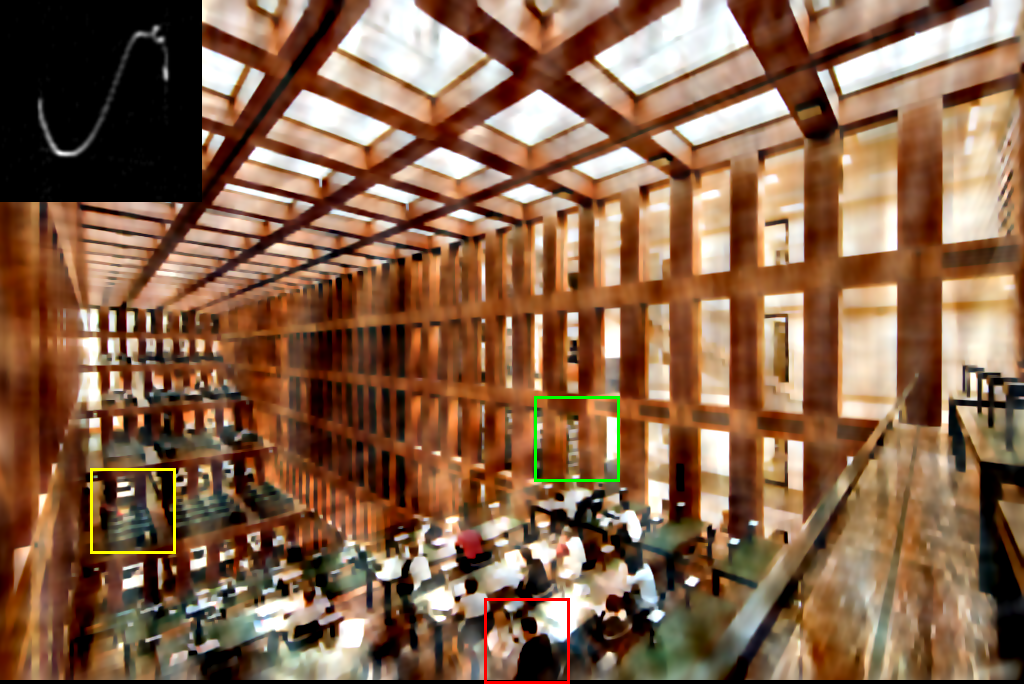}
		\vskip 4pt
		\includegraphics[width=1\linewidth]{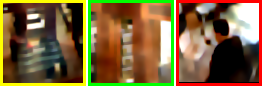}
		\caption*{(d)Xu and Jia\cite{xu2010two} \protect\\ {PSNR: 19.05}\centering}
		\label{lai_manmade_01_kernel_04_xuandjia}%文中引用该图片代号
	\end{minipage}
        \vskip 7pt
        \begin{minipage}{0.22\linewidth}
		\centering
		\includegraphics[width=1\linewidth]{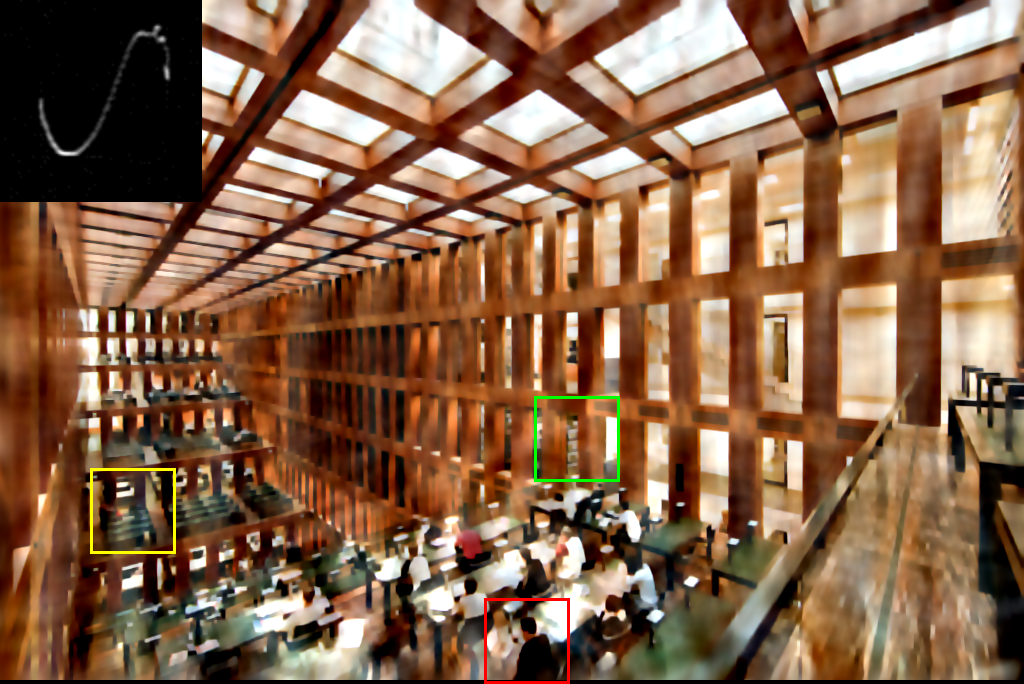}
		\vskip 4pt
		\includegraphics[width=1\linewidth]{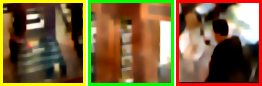}
		\caption*{(e)Xu et al.\cite{xu2013unnatural} \protect\\ {PSNR: 19.17}\centering}
		\label{lai_manmade_01_kernel_04_xuunnatural}%文中引用该图片代号
	\end{minipage} 
        \begin{minipage}{0.22\linewidth}
		\centering
		\includegraphics[width=1\linewidth]{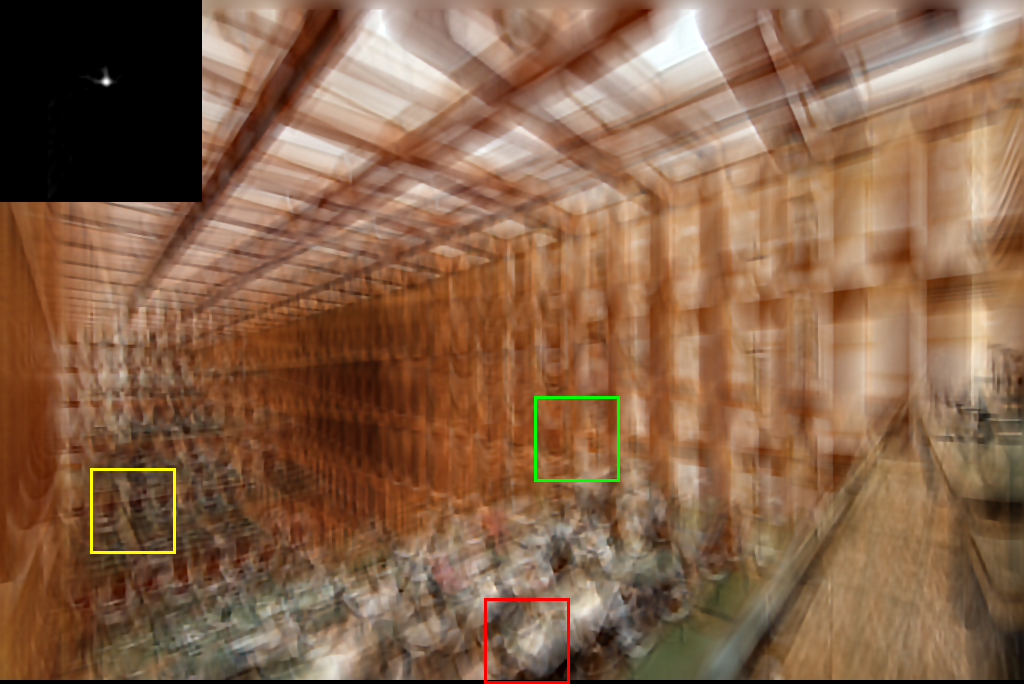}
		\vskip 4pt
		\includegraphics[width=1\linewidth]{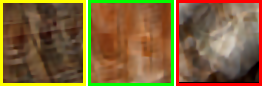}
		\caption*{(f)Michaeli and Irani\cite{michaeli2014blind} \protect\\ {PSNR: 12.61}\centering}
		\label{lai_manmade_01_kernel_04_Michaeli}%文中引用该图片代号
	\end{minipage} 
	\begin{minipage}{0.22\linewidth}
		\centering
		\includegraphics[width=1\linewidth]{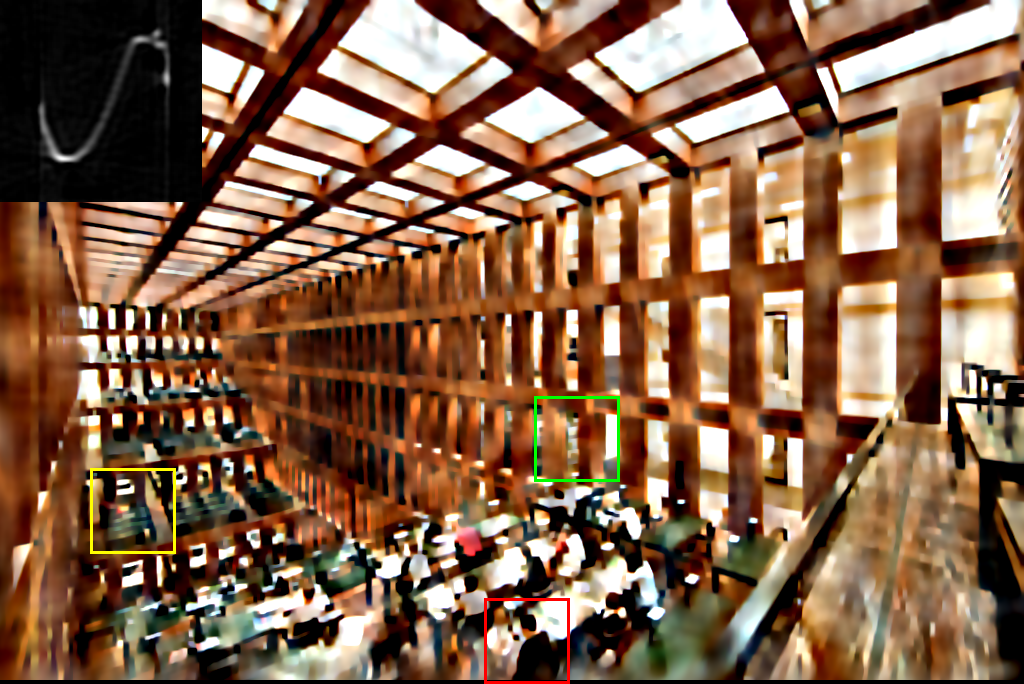}
		\vskip 4pt
		\includegraphics[width=1\linewidth]{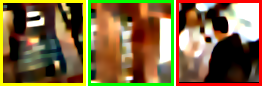}
		\caption*{(g)Perrone and Favaro\cite{perrone2014total} \protect\\ {PSNR: 18.04}\centering}
		\label{lai_manmade_01_kernel_04_Perrone}%文中引用该图片代号
	\end{minipage}
	\begin{minipage}{0.22\linewidth}
		\centering
		\includegraphics[width=1\linewidth]{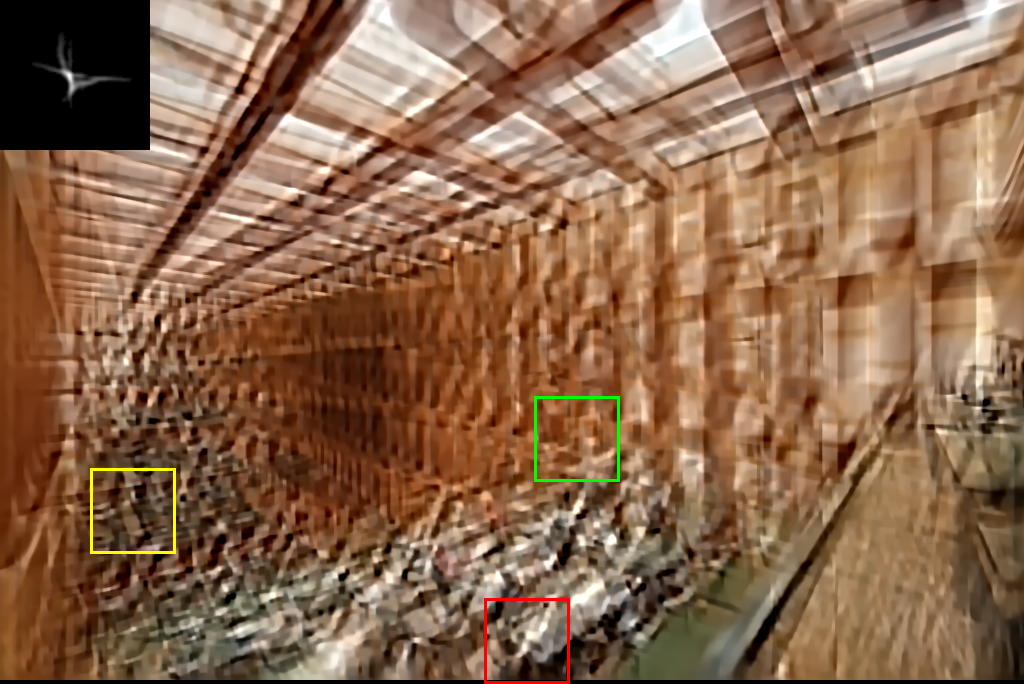}
		\vskip 4pt
		\includegraphics[width=1\linewidth]{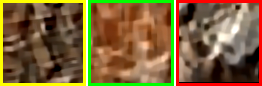}
		\caption*{(h)Pan et al.\cite{pan2018PAMIdarkchannel} \protect\\ {PSNR: 12.27}\centering}
		\label{lai_manmade_01_kernel_04_Pan}%文中引用该图片代号
	\end{minipage}
        \vskip 7pt
        \begin{minipage}{0.22\linewidth}
		\centering
		\includegraphics[width=1\linewidth]{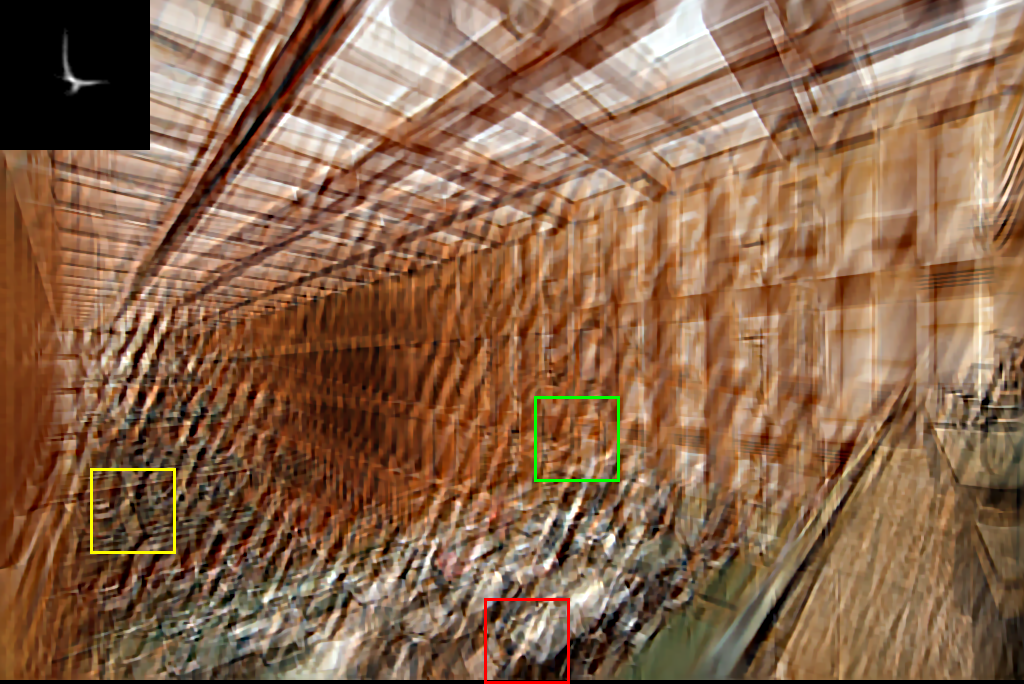}
		\vskip 4pt
		\includegraphics[width=1\linewidth]{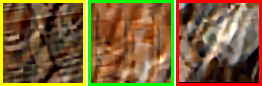}
		\caption*{(i)Wen et al.\cite{wen2021TCSVT} \protect\\ {PSNR: 12.47}\centering}
		\label{lai_manmade_01_kernel_04_wen}%文中引用该图片代号
	\end{minipage}
        \begin{minipage}{0.22\linewidth}
		\centering
		\includegraphics[width=1\linewidth]{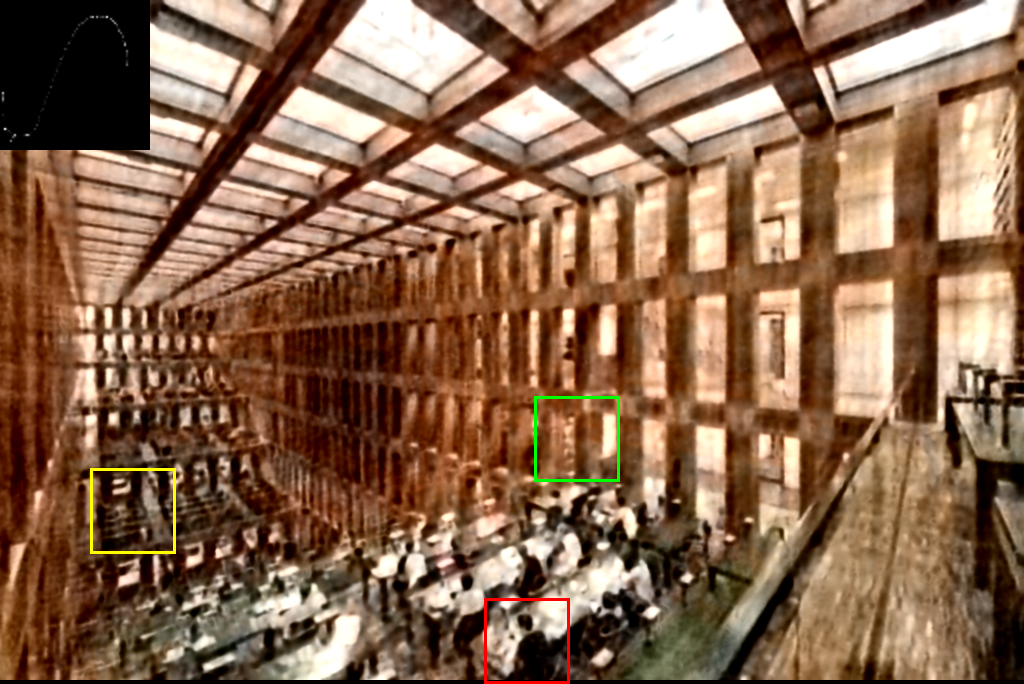}
		\vskip 4pt
		\includegraphics[width=1\linewidth]{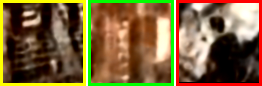}
		\caption*{(j)SelfDeblur\cite{ren2020neural} \protect\\ {PSNR: 19.45}\centering}
		\label{lai_manmade_01_kernel_04_SelfDeblur}%文中引用该图片代号
	\end{minipage}	 
	\begin{minipage}{0.22\linewidth}
		\centering
		\includegraphics[width=1\linewidth]{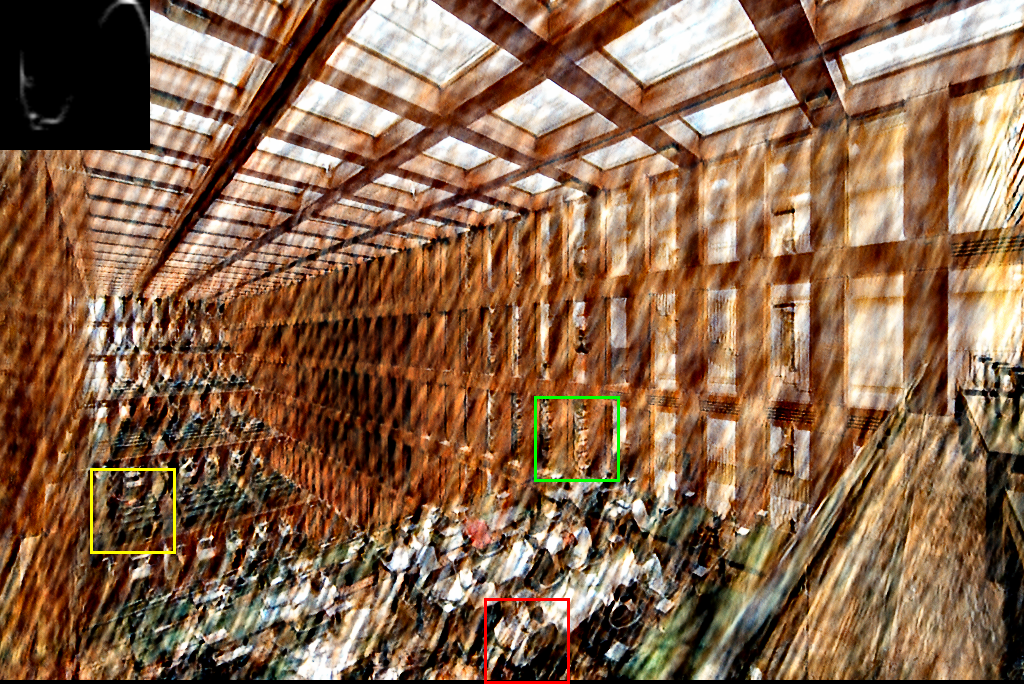}
		\vskip 4pt
		\includegraphics[width=1\linewidth]{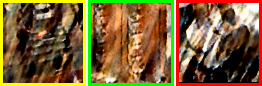}
		\caption*{(k)Fast-SelfDeblur\cite{bai2023fastselfdeblur} \protect\\ {PSNR: 15.40}\centering}
		\label{lai_manmade_01_kernel_04_Fast-SelfDeblur}%文中引用该图片代号
	\end{minipage}
	\begin{minipage}{0.22\linewidth}
		\centering
		\includegraphics[width=1\linewidth]{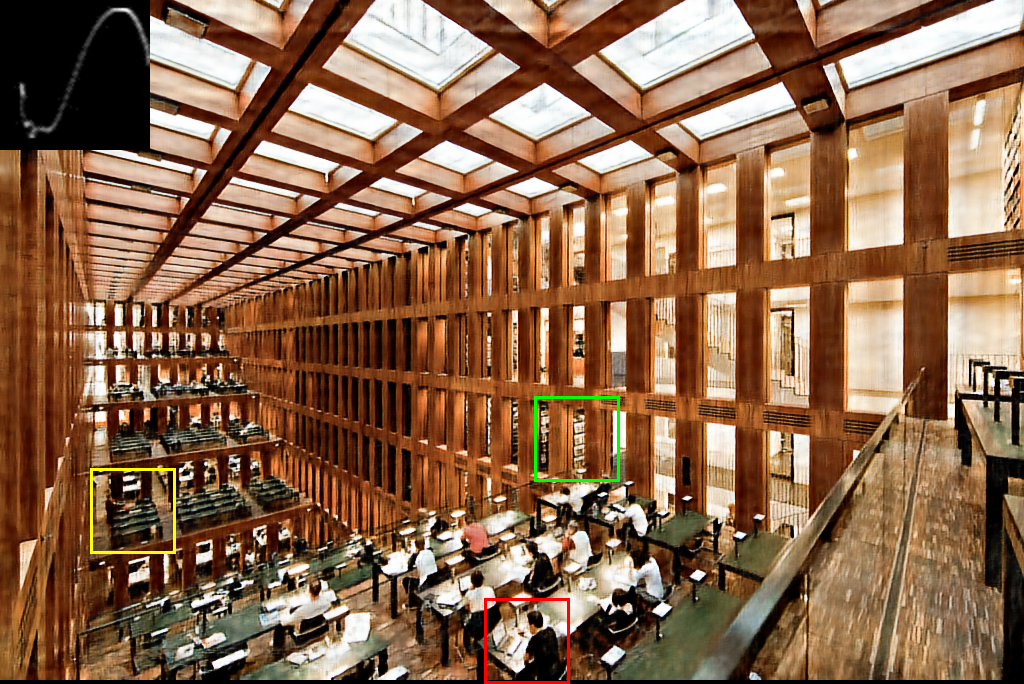}
		\vskip 4pt
		\includegraphics[width=1\linewidth]{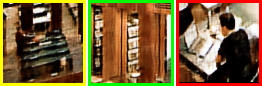}
		\caption*{(l)Self-MSNet (\textbf{Ours}) \protect\\ {PSNR: 24.75}\centering}
		\label{lai_manmade_01_kernel_04_ours}%文中引用该图片代号
	\end{minipage}
        %\vskip -8pt
	\caption{Visual comparison on an example image of the 'man-made' category with 4th blur kernel from Lai's dataset.}
	\label{fig:Lai-visual-manmade_01_kernel_04}
\end{figure*}

\begin{figure*}[!tbp]
	\centering
	\begin{minipage}{0.22\linewidth}
		\centering
		\includegraphics[width=1\linewidth]{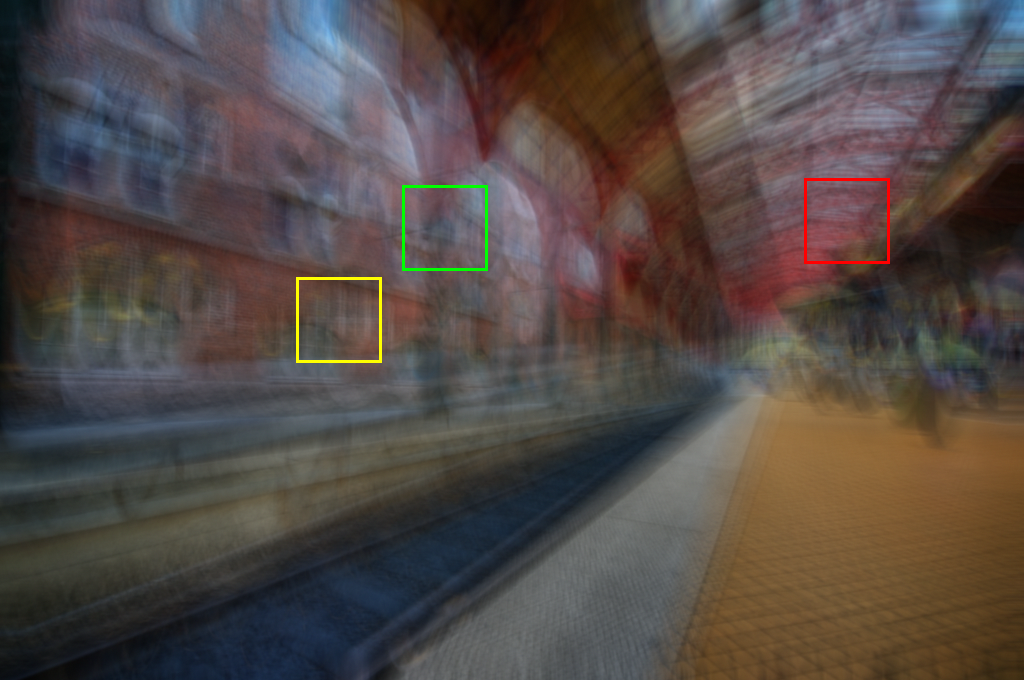}
		\vskip 4pt
		\includegraphics[width=1\linewidth]{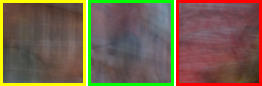}
		\caption*{(a)Blurred image \protect\\ {\textcolor{white}{***}}\centering}
		\label{lai_manmade_05_kernel_04_Blurry image}%文中引用该图片代号
	\end{minipage}
	\begin{minipage}{0.22\linewidth}
		\centering
		\includegraphics[width=1\linewidth]{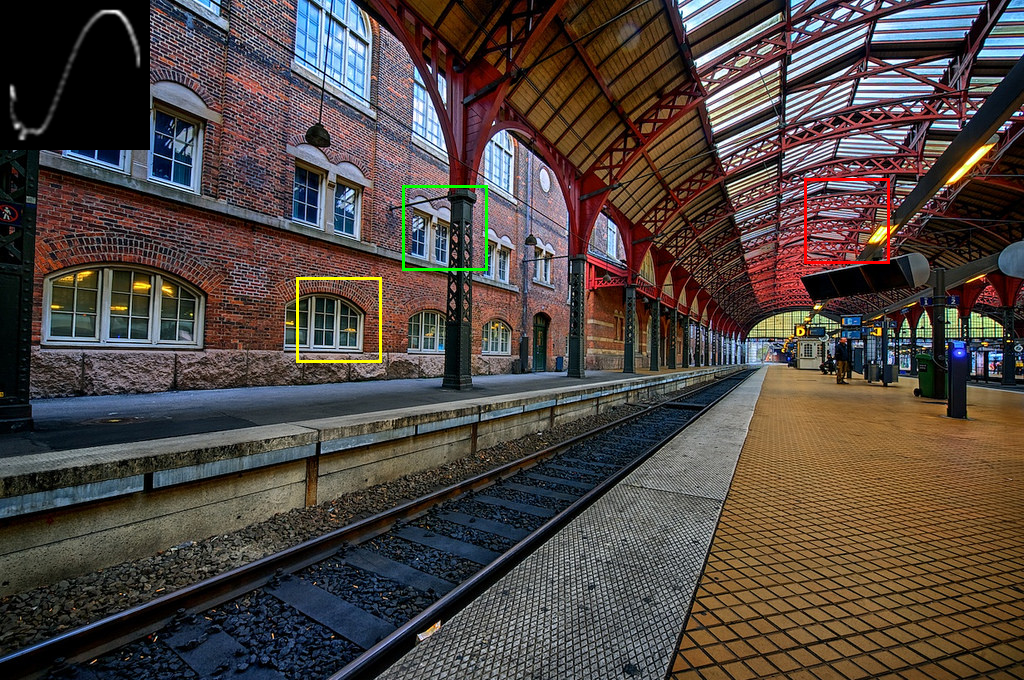}
		\vskip 4pt
		\includegraphics[width=1\linewidth]{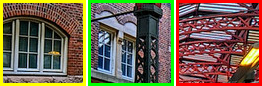}
		\caption*{(b)Ground-truth \protect\\ {\textcolor{white}{***}}\centering}
		\label{lai_manmade_05_kernel_04_Ground-truth}%文中引用该图片代号
	\end{minipage}
	\begin{minipage}{0.22\linewidth}
		\centering
		\includegraphics[width=1\linewidth]{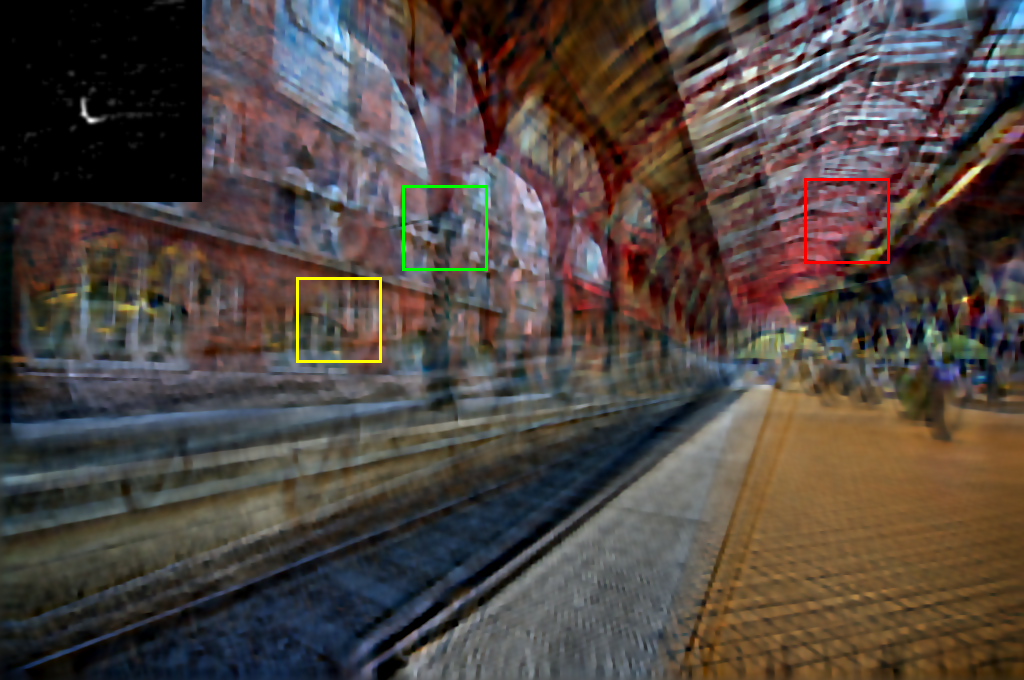}
		\vskip 4pt
		\includegraphics[width=1\linewidth]{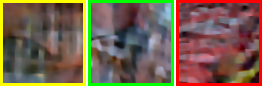}
		\caption*{(c)Cho and Lee\cite{cho2009fast} \protect\\ {PSNR: 15.18}\centering}
		\label{lai_manmade_05_kernel_04_cho}%文中引用该图片代号
	\end{minipage}
        \begin{minipage}{0.22\linewidth}
		\centering
		\includegraphics[width=1\linewidth]{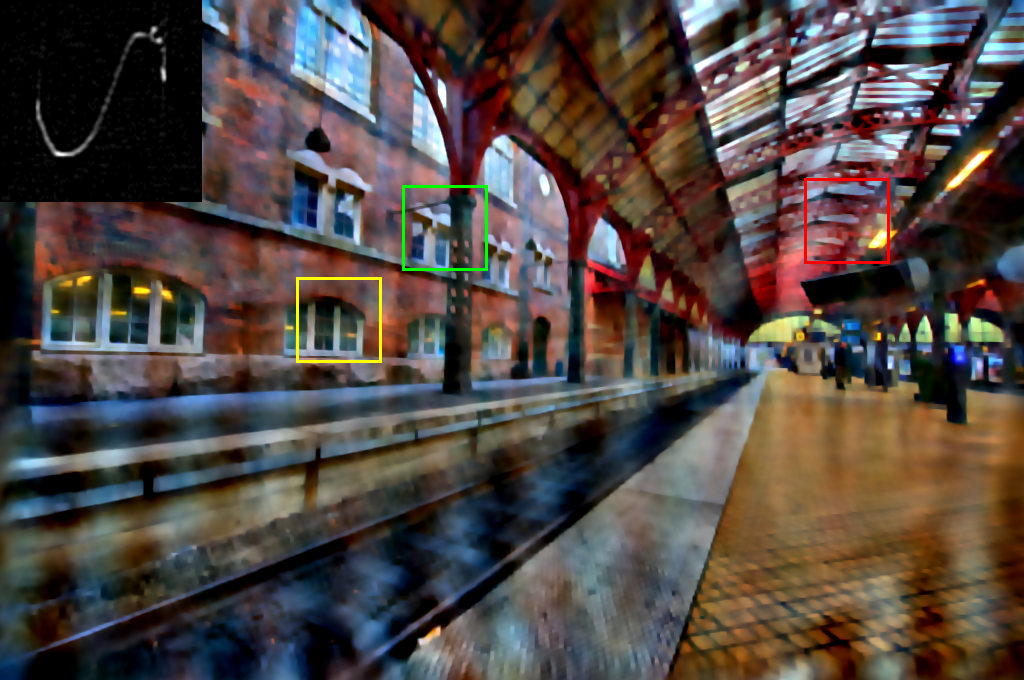}
		\vskip 4pt
		\includegraphics[width=1\linewidth]{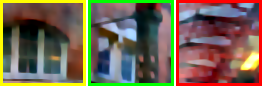}
		\caption*{(d)Xu and Jia\cite{xu2010two} \protect\\ {PSNR: 18.46}\centering}
		\label{lai_manmade_05_kernel_04_xuandjia}%文中引用该图片代号
	\end{minipage}
        \vskip 7pt
        \begin{minipage}{0.22\linewidth}
		\centering
		\includegraphics[width=1\linewidth]{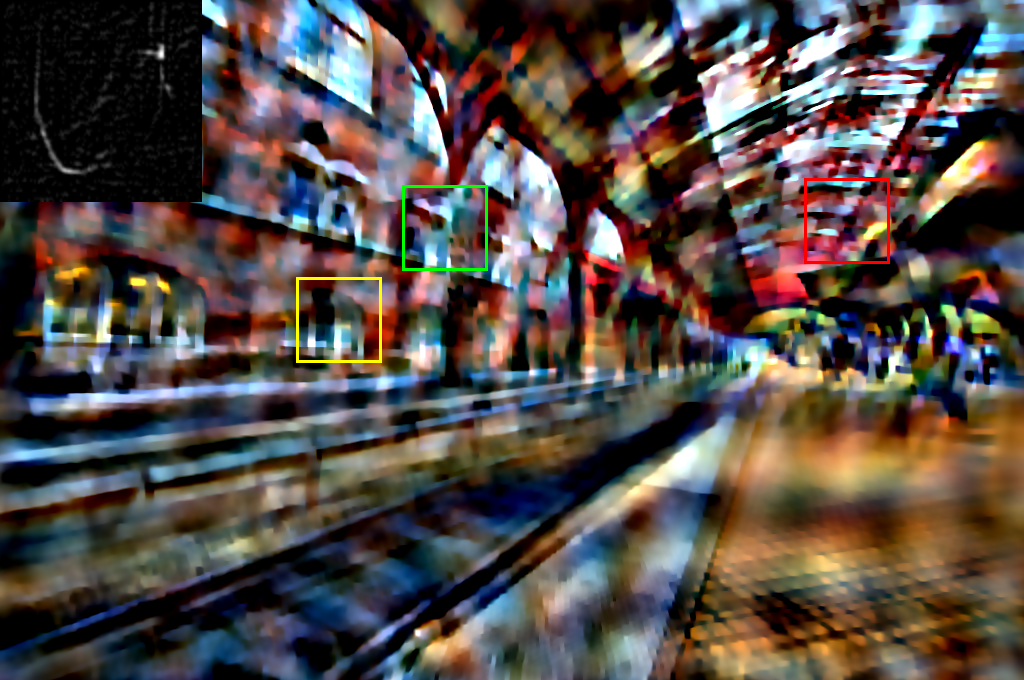}
		\vskip 4pt
		\includegraphics[width=1\linewidth]{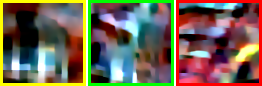}
		\caption*{(e)Xu et al.\cite{xu2013unnatural} \protect\\ {PSNR: 14.43}\centering}
		\label{lai_manmade_05_kernel_04_xuunnatural}%文中引用该图片代号
	\end{minipage} 
        \begin{minipage}{0.22\linewidth}
		\centering
		\includegraphics[width=1\linewidth]{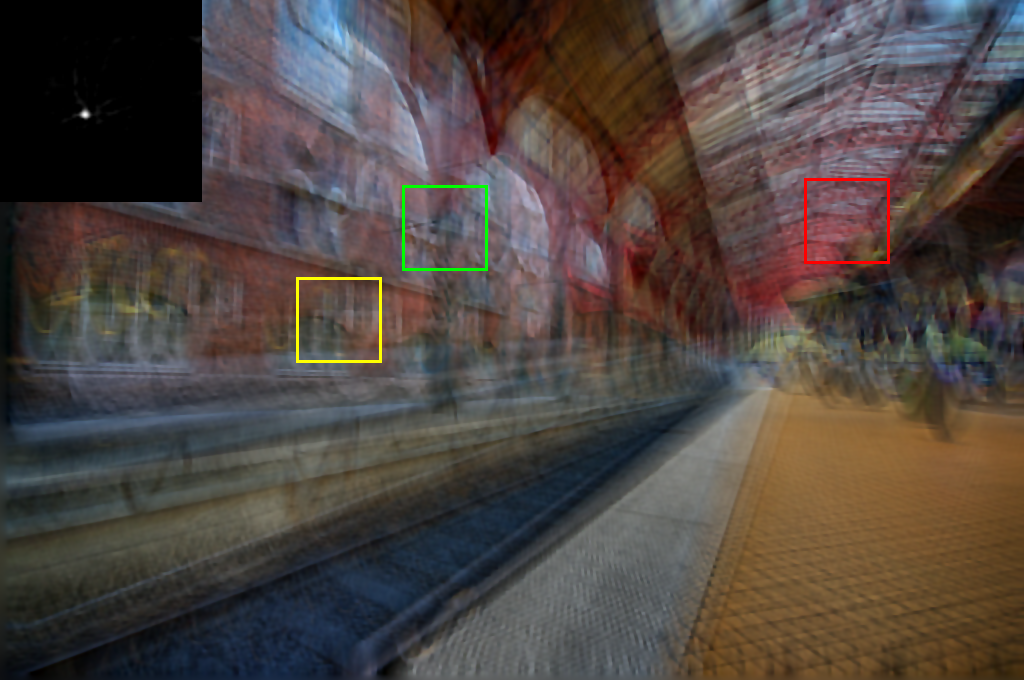}
		\vskip 4pt
		\includegraphics[width=1\linewidth]{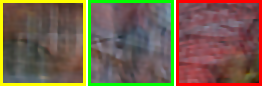}
		\caption*{(f)Michaeli and Irani\cite{michaeli2014blind} \protect\\ {PSNR: 16.02}\centering}
		\label{lai_manmade_05_kernel_04_Michaeli}%文中引用该图片代号
	\end{minipage} 
	\begin{minipage}{0.22\linewidth}
		\centering
		\includegraphics[width=1\linewidth]{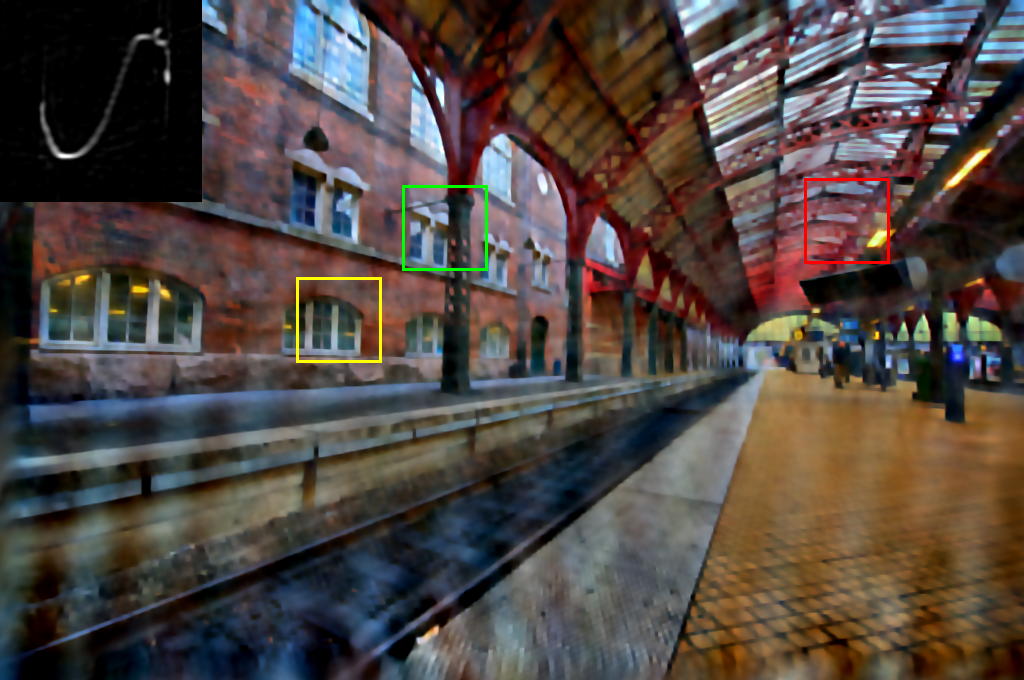}
		\vskip 4pt
		\includegraphics[width=1\linewidth]{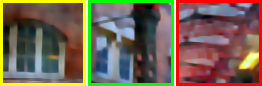}
		\caption*{(g)Perrone and Favaro\cite{perrone2014total} \protect\\ {PSNR: 19.17}\centering}
		\label{lai_manmade_05_kernel_04_Perrone}%文中引用该图片代号
	\end{minipage}
	\begin{minipage}{0.22\linewidth}
		\centering
		\includegraphics[width=1\linewidth]{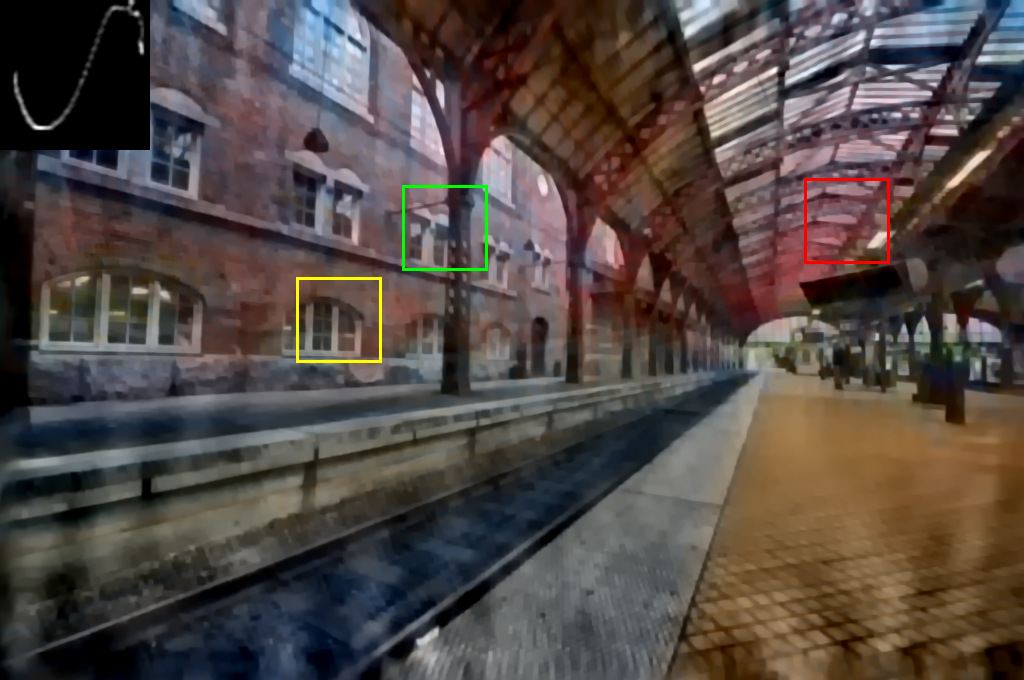}
		\vskip 4pt
		\includegraphics[width=1\linewidth]{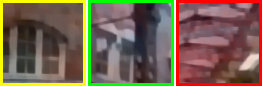}
		\caption*{(h)Pan et al.\cite{pan2018PAMIdarkchannel} \protect\\ {PSNR: 19.23}\centering}
		\label{lai_manmade_05_kernel_04_Pan}%文中引用该图片代号
	\end{minipage}
        \vskip 7pt
        \begin{minipage}{0.22\linewidth}
		\centering
		\includegraphics[width=1\linewidth]{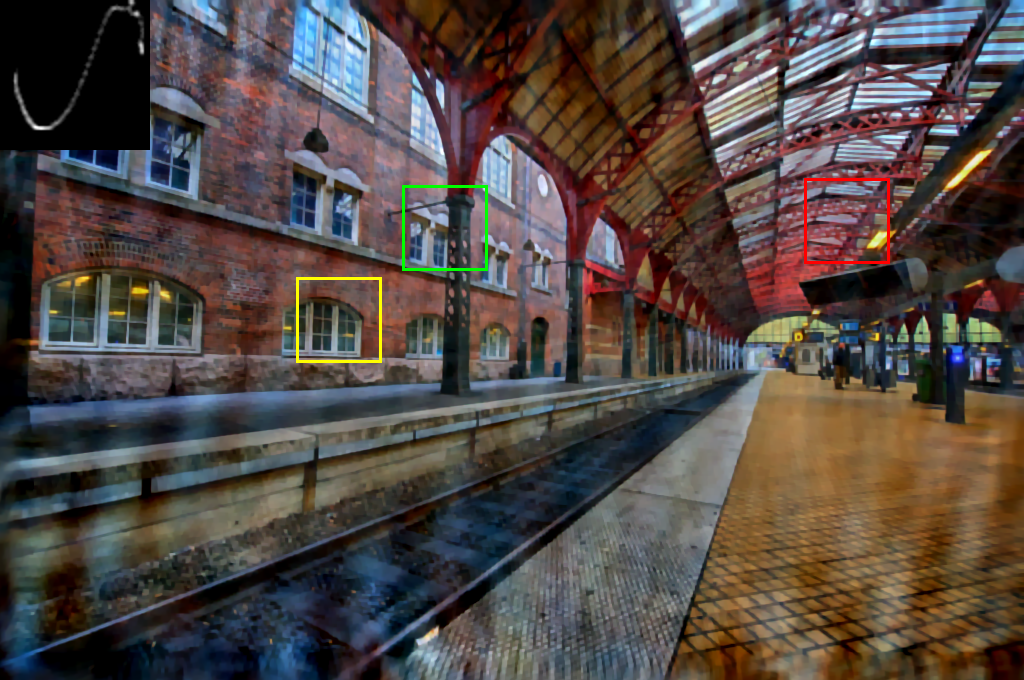}
		\vskip 4pt
		\includegraphics[width=1\linewidth]{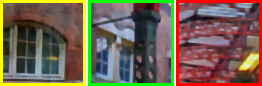}
		\caption*{(i)Wen et al.\cite{wen2021TCSVT} \protect\\ {PSNR: 20.06}\centering}
		\label{lai_manmade_05_kernel_04_wen}%文中引用该图片代号
	\end{minipage}
        \begin{minipage}{0.22\linewidth}
		\centering
		\includegraphics[width=1\linewidth]{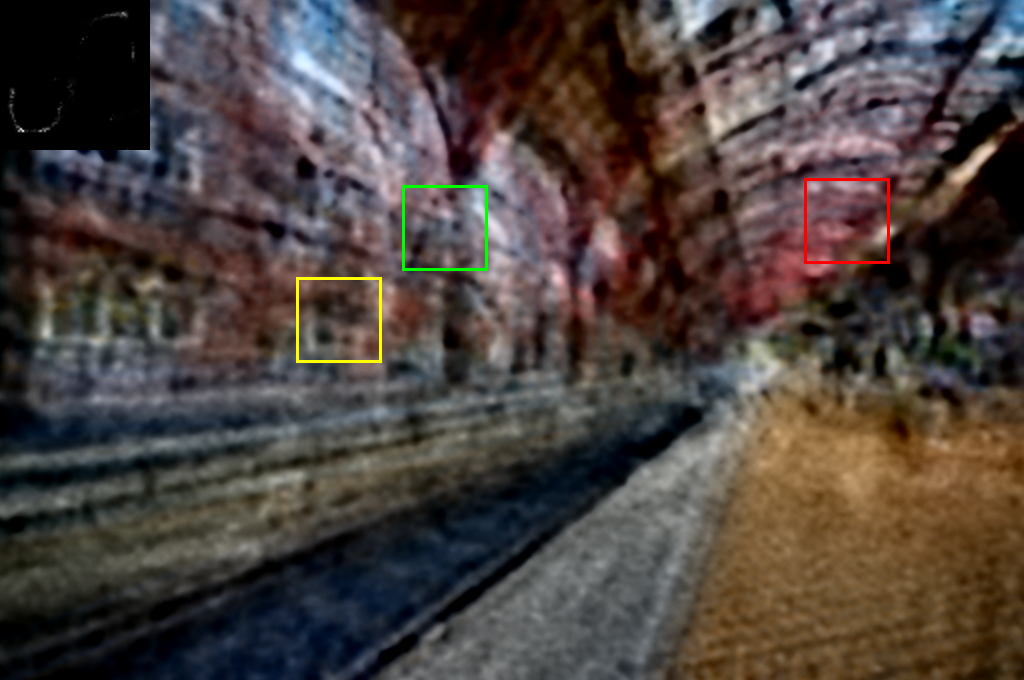}
		\vskip 4pt
		\includegraphics[width=1\linewidth]{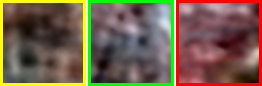}
		\caption*{(j)SelfDeblur\cite{ren2020neural} \protect\\ {PSNR: 16.73}\centering}
		\label{lai_manmade_05_kernel_04_SelfDeblur}%文中引用该图片代号
	\end{minipage}	 
	\begin{minipage}{0.22\linewidth}
		\centering
		\includegraphics[width=1\linewidth]{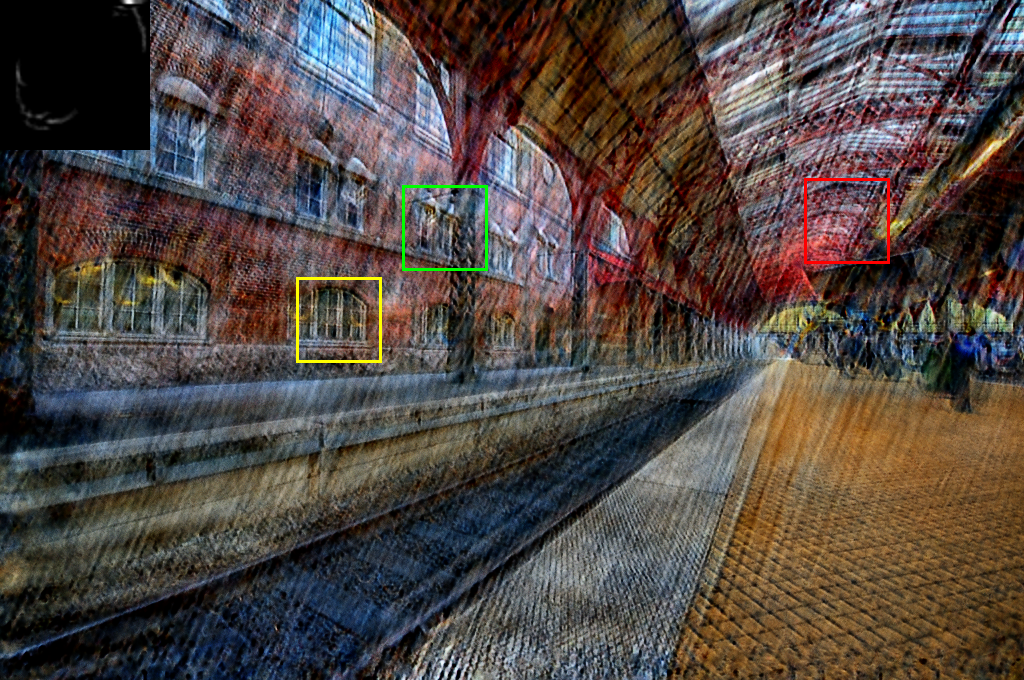}
		\vskip 4pt
		\includegraphics[width=1\linewidth]{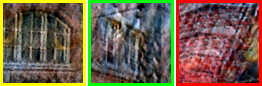}
		\caption*{(k)Fast-SelfDeblur\cite{bai2023fastselfdeblur} \protect\\ {PSNR: 16.92}\centering}
		\label{lai_manmade_05_kernel_04_Fast-SelfDeblur}%文中引用该图片代号
	\end{minipage}
	\begin{minipage}{0.22\linewidth}
		\centering
		\includegraphics[width=1\linewidth]{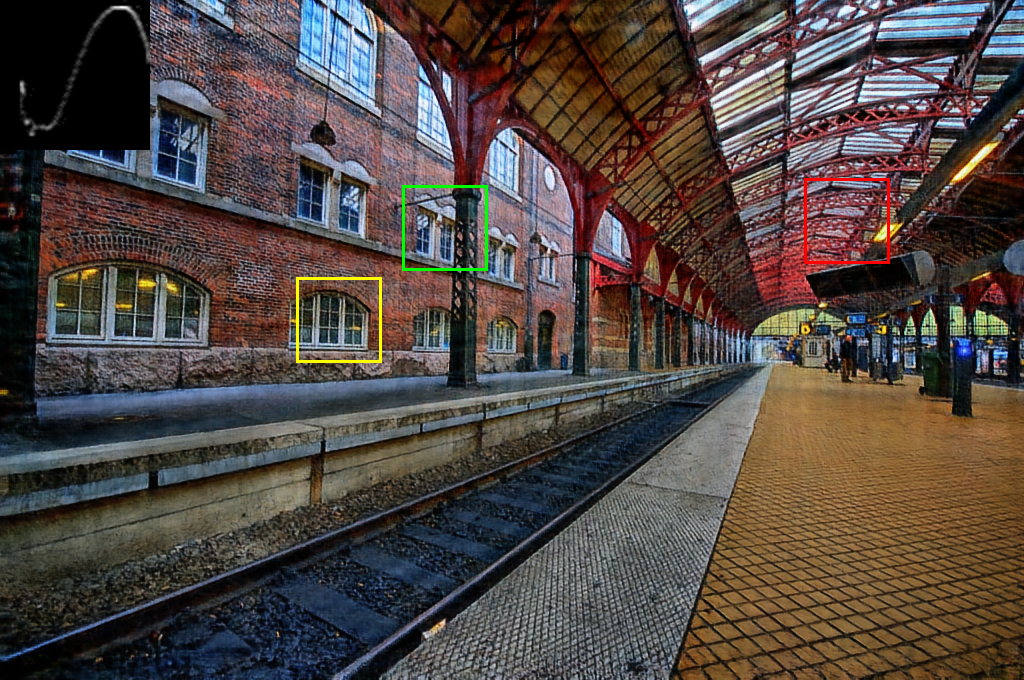}
		\vskip 4pt
		\includegraphics[width=1\linewidth]{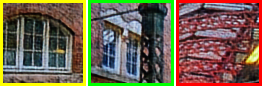}
		\caption*{(l)Self-MSNet (\textbf{Ours}) \protect\\ {PSNR: 22.45}\centering}
		\label{lai_manmade_05_kernel_04_ours}%文中引用该图片代号
	\end{minipage}
        %\vskip -8pt
	\caption{Visual comparison on another example image of the 'man-made' category with 4th blur kernel from Lai's dataset.}
	\label{fig:Lai-visual-manmade_05_kernel_04}
\end{figure*}

\begin{figure*}[!tbp]
	\centering
	\begin{minipage}{0.22\linewidth}
		\centering
		\includegraphics[width=1\linewidth]{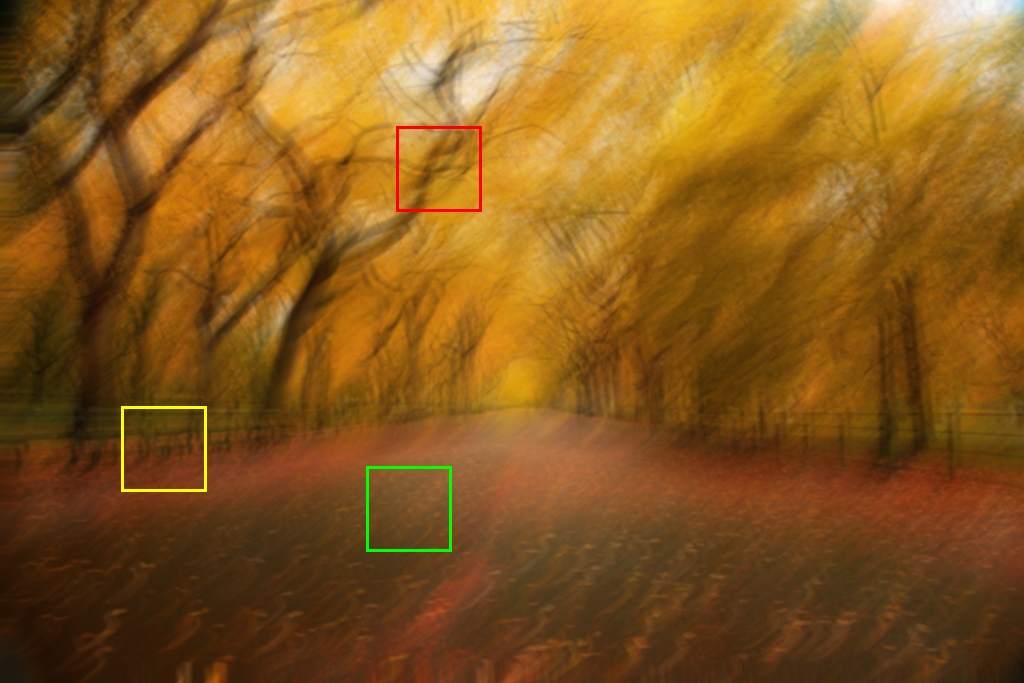}
		\vskip 4pt
		\includegraphics[width=1\linewidth]{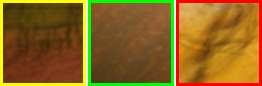}
		\caption*{(a)Blurred image \protect\\ {\textcolor{white}{***}}\centering}
		\label{lai_natural_04_kernel_02_Blurry image}%文中引用该图片代号
	\end{minipage}
	\begin{minipage}{0.22\linewidth}
		\centering
		\includegraphics[width=1\linewidth]{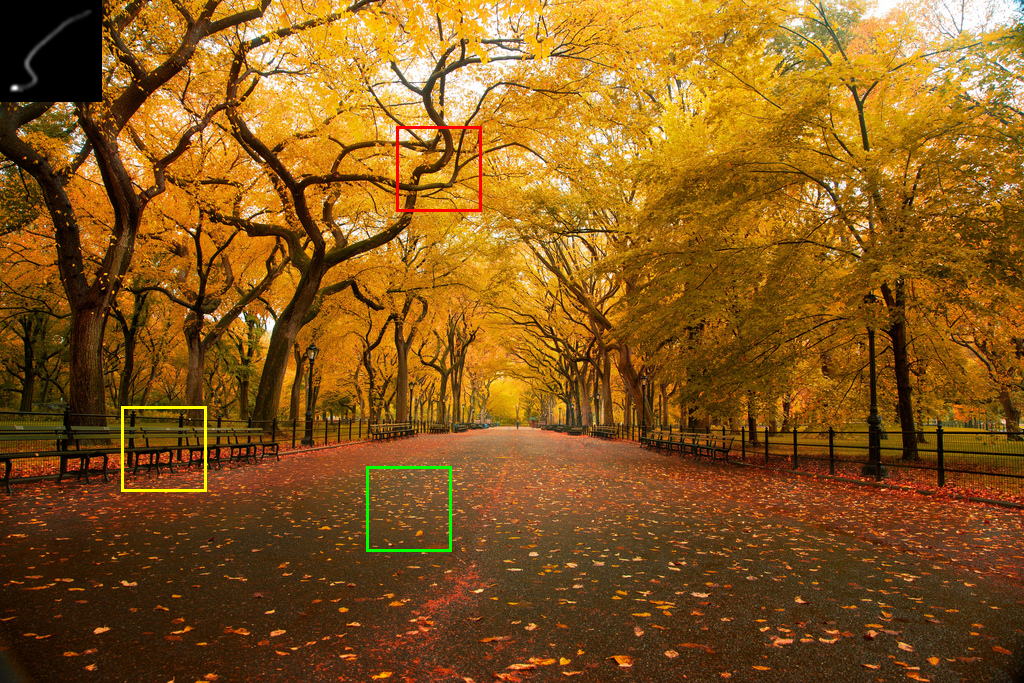}
		\vskip 4pt
		\includegraphics[width=1\linewidth]{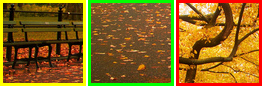}
		\caption*{(b)Ground-truth \protect\\ {\textcolor{white}{***}}\centering}
		\label{lai_natural_04_kernel_02_Ground-truth}%文中引用该图片代号
	\end{minipage}
	\begin{minipage}{0.22\linewidth}
		\centering
		\includegraphics[width=1\linewidth]{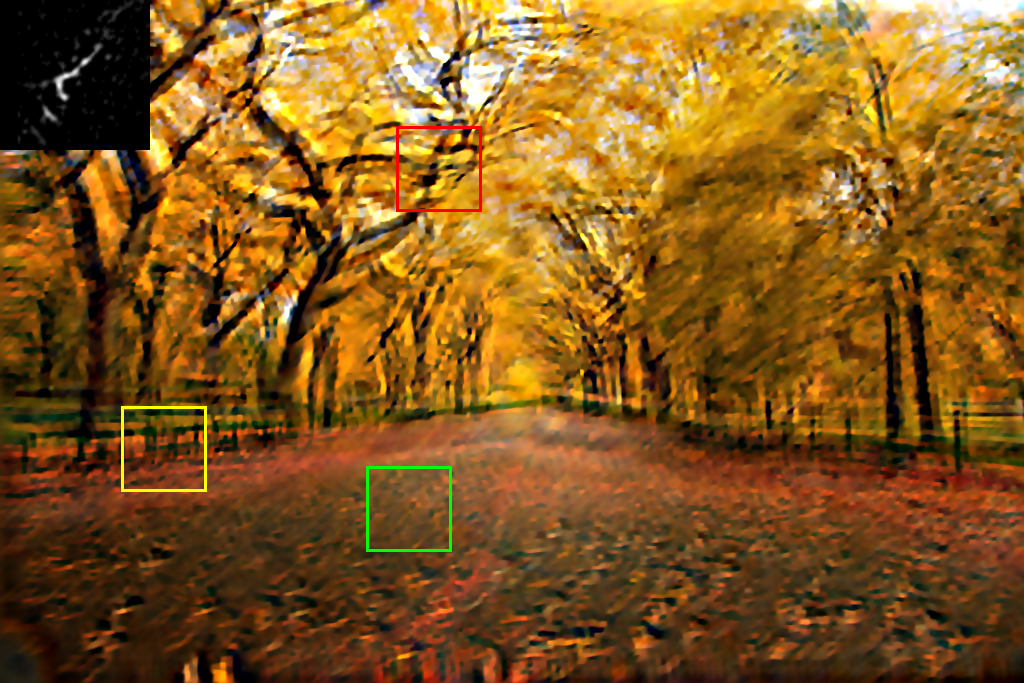}
		\vskip 4pt
		\includegraphics[width=1\linewidth]{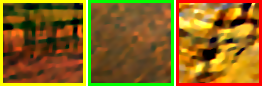}
		\caption*{(c)Cho and Lee\cite{cho2009fast} \protect\\ {PSNR: 16.42}\centering}
		\label{lai_natural_04_kernel_02_cho}%文中引用该图片代号
	\end{minipage}
        \begin{minipage}{0.22\linewidth}
		\centering
		\includegraphics[width=1\linewidth]{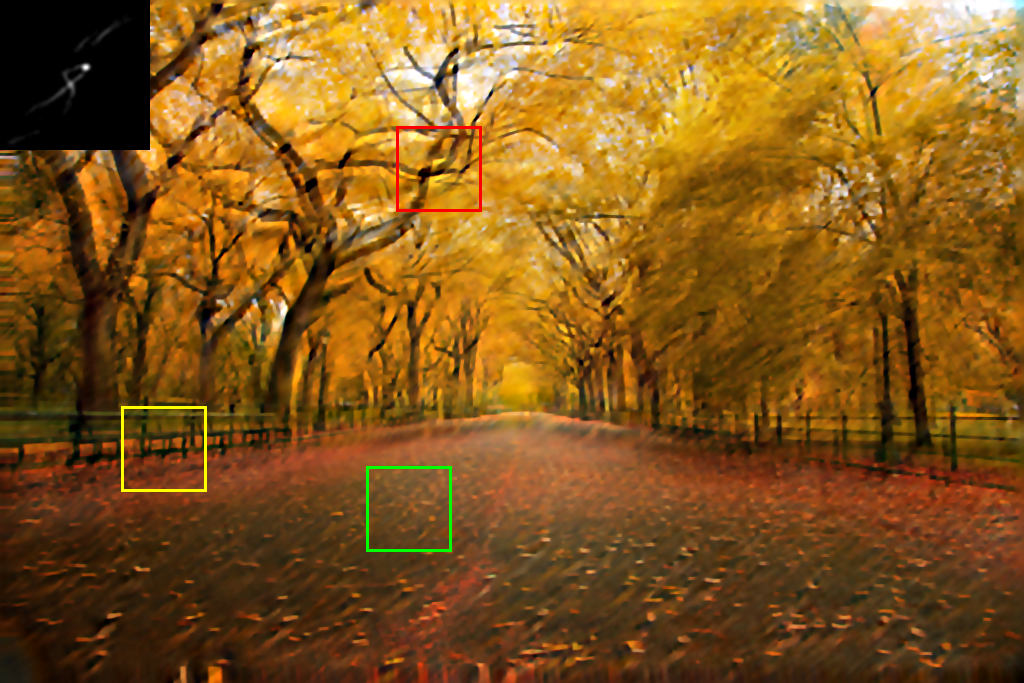}
		\vskip 4pt
		\includegraphics[width=1\linewidth]{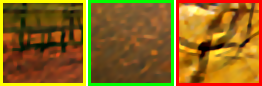}
		\caption*{(d)Xu and Jia\cite{xu2010two} \protect\\ {PSNR: 18.61}\centering}
		\label{lai_natural_04_kernel_02_xuandjia}%文中引用该图片代号
	\end{minipage}
        \vskip 7pt
        \begin{minipage}{0.22\linewidth}
		\centering
		\includegraphics[width=1\linewidth]{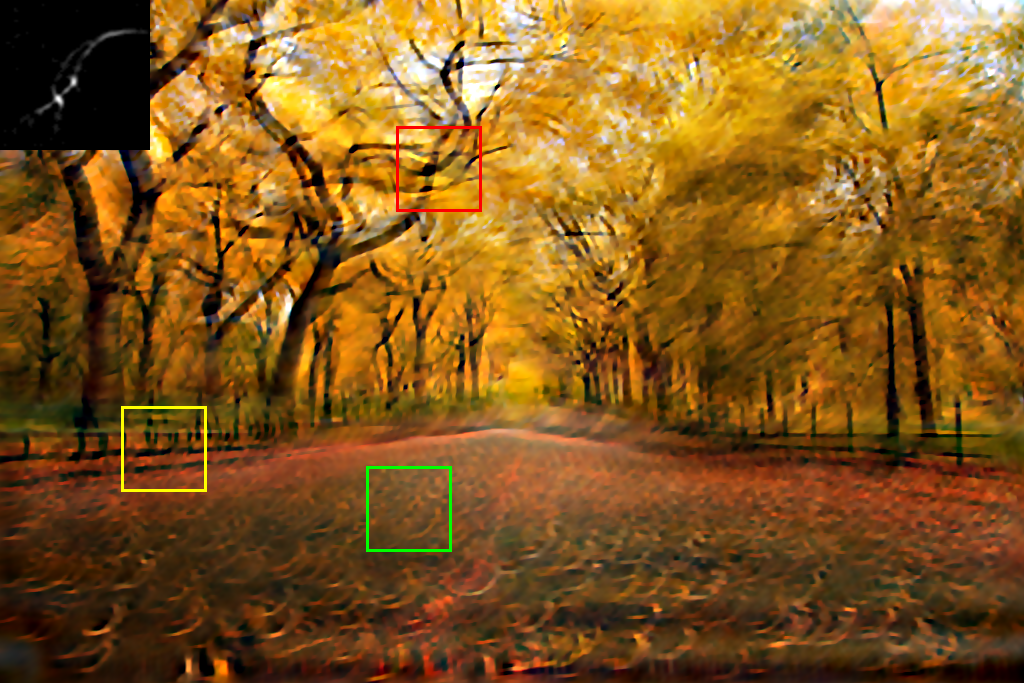}
		\vskip 4pt
		\includegraphics[width=1\linewidth]{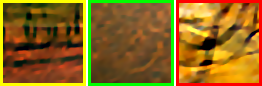}
		\caption*{(e)Xu et al.\cite{xu2013unnatural} \protect\\ {PSNR: 17.60}\centering}
		\label{lai_natural_04_kernel_02_xuunnatural}%文中引用该图片代号
	\end{minipage} 
        \begin{minipage}{0.22\linewidth}
		\centering
		\includegraphics[width=1\linewidth]{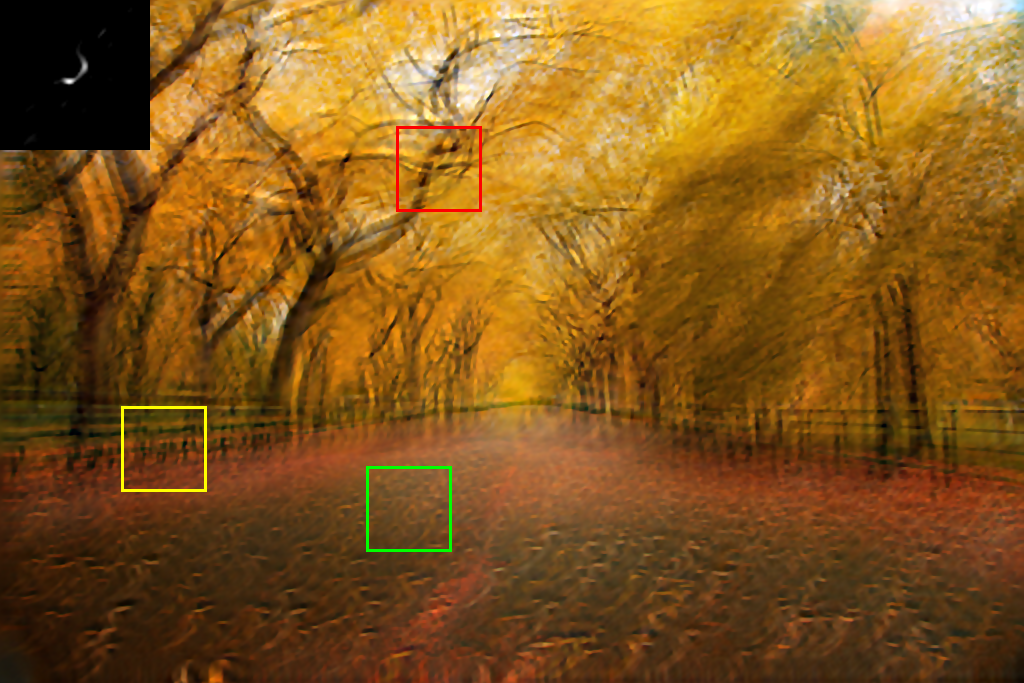}
		\vskip 4pt
		\includegraphics[width=1\linewidth]{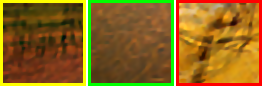}
		\caption*{(f)Michaeli and Irani\cite{michaeli2014blind} \protect\\ {PSNR: 19.34}\centering}
		\label{lai_natural_04_kernel_02_Michaeli}%文中引用该图片代号
	\end{minipage} 
	\begin{minipage}{0.22\linewidth}
		\centering
		\includegraphics[width=1\linewidth]{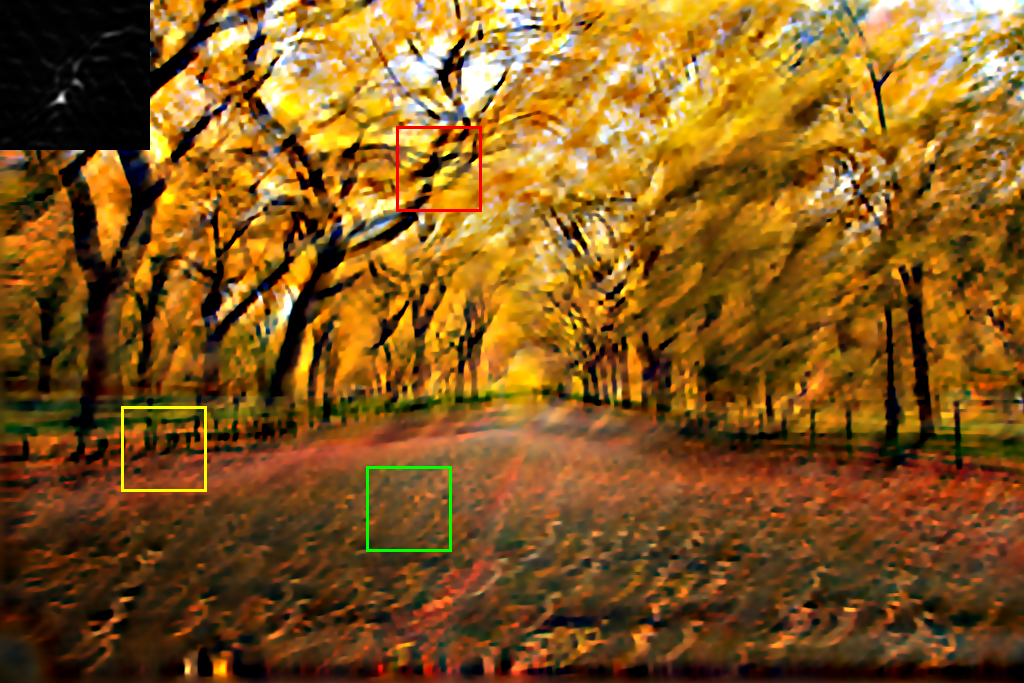}
		\vskip 4pt
		\includegraphics[width=1\linewidth]{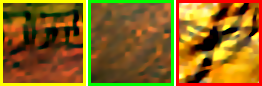}
		\caption*{(g)Perrone and Favaro\cite{perrone2014total} \protect\\ {PSNR: 16.20}\centering}
		\label{lai_natural_04_kernel_02_Perrone}%文中引用该图片代号
	\end{minipage}
	\begin{minipage}{0.22\linewidth}
		\centering
		\includegraphics[width=1\linewidth]{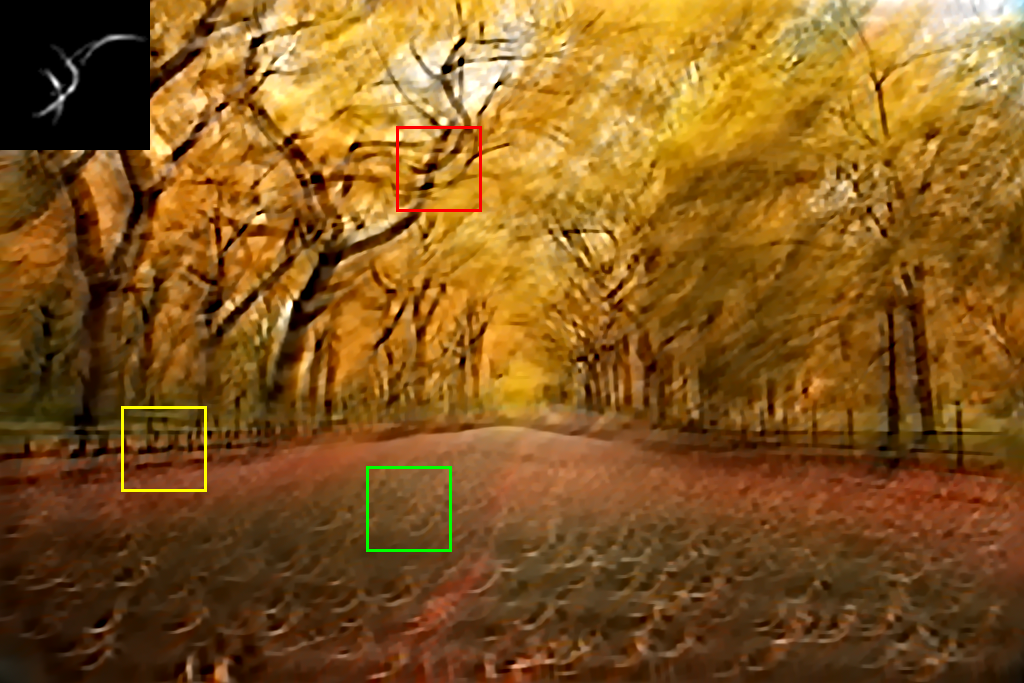}
		\vskip 4pt
		\includegraphics[width=1\linewidth]{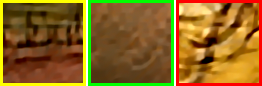}
		\caption*{(h)Pan et al.\cite{pan2018PAMIdarkchannel} \protect\\ {PSNR: 17.88}\centering}
		\label{lai_natural_04_kernel_02_Pan}%文中引用该图片代号
	\end{minipage}
        \vskip 7pt
        \begin{minipage}{0.22\linewidth}
		\centering
		\includegraphics[width=1\linewidth]{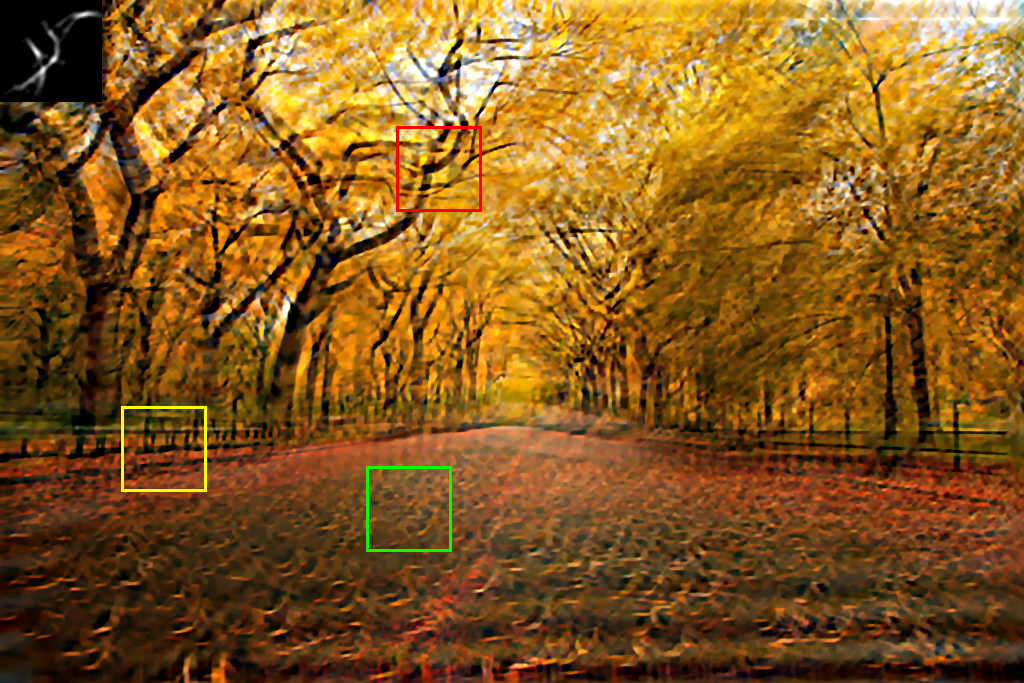}
		\vskip 4pt
		\includegraphics[width=1\linewidth]{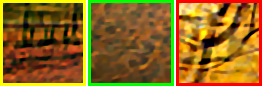}
		\caption*{(i)Wen et al.\cite{wen2021TCSVT} \protect\\ {PSNR: 17.85}\centering}
		\label{lai_natural_04_kernel_02_wen}%文中引用该图片代号
	\end{minipage}
        \begin{minipage}{0.22\linewidth}
		\centering
		\includegraphics[width=1\linewidth]{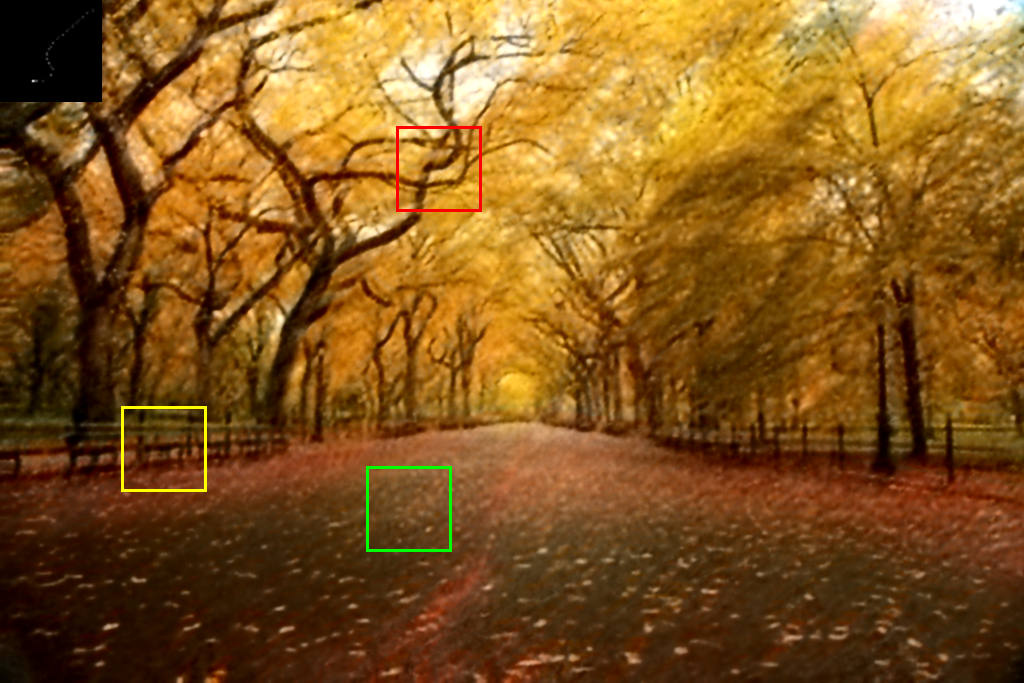}
		\vskip 4pt
		\includegraphics[width=1\linewidth]{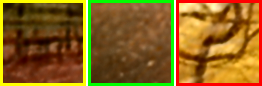}
		\caption*{(j)SelfDeblur\cite{ren2020neural} \protect\\ {PSNR: 19.08}\centering}
		\label{lai_natural_04_kernel_02_SelfDeblur}%文中引用该图片代号
	\end{minipage}	 
	\begin{minipage}{0.22\linewidth}
		\centering
		\includegraphics[width=1\linewidth]{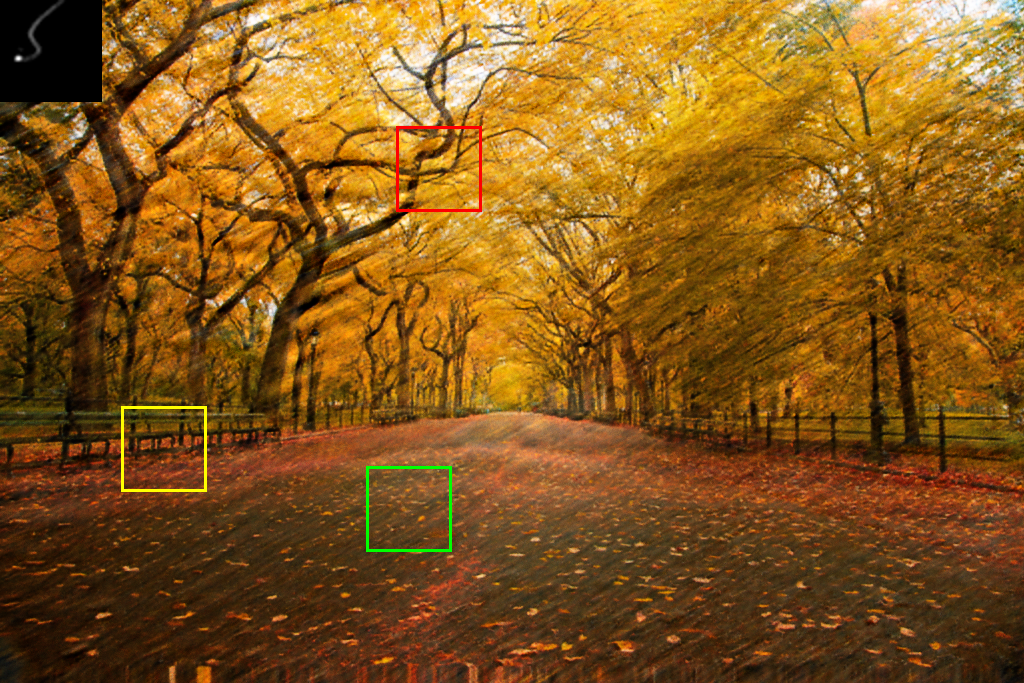}
		\vskip 4pt
		\includegraphics[width=1\linewidth]{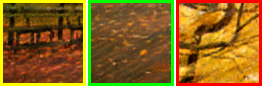}
		\caption*{(k)Fast-SelfDeblur\cite{bai2023fastselfdeblur} \protect\\ {PSNR: 18.22}\centering}
		\label{lai_natural_04_kernel_02_Fast-SelfDeblur}%文中引用该图片代号
	\end{minipage}
	\begin{minipage}{0.22\linewidth}
		\centering
		\includegraphics[width=1\linewidth]{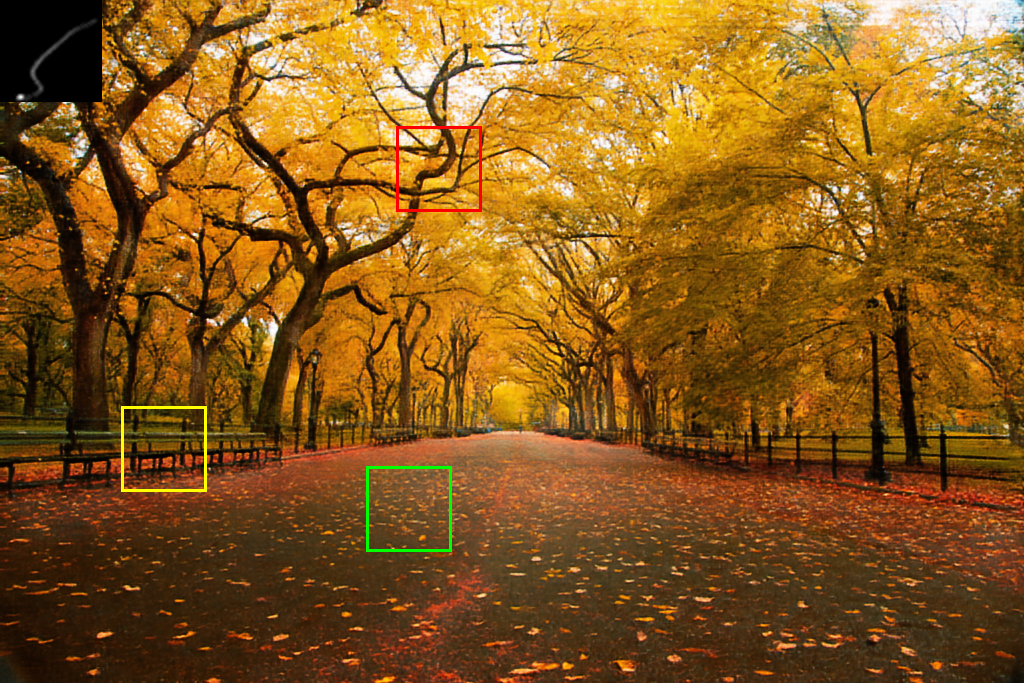}
		\vskip 4pt
		\includegraphics[width=1\linewidth]{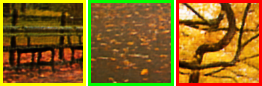}
		\caption*{(l)Self-MSNet (\textbf{Ours}) \protect\\ {PSNR: 25.44}\centering}
		\label{lai_natural_04_kernel_02_ours}%文中引用该图片代号
	\end{minipage}
        %\vskip -8pt
	\caption{Visual comparison on an example image of the 'natural' category with 2th blur kernel from Lai's dataset.}
	\label{fig:Lai-visual-natural_04_kernel_02}
\end{figure*}

\begin{figure*}[!tbp]
	\centering
	\begin{minipage}{0.22\linewidth}
		\centering
		\includegraphics[width=1\linewidth]{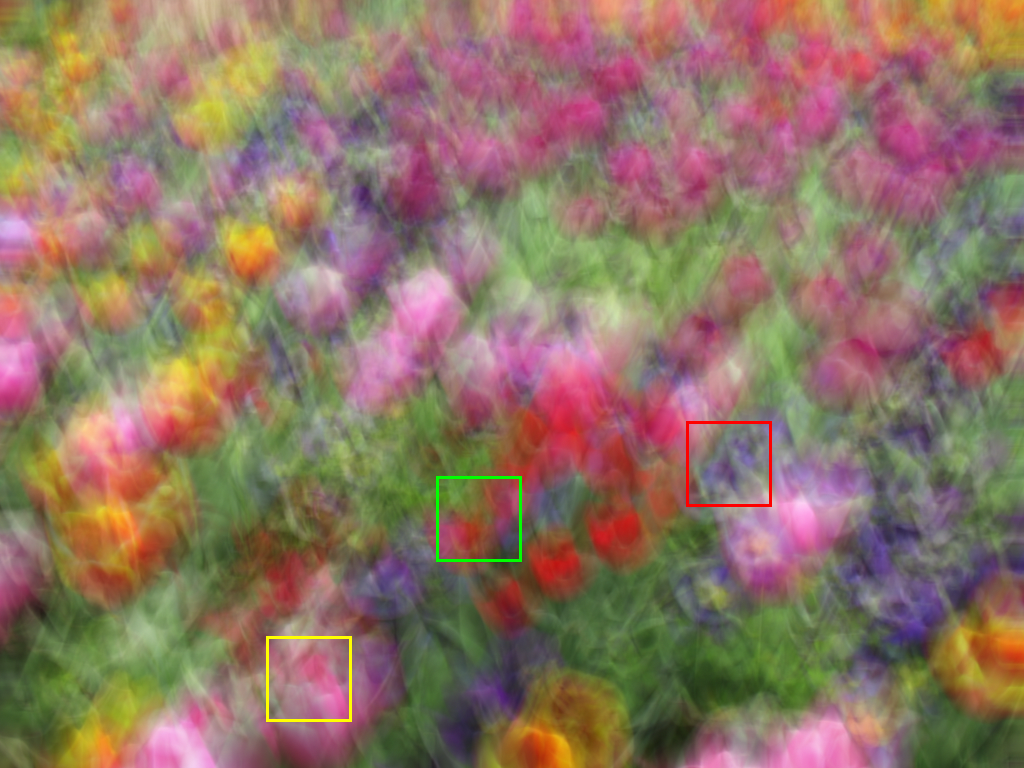}
		\vskip 4pt
		\includegraphics[width=1\linewidth]{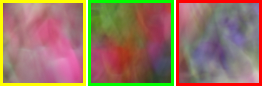}
		\caption*{(a)Blurred image \protect\\ {\textcolor{white}{***}}\centering}
		\label{lai_natural_05_kernel_04_Blurry image}%文中引用该图片代号
	\end{minipage}
	\begin{minipage}{0.22\linewidth}
		\centering
		\includegraphics[width=1\linewidth]{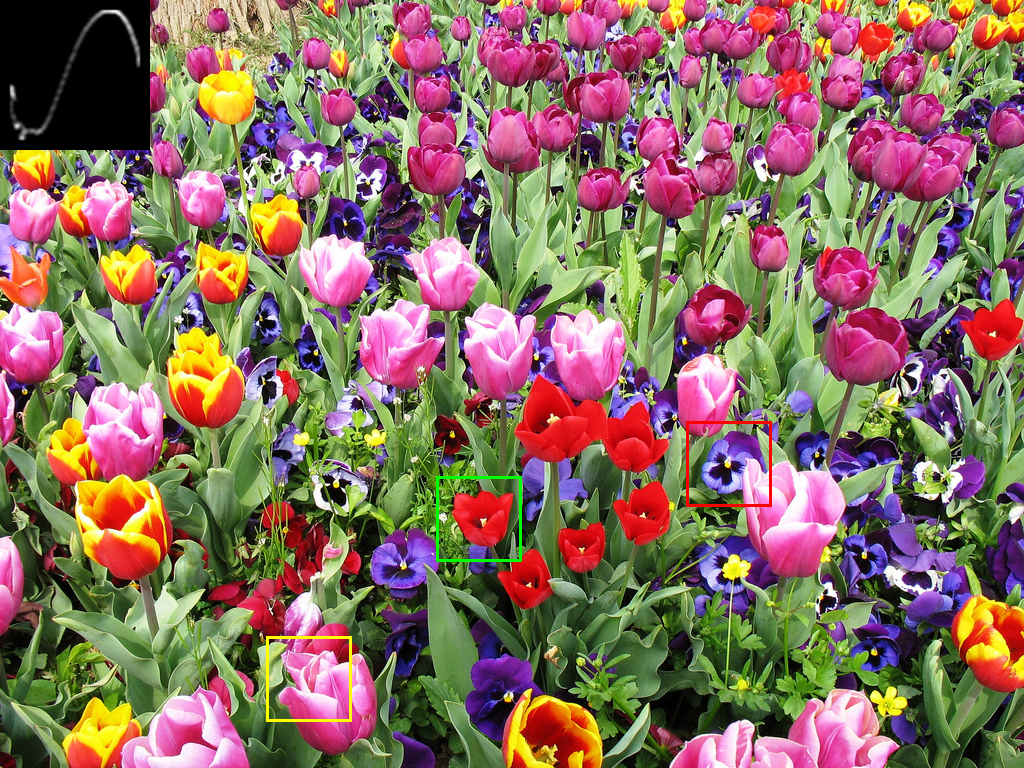}
		\vskip 4pt
		\includegraphics[width=1\linewidth]{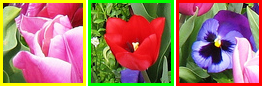}
		\caption*{(b)Ground-truth \protect\\ {\textcolor{white}{***}}\centering}
		\label{lai_natural_05_kernel_04_Ground-truth}%文中引用该图片代号
	\end{minipage}
	\begin{minipage}{0.22\linewidth}
		\centering
		\includegraphics[width=1\linewidth]{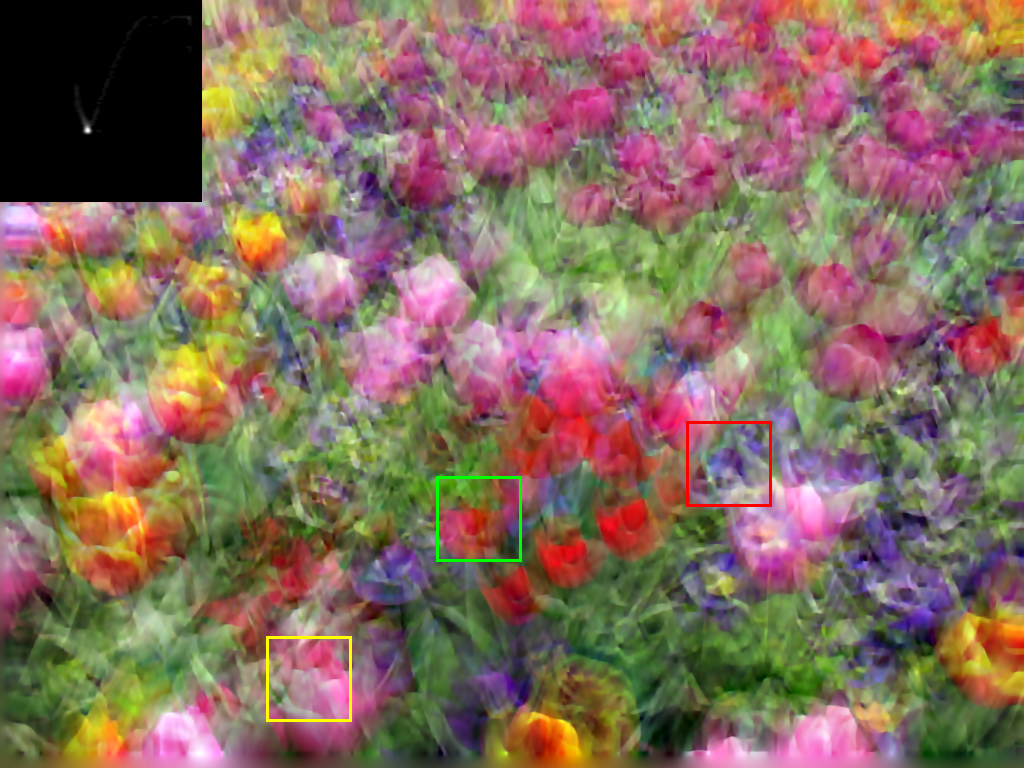}
		\vskip 4pt
		\includegraphics[width=1\linewidth]{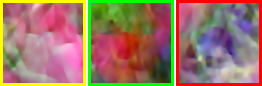}
		\caption*{(c)Cho and Lee\cite{cho2009fast} \protect\\ {PSNR: 14.60}\centering}
		\label{lai_natural_05_kernel_04_cho}%文中引用该图片代号
	\end{minipage}
        \begin{minipage}{0.22\linewidth}
		\centering
		\includegraphics[width=1\linewidth]{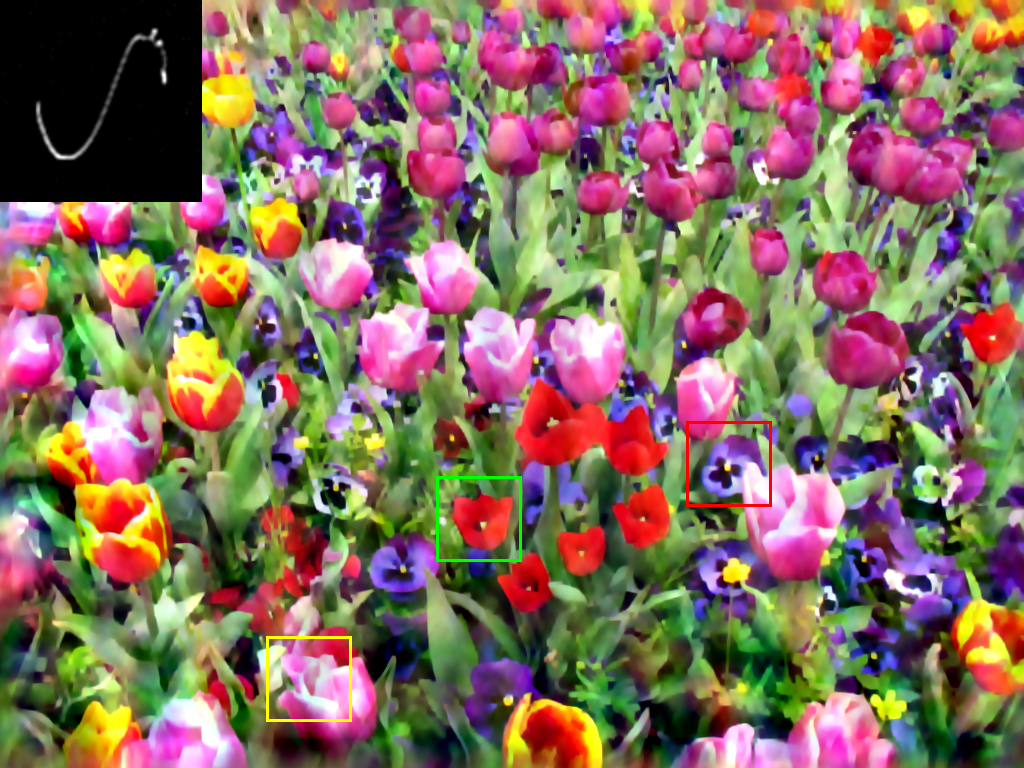}
		\vskip 4pt
		\includegraphics[width=1\linewidth]{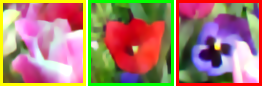}
		\caption*{(d)Xu and Jia\cite{xu2010two} \protect\\ {PSNR: 19.46}\centering}
		\label{lai_natural_05_kernel_04_xuandjia}%文中引用该图片代号
	\end{minipage}
        \vskip 7pt
        \begin{minipage}{0.22\linewidth}
		\centering
		\includegraphics[width=1\linewidth]{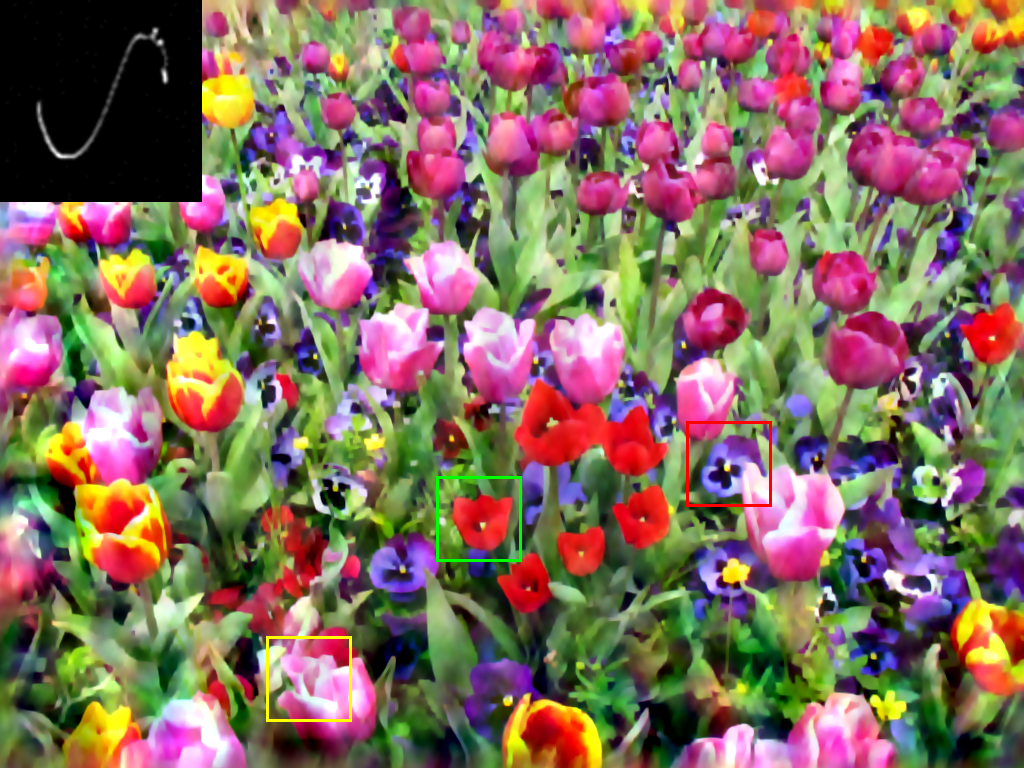}
		\vskip 4pt
		\includegraphics[width=1\linewidth]{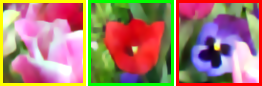}
		\caption*{(e)Xu et al.\cite{xu2013unnatural} \protect\\ {PSNR: 19.76}\centering}
		\label{lai_natural_05_kernel_04_xuunnatural}%文中引用该图片代号
	\end{minipage} 
        \begin{minipage}{0.22\linewidth}
		\centering
		\includegraphics[width=1\linewidth]{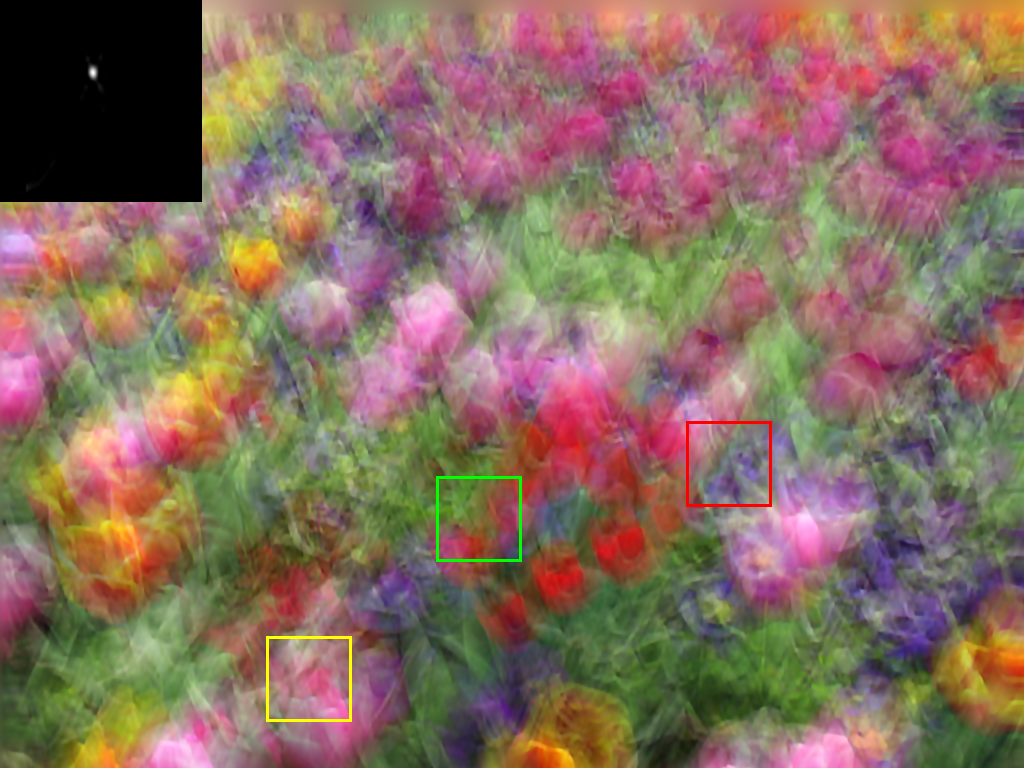}
		\vskip 4pt
		\includegraphics[width=1\linewidth]{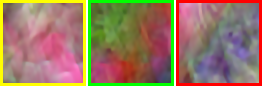}
		\caption*{(f)Michaeli and Irani\cite{michaeli2014blind} \protect\\ {PSNR: 14.29}\centering}
		\label{lai_natural_05_kernel_04_Michaeli}%文中引用该图片代号
	\end{minipage} 
	\begin{minipage}{0.22\linewidth}
		\centering
		\includegraphics[width=1\linewidth]{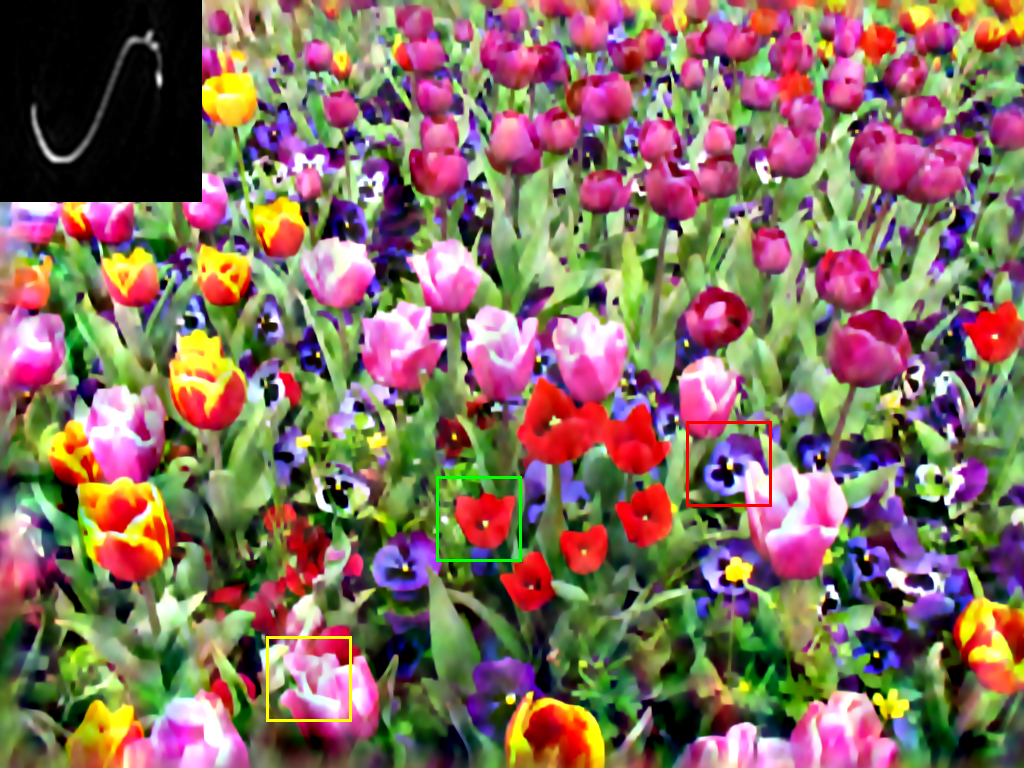}
		\vskip 4pt
		\includegraphics[width=1\linewidth]{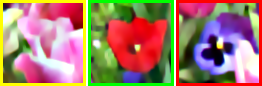}
		\caption*{(g)Perrone and Favaro\cite{perrone2014total} \protect\\ {PSNR: 19.78}\centering}
		\label{lai_natural_05_kernel_04_Perrone}%文中引用该图片代号
	\end{minipage}
	\begin{minipage}{0.22\linewidth}
		\centering
		\includegraphics[width=1\linewidth]{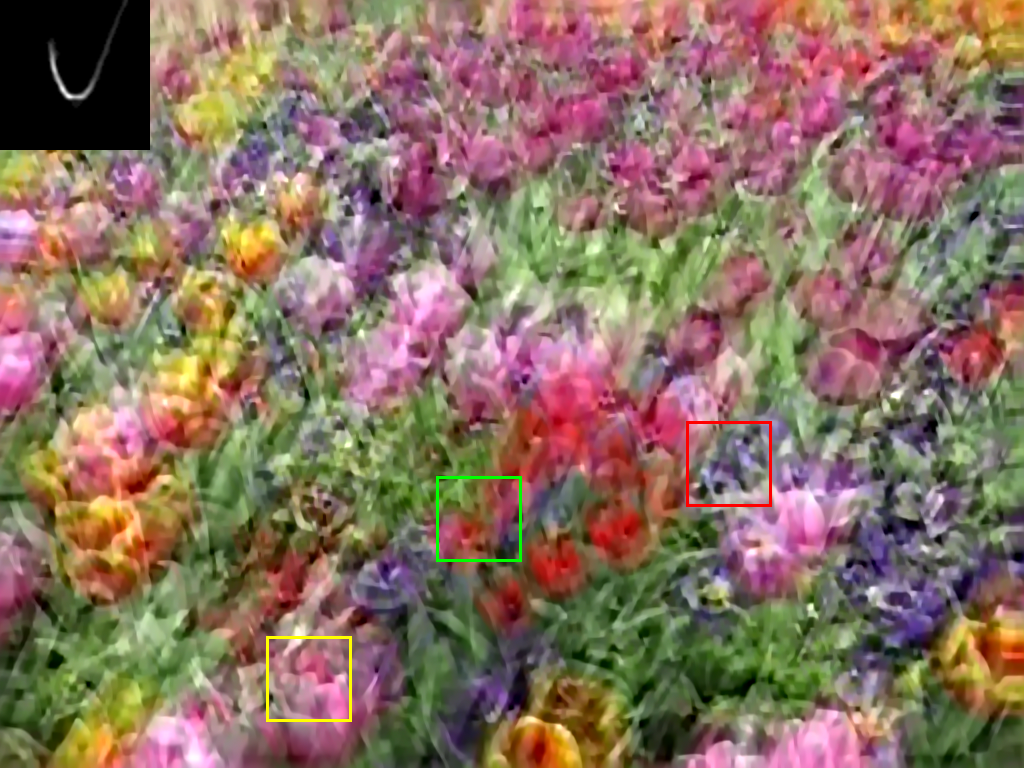}
		\vskip 4pt
		\includegraphics[width=1\linewidth]{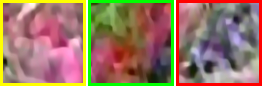}
		\caption*{(h)Pan et al.\cite{pan2018PAMIdarkchannel} \protect\\ {PSNR: 13.66}\centering}
		\label{lai_natural_05_kernel_04_Pan}%文中引用该图片代号
	\end{minipage}
        \vskip 7pt
        \begin{minipage}{0.22\linewidth}
		\centering
		\includegraphics[width=1\linewidth]{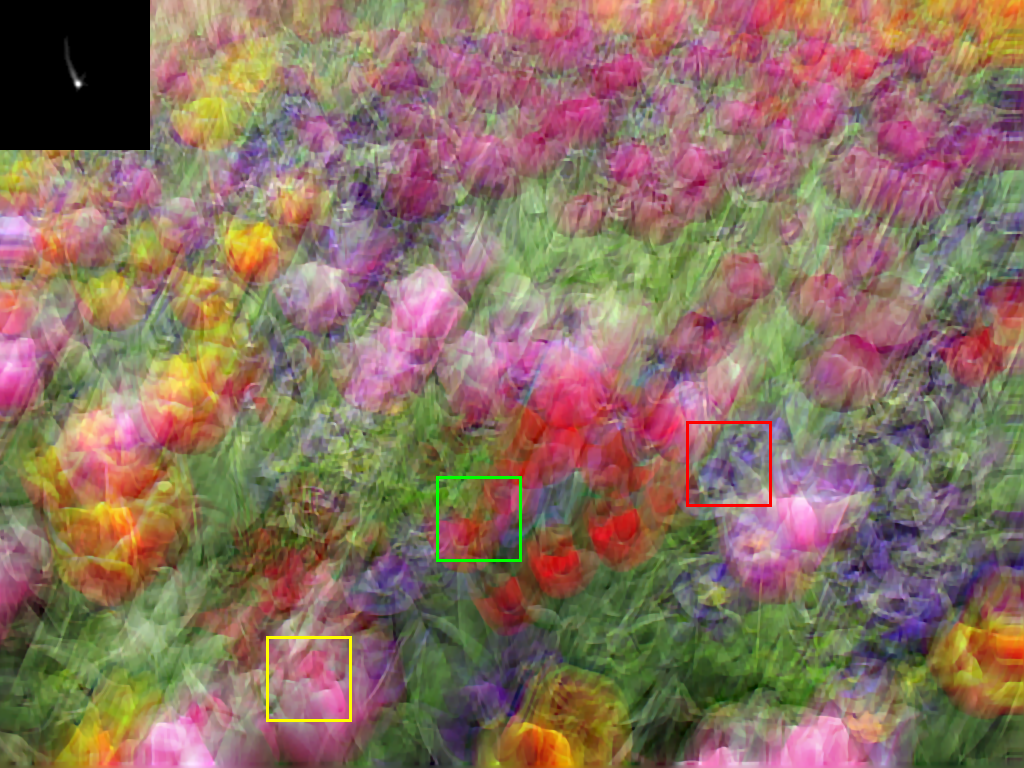}
		\vskip 4pt
		\includegraphics[width=1\linewidth]{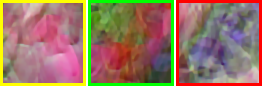}
		\caption*{(i)Wen et al.\cite{wen2021TCSVT} \protect\\ {PSNR: 14.60}\centering}
		\label{lai_natural_05_kernel_04_wen}%文中引用该图片代号
	\end{minipage}
        \begin{minipage}{0.22\linewidth}
		\centering
		\includegraphics[width=1\linewidth]{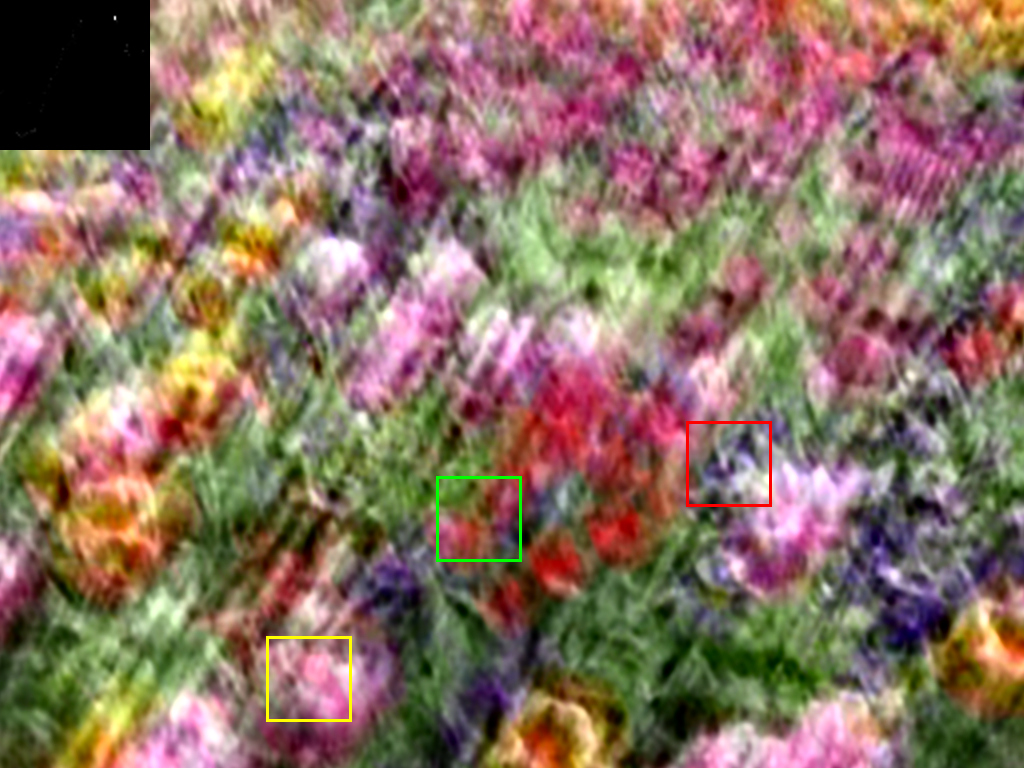}
		\vskip 4pt
		\includegraphics[width=1\linewidth]{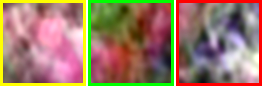}
		\caption*{(j)SelfDeblur\cite{ren2020neural} \protect\\ {PSNR: 16.36}\centering}
		\label{lai_natural_05_kernel_04_SelfDeblur}%文中引用该图片代号
	\end{minipage}	 
	\begin{minipage}{0.22\linewidth}
		\centering
		\includegraphics[width=1\linewidth]{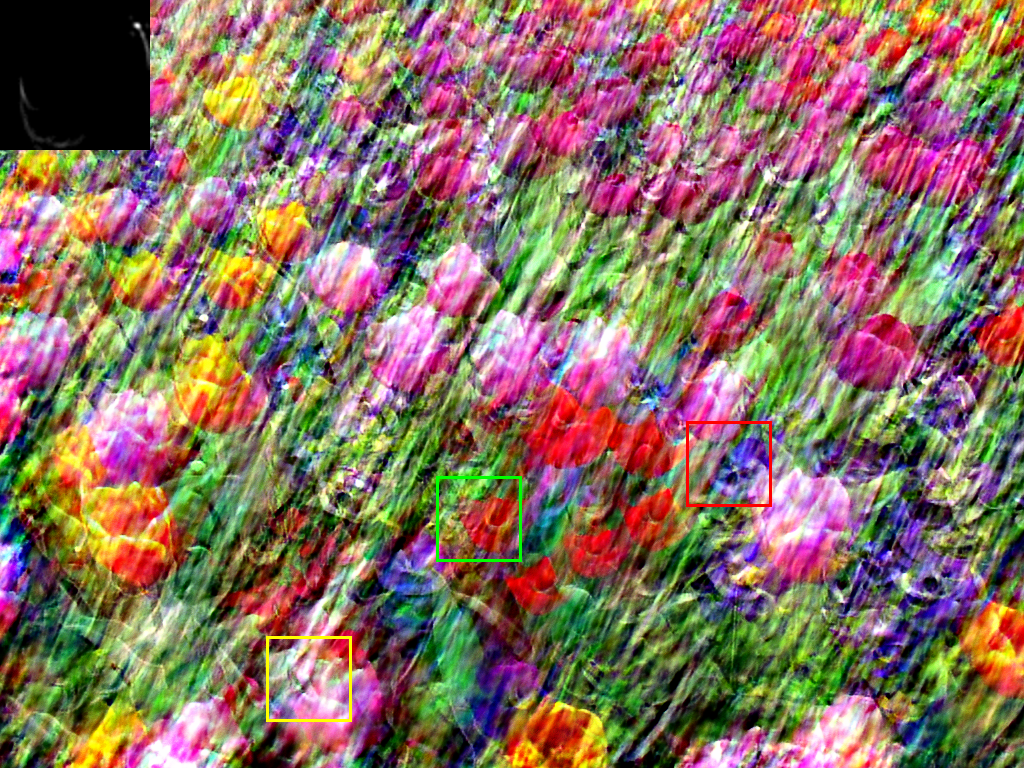}
		\vskip 4pt
		\includegraphics[width=1\linewidth]{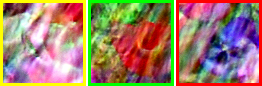}
		\caption*{(k)Fast-SelfDeblur\cite{bai2023fastselfdeblur} \protect\\ {PSNR: 14.49}\centering}
		\label{lai_natural_05_kernel_04_Fast-SelfDeblur}%文中引用该图片代号
	\end{minipage}
	\begin{minipage}{0.22\linewidth}
		\centering
		\includegraphics[width=1\linewidth]{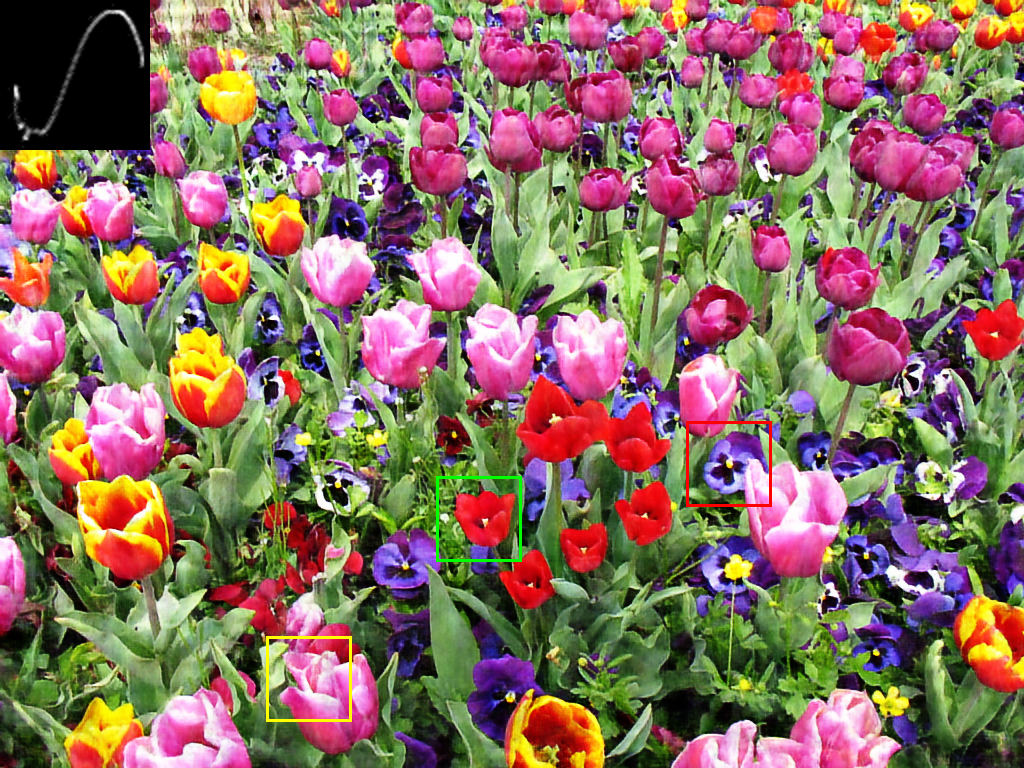}
		\vskip 4pt
		\includegraphics[width=1\linewidth]{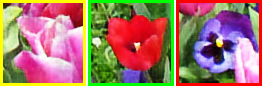}
		\caption*{(l)Self-MSNet (\textbf{Ours}) \protect\\ {PSNR: 23.81}\centering}
		\label{lai_natural_05_kernel_04_ours}%文中引用该图片代号
	\end{minipage}
        %\vskip -8pt
	\caption{Visual comparison on an example image of the 'natural' category with 4th blur kernel from Lai's dataset.}
	\label{fig:Lai-visual-natural_05_kernel_04}
\end{figure*}

\begin{figure*}[!tbp]
	\centering
	\begin{minipage}{0.2\linewidth}
		\centering
		\includegraphics[width=1\linewidth]{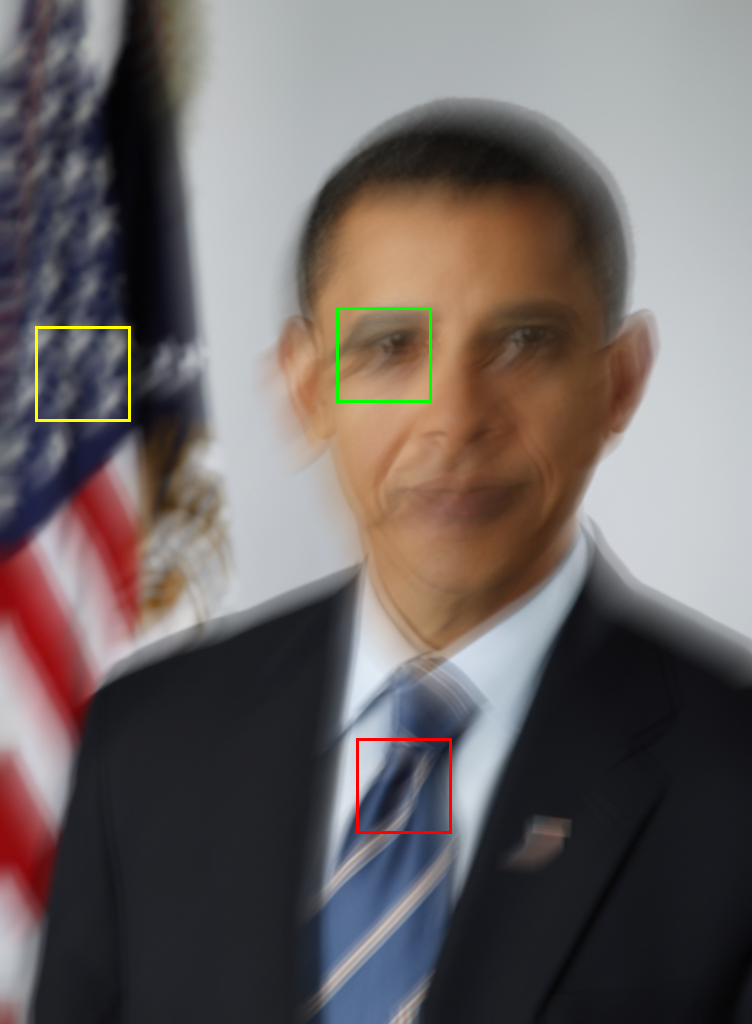}
		\vskip 4pt
		\includegraphics[width=1\linewidth]{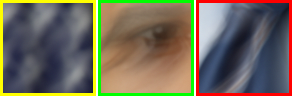}
		\caption*{(a)Blurred image \protect\\ {\textcolor{white}{***}}\centering}
		\label{lai_people_03_kernel_02_Blurry image}%文中引用该图片代号
	\end{minipage}
	\begin{minipage}{0.2\linewidth}
		\centering
		\includegraphics[width=1\linewidth]{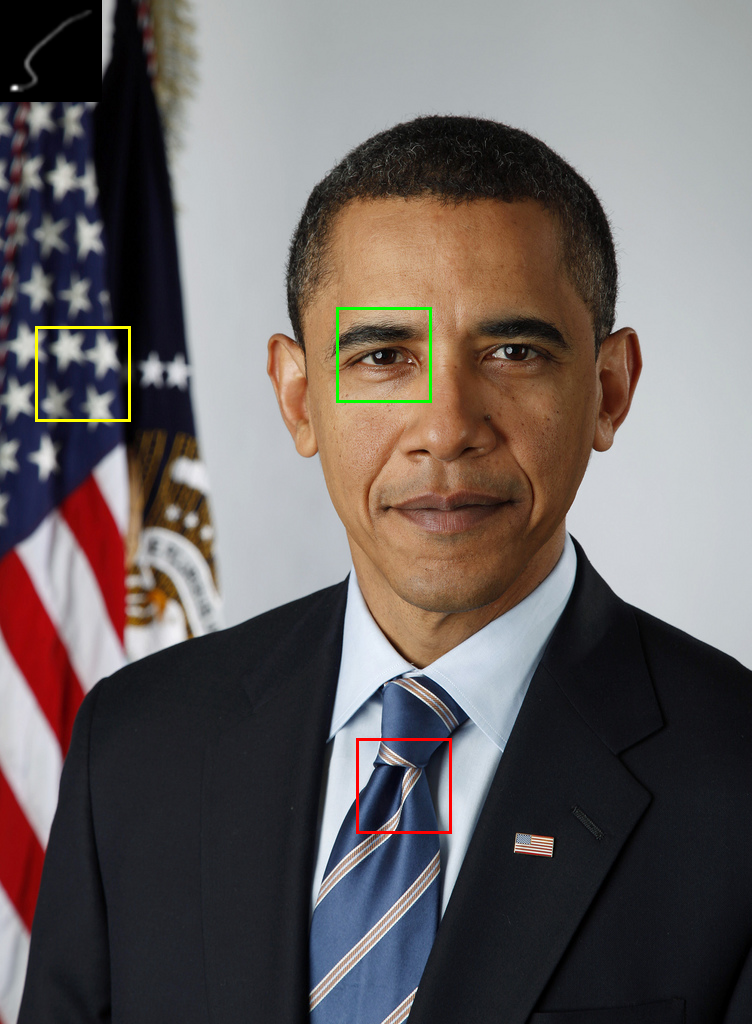}
		\vskip 4pt
		\includegraphics[width=1\linewidth]{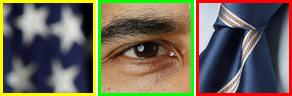}
		\caption*{(b)Ground-truth \protect\\ {\textcolor{white}{***}}\centering}
		\label{lai_people_03_kernel_02_Ground-truth}%文中引用该图片代号
	\end{minipage}
	\begin{minipage}{0.2\linewidth}
		\centering
		\includegraphics[width=1\linewidth]{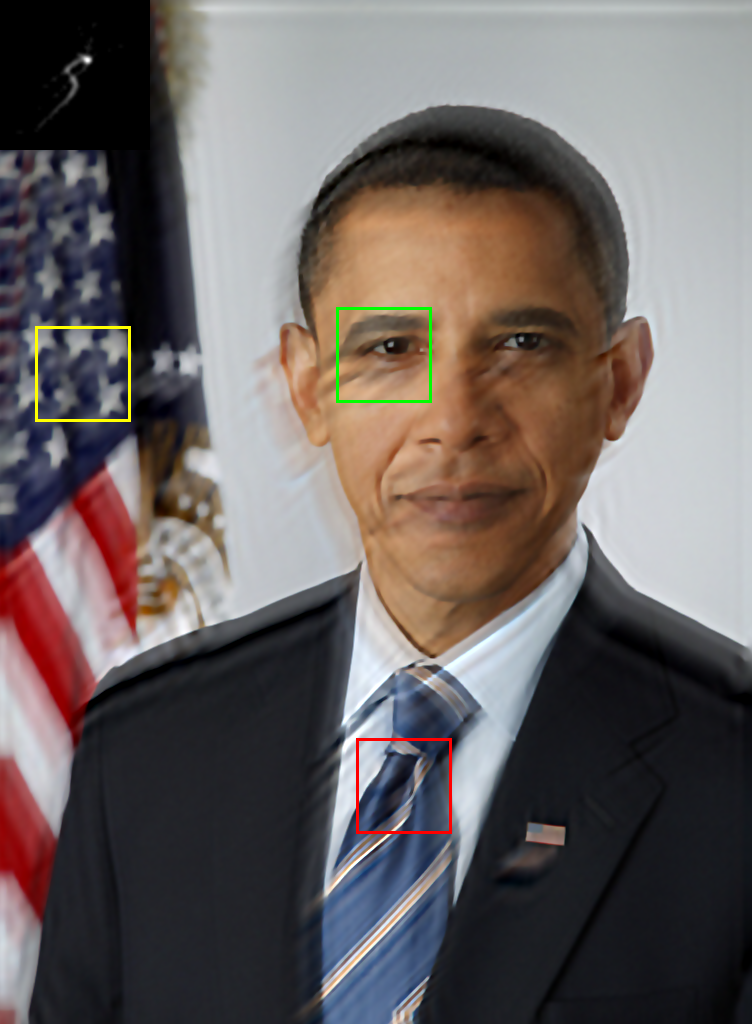}
		\vskip 4pt
		\includegraphics[width=1\linewidth]{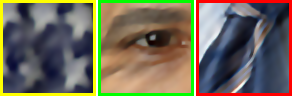}
		\caption*{(c)Cho and Lee\cite{cho2009fast} \protect\\ {PSNR: 20.99}\centering}
		\label{lai_people_03_kernel_02_cho}%文中引用该图片代号
	\end{minipage}
        \begin{minipage}{0.2\linewidth}
		\centering
		\includegraphics[width=1\linewidth]{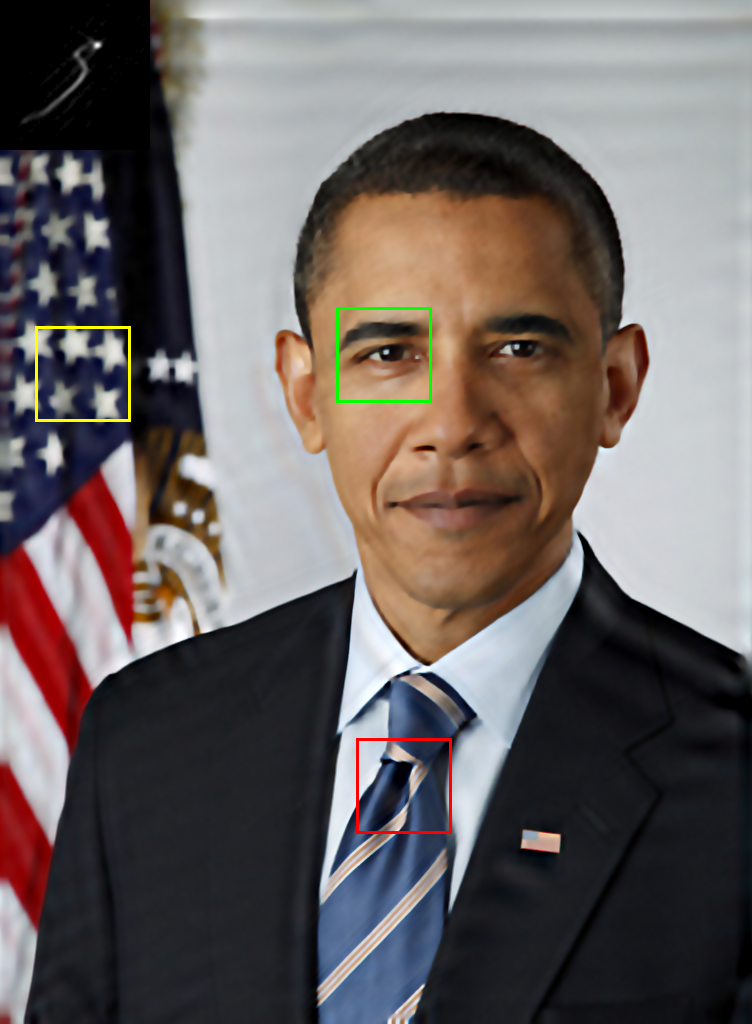}
		\vskip 4pt
		\includegraphics[width=1\linewidth]{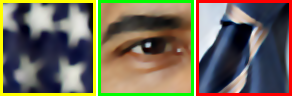}
		\caption*{(d)Xu and Jia\cite{xu2010two} \protect\\ {PSNR: 25.26}\centering}
		\label{lai_people_03_kernel_02_xuandjia}%文中引用该图片代号
	\end{minipage}
        \vskip 7pt
        \begin{minipage}{0.2\linewidth}
		\centering
		\includegraphics[width=1\linewidth]{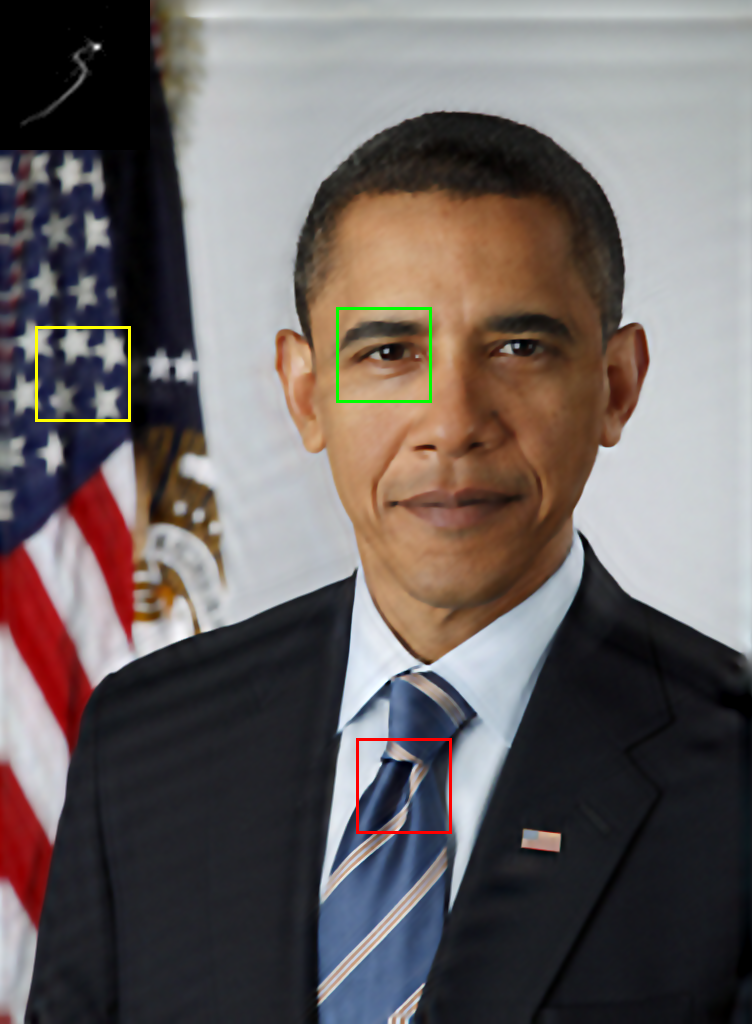}
		\vskip 4pt
		\includegraphics[width=1\linewidth]{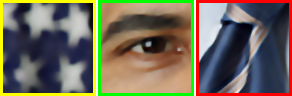}
		\caption*{(e)Xu et al.\cite{xu2013unnatural} \protect\\ {PSNR: 25.50}\centering}
		\label{lai_people_03_kernel_02_xuunnatural}%文中引用该图片代号
	\end{minipage} 
        \begin{minipage}{0.2\linewidth}
		\centering
		\includegraphics[width=1\linewidth]{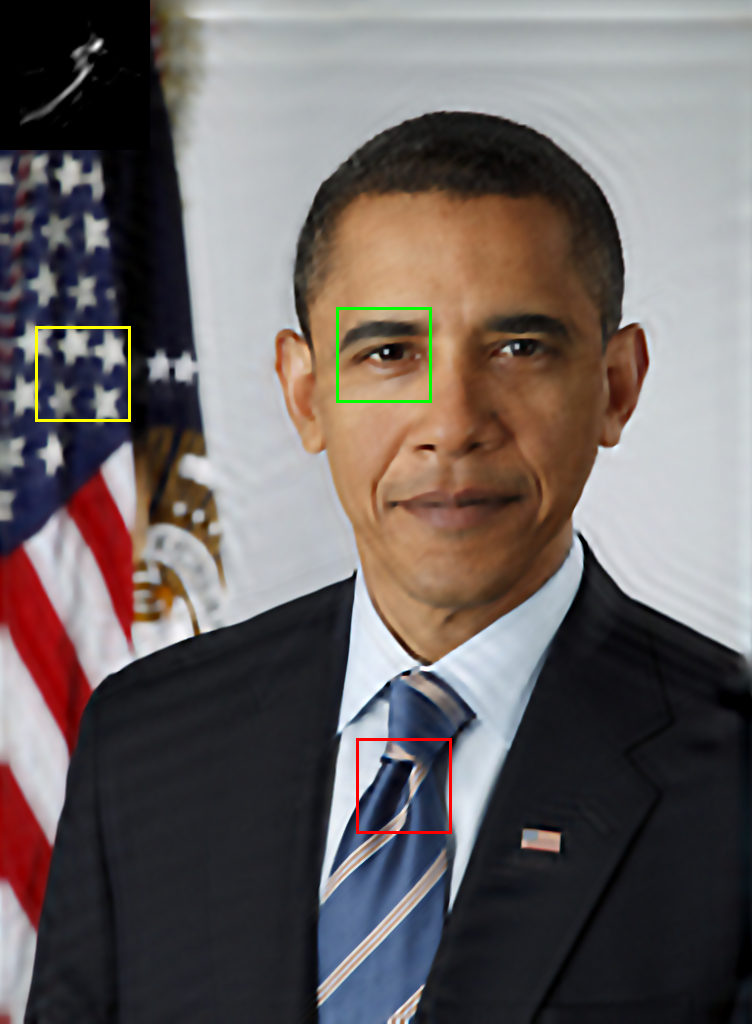}
		\vskip 4pt
		\includegraphics[width=1\linewidth]{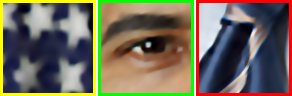}
		\caption*{\leftline{(f)Michaeli and Irani\cite{michaeli2014blind}} \protect\\ {PSNR: 26.17}\centering}
		\label{lai_people_03_kernel_02_Michaeli}%文中引用该图片代号
	\end{minipage} 
	\begin{minipage}{0.2\linewidth}
		\centering
		\includegraphics[width=1\linewidth]{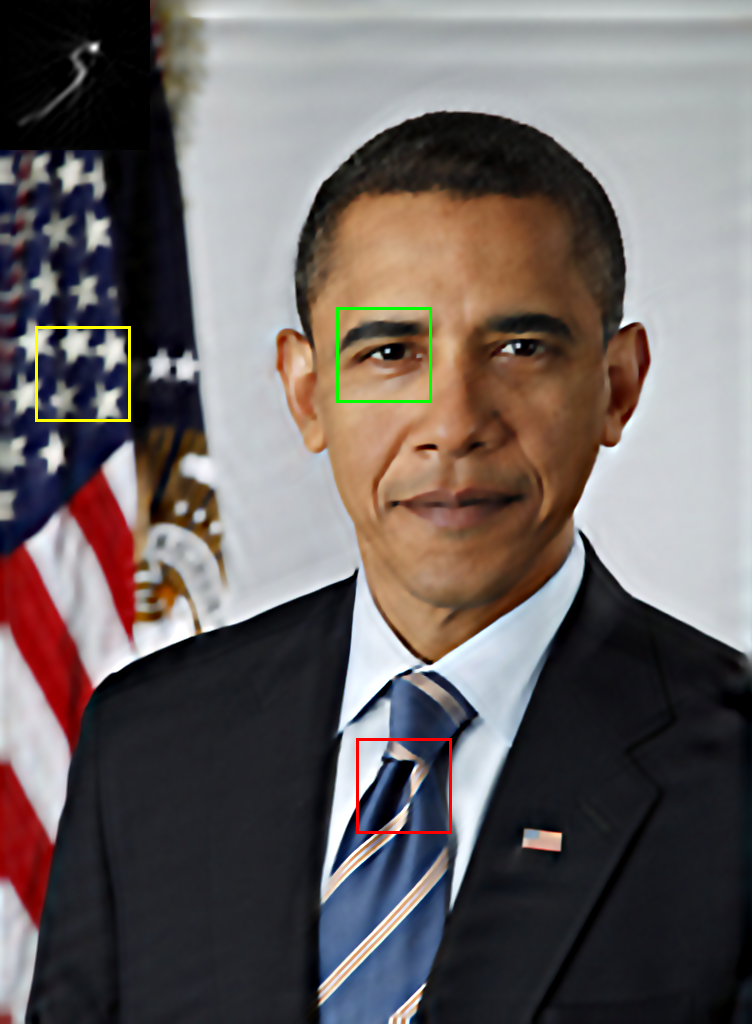}
		\vskip 4pt
		\includegraphics[width=1\linewidth]{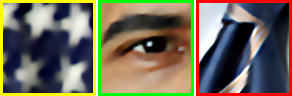}
		\caption*{\leftline{(g)Perrone and Favaro\cite{perrone2014total}} \protect\\ {PSNR: 22.39}\centering}
		\label{lai_people_03_kernel_02_Perrone}%文中引用该图片代号
	\end{minipage}
	\begin{minipage}{0.2\linewidth}
		\centering
		\includegraphics[width=1\linewidth]{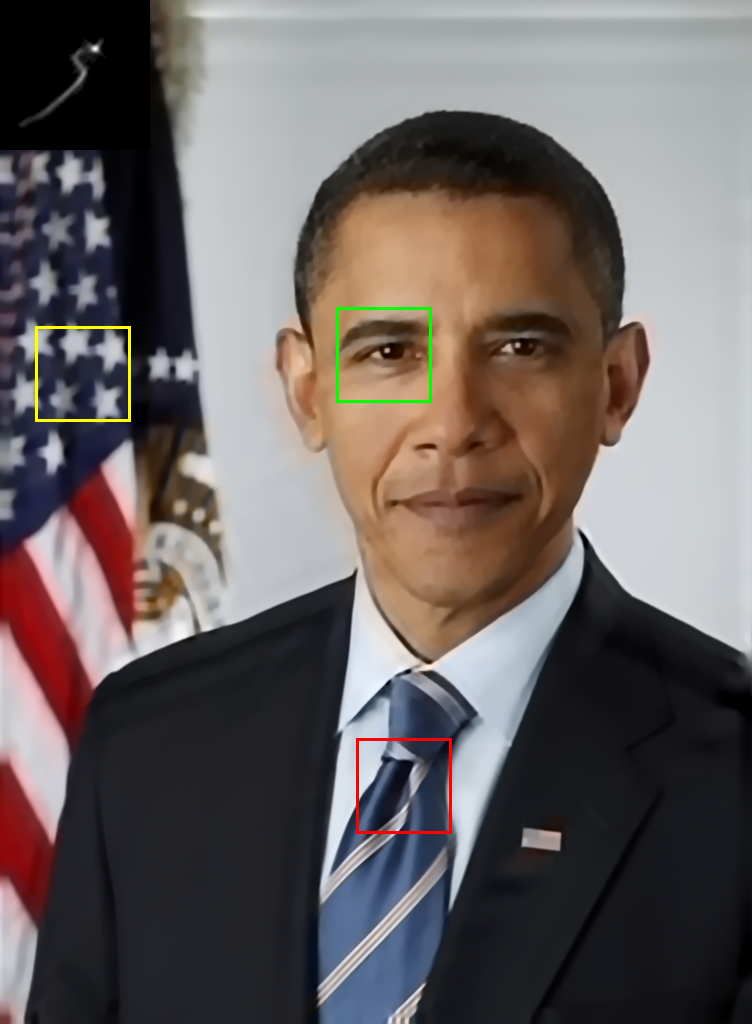}
		\vskip 4pt
		\includegraphics[width=1\linewidth]{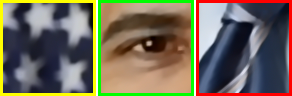}
		\caption*{(h)Pan et al.\cite{pan2018PAMIdarkchannel} \protect\\ {PSNR: 24.89}\centering}
		\label{lai_people_03_kernel_02_Pan}%文中引用该图片代号
	\end{minipage}
        \vskip 7pt
        \begin{minipage}{0.2\linewidth}
		\centering
		\includegraphics[width=1\linewidth]{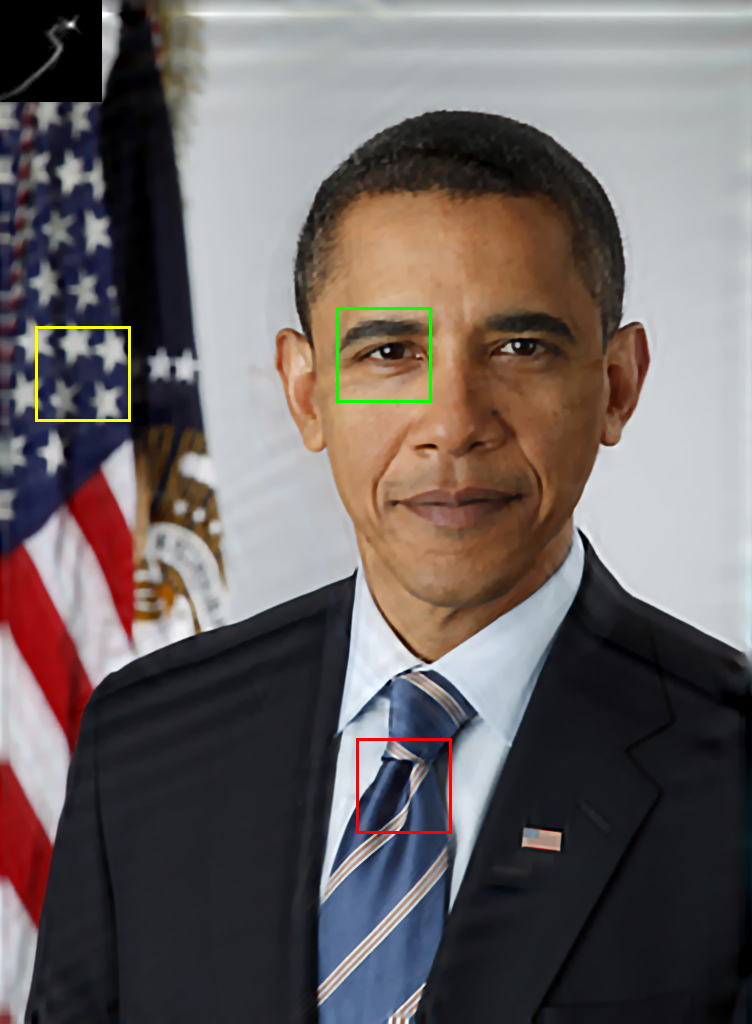}
		\vskip 4pt
		\includegraphics[width=1\linewidth]{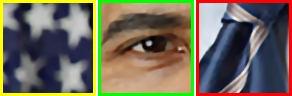}
		\caption*{(i)Wen et al.\cite{wen2021TCSVT} \protect\\ {PSNR: 31.24}\centering}
		\label{lai_people_03_kernel_02_wen}%文中引用该图片代号
	\end{minipage}
        \begin{minipage}{0.2\linewidth}
		\centering
		\includegraphics[width=1\linewidth]{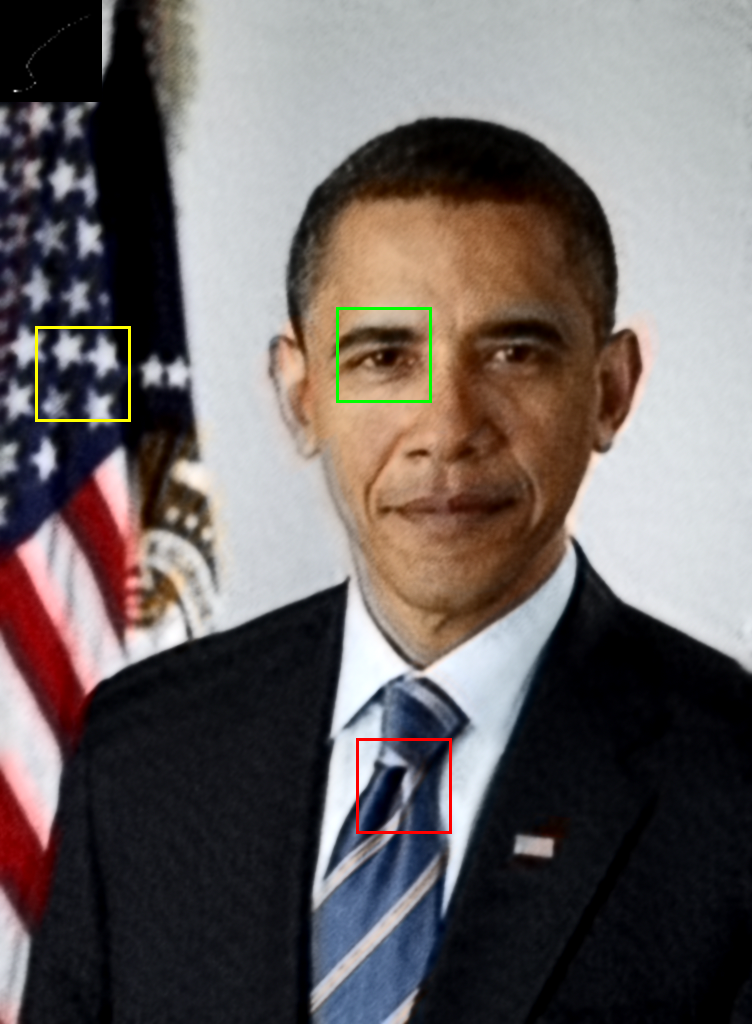}
		\vskip 4pt
		\includegraphics[width=1\linewidth]{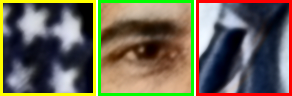}
		\caption*{(j)SelfDeblur\cite{ren2020neural} \protect\\ {PSNR: 26.82}\centering}
		\label{lai_people_03_kernel_02_SelfDeblur}%文中引用该图片代号
	\end{minipage}	 
	\begin{minipage}{0.2\linewidth}
		\centering
		\includegraphics[width=1\linewidth]{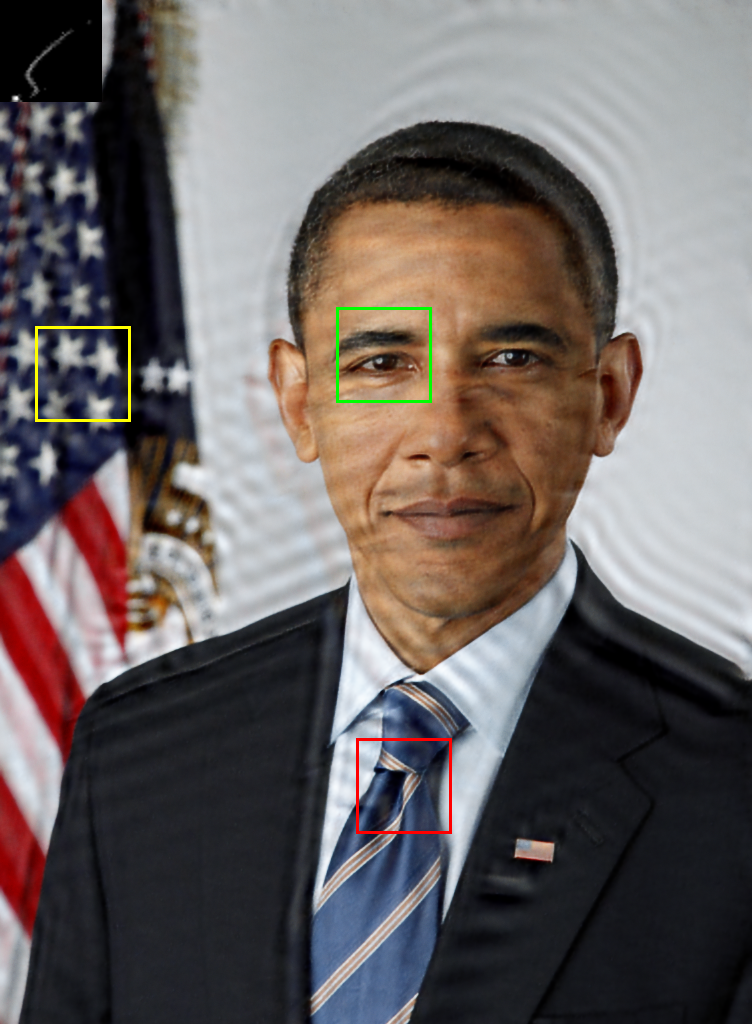}
		\vskip 4pt
		\includegraphics[width=1\linewidth]{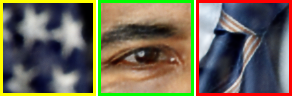}
		\caption*{(k)Fast-SelfDeblur\cite{bai2023fastselfdeblur} \protect\\ {PSNR: 28.55}\centering}
		\label{lai_people_03_kernel_02_Fast-SelfDeblur}%文中引用该图片代号
	\end{minipage}
	\begin{minipage}{0.2\linewidth}
		\centering
		\includegraphics[width=1\linewidth]{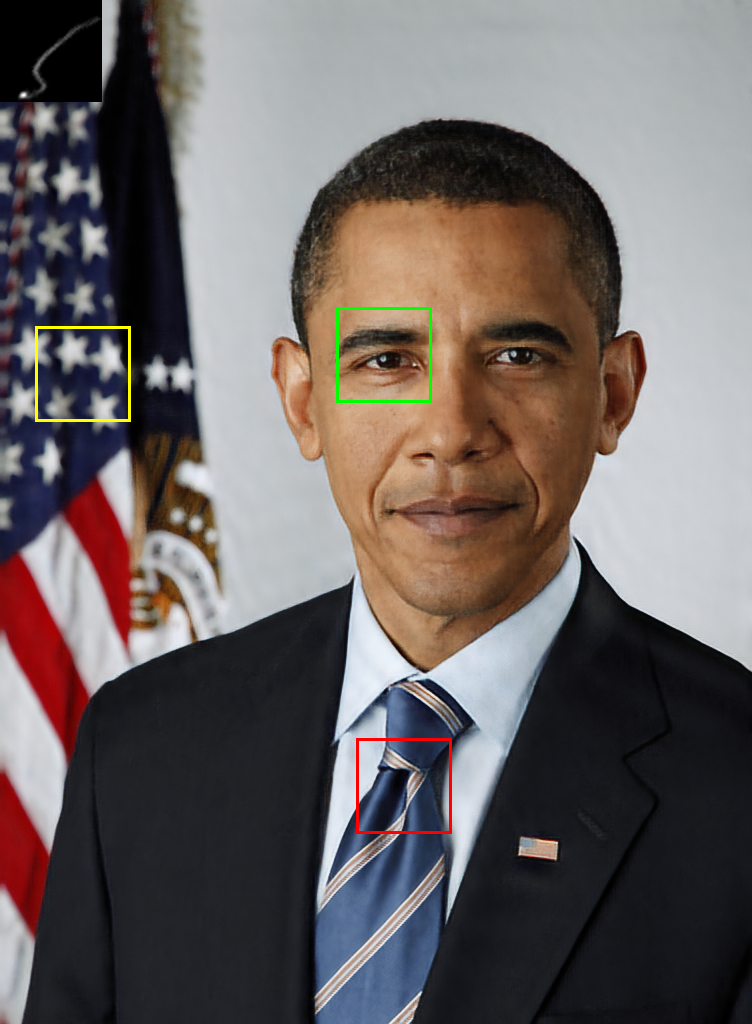}
		\vskip 4pt
		\includegraphics[width=1\linewidth]{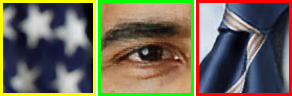}
		\caption*{(l)Self-MSNet (\textbf{Ours}) \protect\\ {PSNR: 35.10}\centering}
		\label{lai_people_03_kernel_02_ours}%文中引用该图片代号
	\end{minipage}
        %\vskip -8pt
	\caption{Visual comparison on an example image of the 'people' category with 2th blur kernel from Lai's dataset.}
	\label{fig:Lai-visual-people_03_kernel_02}
\end{figure*}

\begin{figure*}[!tbp]
	\centering
	\begin{minipage}{0.22\linewidth}
		\centering
		\includegraphics[width=1\linewidth]{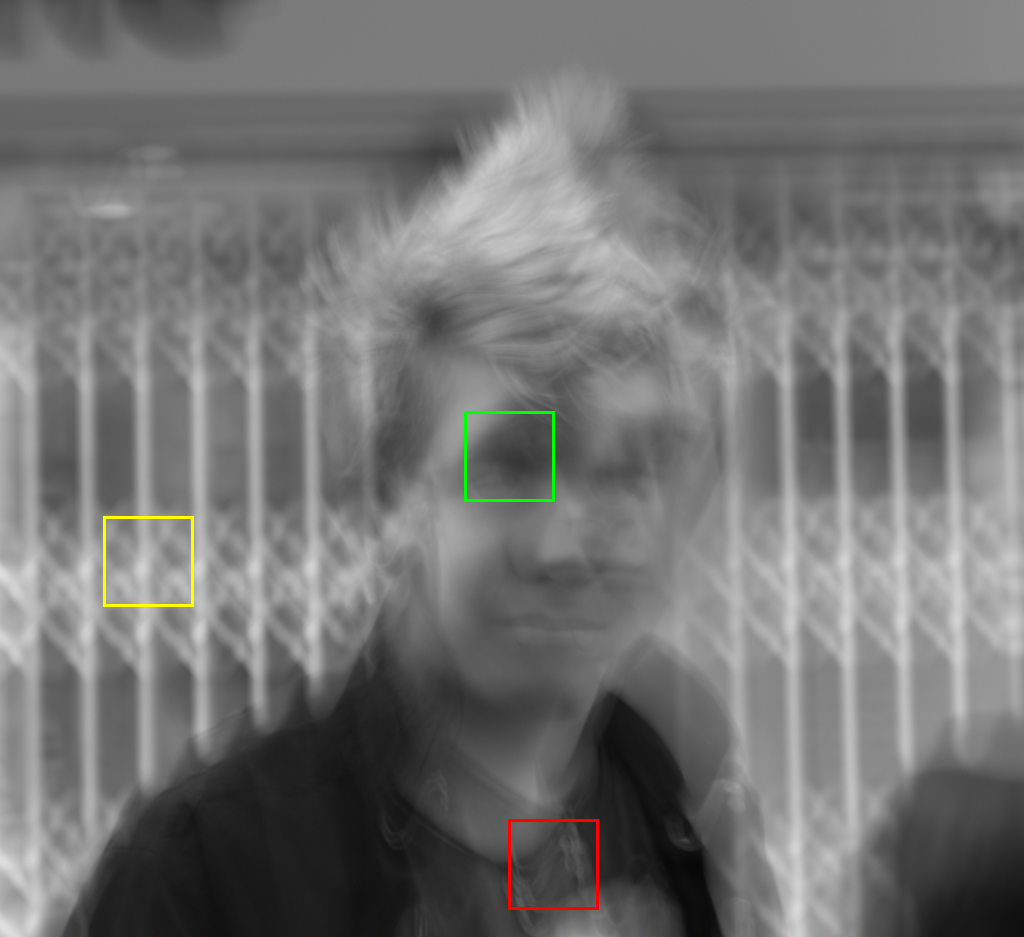}
		\vskip 4pt
		\includegraphics[width=1\linewidth]{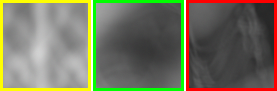}
		\caption*{(a)Blurred image \protect\\ {\textcolor{white}{***}}\centering}
		\label{lai_people_05_kernel_04_Blurry image}%文中引用该图片代号
	\end{minipage}
	\begin{minipage}{0.22\linewidth}
		\centering
		\includegraphics[width=1\linewidth]{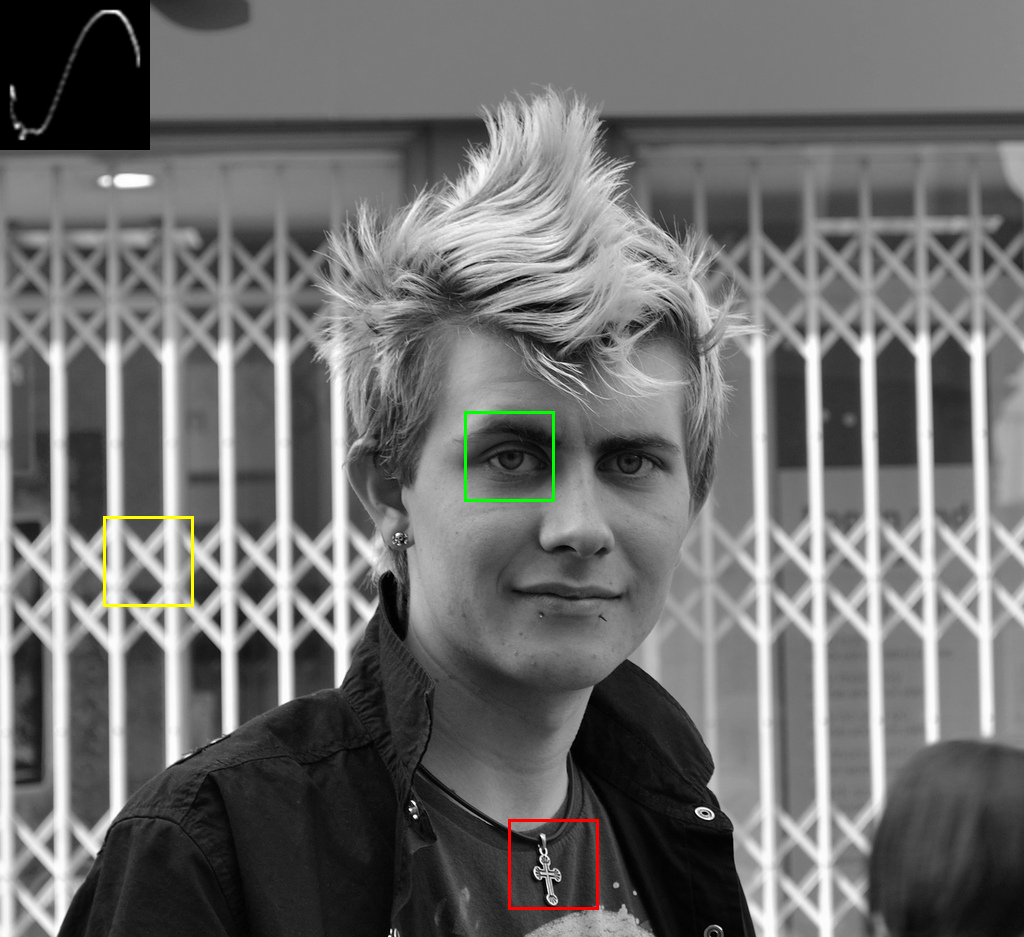}
		\vskip 4pt
		\includegraphics[width=1\linewidth]{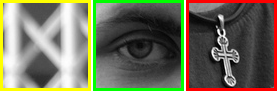}
		\caption*{(b)Ground-truth \protect\\ {\textcolor{white}{***}}\centering}
		\label{lai_people_05_kernel_04_Ground-truth}%文中引用该图片代号
	\end{minipage}
	\begin{minipage}{0.22\linewidth}
		\centering
		\includegraphics[width=1\linewidth]{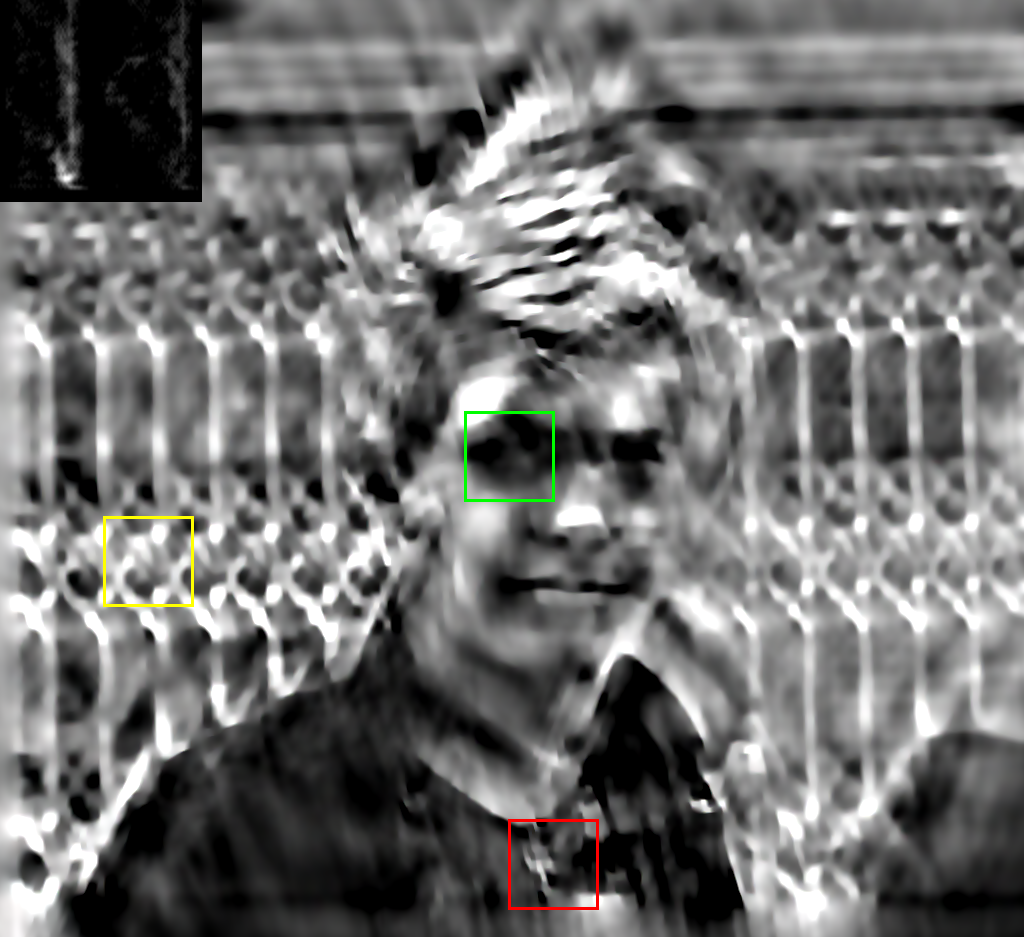}
		\vskip 4pt
		\includegraphics[width=1\linewidth]{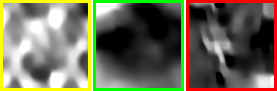}
		\caption*{(c)Cho and Lee\cite{cho2009fast} \protect\\ {PSNR: 15.92}\centering}
		\label{lai_people_05_kernel_04_cho}%文中引用该图片代号
	\end{minipage}
        \begin{minipage}{0.22\linewidth}
		\centering
		\includegraphics[width=1\linewidth]{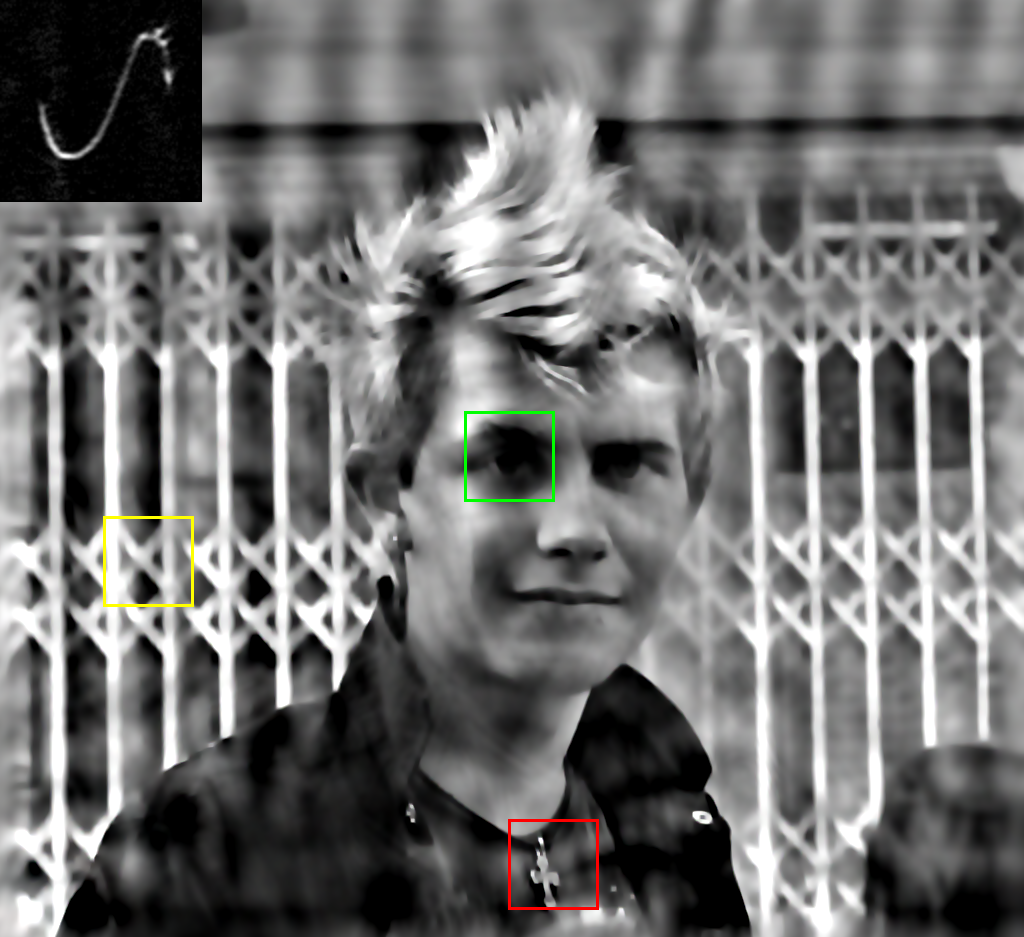}
		\vskip 4pt
		\includegraphics[width=1\linewidth]{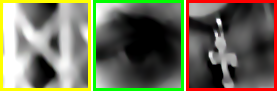}
		\caption*{(d)Xu and Jia\cite{xu2010two} \protect\\ {PSNR: 20.26}\centering}
		\label{lai_people_05_kernel_04_xuandjia}%文中引用该图片代号
	\end{minipage}
        \vskip 7pt
        \begin{minipage}{0.22\linewidth}
		\centering
		\includegraphics[width=1\linewidth]{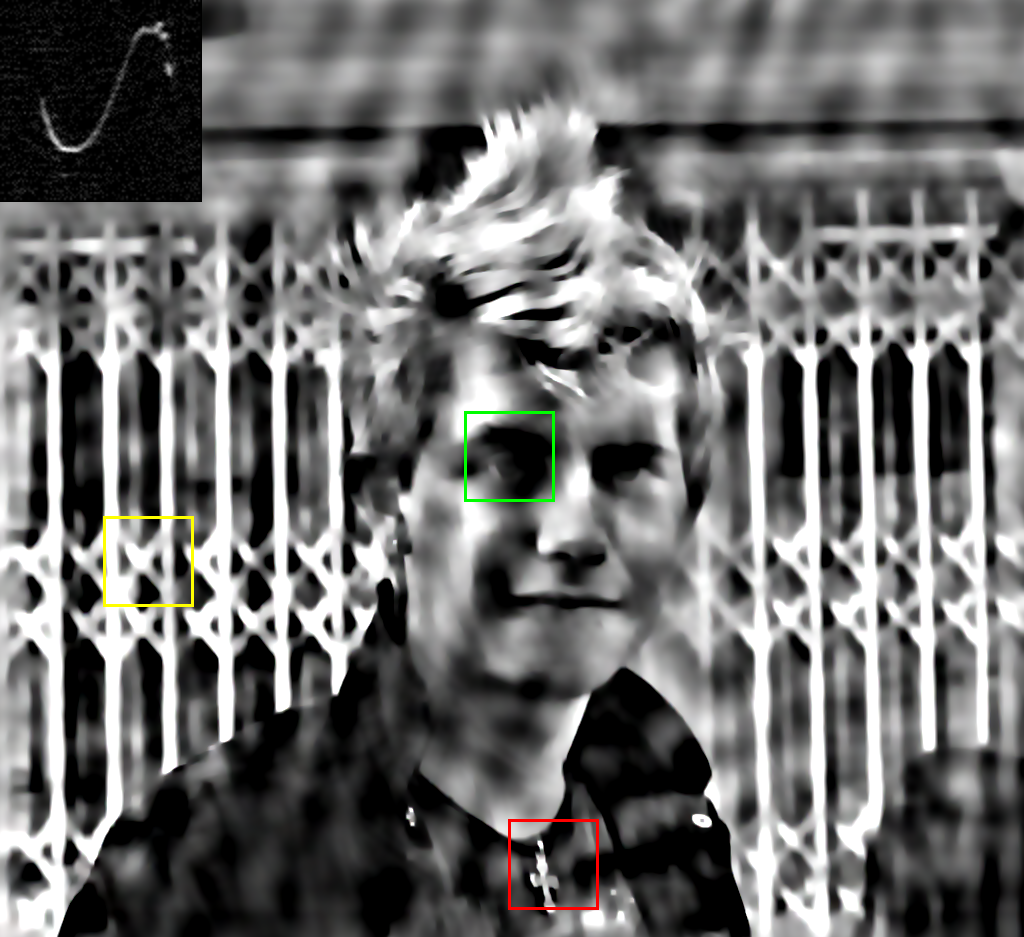}
		\vskip 4pt
		\includegraphics[width=1\linewidth]{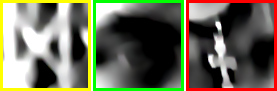}
		\caption*{(e)Xu et al.\cite{xu2013unnatural} \protect\\ {PSNR: 16.73}\centering}
		\label{lai_people_05_kernel_04_xuunnatural}%文中引用该图片代号
	\end{minipage} 
        \begin{minipage}{0.22\linewidth}
		\centering
		\includegraphics[width=1\linewidth]{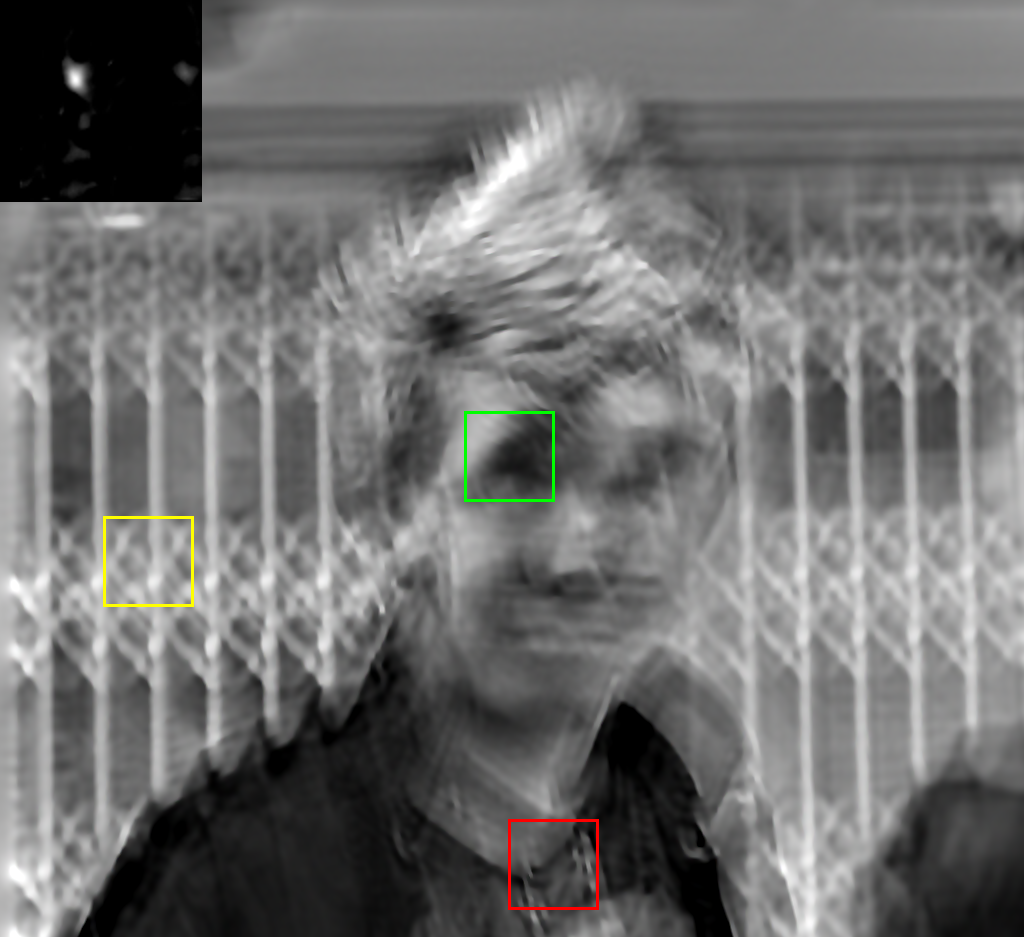}
		\vskip 4pt
		\includegraphics[width=1\linewidth]{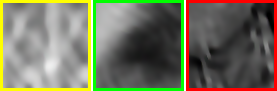}
		\caption*{(f)Michaeli and Irani\cite{michaeli2014blind} \protect\\ {PSNR: 15.93}\centering}
		\label{lai_people_05_kernel_04_Michaeli}%文中引用该图片代号
	\end{minipage} 
	\begin{minipage}{0.22\linewidth}
		\centering
		\includegraphics[width=1\linewidth]{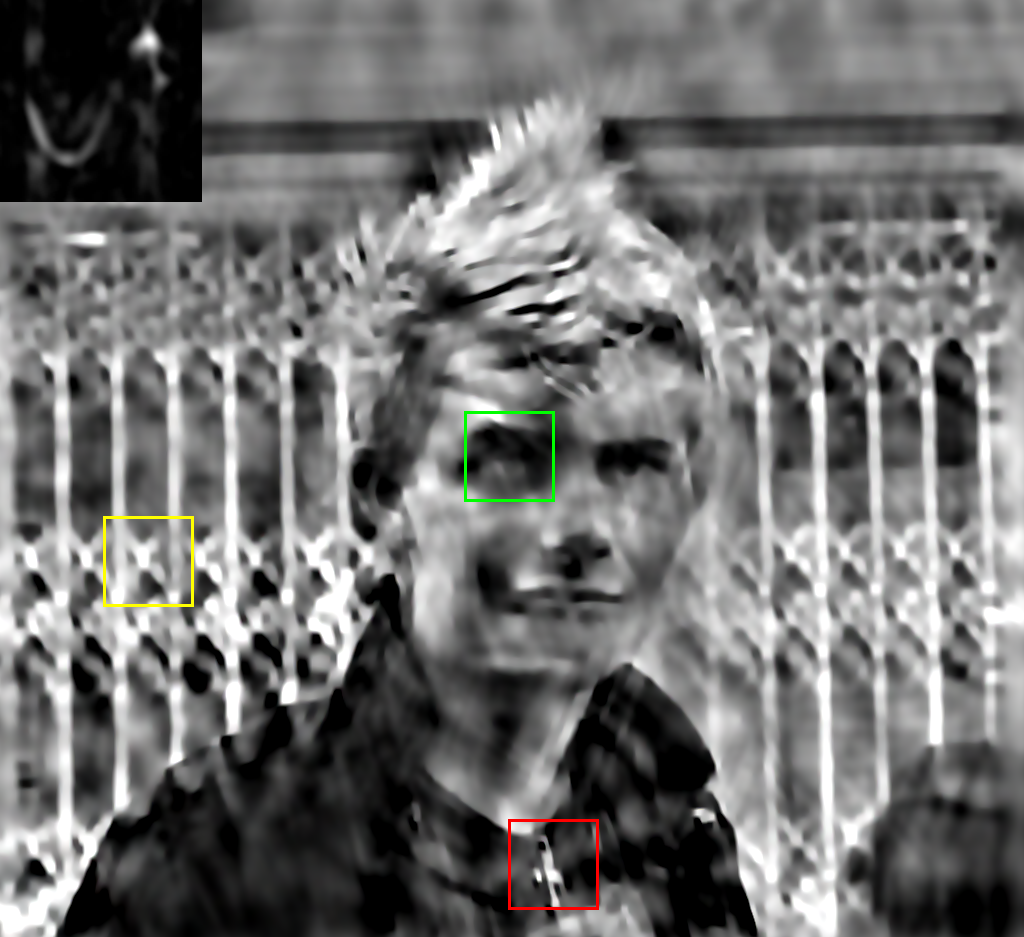}
		\vskip 4pt
		\includegraphics[width=1\linewidth]{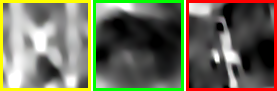}
		\caption*{(g)Perrone and Favaro\cite{perrone2014total} \protect\\ {PSNR: 18.90}\centering}
		\label{lai_people_05_kernel_04_Perrone}%文中引用该图片代号
	\end{minipage}
	\begin{minipage}{0.22\linewidth}
		\centering
		\includegraphics[width=1\linewidth]{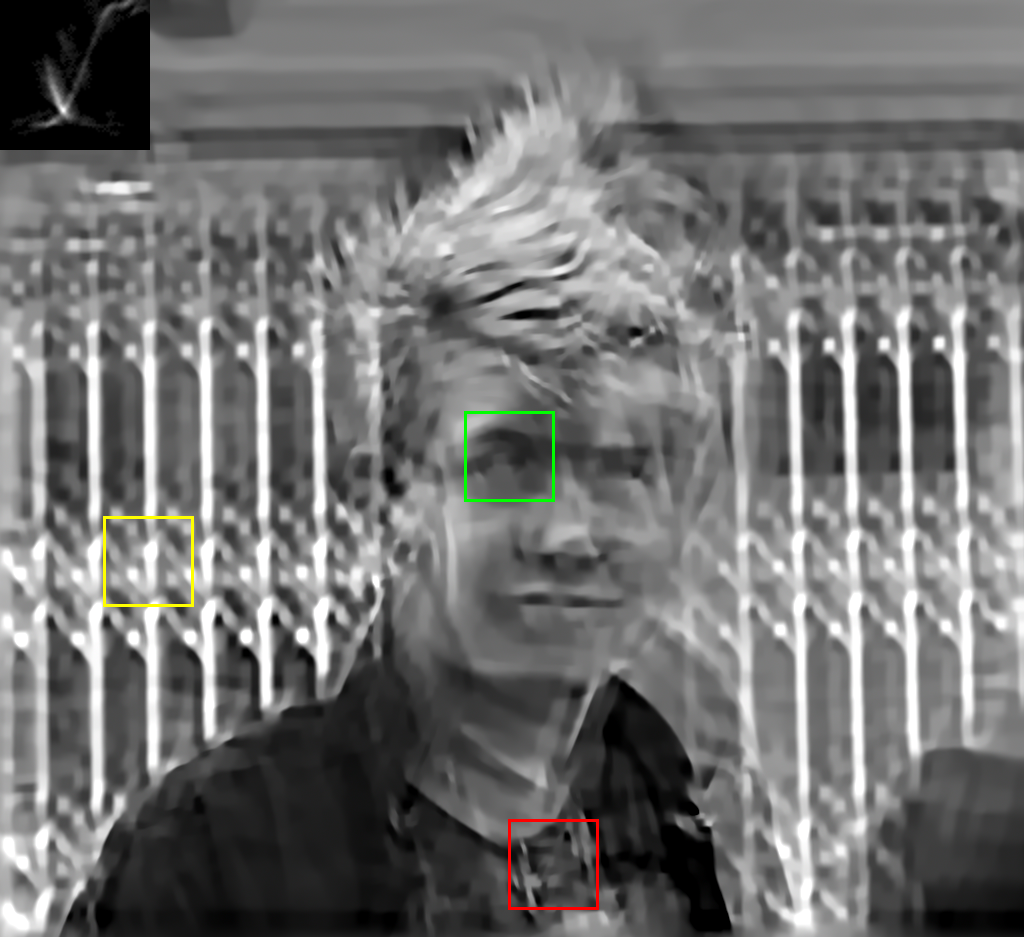}
		\vskip 4pt
		\includegraphics[width=1\linewidth]{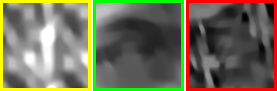}
		\caption*{(h)Pan et al.\cite{pan2018PAMIdarkchannel} \protect\\ {PSNR: 14.24}\centering}
		\label{lai_people_05_kernel_04_Pan}%文中引用该图片代号
	\end{minipage}
        \vskip 7pt
        \begin{minipage}{0.22\linewidth}
		\centering
		\includegraphics[width=1\linewidth]{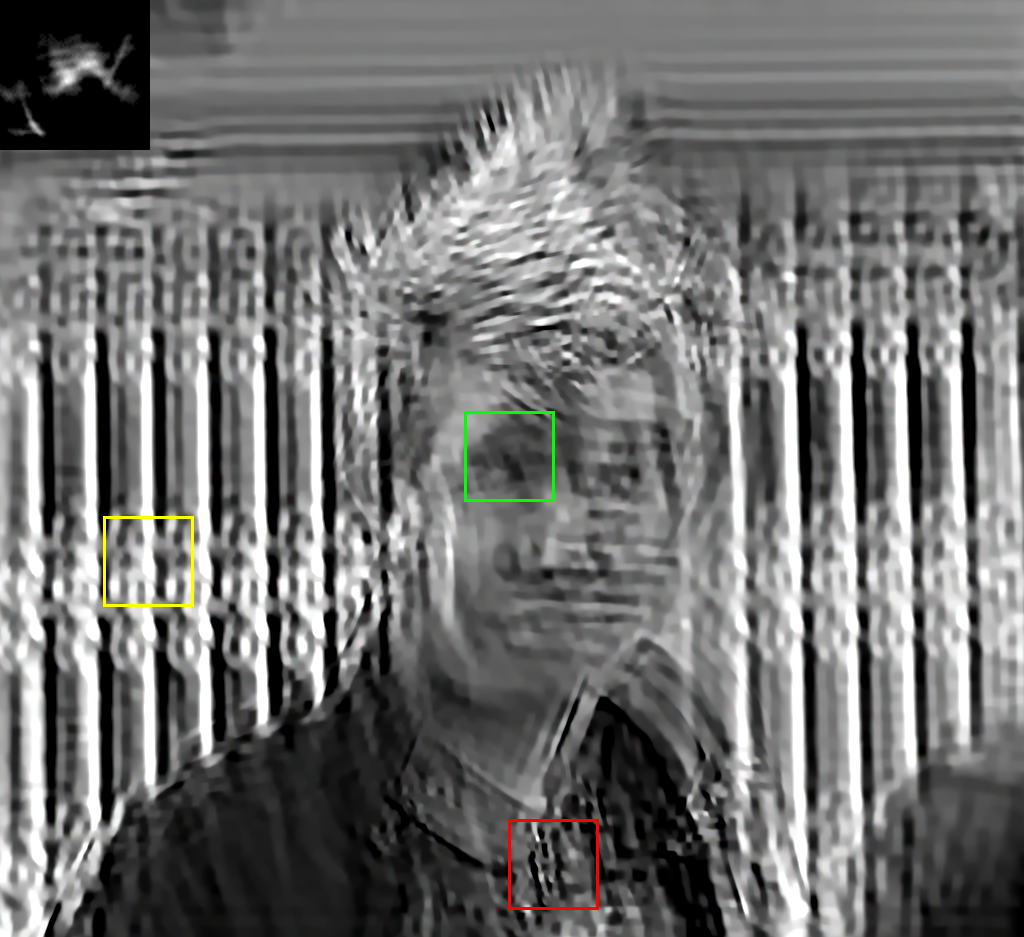}
		\vskip 4pt
		\includegraphics[width=1\linewidth]{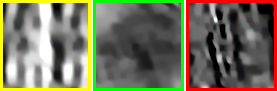}
		\caption*{(i)Wen et al.\cite{wen2021TCSVT} \protect\\ {PSNR: 12.97}\centering}
		\label{lai_people_05_kernel_04_wen}%文中引用该图片代号
	\end{minipage}
        \begin{minipage}{0.22\linewidth}
		\centering
		\includegraphics[width=1\linewidth]{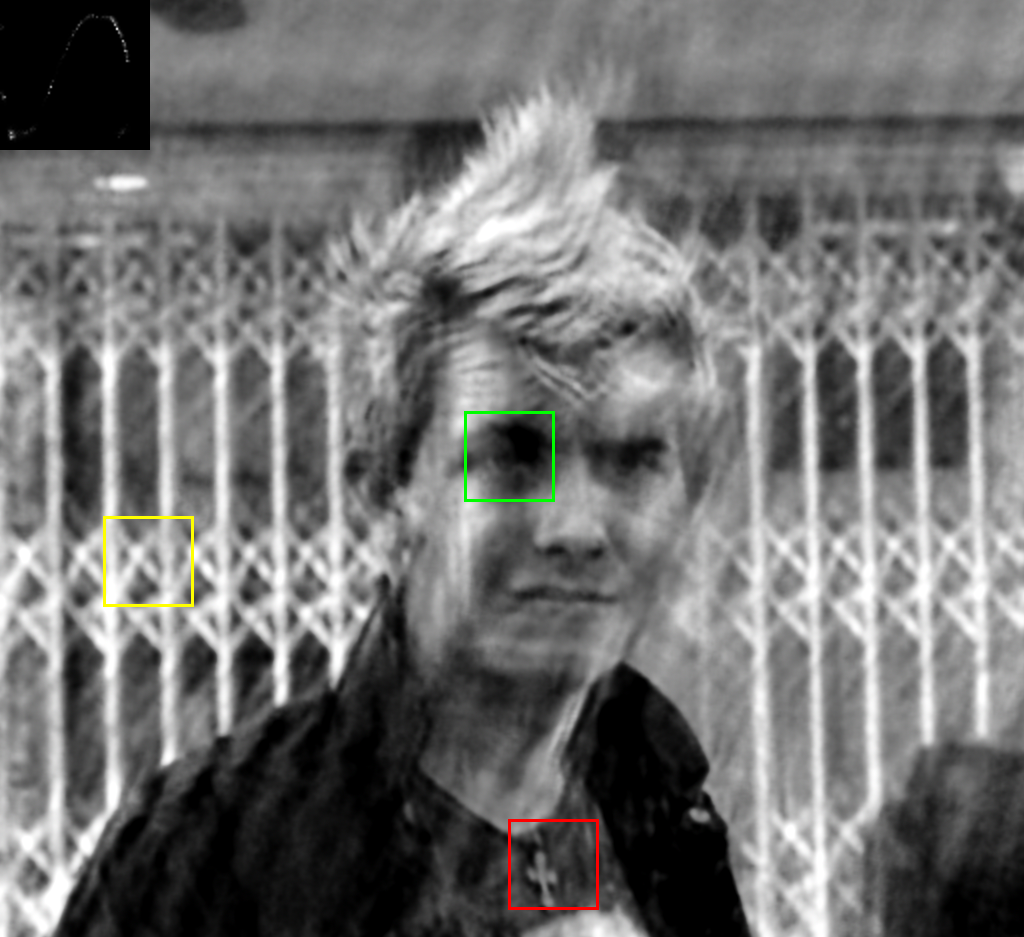}
		\vskip 4pt
		\includegraphics[width=1\linewidth]{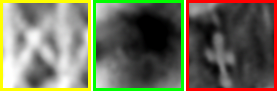}
		\caption*{(j)SelfDeblur\cite{ren2020neural} \protect\\ {PSNR: 22.02}\centering}
		\label{lai_people_05_kernel_04_SelfDeblur}%文中引用该图片代号
	\end{minipage}	 
	\begin{minipage}{0.22\linewidth}
		\centering
		\includegraphics[width=1\linewidth]{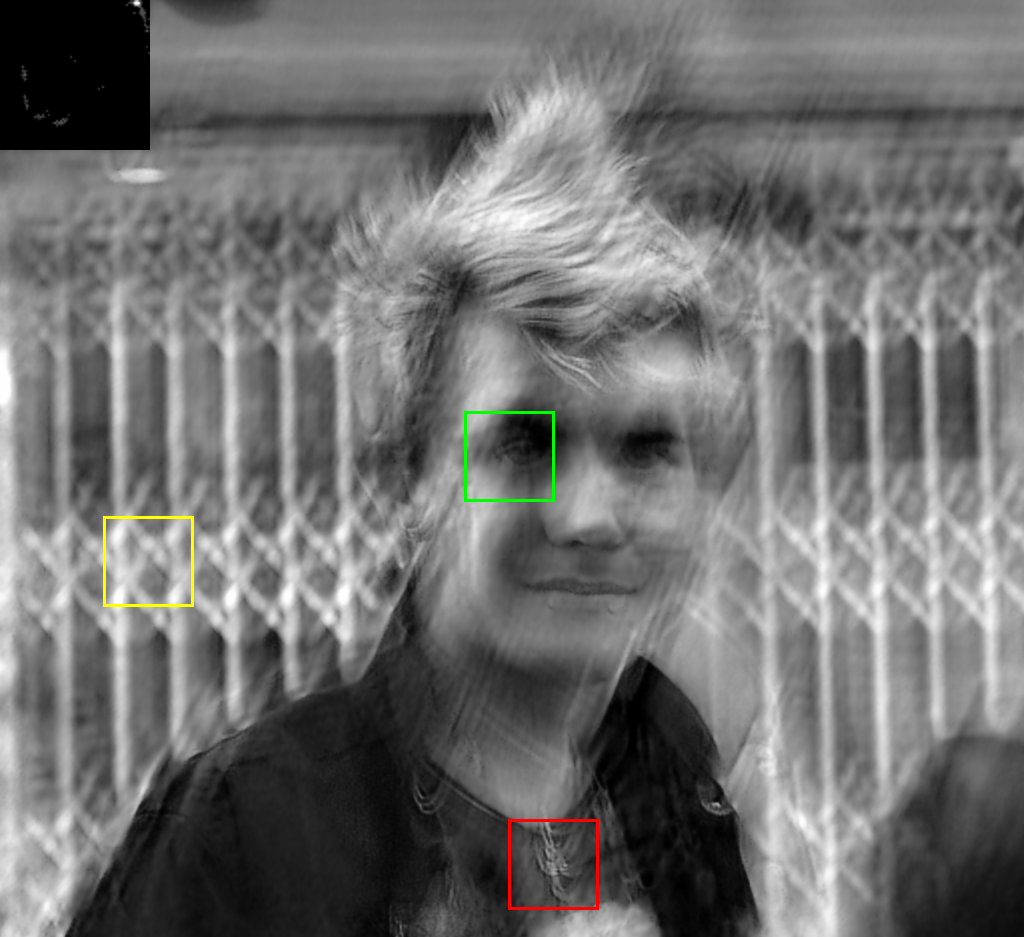}
		\vskip 4pt
		\includegraphics[width=1\linewidth]{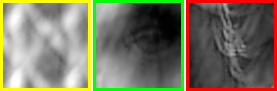}
		\caption*{(k)Fast-SelfDeblur\cite{bai2023fastselfdeblur} \protect\\ {PSNR: 20.16}\centering}
		\label{lai_people_05_kernel_04_Fast-SelfDeblur}%文中引用该图片代号
	\end{minipage}
	\begin{minipage}{0.22\linewidth}
		\centering
		\includegraphics[width=1\linewidth]{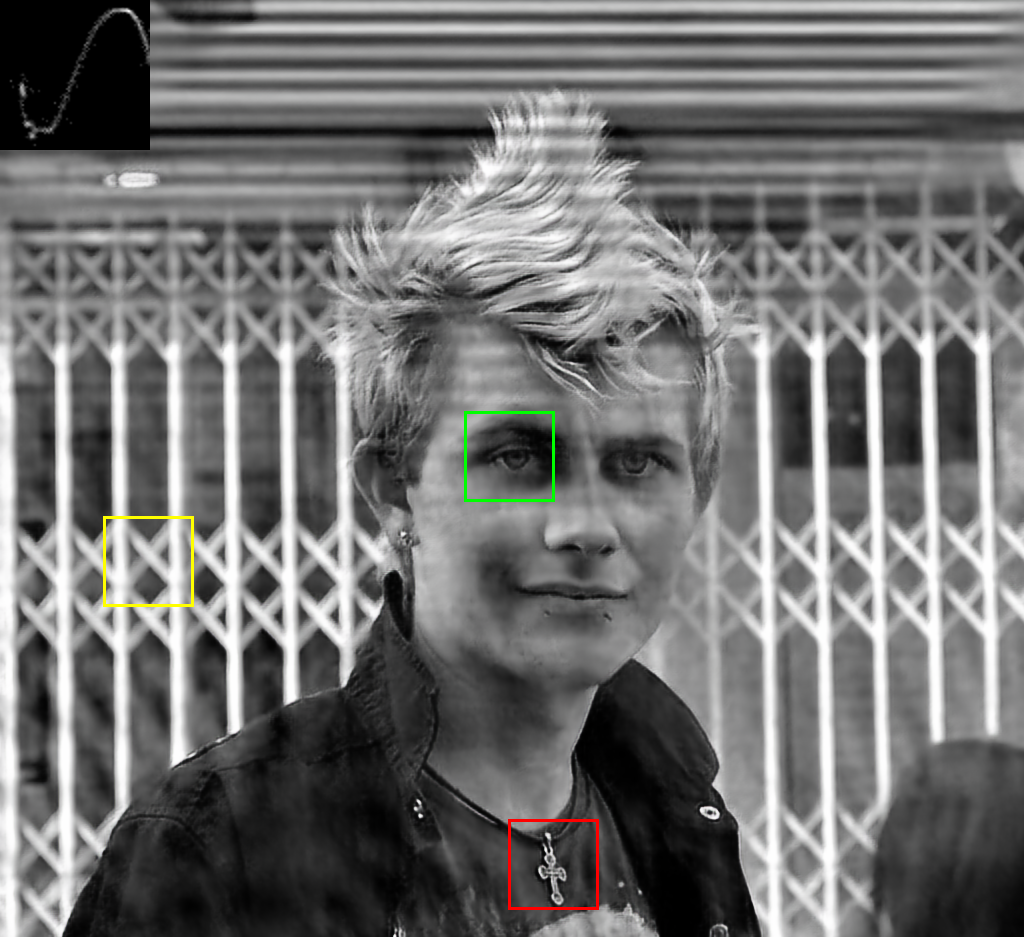}
		\vskip 4pt
		\includegraphics[width=1\linewidth]{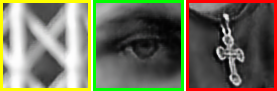}
		\caption*{(l)Self-MSNet (\textbf{Ours}) \protect\\ {PSNR: 24.03}\centering}
		\label{lai_people_05_kernel_04_ours}%文中引用该图片代号
	\end{minipage}
        %\vskip -8pt
	\caption{Visual comparison on another example image of the 'people' category with 4th blur kernel from Lai's dataset.}
	\label{fig:Lai-visual-people_05_kernel_04}
\end{figure*}

\begin{figure*}[!tbp]
	\centering
	\begin{minipage}{0.22\linewidth}
		\centering
		\includegraphics[width=1\linewidth]{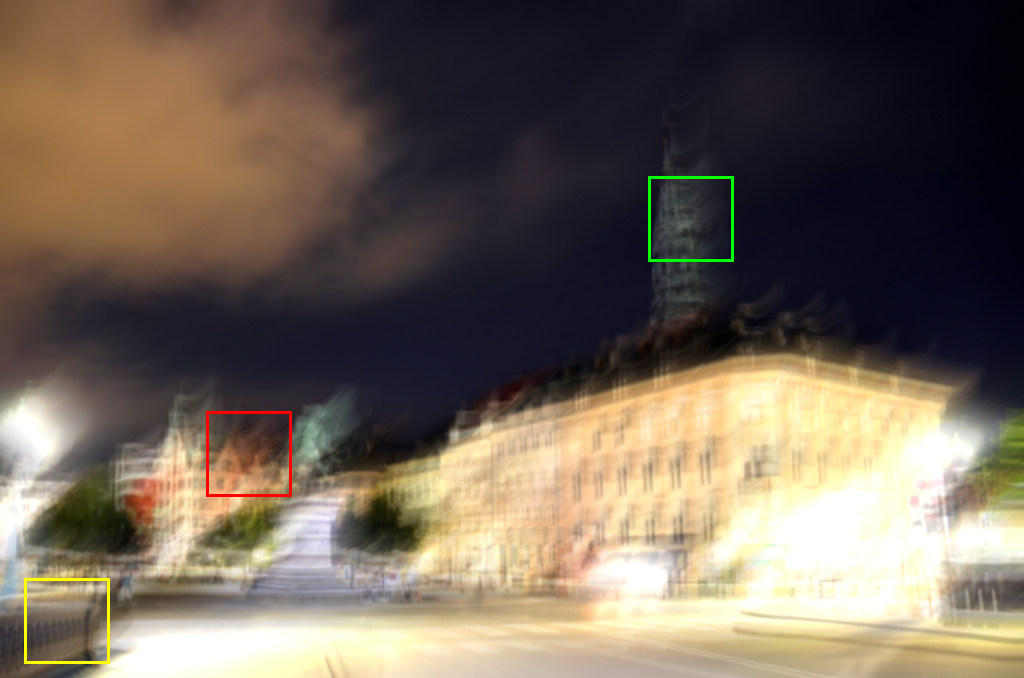}
		\vskip 4pt
		\includegraphics[width=1\linewidth]{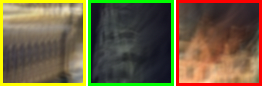}
		\caption*{(a)Blurred image \protect\\ {\textcolor{white}{***}}\centering}
		\label{lai_saturated_05_kernel_03_Blurry image}%文中引用该图片代号
	\end{minipage}
	\begin{minipage}{0.22\linewidth}
		\centering
		\includegraphics[width=1\linewidth]{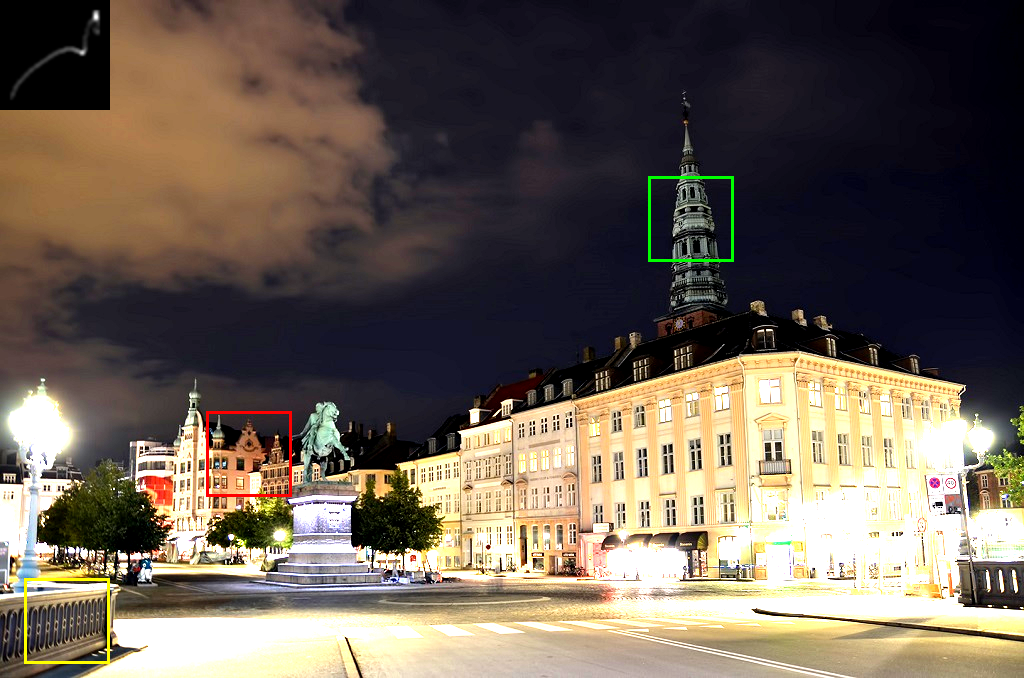}
		\vskip 4pt
		\includegraphics[width=1\linewidth]{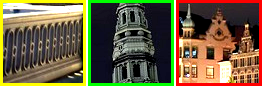}
		\caption*{(b)Ground-truth \protect\\ {\textcolor{white}{***}}\centering}
		\label{lai_saturated_05_kernel_03_Ground-truth}%文中引用该图片代号
	\end{minipage}
	\begin{minipage}{0.22\linewidth}
		\centering
		\includegraphics[width=1\linewidth]{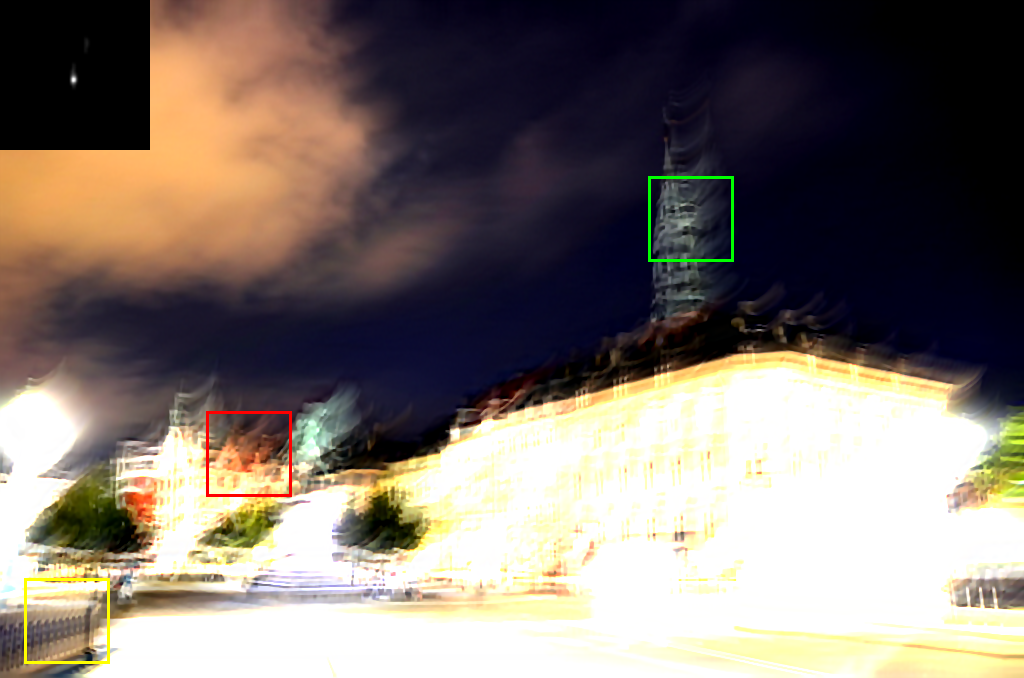}
		\vskip 4pt
		\includegraphics[width=1\linewidth]{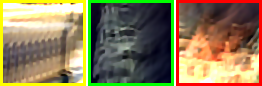}
		\caption*{(c)Cho and Lee\cite{cho2009fast} \protect\\ {PSNR: 15.34}\centering}
		\label{lai_saturated_05_kernel_03_cho}%文中引用该图片代号
	\end{minipage}
        \begin{minipage}{0.22\linewidth}
		\centering
		\includegraphics[width=1\linewidth]{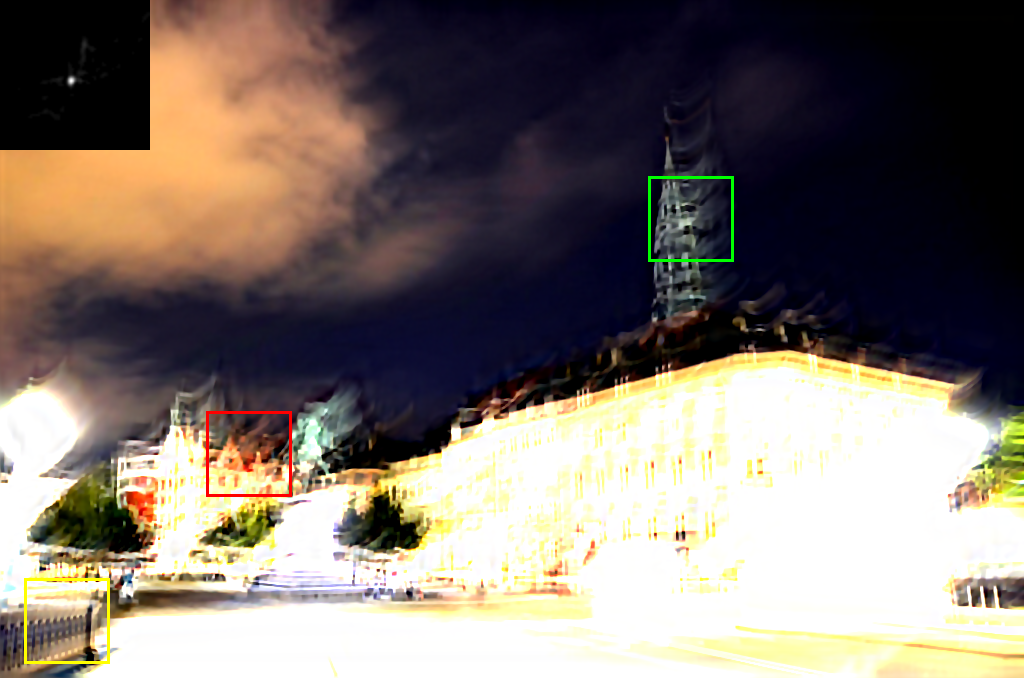}
		\vskip 4pt
		\includegraphics[width=1\linewidth]{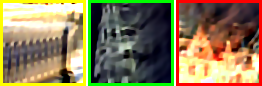}
		\caption*{(d)Xu and Jia\cite{xu2010two} \protect\\ {PSNR: 15.33}\centering}
		\label{lai_saturated_05_kernel_03_xuandjia}%文中引用该图片代号
	\end{minipage}
        \vskip 7pt
        \begin{minipage}{0.22\linewidth}
		\centering
		\includegraphics[width=1\linewidth]{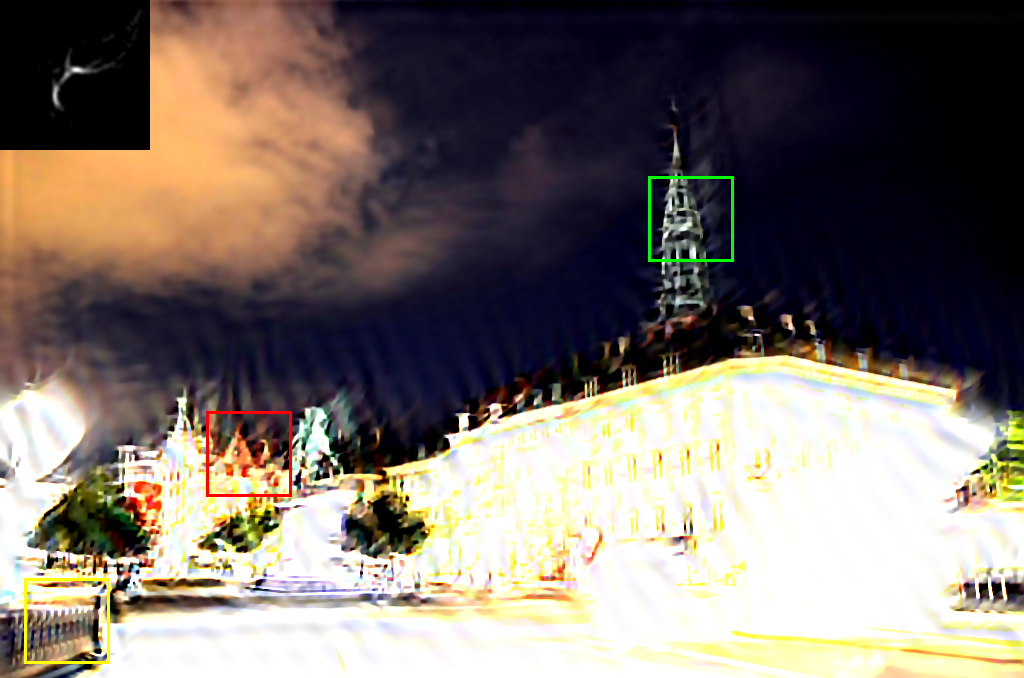}
		\vskip 4pt
		\includegraphics[width=1\linewidth]{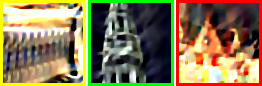}
		\caption*{(e)Xu et al.\cite{xu2013unnatural} \protect\\ {PSNR: 15.08}\centering}
		\label{lai_saturated_05_kernel_03_xuunnatural}%文中引用该图片代号
	\end{minipage} 
        \begin{minipage}{0.22\linewidth}
		\centering
		\includegraphics[width=1\linewidth]{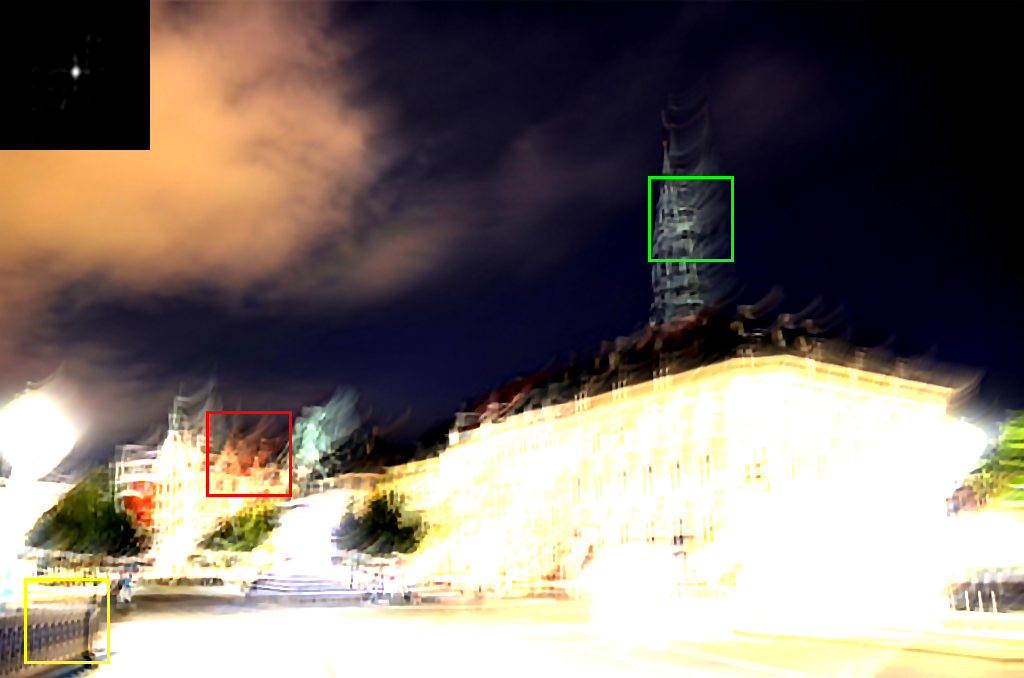}
		\vskip 4pt
		\includegraphics[width=1\linewidth]{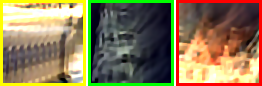}
		\caption*{(f)Michaeli and Irani\cite{michaeli2014blind} \protect\\ {PSNR: 15.25}\centering}
		\label{lai_saturated_05_kernel_03_Michaeli}%文中引用该图片代号
	\end{minipage} 
	\begin{minipage}{0.22\linewidth}
		\centering
		\includegraphics[width=1\linewidth]{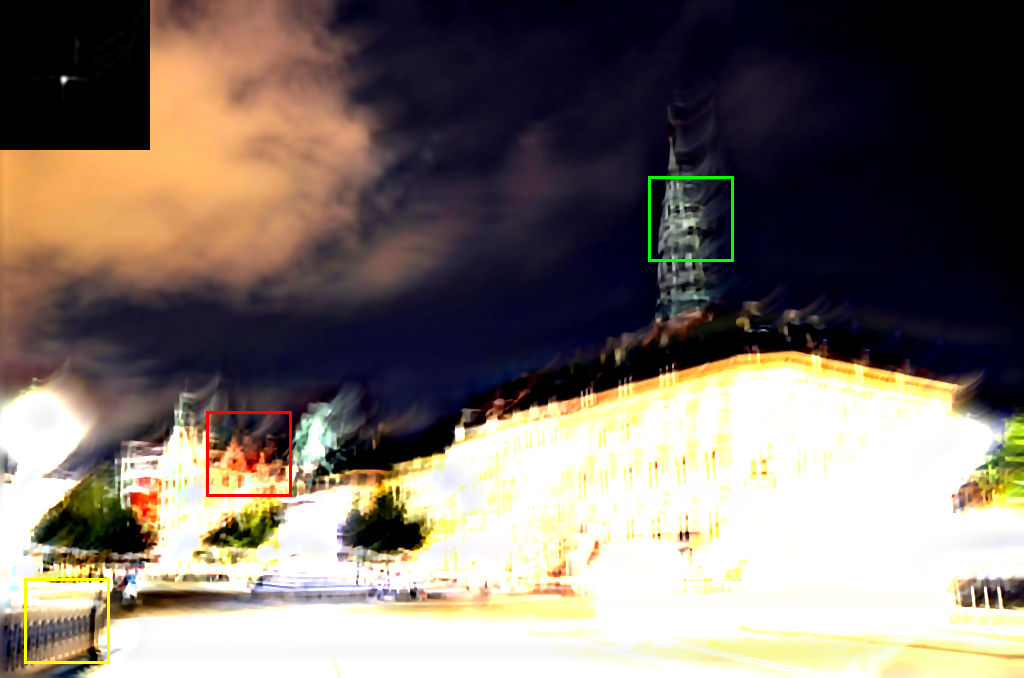}
		\vskip 4pt
		\includegraphics[width=1\linewidth]{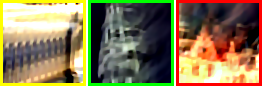}
		\caption*{(g)Perrone and Favaro\cite{perrone2014total} \protect\\ {PSNR: 15.39}\centering}
		\label{lai_saturated_05_kernel_03_Perrone}%文中引用该图片代号
	\end{minipage}
	\begin{minipage}{0.22\linewidth}
		\centering
		\includegraphics[width=1\linewidth]{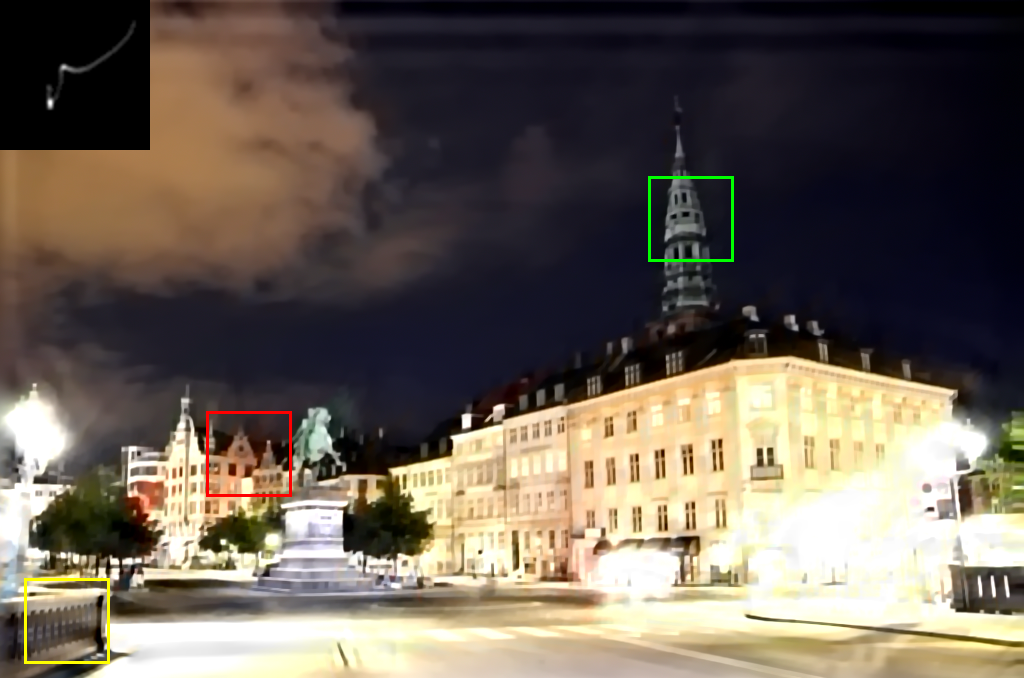}
		\vskip 4pt
		\includegraphics[width=1\linewidth]{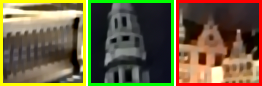}
		\caption*{(h)Pan et al.\cite{pan2018PAMIdarkchannel} \protect\\ {PSNR: 20.04}\centering}
		\label{lai_saturated_05_kernel_03_Pan}%文中引用该图片代号
	\end{minipage}
        \vskip 7pt
        \begin{minipage}{0.22\linewidth}
		\centering
		\includegraphics[width=1\linewidth]{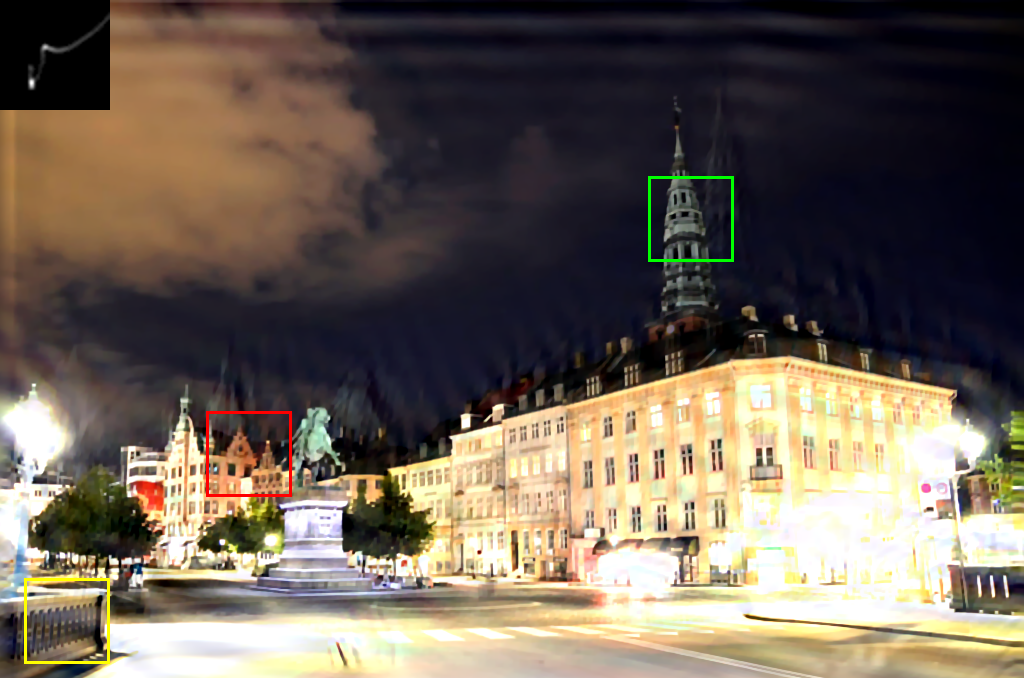}
		\vskip 4pt
		\includegraphics[width=1\linewidth]{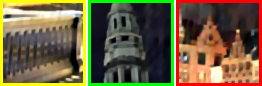}
		\caption*{(i)Wen et al.\cite{wen2021TCSVT} \protect\\ {PSNR: 24.09}\centering}
		\label{lai_saturated_05_kernel_03_wen}%文中引用该图片代号
	\end{minipage}
        \begin{minipage}{0.22\linewidth}
		\centering
		\includegraphics[width=1\linewidth]{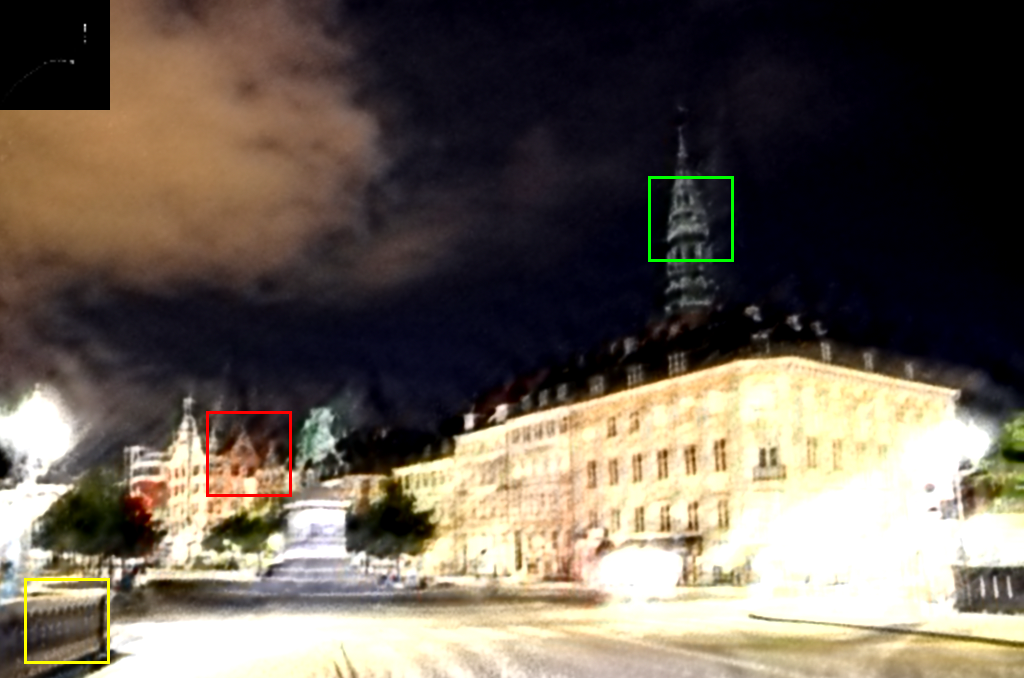}
		\vskip 4pt
		\includegraphics[width=1\linewidth]{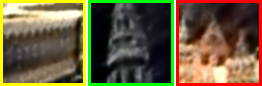}
		\caption*{(j)SelfDeblur\cite{ren2020neural} \protect\\ {PSNR: 21.03}\centering}
		\label{lai_saturated_05_kernel_03_SelfDeblur}%文中引用该图片代号
	\end{minipage}	 
	\begin{minipage}{0.22\linewidth}
		\centering
		\includegraphics[width=1\linewidth]{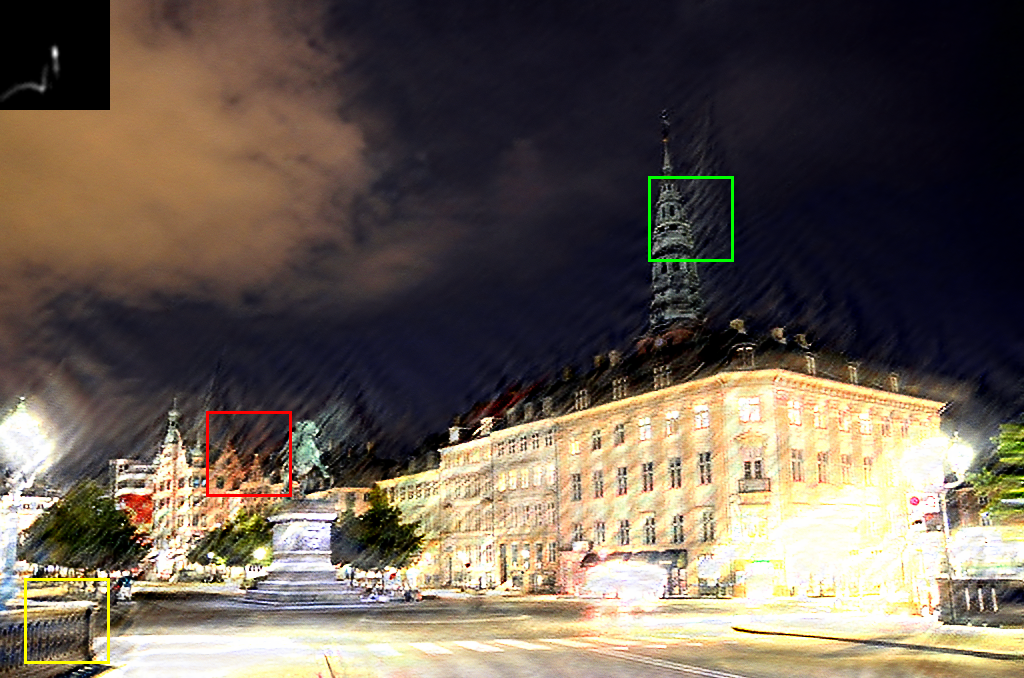}
		\vskip 4pt
		\includegraphics[width=1\linewidth]{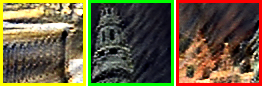}
		\caption*{(k)Fast-SelfDeblur\cite{bai2023fastselfdeblur} \protect\\ {PSNR: 16.63}\centering}
		\label{lai_saturated_05_kernel_03_Fast-SelfDeblur}%文中引用该图片代号
	\end{minipage}
	\begin{minipage}{0.22\linewidth}
		\centering
		\includegraphics[width=1\linewidth]{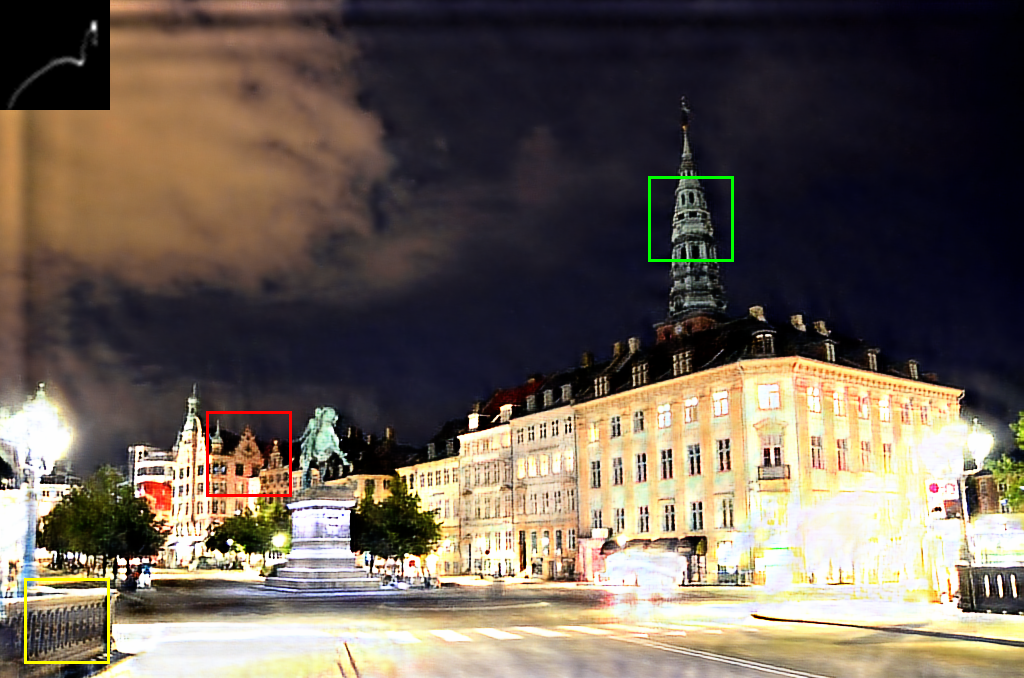}
		\vskip 4pt
		\includegraphics[width=1\linewidth]{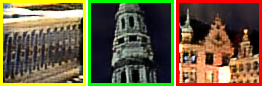}
		\caption*{(l)Self-MSNet (\textbf{Ours}) \protect\\ {PSNR: 25.23}\centering}
		\label{lai_saturated_05_kernel_03_ours}%文中引用该图片代号
	\end{minipage}
        %\vskip -8pt
	\caption{Visual comparison on an example image of the 'saturated' category with 3th blur kernel from Lai's dataset.}
	\label{fig:Lai-visual-saturated_05_kernel_03}
\end{figure*}

\begin{figure*}[!tbp]
	\centering
	\begin{minipage}{0.22\linewidth}
		\centering
		\includegraphics[width=1\linewidth]{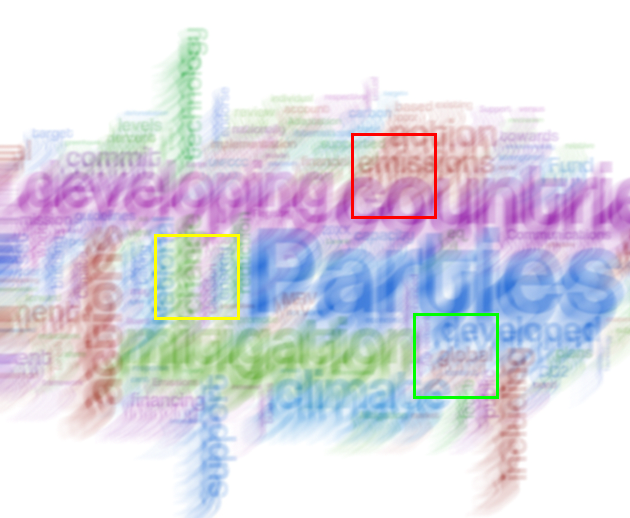}
		\vskip 4pt
		\includegraphics[width=1\linewidth]{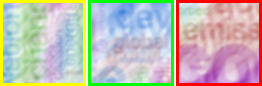}
		\caption*{(a)Blurred image \protect\\ {\textcolor{white}{***}}\centering}
		\label{lai_text_01_kernel_02_Blurry image}%文中引用该图片代号
	\end{minipage}
	\begin{minipage}{0.22\linewidth}
		\centering
		\includegraphics[width=1\linewidth]{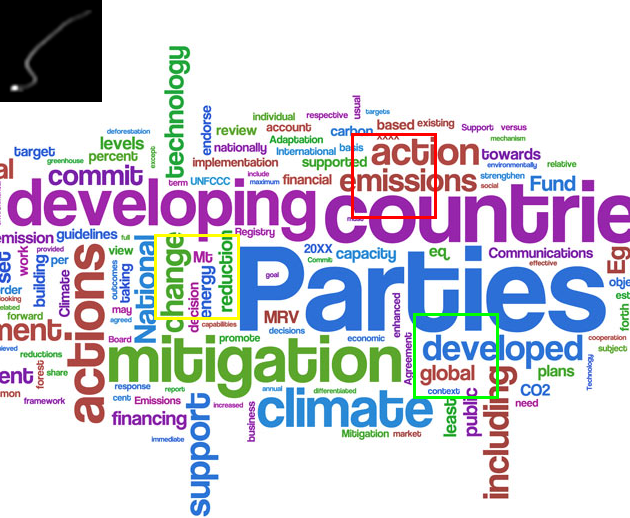}
		\vskip 4pt
		\includegraphics[width=1\linewidth]{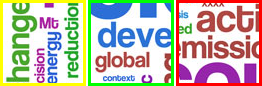}
		\caption*{(b)Ground-truth \protect\\ {\textcolor{white}{***}}\centering}
		\label{lai_text_01_kernel_02_Ground-truth}%文中引用该图片代号
	\end{minipage}
	\begin{minipage}{0.22\linewidth}
		\centering
		\includegraphics[width=1\linewidth]{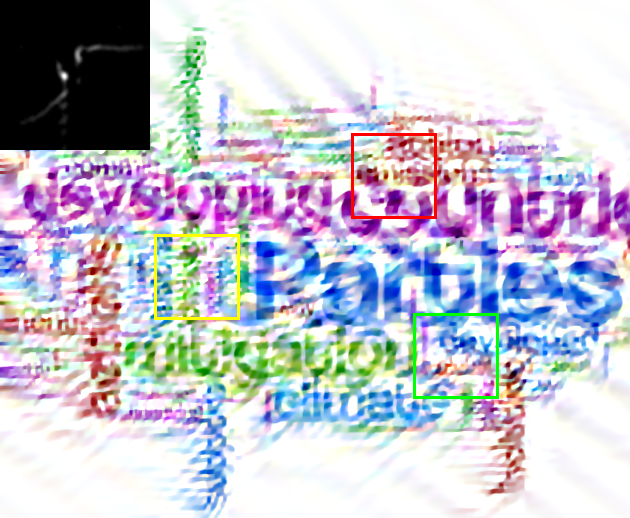}
		\vskip 4pt
		\includegraphics[width=1\linewidth]{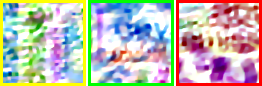}
		\caption*{(c)Cho and Lee\cite{cho2009fast} \protect\\ {PSNR: 13.87}\centering}
		\label{lai_text_01_kernel_02_cho}%文中引用该图片代号
	\end{minipage}
        \begin{minipage}{0.22\linewidth}
		\centering
		\includegraphics[width=1\linewidth]{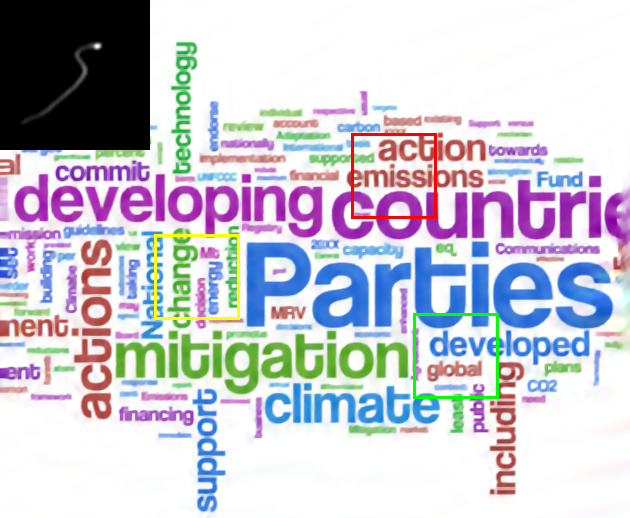}
		\vskip 4pt
		\includegraphics[width=1\linewidth]{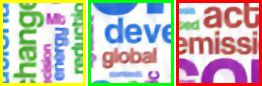}
		\caption*{(d)Xu and Jia\cite{xu2010two} \protect\\ {PSNR: 18.20}\centering}
		\label{lai_text_01_kernel_02_xuandjia}%文中引用该图片代号
	\end{minipage}
        \vskip 7pt
        \begin{minipage}{0.22\linewidth}
		\centering
		\includegraphics[width=1\linewidth]{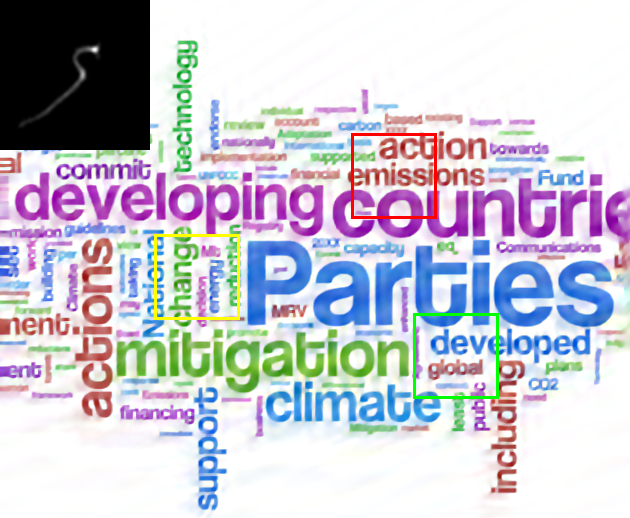}
		\vskip 4pt
		\includegraphics[width=1\linewidth]{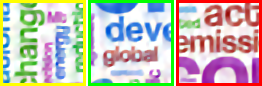}
		\caption*{(e)Xu et al.\cite{xu2013unnatural} \protect\\ {PSNR: 16.12}\centering}
		\label{lai_text_01_kernel_02_xuunnatural}%文中引用该图片代号
	\end{minipage} 
        \begin{minipage}{0.22\linewidth}
		\centering
		\includegraphics[width=1\linewidth]{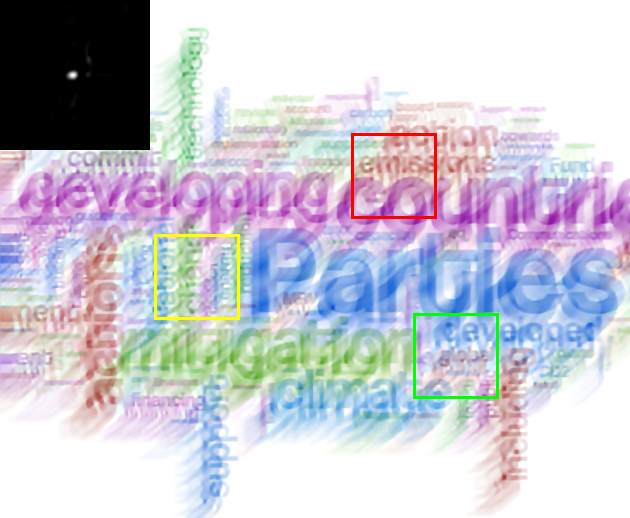}
		\vskip 4pt
		\includegraphics[width=1\linewidth]{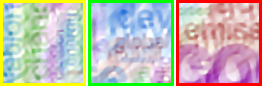}
		\caption*{(f)Michaeli and Irani\cite{michaeli2014blind} \protect\\ {PSNR: 14.69}\centering}
		\label{lai_text_01_kernel_02_Michaeli}%文中引用该图片代号
	\end{minipage} 
	\begin{minipage}{0.22\linewidth}
		\centering
		\includegraphics[width=1\linewidth]{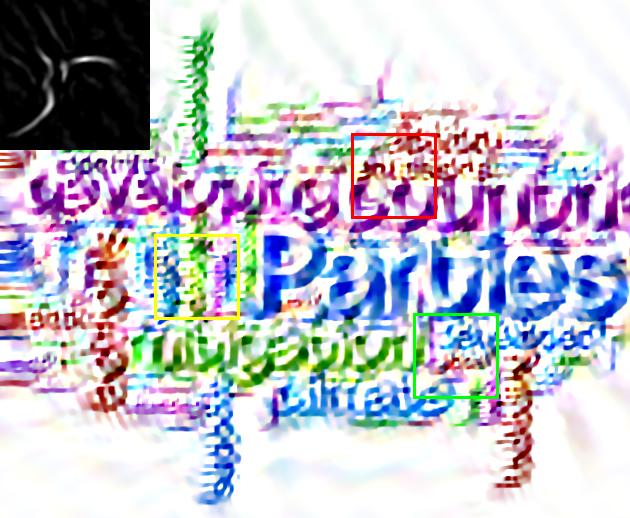}
		\vskip 4pt
		\includegraphics[width=1\linewidth]{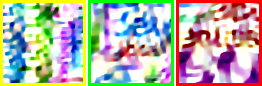}
		\caption*{(g)Perrone and Favaro\cite{perrone2014total} \protect\\ {PSNR: 11.98}\centering}
		\label{lai_text_01_kernel_02_Perrone}%文中引用该图片代号
	\end{minipage}
	\begin{minipage}{0.22\linewidth}
		\centering
		\includegraphics[width=1\linewidth]{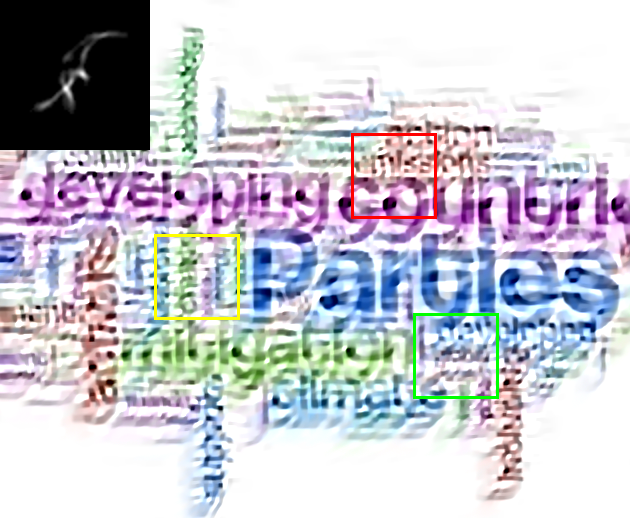}
		\vskip 4pt
		\includegraphics[width=1\linewidth]{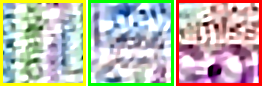}
		\caption*{(h)Pan et al.\cite{pan2018PAMIdarkchannel} \protect\\ {PSNR: 13.97}\centering}
		\label{lai_text_01_kernel_02_Pan}%文中引用该图片代号
	\end{minipage}
        \vskip 7pt
        \begin{minipage}{0.22\linewidth}
		\centering
		\includegraphics[width=1\linewidth]{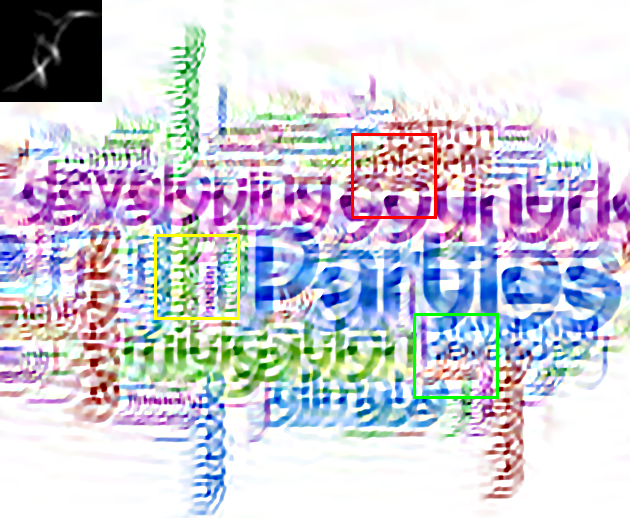}
		\vskip 4pt
		\includegraphics[width=1\linewidth]{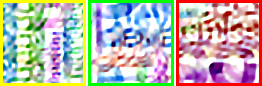}
		\caption*{(i)Wen et al.\cite{wen2021TCSVT} \protect\\ {PSNR: 14.56}\centering}
		\label{lai_text_01_kernel_02_wen}%文中引用该图片代号
	\end{minipage}
        \begin{minipage}{0.22\linewidth}
		\centering
		\includegraphics[width=1\linewidth]{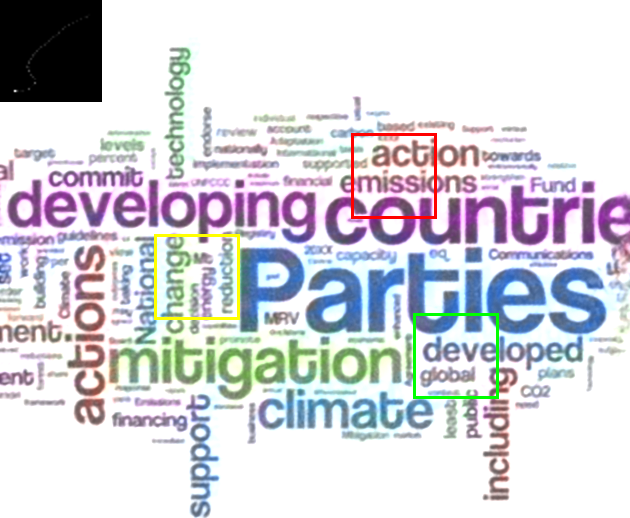}
		\vskip 4pt
		\includegraphics[width=1\linewidth]{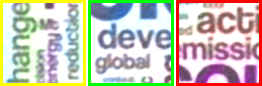}
		\caption*{(j)SelfDeblur\cite{ren2020neural} \protect\\ {PSNR: 22.45}\centering}
		\label{lai_text_01_kernel_02_SelfDeblur}%文中引用该图片代号
	\end{minipage}	 
	\begin{minipage}{0.22\linewidth}
		\centering
		\includegraphics[width=1\linewidth]{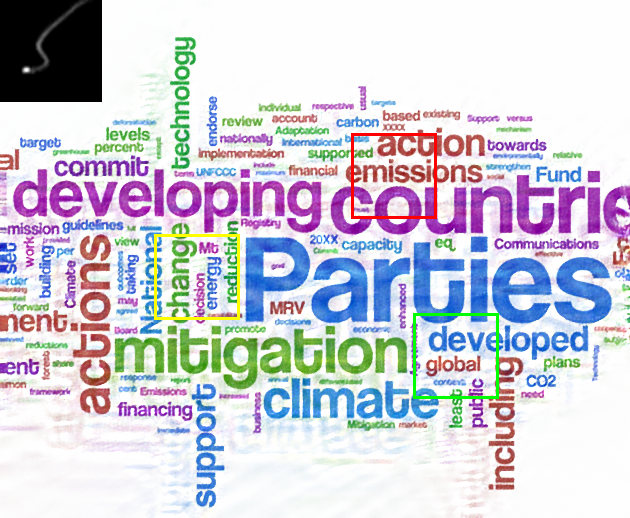}
		\vskip 4pt
		\includegraphics[width=1\linewidth]{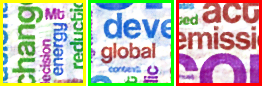}
		\caption*{(k)Fast-SelfDeblur\cite{bai2023fastselfdeblur} \protect\\ {PSNR: 25.20}\centering}
		\label{lai_text_01_kernel_02_Fast-SelfDeblur}%文中引用该图片代号
	\end{minipage}
	\begin{minipage}{0.22\linewidth}
		\centering
		\includegraphics[width=1\linewidth]{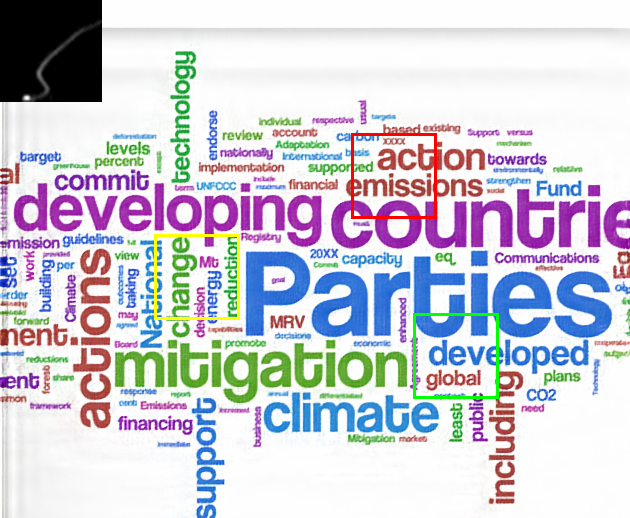}
		\vskip 4pt
		\includegraphics[width=1\linewidth]{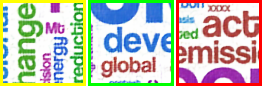}
		\caption*{(l)Self-MSNet (\textbf{Ours}) \protect\\ {PSNR: 28.03}\centering}
		\label{lai_text_01_kernel_02_ours}%文中引用该图片代号
	\end{minipage}
        %\vskip -8pt
	\caption{Visual comparison on an example image of the 'text' category with 2th blur kernel from Lai's dataset.}
	\label{fig:Lai-visual-text_01_kernel_02}
\end{figure*}

\begin{figure*}[!tbp]
	\centering
	\begin{minipage}{0.22\linewidth}
		\centering
		\includegraphics[width=1\linewidth]{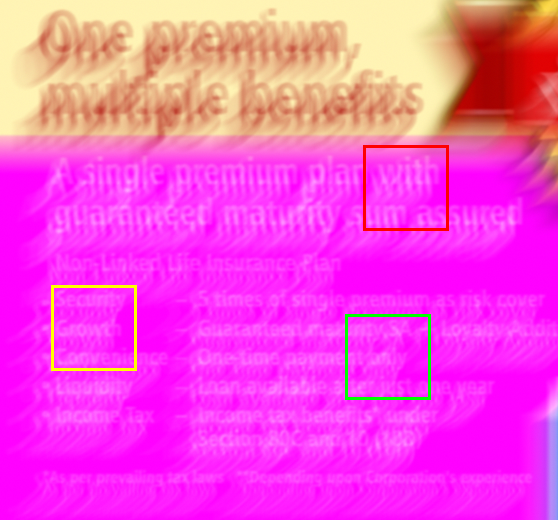}
		\vskip 4pt
		\includegraphics[width=1\linewidth]{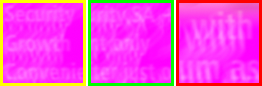}
		\caption*{(a)Blurred image \protect\\ {\textcolor{white}{***}}\centering}
		\label{lai_text_04_kernel_02_Blurry image}%文中引用该图片代号
	\end{minipage}
	\begin{minipage}{0.22\linewidth}
		\centering
		\includegraphics[width=1\linewidth]{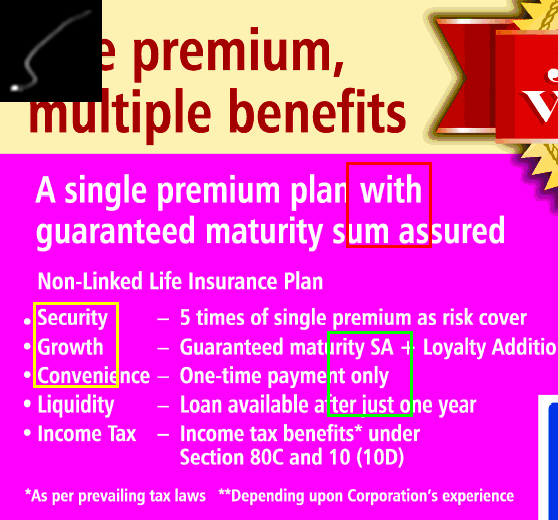}
		\vskip 4pt
		\includegraphics[width=1\linewidth]{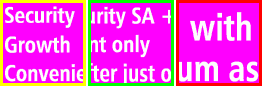}
		\caption*{(b)Ground-truth \protect\\ {\textcolor{white}{***}}\centering}
		\label{lai_text_04_kernel_02_Ground-truth}%文中引用该图片代号
	\end{minipage}
	\begin{minipage}{0.22\linewidth}
		\centering
		\includegraphics[width=1\linewidth]{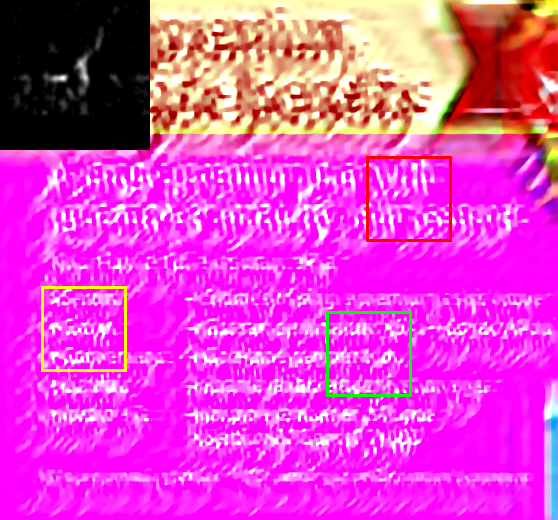}
		\vskip 4pt
		\includegraphics[width=1\linewidth]{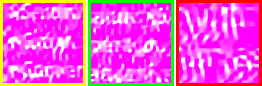}
		\caption*{(c)Cho and Lee\cite{cho2009fast} \protect\\ {PSNR: 13.48}\centering}
		\label{lai_text_04_kernel_02_cho}%文中引用该图片代号
	\end{minipage}
        \begin{minipage}{0.22\linewidth}
		\centering
		\includegraphics[width=1\linewidth]{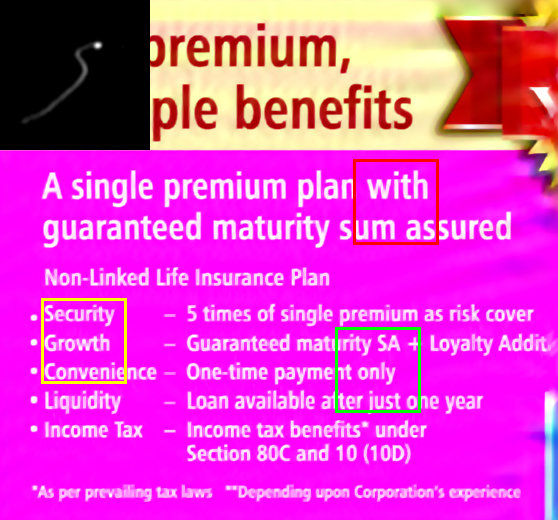}
		\vskip 4pt
		\includegraphics[width=1\linewidth]{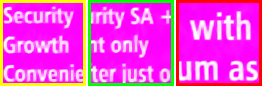}
		\caption*{(d)Xu and Jia\cite{xu2010two} \protect\\ {PSNR: 17.48}\centering}
		\label{lai_text_04_kernel_02_xuandjia}%文中引用该图片代号
	\end{minipage}
        \vskip 7pt
        \begin{minipage}{0.22\linewidth}
		\centering
		\includegraphics[width=1\linewidth]{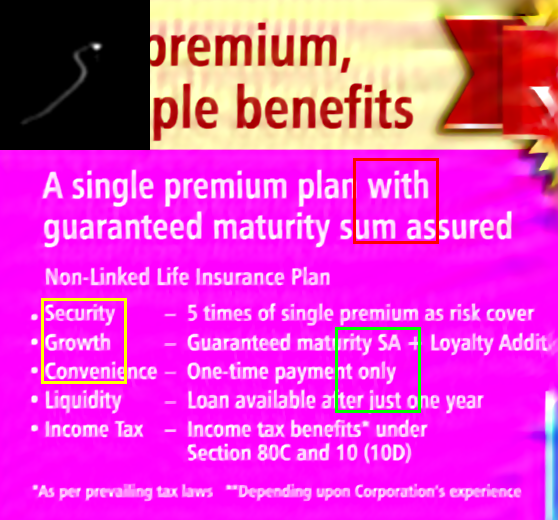}
		\vskip 4pt
		\includegraphics[width=1\linewidth]{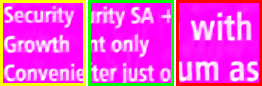}
		\caption*{(e)Xu et al.\cite{xu2013unnatural} \protect\\ {PSNR: 16.18}\centering}
		\label{lai_text_04_kernel_02_xuunnatural}%文中引用该图片代号
	\end{minipage} 
        \begin{minipage}{0.22\linewidth}
		\centering
		\includegraphics[width=1\linewidth]{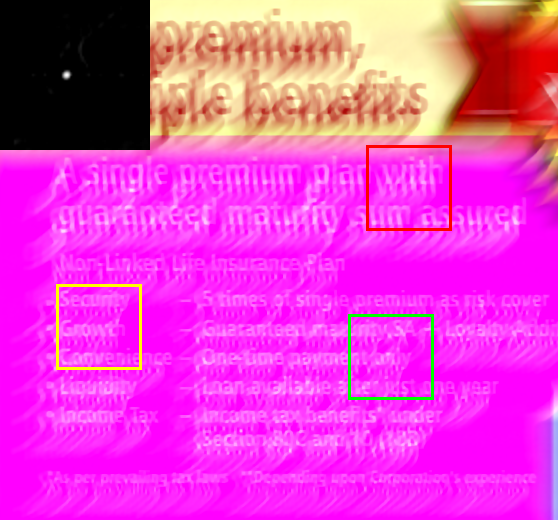}
		\vskip 4pt
		\includegraphics[width=1\linewidth]{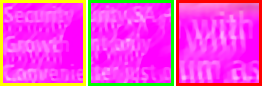}
		\caption*{(f)Michaeli and Irani\cite{michaeli2014blind} \protect\\ {PSNR: 14.39}\centering}
		\label{lai_text_04_kernel_02_Michaeli}%文中引用该图片代号
	\end{minipage} 
	\begin{minipage}{0.22\linewidth}
		\centering
		\includegraphics[width=1\linewidth]{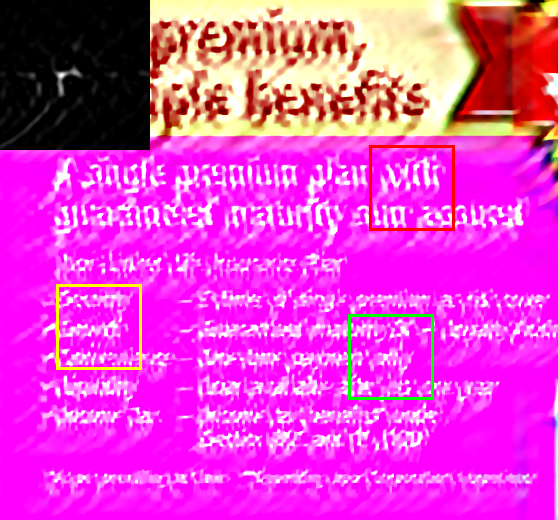}
		\vskip 4pt
		\includegraphics[width=1\linewidth]{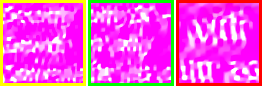}
		\caption*{(g)Perrone and Favaro\cite{perrone2014total} \protect\\ {PSNR: 12.37}\centering}
		\label{lai_text_04_kernel_02_Perrone}%文中引用该图片代号
	\end{minipage}
	\begin{minipage}{0.22\linewidth}
		\centering
		\includegraphics[width=1\linewidth]{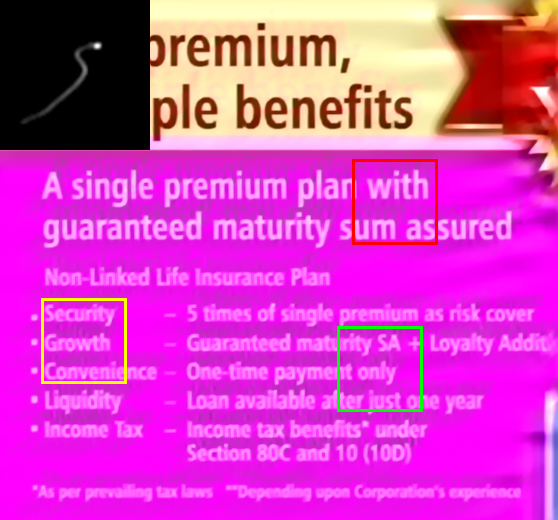}
		\vskip 4pt
		\includegraphics[width=1\linewidth]{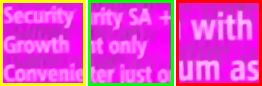}
		\caption*{(h)Pan et al.\cite{pan2018PAMIdarkchannel} \protect\\ {PSNR: 17.46}\centering}
		\label{lai_text_04_kernel_02_Pan}%文中引用该图片代号
	\end{minipage}
        \vskip 7pt
        \begin{minipage}{0.22\linewidth}
		\centering
		\includegraphics[width=1\linewidth]{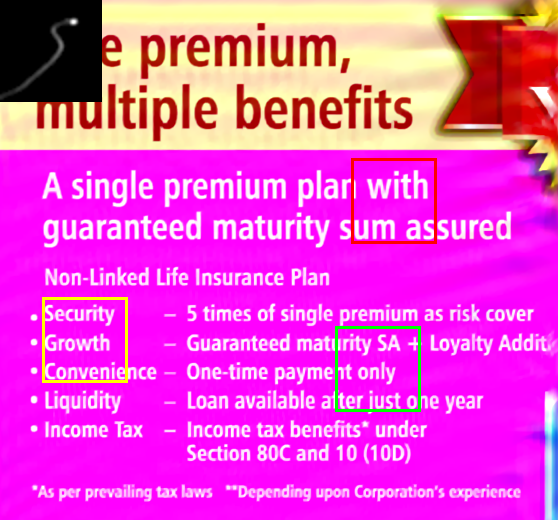}
		\vskip 4pt
		\includegraphics[width=1\linewidth]{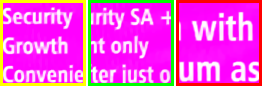}
		\caption*{(i)Wen et al.\cite{wen2021TCSVT} \protect\\ {PSNR: 25.94}\centering}
		\label{lai_text_04_kernel_02_wen}%文中引用该图片代号
	\end{minipage}
        \begin{minipage}{0.22\linewidth}
		\centering
		\includegraphics[width=1\linewidth]{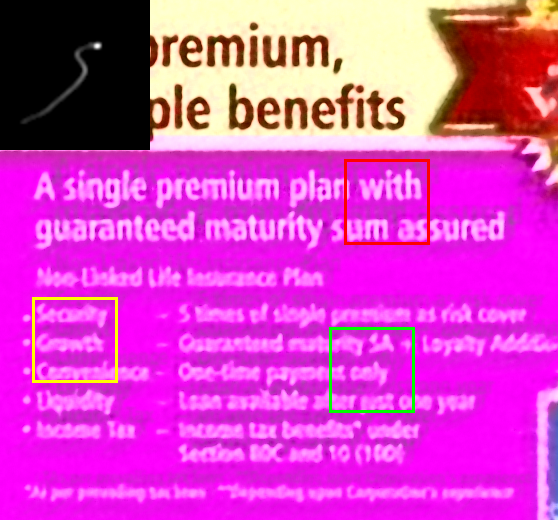}
		\vskip 4pt
		\includegraphics[width=1\linewidth]{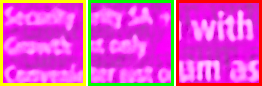}
		\caption*{(j)SelfDeblur\cite{ren2020neural} \protect\\ {PSNR: 20.23}\centering}
		\label{lai_text_04_kernel_02_SelfDeblur}%文中引用该图片代号
	\end{minipage}	 
	\begin{minipage}{0.22\linewidth}
		\centering
		\includegraphics[width=1\linewidth]{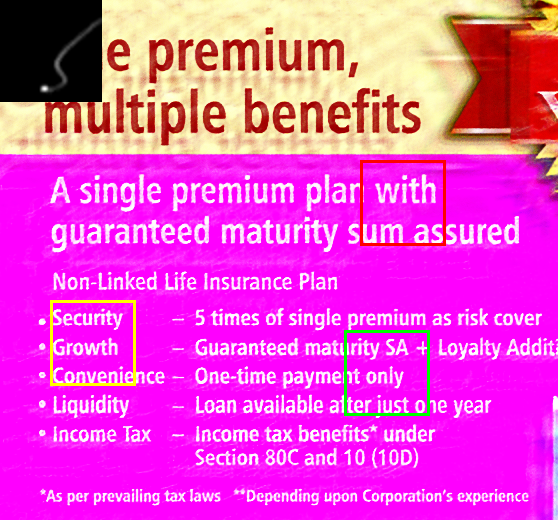}
		\vskip 4pt
		\includegraphics[width=1\linewidth]{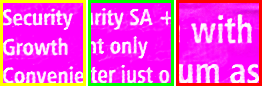}
		\caption*{(k)Fast-SelfDeblur\cite{bai2023fastselfdeblur} \protect\\ {PSNR: 14.48}\centering}
		\label{lai_text_04_kernel_02_Fast-SelfDeblur}%文中引用该图片代号
	\end{minipage}
	\begin{minipage}{0.22\linewidth}
		\centering
		\includegraphics[width=1\linewidth]{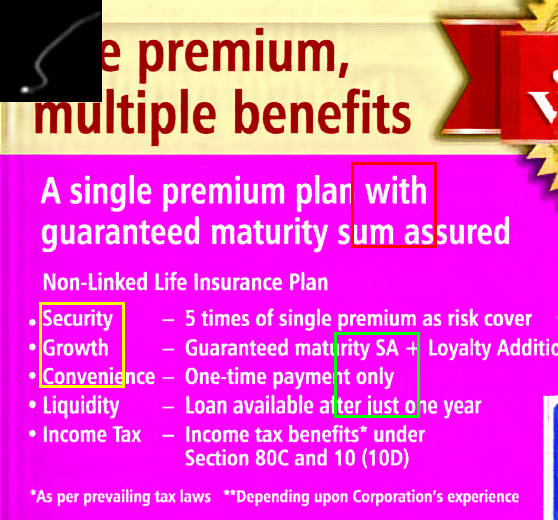}
		\vskip 4pt
		\includegraphics[width=1\linewidth]{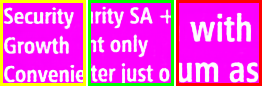}
		\caption*{(l)Self-MSNet (\textbf{Ours}) \protect\\ {PSNR: 30.01}\centering}
		\label{lai_text_04_kernel_02_ours}%文中引用该图片代号
	\end{minipage}
        %\vskip -8pt
	\caption{Visual comparison on another example image of the 'text' category with 2th blur kernel from Lai's dataset.}
        \label{fig:Lai-visual-text_04_kernel_02}
\end{figure*}

%\subsection{Qualitative results - Lai}
In addition, we provide visual comparisons of various state-of-the-art methods on Lai et al.'s dataset. The example images cover five categories (man-made, natural, people/face, saturated, and text) and different types of kernels. The deblurred results are illustrated in Figs.\ref{fig:Lai-visual-manmade_01_kernel_04}-\ref{fig:Lai-visual-text_04_kernel_02}, where the blur kernels estimated by various methods are displayed in the top left corner, and the close-ups of different image regions are shown at the bottom. The proposed method demonstrates the ability to accurately estimate the blur kernels, whereas other methods fail to reliably estimate the kernels in one or more cases. Moreover, our method can restore detailed textures and edge information of the image, as demonstrated by examples such as the leaf in the green box of Fig.\ref{fig:Lai-visual-natural_04_kernel_02} and the windows in the red box of Fig.\ref{fig:Lai-visual-saturated_05_kernel_03}. As shown in Fig.\ref{fig:Lai-visual-people_05_kernel_04}, the deblurred images by SelfDeblur \cite{ren2020neural} and Fast-SelfDeblur \cite{bai2023fastselfdeblur} exhibit significant ringing artifacts, whereas our Self-MSNet obtains an accurate blur kernel leading to a clearer result with fewer ringing artifacts. The superior visual performance further validates the efficacy of the proposed method.

\subsection{Experimental Results on Kohler's Dataset}
\begin{figure*}[!tbp]
    \centering
    \includegraphics[width=0.98\linewidth]{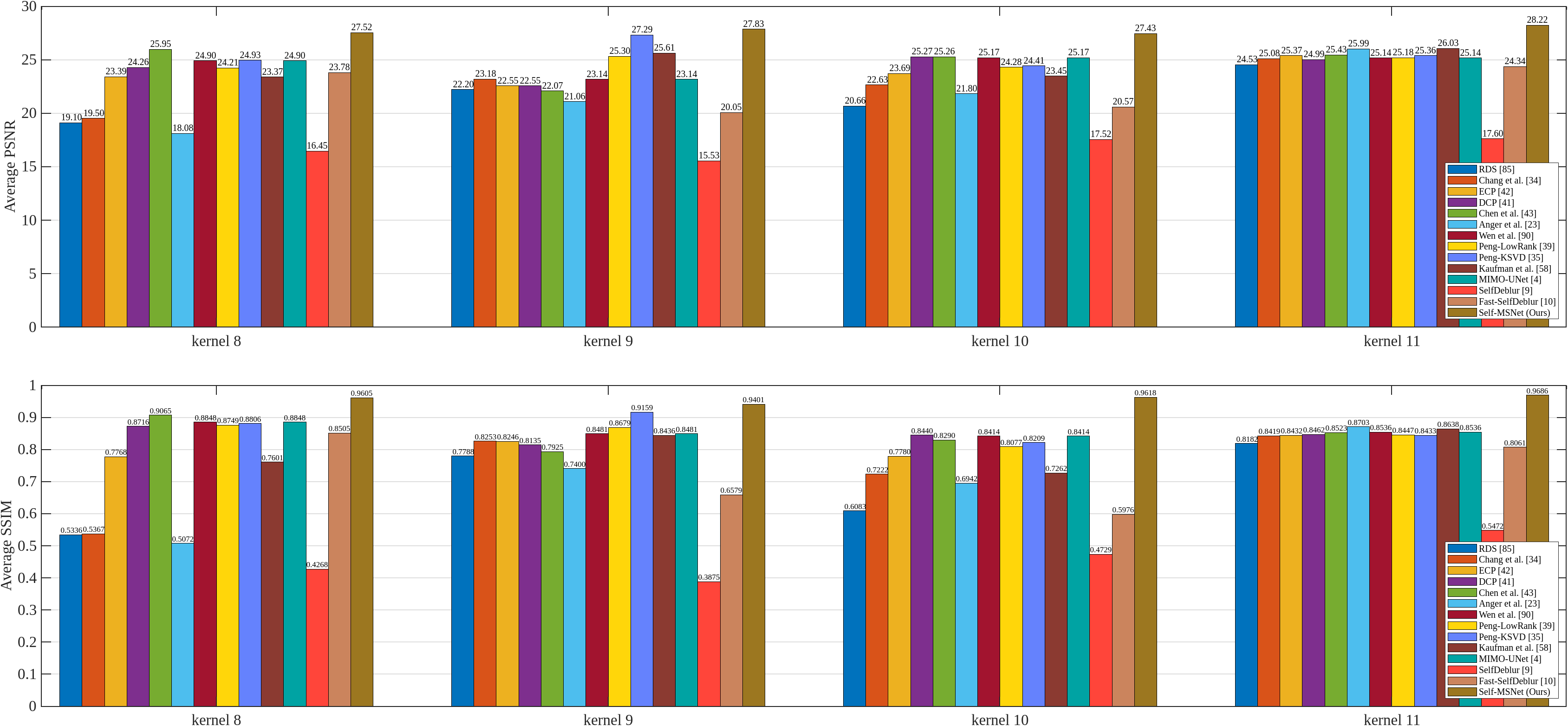}
    \caption{Quantitative comparisons of four large blur kernels (8th-11th) on Kohler's dataset.}
    \label{fig:kohler-bar-ker8-11}
\end{figure*}

The dataset, provided by Kohler et al. \cite{Kohler2012}, consists of 48 real blurred images. The image blurring is realized by recording camera motion trajectories for 12 blur kernels (where the 8th-11th kernels are large-sized), and then applying these trajectories on 4 pictures hanging on the wall using a robotic platform with camera motion control. Generated from 3D motion trajectories, the blur is no longer uniform across the image. The kernel size ranges from $41 \times 41$ to $151 \times 151$ pixels. The parameters are set as follows: the maximum allowed number of iterations is $K=2000$, the learning rate is $\eta = 0.01$, the regularization weights are set to $\lambda_{h} = 120$, $\lambda_{x} = 1$ and $\gamma = 10$ respectively.
We compare the proposed Self-MSNet with the mathematical optimization-based methods \cite{fergus2006removing, shan2008highquality, cho2009fast, xu2010two, whyte2011shaken, krishnan2011normalized, Hirsch2011fastremoval, Goldstein2012ECCVspectralirregularities, xu2013unnatural, whyte2014IJCVshaken, RDS2014, yue2014ECCVhybrid, gong2016CVPRautomatic, czc2017sparse, yan2017extremechannel, pan2018PAMIdarkchannel, chen2019blind, Anger2019L0, bai2019TIPgraph, chen2020enhanced, chen2020OID, shao2020gradient, peng2021ksvd, wen2021TCSVT, peng2022lowrank, shao2023Revisiting}, the supervised learning-based methods \cite{sun2015CVPRlearning, Ramakrishnan2017ICCVWGFMD, gong2017CVPRfromMBMF, nah2017deep, tao2018scale, kupyn2018deblurgan, kupyn2019deblurganv2, kaufman2020deblurring, cho2021rethinking}, and the self-supervised learning-based methods \cite{ren2020neural, bai2023fastselfdeblur, li2023deepEM}. The quantitative comparison results of these methods, including the average PSNR and SSIM scores on Kohler's dataset, are summarized in Tab.\ref{tab:Kohler-ave-PSNRssim}. The results of \cite{kaufman2020deblurring, cho2021rethinking,krishnan2011normalized, RDS2014,Anger2019L0,ren2020neural} are generated using their released codes, those of \cite{gong2016CVPRautomatic, czc2017sparse, yan2017extremechannel, pan2018PAMIdarkchannel,chen2019blind,shao2020gradient, peng2021ksvd, wen2021TCSVT, peng2022lowrank, shao2023Revisiting,Ramakrishnan2017ICCVWGFMD,kupyn2019deblurganv2,bai2023fastselfdeblur, li2023deepEM} are quoted from their original papers, and the rest \cite{fergus2006removing, shan2008highquality, cho2009fast, xu2010two, whyte2011shaken, Hirsch2011fastremoval, Goldstein2012ECCVspectralirregularities, xu2013unnatural, whyte2014IJCVshaken,yue2014ECCVhybrid,bai2019TIPgraph, chen2020enhanced, chen2020OID,sun2015CVPRlearning,gong2017CVPRfromMBMF, nah2017deep, tao2018scale, kupyn2018deblurgan} are quoted from other literature. \underline{Underlined} refers to the suboptimal result. As we can see, our Self-MSNet demonstrates superior performance to all the aforementioned supervised learning-based methods and all the self-supervised learning-based methods mentioned above. In the scenarios involving large and real-world blurs, deep learning-based methods tend to underperform mathematical optimization-based image deblurring methods. However, the proposed Self-MSNet achieves superior results to all of mathematical optimization-based methods, except that it falls behind \cite{chen2019blind} only by 0.02dB in terms of PSNR. Figure \ref{fig:kohler-bar-ker8-11} shows the evaluation result comparisons among various methods in the cases of four large-sized blur kernels (8th-11th) from Kohler's dataset \cite{Kohler2012}. The proposed Self-MSNet outperforms those competing methods in terms of both PSNR and SSIM, which confirms the efficacy of our method in handling large-scale blur kernels.

\begin{table}[h!]
    \caption{Average PSNR/SSIM Comparisons with State-of-the-art Methods on Kohler's Dataset.}
    \label{tab:Kohler-ave-PSNRssim}
    \centering
    \begin{tabular}{p{3.3cm}p{1.3cm}<{\centering}p{1.3cm}<{\centering}}
	\hline%\toprule
	Method    &PSNR(dB)   &SSIM \\
	\hline%\midrule
        Fergus et al.${}^\triangle$ \cite{fergus2006removing}    &22.73    &0.6820  \\      
        Shan et al.${}^\triangle$ \cite{shan2008highquality}    &25.89    &0.8420  \\       
        Cho and Lee${}^\triangle$ \cite{cho2009fast}    &28.98    &0.9330 \\                
        Xu and Jia${}^\triangle$ \cite{xu2010two}    &29.53    &0.9440 \\                   
        Whyte et al.${}^\triangle$ \cite{whyte2011shaken}    &28.07    &0.8480 \\           
        Krishnan et al.${}^\triangle$ \cite{krishnan2011normalized}    &24.23    &0.8132 \\ 
        Hirsch et al.${}^\triangle$ \cite{Hirsch2011fastremoval}    &27.77    &0.8520   \\  
        Goldstein et al.${}^\triangle$ \cite{Goldstein2012ECCVspectralirregularities} &23.74 &0.6917  \\
	Xu et al.${}^\triangle$ \cite{xu2013unnatural} &27.47	&0.7506 \\                  
        Whyte et al.${}^\triangle$ \cite{whyte2014IJCVshaken} &27.03	&0.7467  \\         
	RDS${}^\triangle$ \cite{RDS2014} &27.76	&0.8629 \\                                  
	Yue et al.${}^\triangle$ \cite{yue2014ECCVhybrid} &28.12	&0.8484 \\              
	Gong et al.${}^\triangle$ \cite{gong2016CVPRautomatic} &28.75	&0.8768 \\          
        Chang et al.${}^\triangle$ \cite{czc2017sparse}    &27.48    &0.8863  \\            
        ECP${}^\triangle$ \cite{yan2017extremechannel}    &28.81    &0.9129  \\             
        DCP${}^\triangle$ \cite{pan2018PAMIdarkchannel}    &29.95    &0.9337  \\
        Chen et al.${}^\triangle$ \cite{chen2019blind}    &\textbf{30.38}    &0.9343\\            
        Anger et al.${}^\triangle$ \cite{Anger2019L0}    &28.86    &0.8848  \\             
        Bai et al.${}^\triangle$ \cite{bai2019TIPgraph}    &26.11    &0.8111  \\            
        ESM${}^\triangle$ \cite{chen2020enhanced}    &27.44    &0.8917 \\                   
        OID${}^\triangle$ \cite{chen2020OID}    &26.88    &0.8558  \\ 
        Shao et al.${}^\triangle$ \cite{shao2020gradient} &27.38	&0.8942\\                      
        Peng-KSVD${}^\triangle$ \cite{peng2021ksvd}    &29.52    &0.9370\\
        Wen et al.${}^\triangle$ \cite{wen2021TCSVT}    &29.98    &0.9392\\
        Peng-LowRank${}^\triangle$ \cite{peng2022lowrank}    &29.18    &0.9251\\
        RDP${}^\triangle$ \cite{shao2023Revisiting}    &27.44    &0.8941\\
        Sun et al.${}^{\dag}$ \cite{sun2015CVPRlearning}    &25.12    &0.7281\\
        GFMD${}^{\dag}$ \cite{Ramakrishnan2017ICCVWGFMD} &27.08	&0.7510\\
        MBMF${}^{\dag}$ \cite{gong2017CVPRfromMBMF} &26.59	&0.7418\\
        DeepDeblur${}^{\dag}$ \cite{nah2017deep}    &26.48    &0.8070\\
        SRN-DeblurNet${}^{\dag}$ \cite{tao2018scale}    &26.75    &0.8370\\
        DeblurGAN${}^{\dag}$ \cite{kupyn2018deblurgan}    &26.10    &0.8160\\
        DeblurGAN-v2${}^{\dag}$ \cite{kupyn2019deblurganv2}    &26.72    &0.8360\\
        Kaufman et al.${}^{\dag}$\cite{kaufman2020deblurring}    &30.17    &0.9164\\
        MIMO-UNet${}^{\dag}$\cite{cho2021rethinking}    &25.34    &0.7917\\
        SelfDeblur \cite{ren2020neural}    &20.26    &0.6609\\
        Fast-SelfDeblur \cite{bai2023fastselfdeblur}    &27.92    &0.8739\\
        Li et al.\cite{li2023deepEM}    &30.26    &0.9400\\
        Self-MSNet-S &28.13	&\underline{0.9517}\\
        Self-MSNet ({\textbf{Ours}})    &\underline{30.36}    &\textbf{0.9660}\\
    \hline%\bottomrule
    \end{tabular}
\end{table}

\begin{figure*}[!tbp]
    \centering
    \begin{minipage}{0.16\linewidth}
		\centering
		\includegraphics[width=1\linewidth]{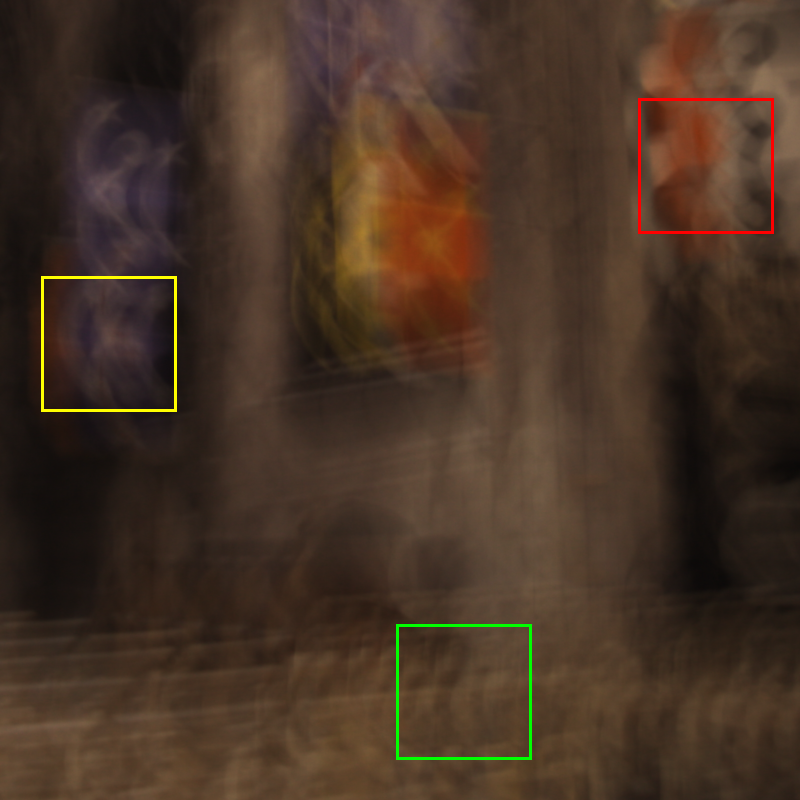}
		\vskip 2pt
		\includegraphics[width=1\linewidth]{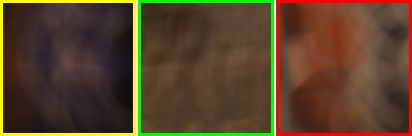}
        %\vskip -5pt
        \caption*{}
    \end{minipage}
    \begin{minipage}{0.16\linewidth}
		\centering
		\includegraphics[width=1\linewidth]{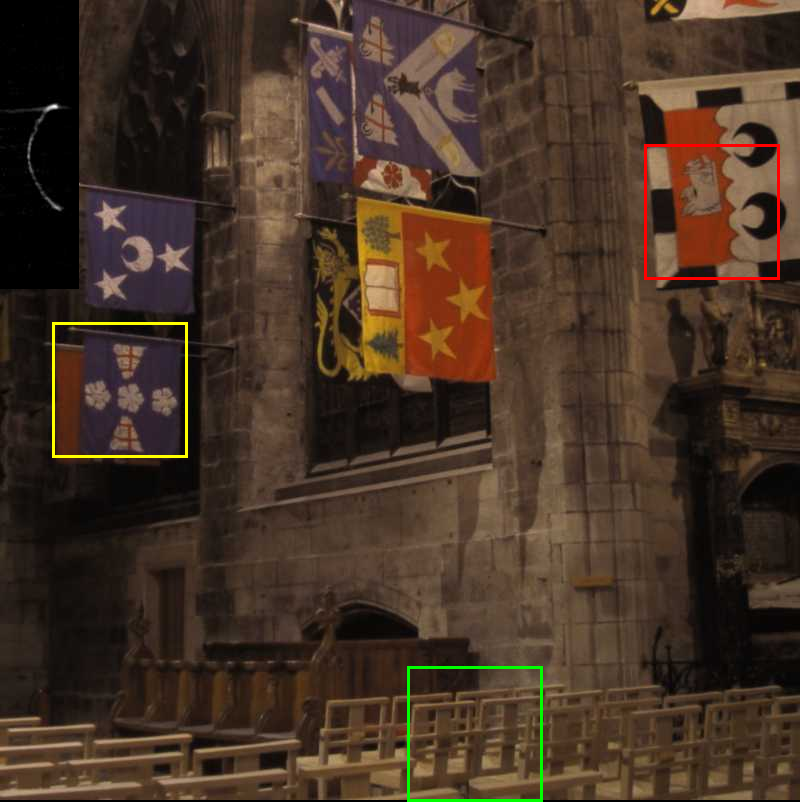}
		\vskip 2pt
		\includegraphics[width=1\linewidth]{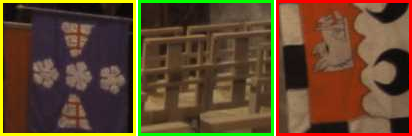}
        %\vskip -5pt
        \caption*{}
    \end{minipage}
    \begin{minipage}{0.16\linewidth}
		\centering
		\includegraphics[width=1\linewidth]{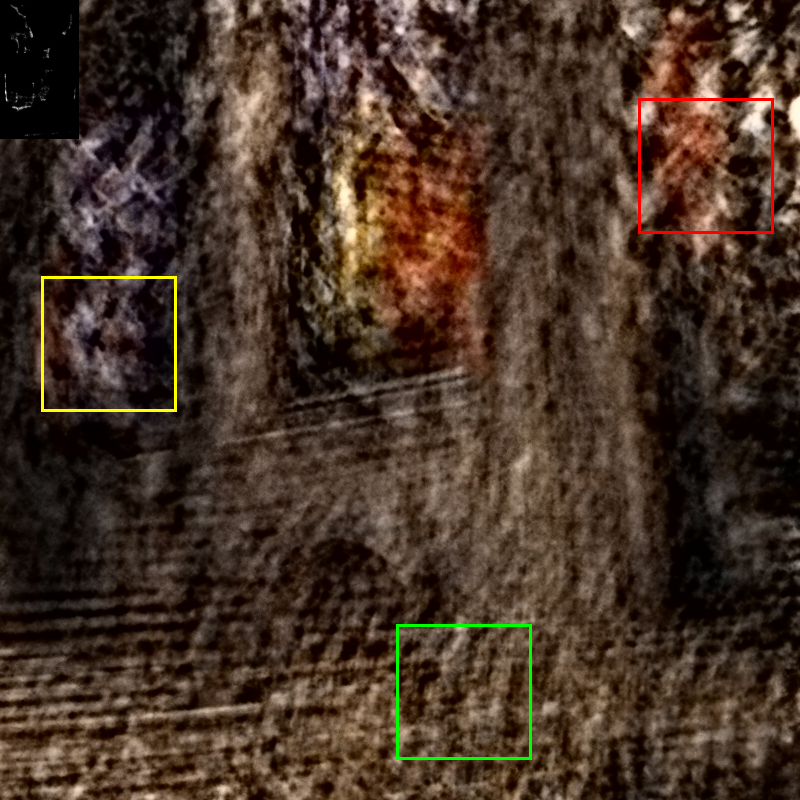}
		\vskip 2pt
		\includegraphics[width=1\linewidth]{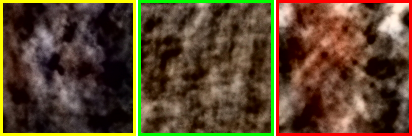}
        %\vskip -5pt
        \caption*{PSNR: 19.38}
    \end{minipage}
    \begin{minipage}{0.16\linewidth}
		\centering
		\includegraphics[width=1\linewidth]{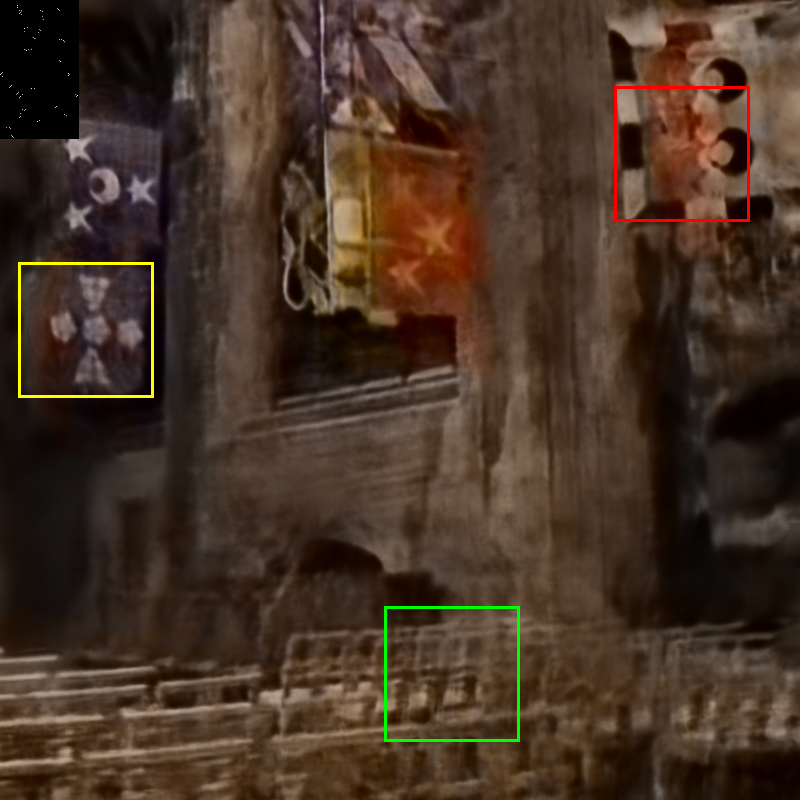}
		\vskip 2pt
		\includegraphics[width=1\linewidth]{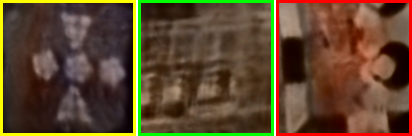}
        %\vskip -5pt
        \caption*{PSNR: 25.69}
    \end{minipage}
    \begin{minipage}{0.16\linewidth}
		\centering
		\includegraphics[width=1\linewidth]{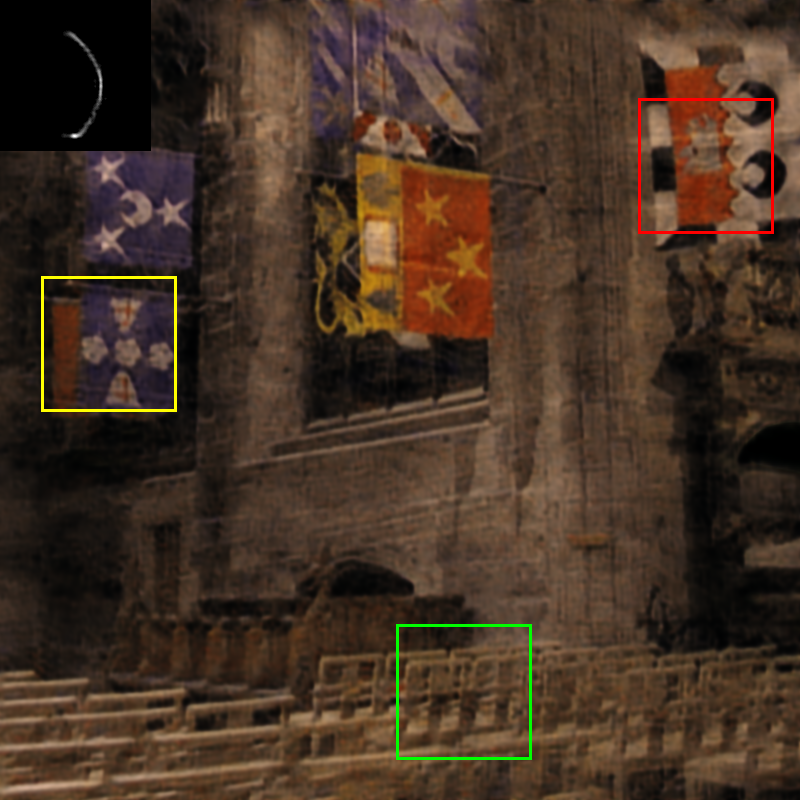}
		\vskip 2pt
		\includegraphics[width=1\linewidth]{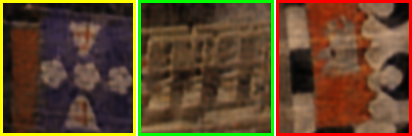}
        %\vskip -5pt
        \caption*{PSNR: 27.64}
    \end{minipage}
    \vskip 3pt
    \begin{minipage}{0.16\linewidth}
		\centering
		\includegraphics[width=1\linewidth]{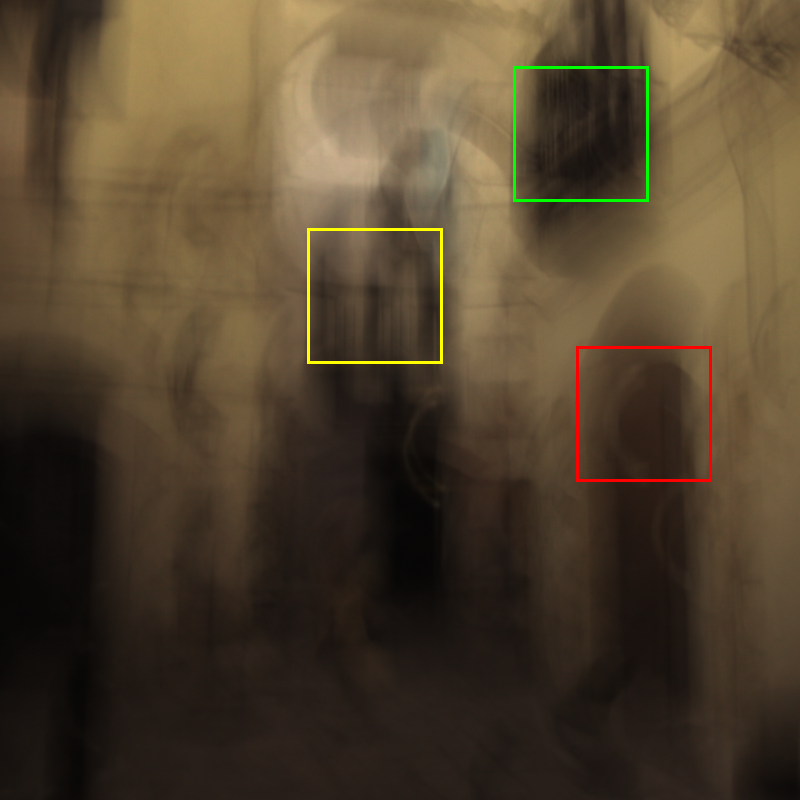}
		\vskip 2pt
		\includegraphics[width=1\linewidth]{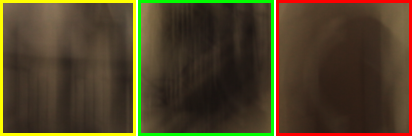}
        %\vskip -5pt
        \caption*{}
        %\vskip -10pt
        \caption*{(a)Blurred image\centering} 
	\label{Blurry image}%文中引用该图片代号
    \end{minipage}
    \begin{minipage}{0.16\linewidth}
		\centering
		\includegraphics[width=1\linewidth]{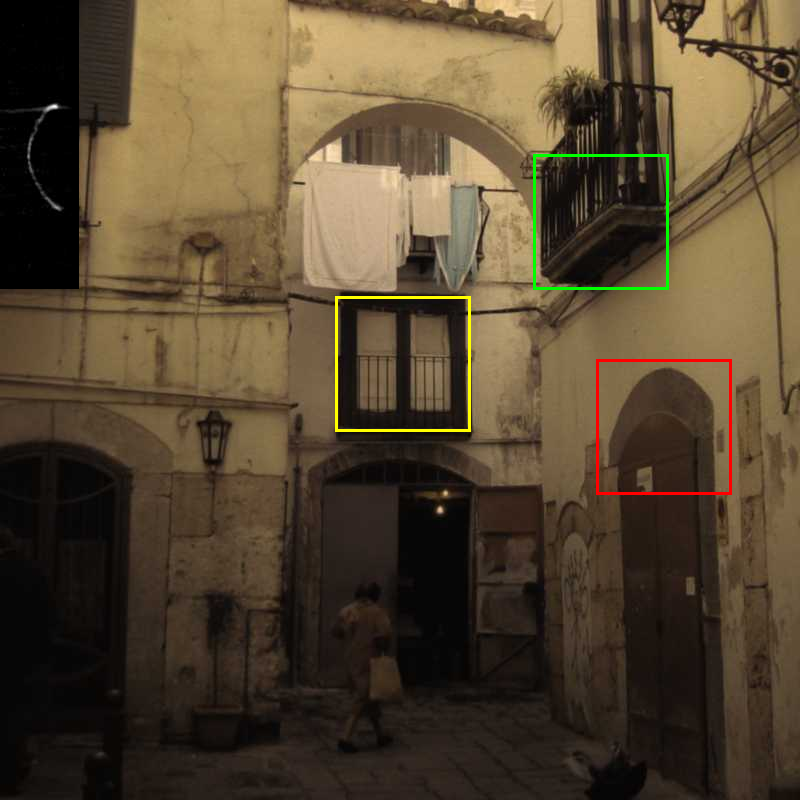}
		\vskip 2pt
		\includegraphics[width=1\linewidth]{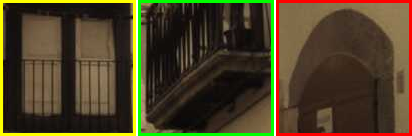}
        %\vskip -5pt
        \caption*{}
        %\vskip -10pt
        \caption*{(b)Ground-truth\centering}
        \label{Ground-truth}%文中引用该图片代号
    \end{minipage}
    \begin{minipage}{0.16\linewidth}
		\centering
		\includegraphics[width=1\linewidth]{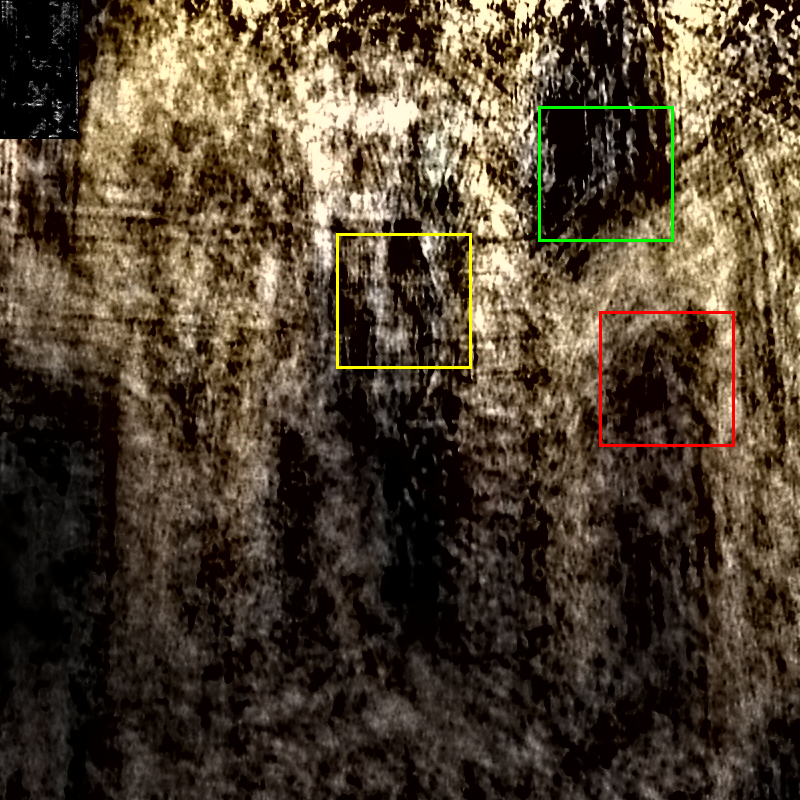}
		\vskip 2pt
		\includegraphics[width=1\linewidth]{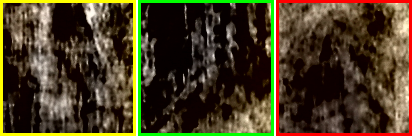}
        %\vskip -5pt
        \caption*{PSNR: 16.45}
        %\vskip -10pt
        \caption*{(c)SelfDeblur\cite{ren2020neural}\centering} %SelfDeblur
	\label{SelfDeblur}%文中引用该图片代号
    \end{minipage}
    \begin{minipage}{0.16\linewidth}
		\centering
		\includegraphics[width=1\linewidth]{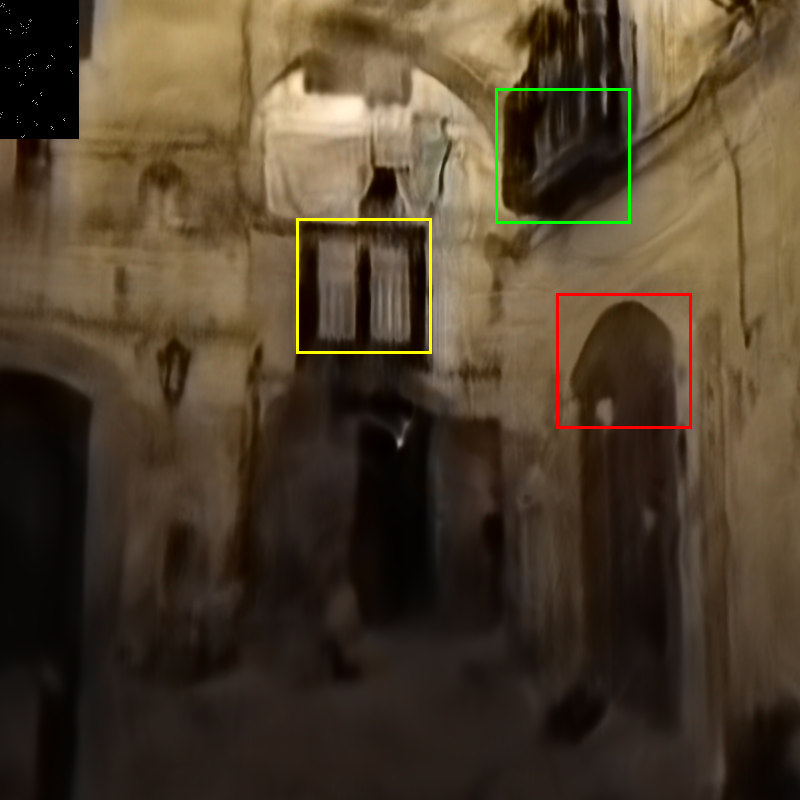}
		\vskip 2pt
		\includegraphics[width=1\linewidth]{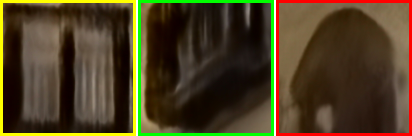}
        %\vskip -5pt
        \caption*{PSNR: 25.76}
        %\vskip -10pt
        \caption*{\leftline{(d)Fast-SelfDeblur\cite{bai2023fastselfdeblur}}} %Fast-SelfDeblur
	\label{Fast-SelfDeblur}%文中引用该图片代号
    \end{minipage}
    \begin{minipage}{0.16\linewidth}
		\centering
		\includegraphics[width=1\linewidth]{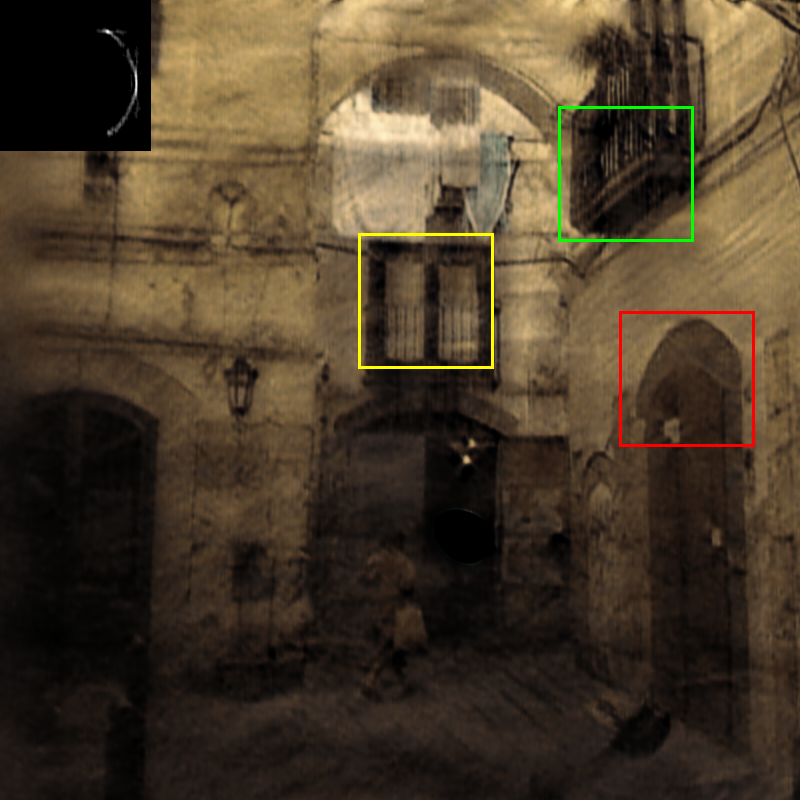}
		\vskip 2pt
		\includegraphics[width=1\linewidth]{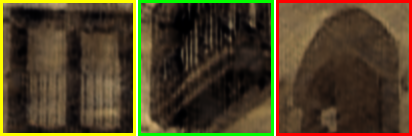}
        %\vskip -5pt
        \caption*{PSNR: 26.01}
        %\vskip -10pt
        \caption*{(e){\textbf{Ours}}}
	\label{Ours}%文中引用该图片代号
    \end{minipage}
    %\vskip -8pt
    \caption{Visual comparison on example images with 8th blur kernel from Kohler's dataset.}
    \label{fig:Kohler-visual-ker8}
\end{figure*}

\begin{figure*}[htbp]
    \centering
    \begin{minipage}{0.16\linewidth}
		\centering
		\includegraphics[width=1\linewidth]{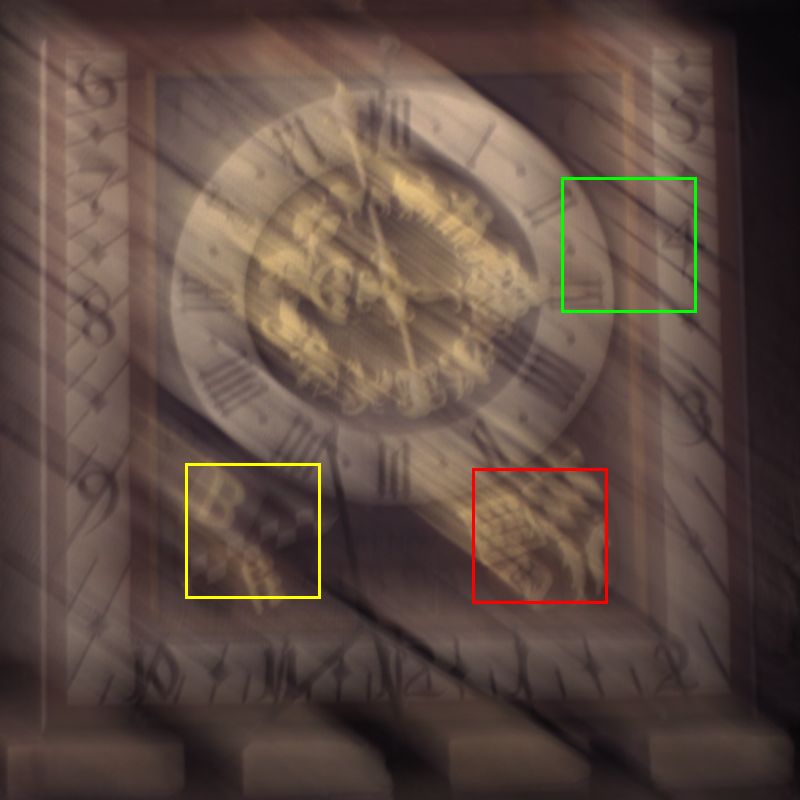}
		\vskip 2pt
		\includegraphics[width=1\linewidth]{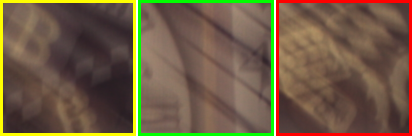}
        %\vskip -5pt
        \caption*{}
    \end{minipage}
    \begin{minipage}{0.16\linewidth}
		\centering
		\includegraphics[width=1\linewidth]{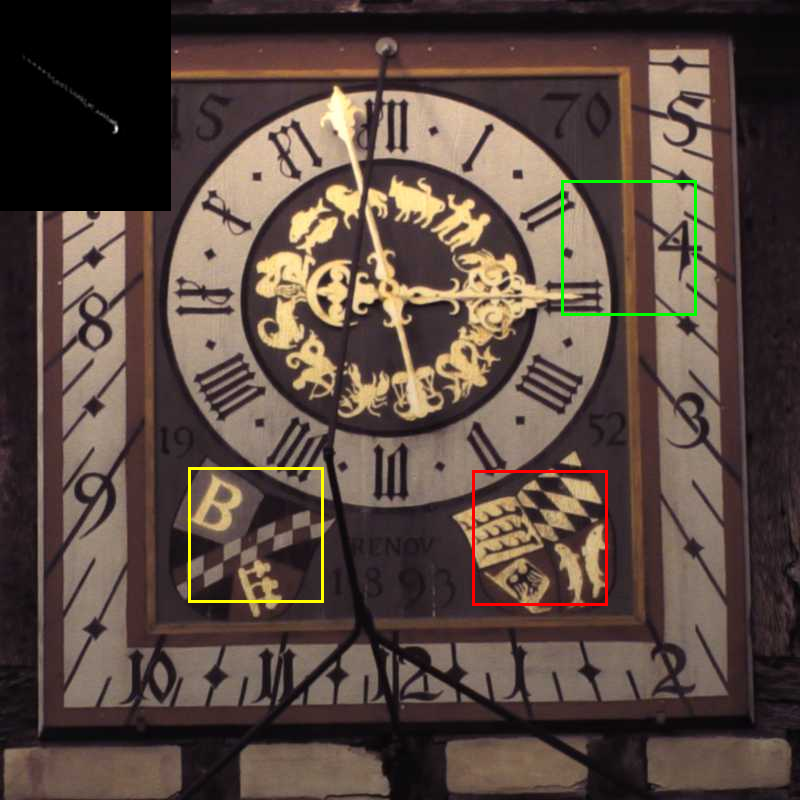}
		\vskip 2pt
		\includegraphics[width=1\linewidth]{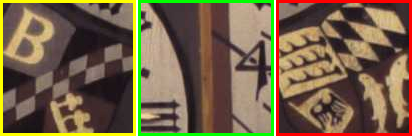}
        %\vskip -5pt
        \caption*{}
    \end{minipage}
    \begin{minipage}{0.16\linewidth}
		\centering
		\includegraphics[width=1\linewidth]{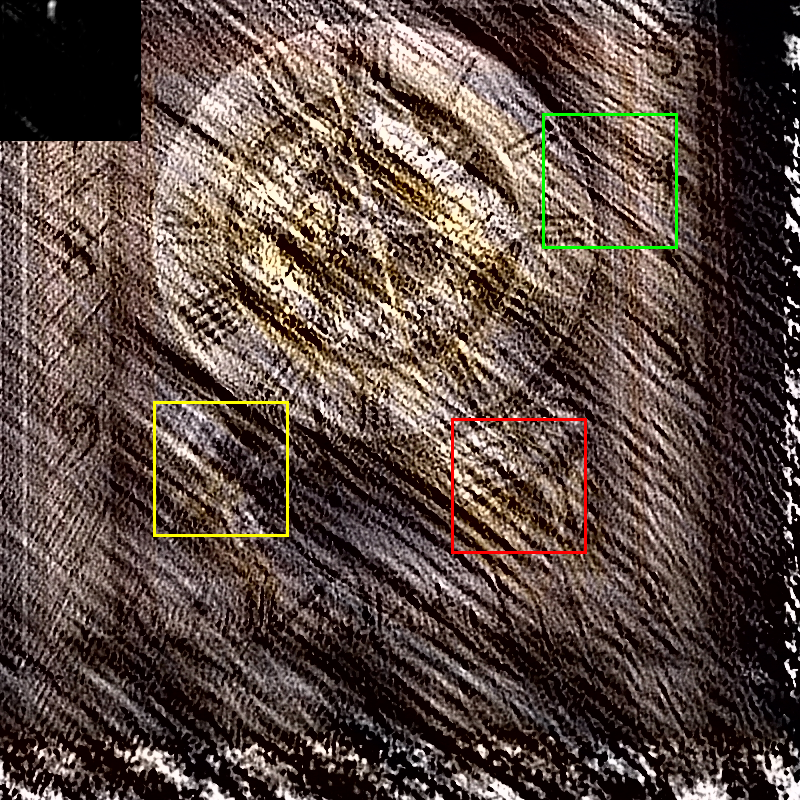}
		\vskip 2pt
		\includegraphics[width=1\linewidth]{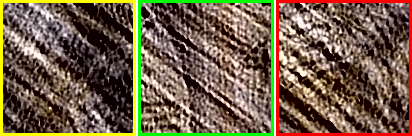}
        %\vskip -5pt
        \caption*{PSNR: 13.37}
    \end{minipage}
    \begin{minipage}{0.16\linewidth}
		\centering
		\includegraphics[width=1\linewidth]{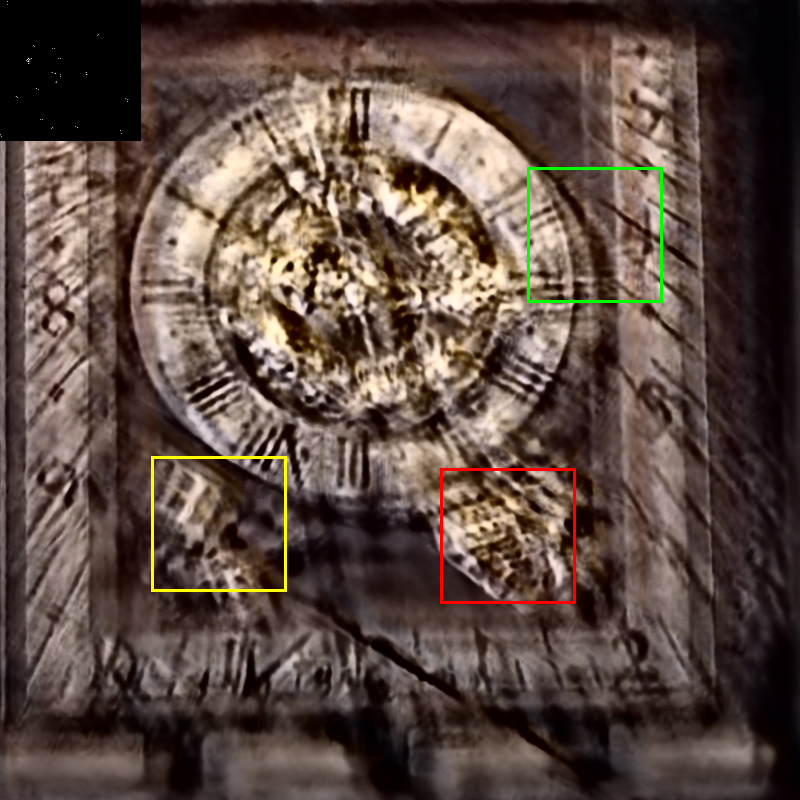}
		\vskip 2pt
		\includegraphics[width=1\linewidth]{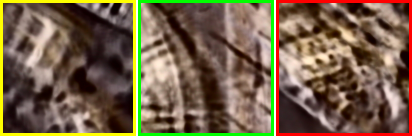}
        %\vskip -5pt
        \caption*{PSNR: 17.17}
    \end{minipage}
    \begin{minipage}{0.16\linewidth}
		\centering
		\includegraphics[width=1\linewidth]{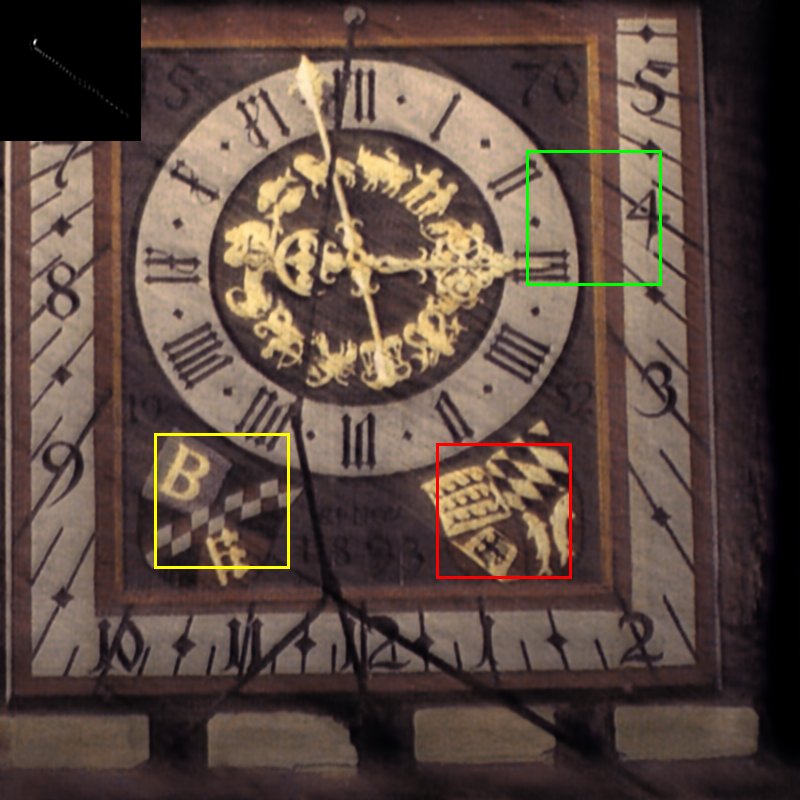}
		\vskip 2pt
		\includegraphics[width=1\linewidth]{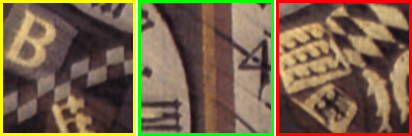}
        %\vskip -5pt
        \caption*{PSNR: 24.21}
    \end{minipage}
    \vskip 3pt
    \begin{minipage}{0.16\linewidth}
		\centering
		\includegraphics[width=1\linewidth]{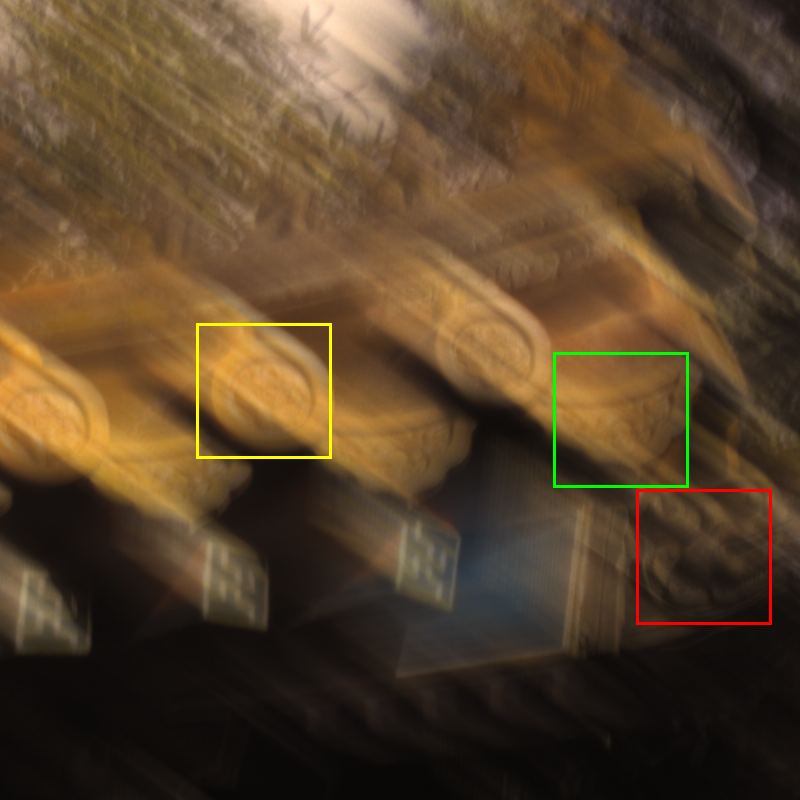}
		\vskip 2pt
		\includegraphics[width=1\linewidth]{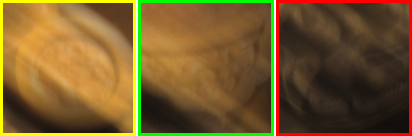}
        %\vskip -5pt
        \caption*{}
        %\vskip -10pt
        \caption*{(a)Blurred image\centering} 
	\label{Blurry image}%文中引用该图片代号
    \end{minipage}
    \begin{minipage}{0.16\linewidth}
		\centering
		\includegraphics[width=1\linewidth]{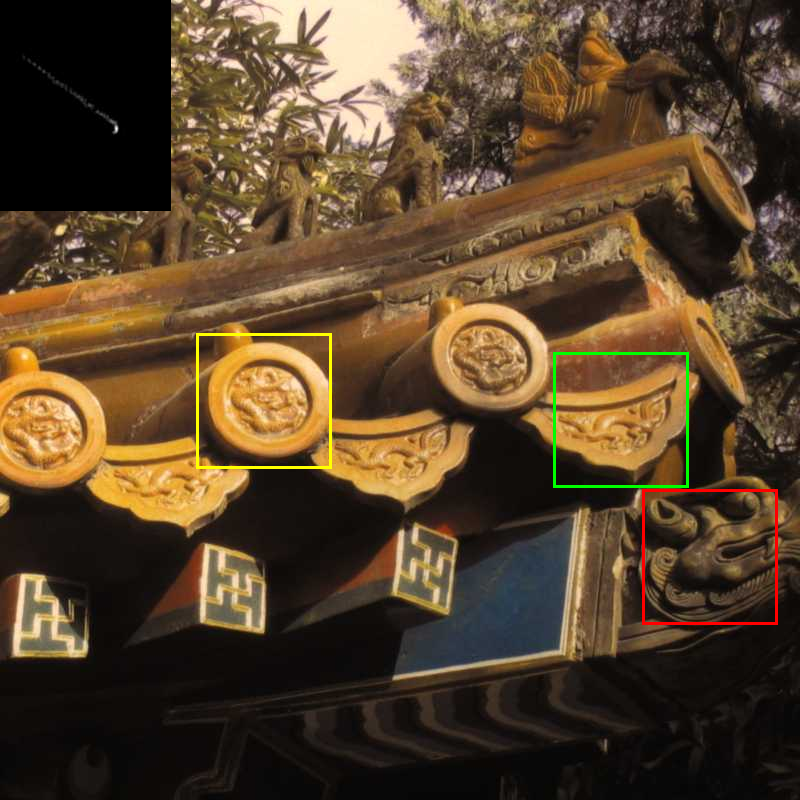}
		\vskip 2pt
		\includegraphics[width=1\linewidth]{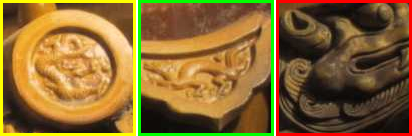}
        %\vskip -5pt
        \caption*{}
        %\vskip -10pt
        \caption*{(b)Ground-truth\centering}
        \label{Ground-truth}%文中引用该图片代号
    \end{minipage}
    \begin{minipage}{0.16\linewidth}
		\centering
		\includegraphics[width=1\linewidth]{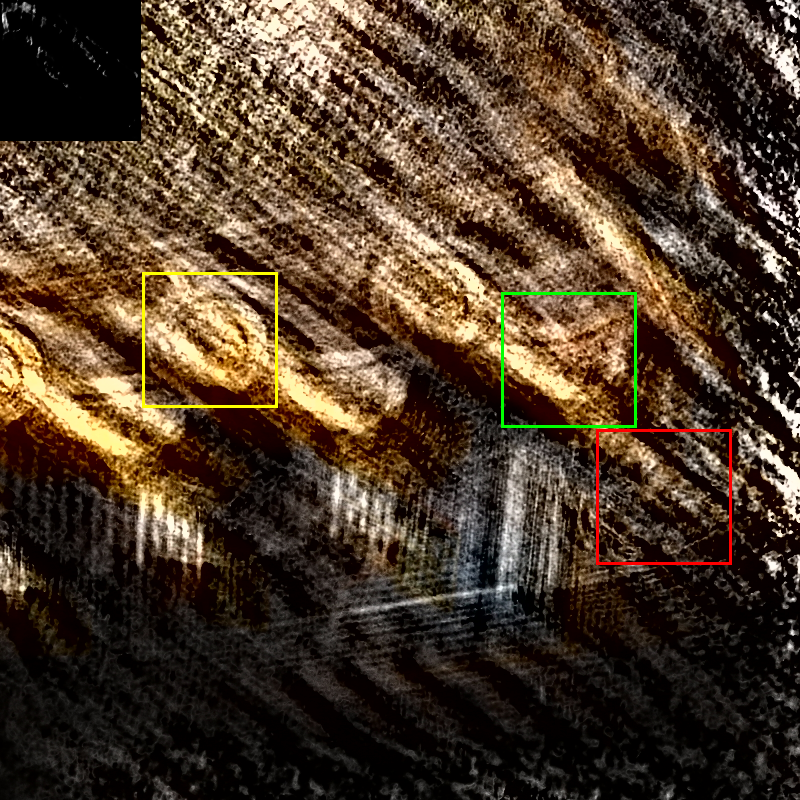}
		\vskip 2pt
		\includegraphics[width=1\linewidth]{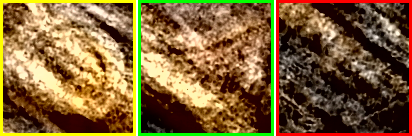}
        %\vskip -5pt
        \caption*{PSNR: 15.08}
        %\vskip -10pt
        \caption*{(c)SelfDeblur\cite{ren2020neural}\centering} %SelfDeblur
	\label{SelfDeblur}%文中引用该图片代号
    \end{minipage}
    \begin{minipage}{0.16\linewidth}
		\centering
		\includegraphics[width=1\linewidth]{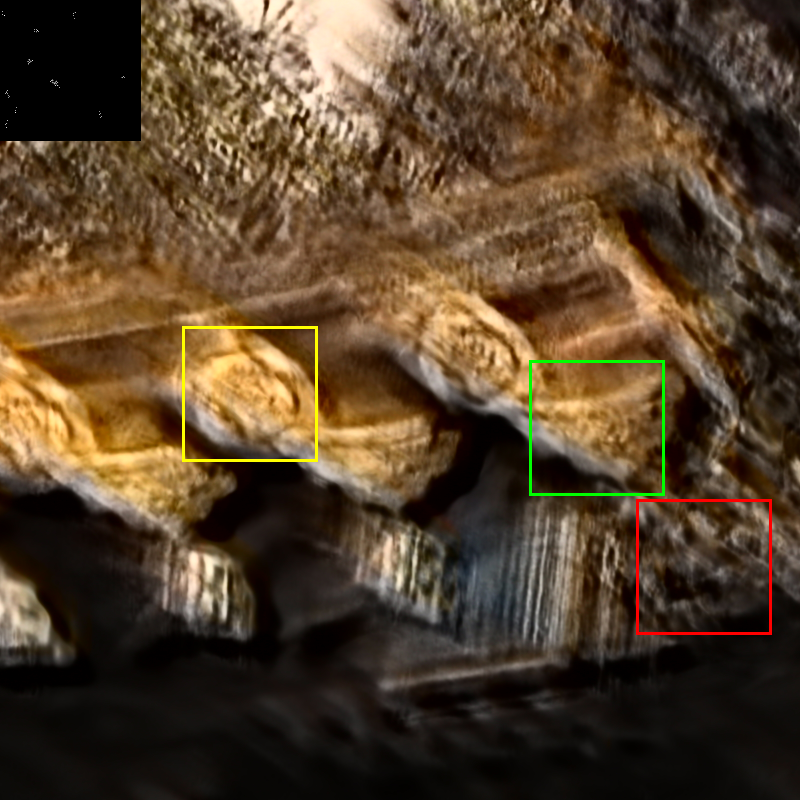}
		\vskip 2pt
		\includegraphics[width=1\linewidth]{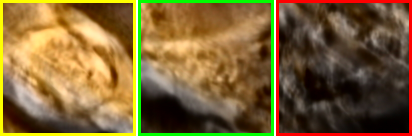}
        %\vskip -5pt
        \caption*{PSNR: 17.73}
        %\vskip -10pt
        \caption*{\leftline{(d)Fast-SelfDeblur\cite{bai2023fastselfdeblur}}} %Fast-SelfDeblur
	\label{Fast-SelfDeblur}%文中引用该图片代号
    \end{minipage}
    \begin{minipage}{0.16\linewidth}
		\centering
		\includegraphics[width=1\linewidth]{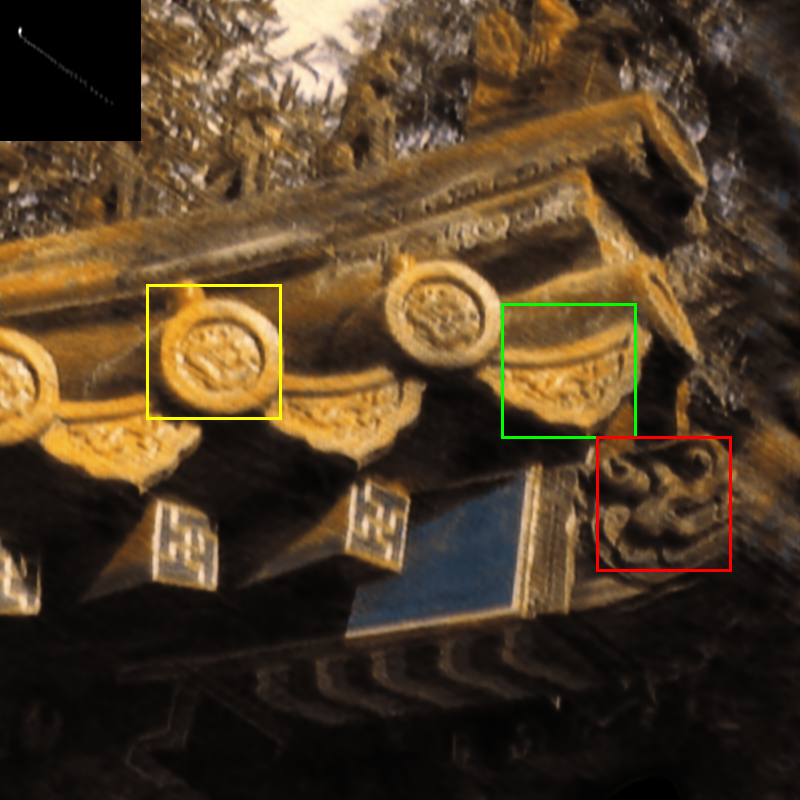}
		\vskip 2pt
		\includegraphics[width=1\linewidth]{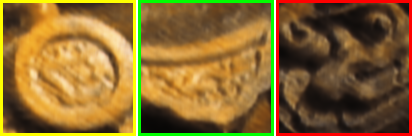}
        %\vskip -5pt
        \caption*{PSNR: 25.62}
        %\vskip -10pt
        \caption*{(e){\textbf{Ours}}}
	\label{Ours}%文中引用该图片代号
    \end{minipage}
    %\vskip -8pt
    \caption{Visual comparison on example images with 9th blur kernel from Kohler's dataset.}
    \label{fig:Kohler-visual-ker9}
\end{figure*}

\begin{figure*}[!tbp]
    \centering
    \begin{minipage}{0.16\linewidth}
		\centering
		\includegraphics[width=1\linewidth]{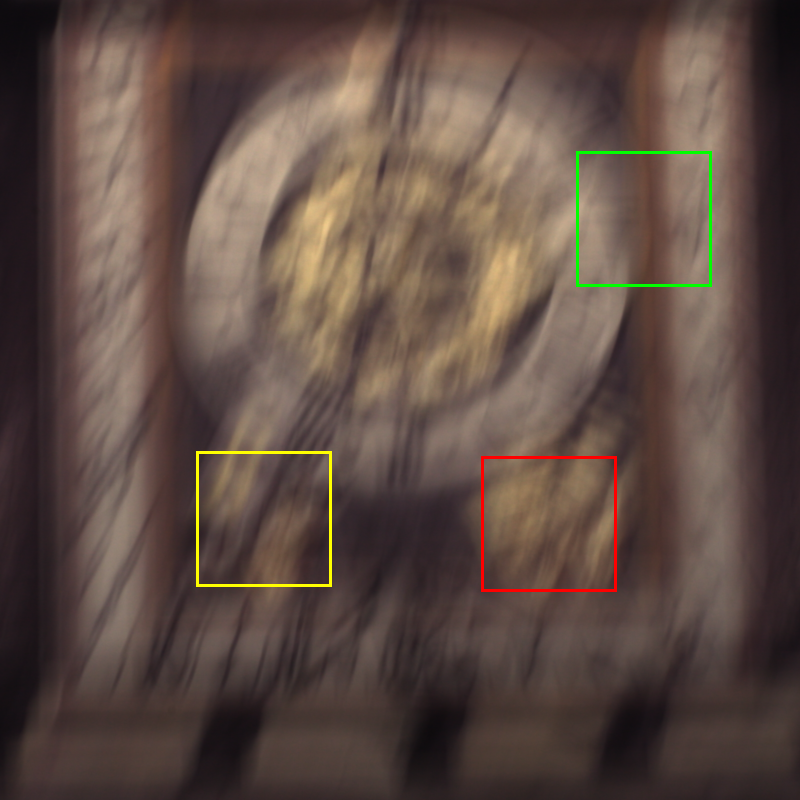}
		\vskip 2pt
		\includegraphics[width=1\linewidth]{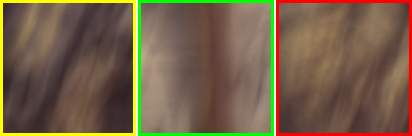}
        %\vskip -5pt
        \caption*{}
    \end{minipage}
    \begin{minipage}{0.16\linewidth}
		\centering
		\includegraphics[width=1\linewidth]{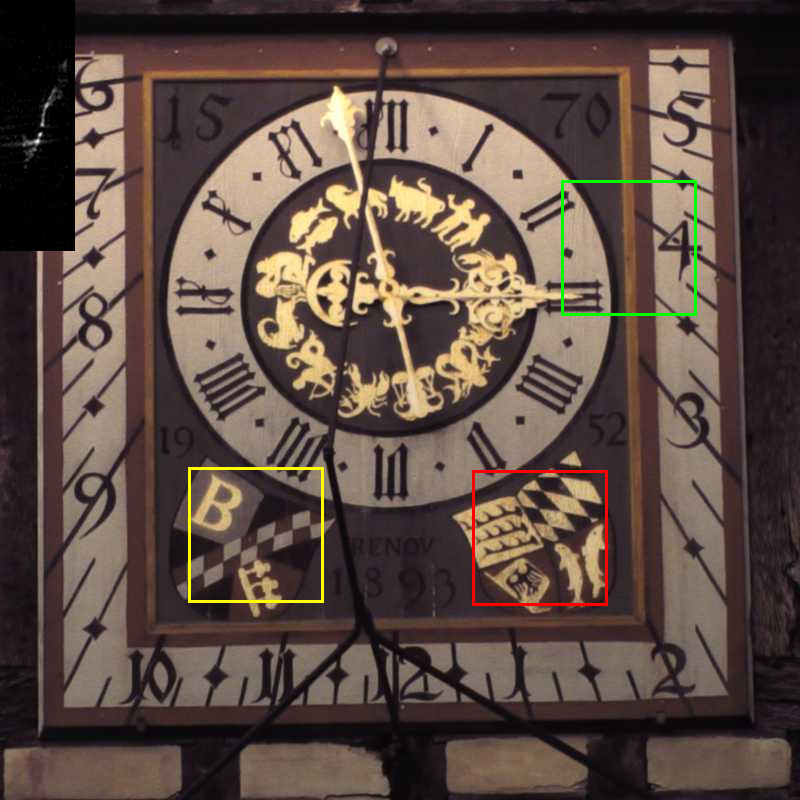}
		\vskip 2pt
		\includegraphics[width=1\linewidth]{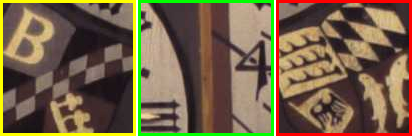}
        %\vskip -5pt
        \caption*{}
    \end{minipage}
    \begin{minipage}{0.16\linewidth}
		\centering
		\includegraphics[width=1\linewidth]{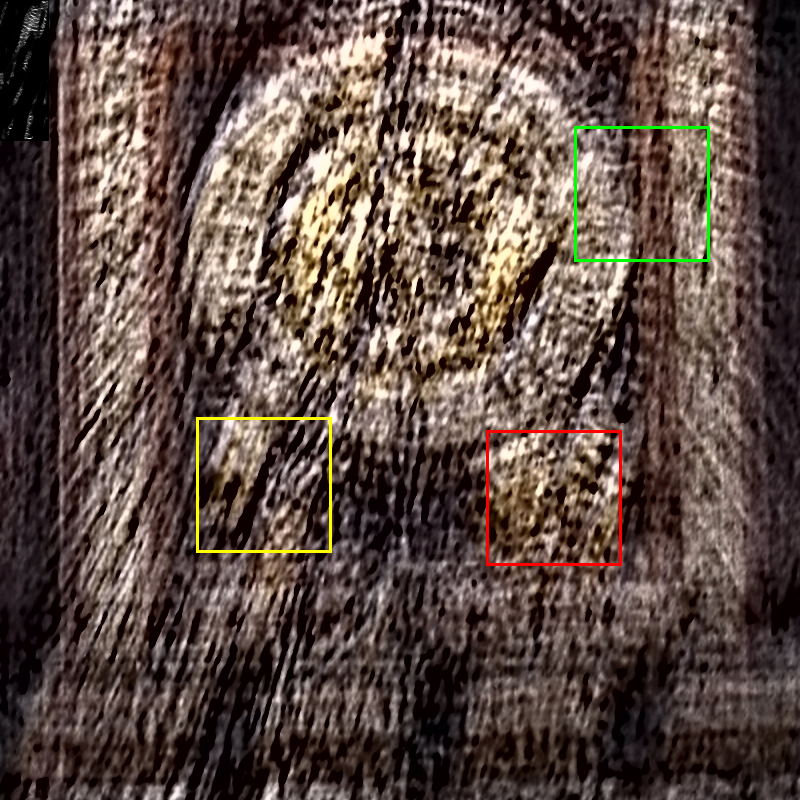}
		\vskip 2pt
		\includegraphics[width=1\linewidth]{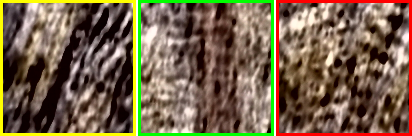}
        %\vskip -5pt
        \caption*{PSNR: 14.64}
    \end{minipage}
    \begin{minipage}{0.16\linewidth}
		\centering
		\includegraphics[width=1\linewidth]{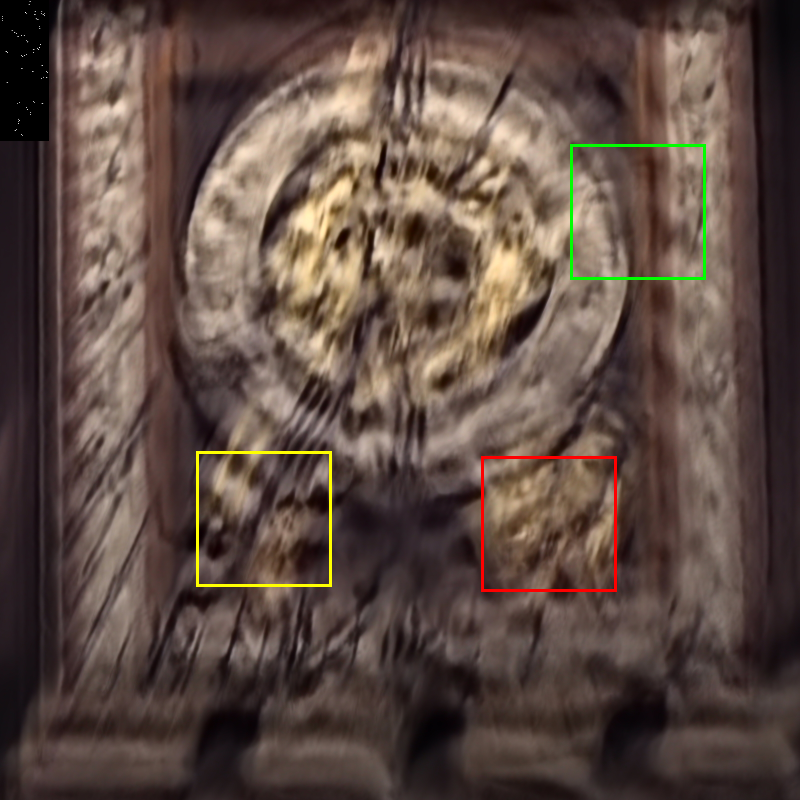}
		\vskip 2pt
		\includegraphics[width=1\linewidth]{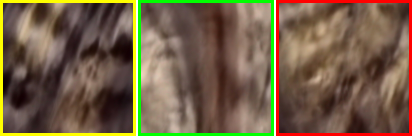}
        %\vskip -5pt
        \caption*{PSNR: 16.64}
    \end{minipage}
    \begin{minipage}{0.16\linewidth}
		\centering
		\includegraphics[width=1\linewidth]{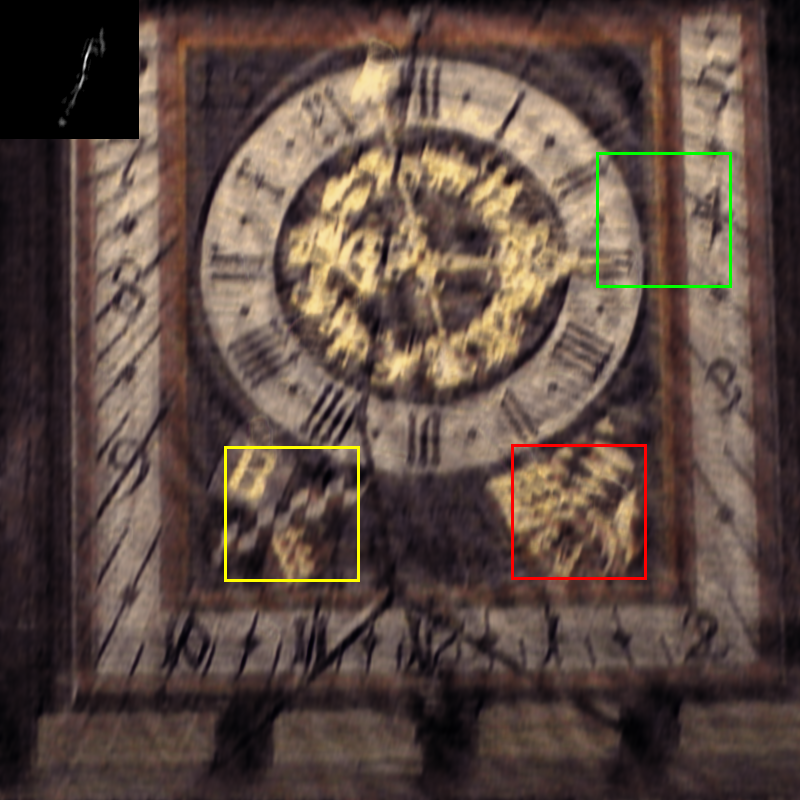}
		\vskip 2pt
		\includegraphics[width=1\linewidth]{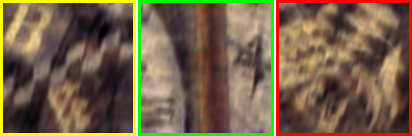}
        %\vskip -5pt
        \caption*{PSNR: 20.65}
    \end{minipage}
    \vskip 3pt
    \begin{minipage}{0.16\linewidth}
		\centering
		\includegraphics[width=1\linewidth]{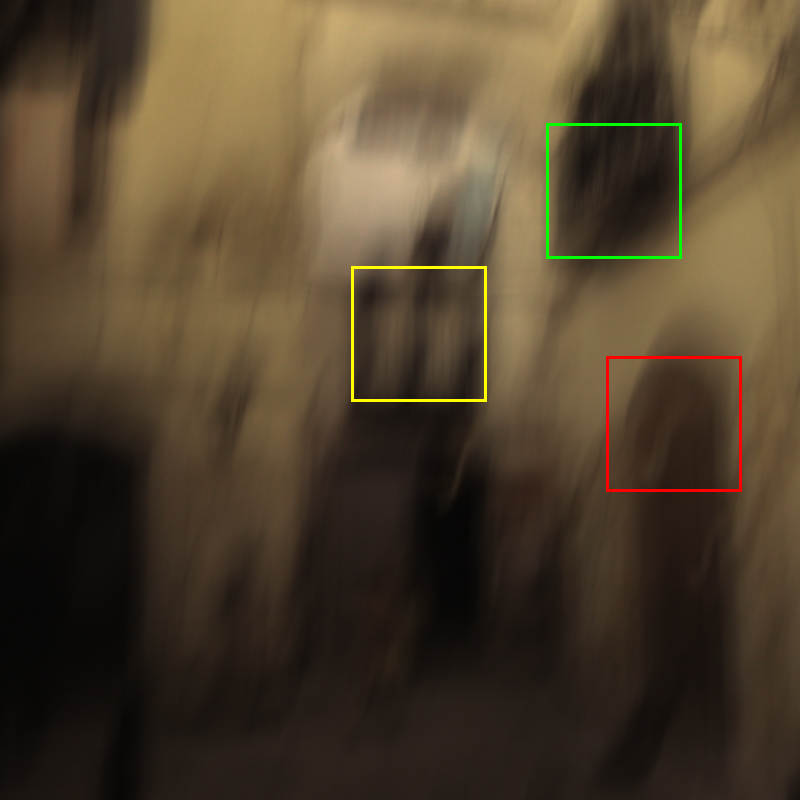}
		\vskip 2pt
		\includegraphics[width=1\linewidth]{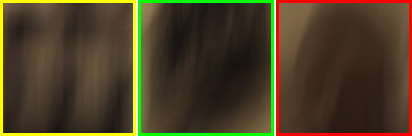}
        %\vskip -5pt
        \caption*{}
        %\vskip -10pt
        \caption*{(a)Blurred image\centering} 
	\label{Blurry image}%文中引用该图片代号
    \end{minipage}
    \begin{minipage}{0.16\linewidth}
		\centering
		\includegraphics[width=1\linewidth]{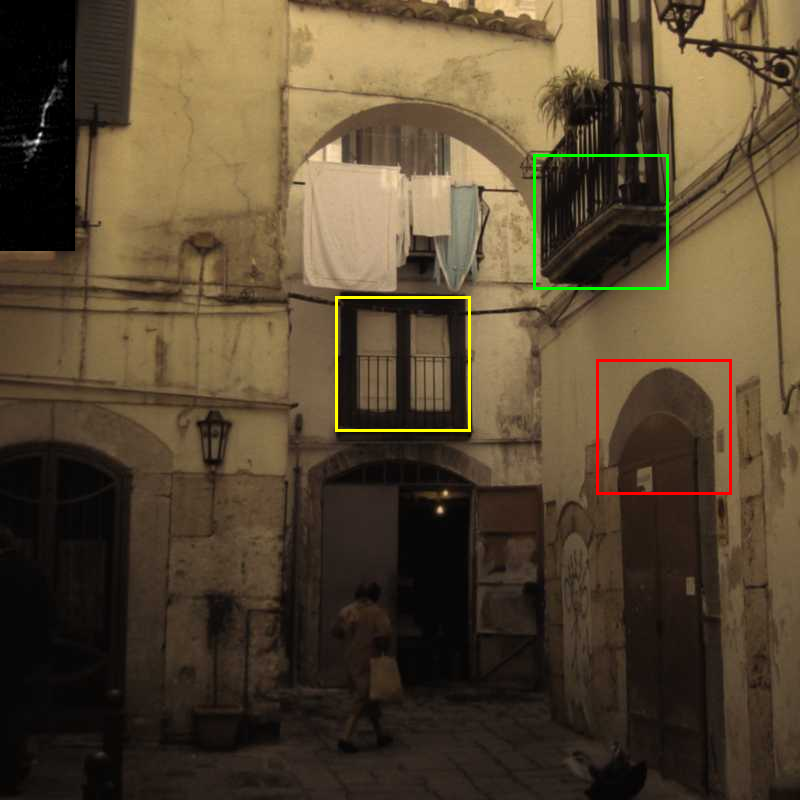}
		\vskip 2pt
		\includegraphics[width=1\linewidth]{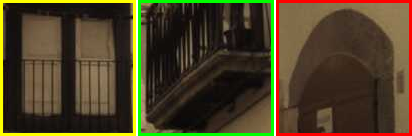}
        %\vskip -5pt
        \caption*{}
        %\vskip -10pt
        \caption*{(b)Ground-truth\centering}
        \label{Ground-truth}%文中引用该图片代号
    \end{minipage}
    \begin{minipage}{0.16\linewidth}
		\centering
		\includegraphics[width=1\linewidth]{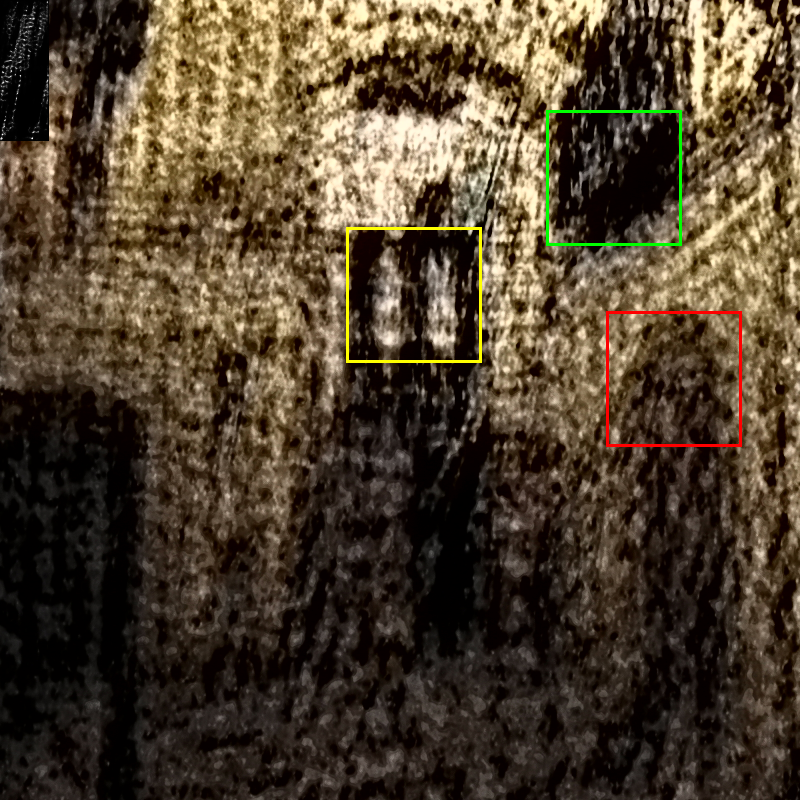}
		\vskip 2pt
		\includegraphics[width=1\linewidth]{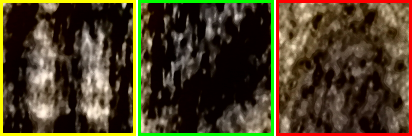}
        %%\vskip -5pt
        \caption*{PSNR: 17.45}
        %\vskip -10pt
        \caption*{(c)SelfDeblur\cite{ren2020neural}\centering} %SelfDeblur
	\label{SelfDeblur}%文中引用该图片代号
    \end{minipage}
    \begin{minipage}{0.16\linewidth}
		\centering
		\includegraphics[width=1\linewidth]{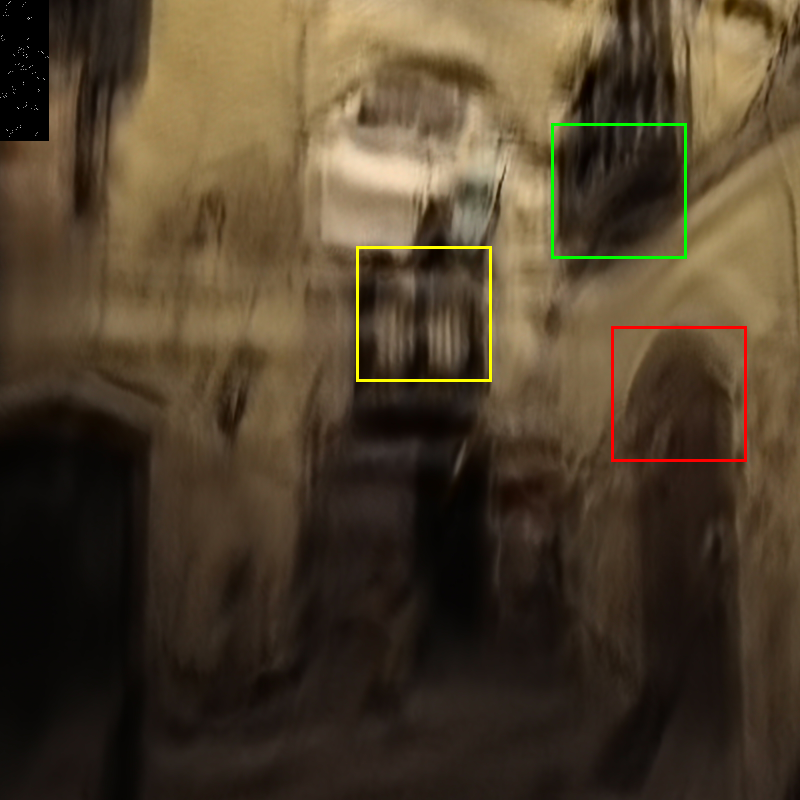}
		\vskip 2pt
		\includegraphics[width=1\linewidth]{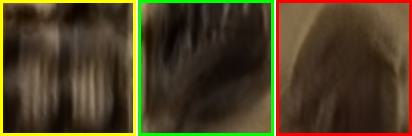}
        %\vskip -5pt
        \caption*{PSNR: 24.33}
        %\vskip -10pt
        \caption*{\leftline{(d)Fast-SelfDeblur\cite{bai2023fastselfdeblur}}} %Fast-SelfDeblur
	\label{Fast-SelfDeblur}%文中引用该图片代号
    \end{minipage}
    \begin{minipage}{0.16\linewidth}
		\centering
		\includegraphics[width=1\linewidth]{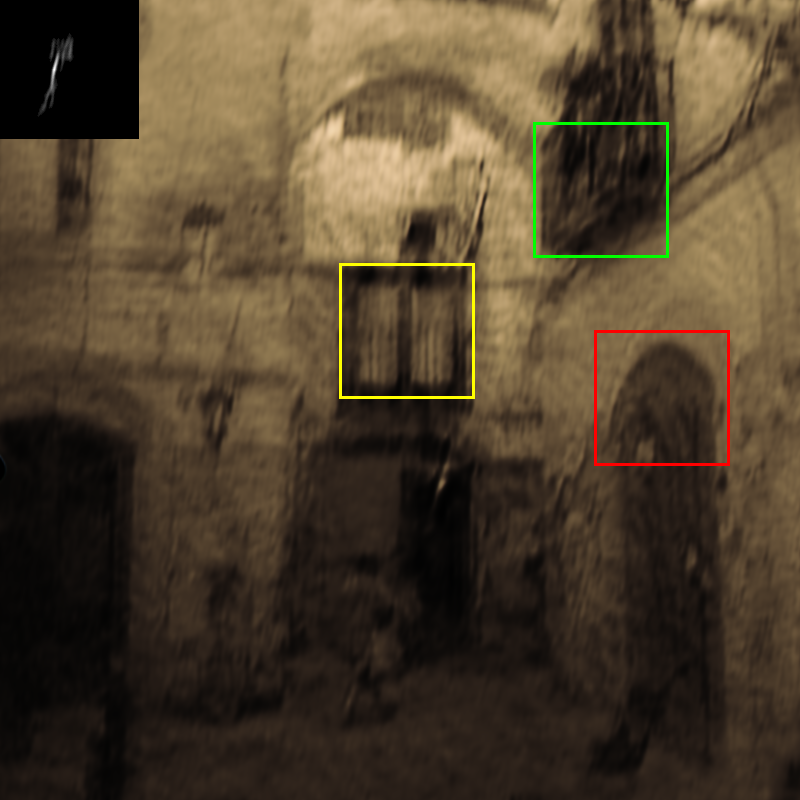}
		\vskip 2pt
		\includegraphics[width=1\linewidth]{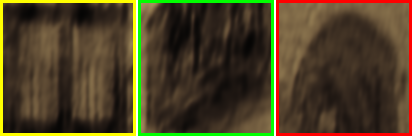}
        %\vskip -5pt
        \caption*{PSNR: 25.83}
        %\vskip -10pt
        \caption*{(e){\textbf{Ours}}}
	\label{Ours}%文中引用该图片代号
    \end{minipage}
    %\vskip -8pt
    \caption{Visual comparison on example images with 10th blur kernel from Kohler's dataset.}
    \label{fig:Kohler-visual-ker10}
\end{figure*}

\begin{figure*}[!tbp]
    \centering
    \begin{minipage}{0.16\linewidth}
		\centering
		\includegraphics[width=1\linewidth]{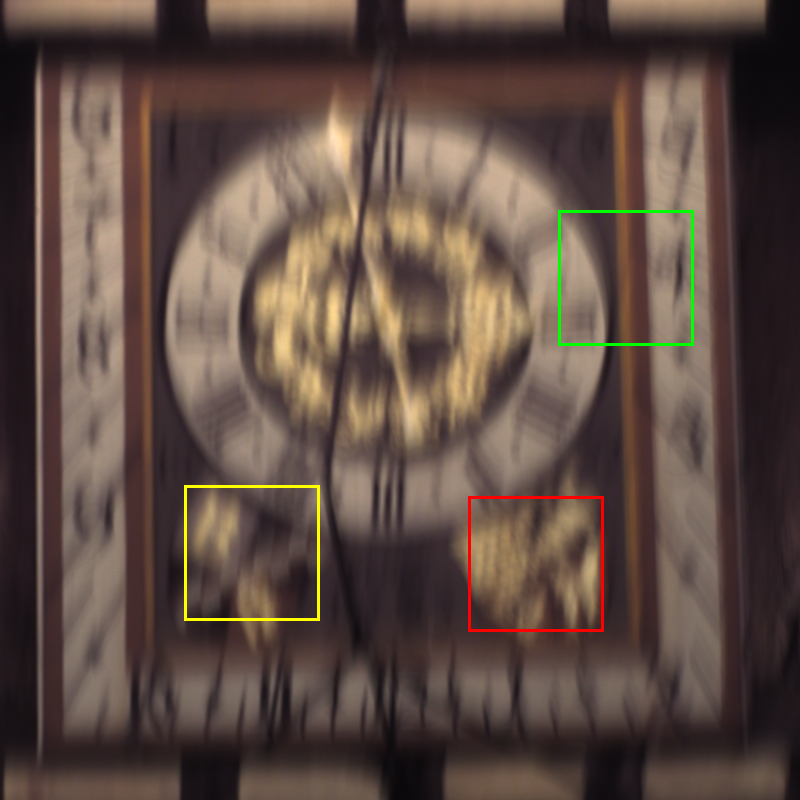}
		\vskip 2pt
		\includegraphics[width=1\linewidth]{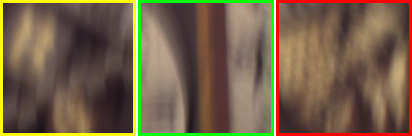}
        %\vskip -5pt
        \caption*{}
    \end{minipage}
    \begin{minipage}{0.16\linewidth}
		\centering
		\includegraphics[width=1\linewidth]{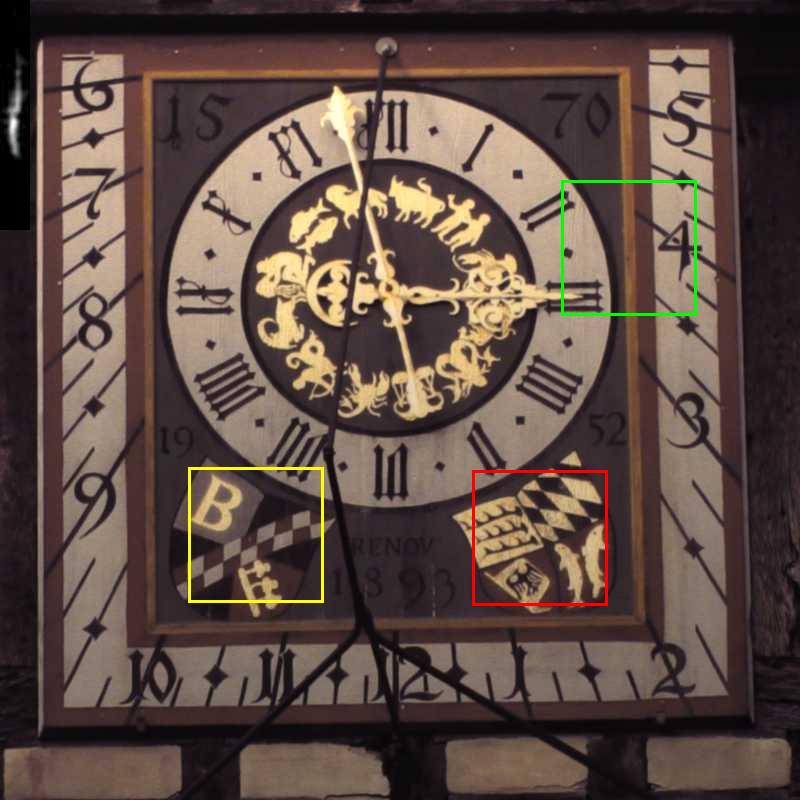}
		\vskip 2pt
		\includegraphics[width=1\linewidth]{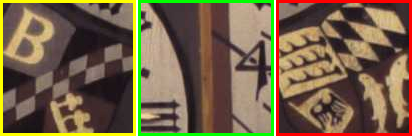}
        %\vskip -5pt
        \caption*{}
    \end{minipage}
    \begin{minipage}{0.16\linewidth}
		\centering
		\includegraphics[width=1\linewidth]{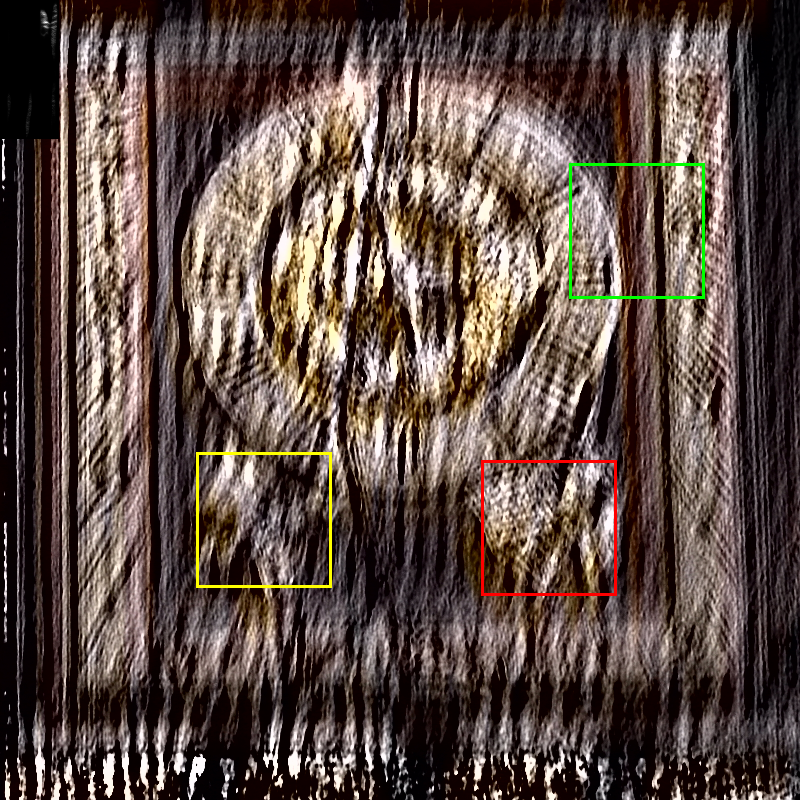}
		\vskip 2pt
		\includegraphics[width=1\linewidth]{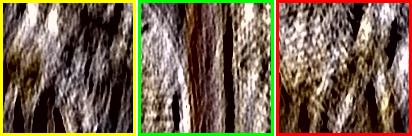}
        %\vskip -5pt
        \caption*{PSNR: 14.40}
    \end{minipage}
    \begin{minipage}{0.16\linewidth}
		\centering
		\includegraphics[width=1\linewidth]{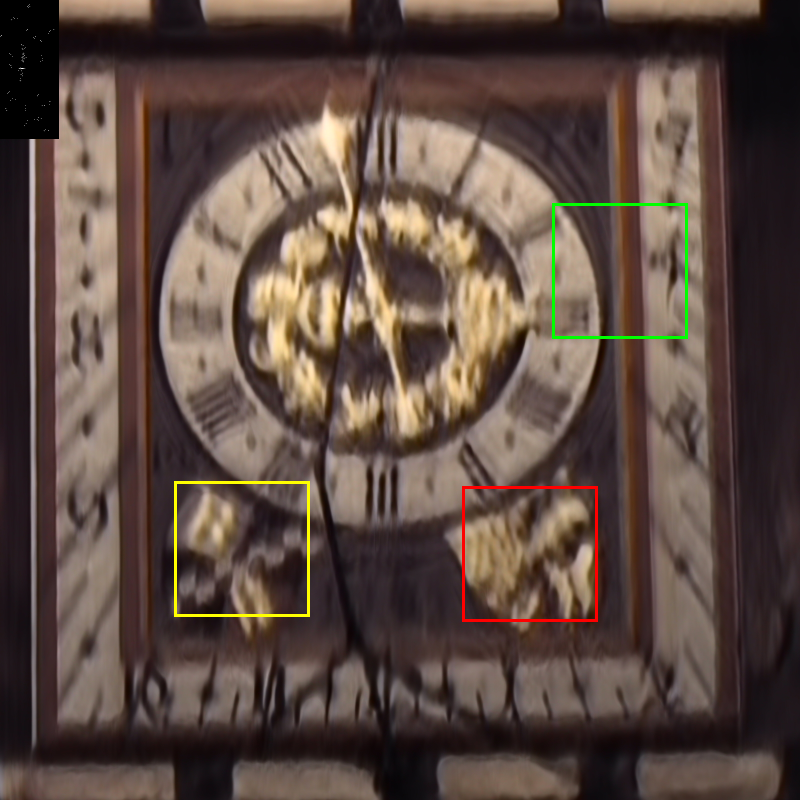}
		\vskip 2pt
		\includegraphics[width=1\linewidth]{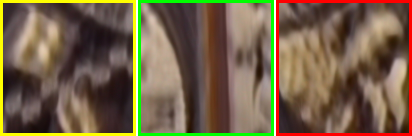}
        %\vskip -5pt
        \caption*{PSNR: 21.98}
    \end{minipage}
    \begin{minipage}{0.16\linewidth}
		\centering
		\includegraphics[width=1\linewidth]{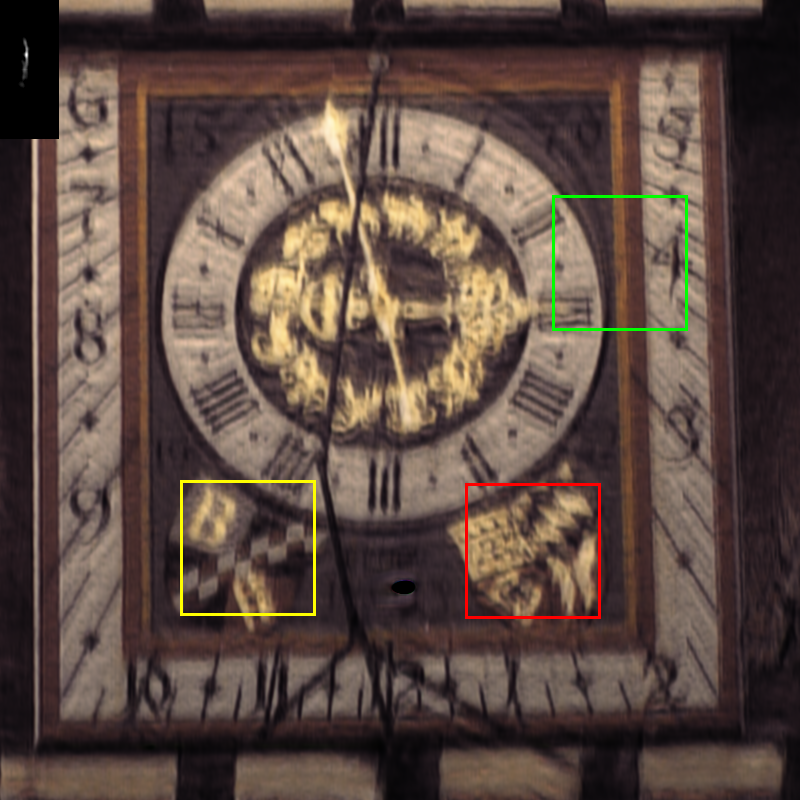}
		\vskip 2pt
		\includegraphics[width=1\linewidth]{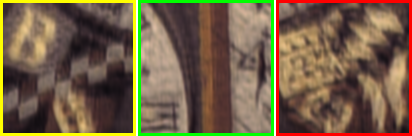}
        %\vskip -5pt
        \caption*{PSNR: 22.87}
    \end{minipage}
    \vskip 3pt
    \begin{minipage}{0.16\linewidth}
		\centering
		\includegraphics[width=1\linewidth]{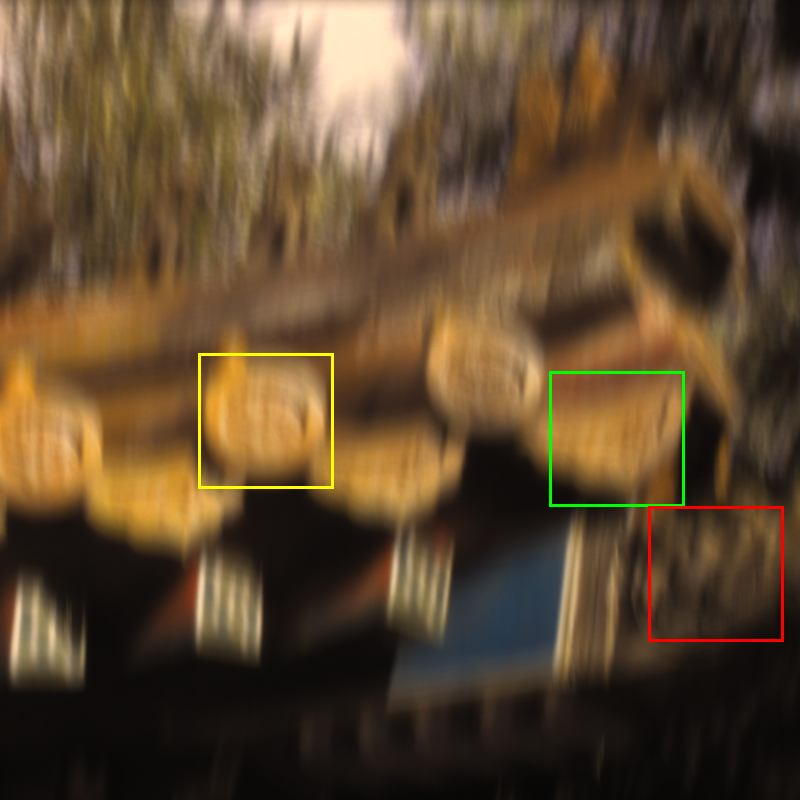}
		\vskip 2pt
		\includegraphics[width=1\linewidth]{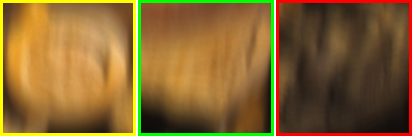}
        %\vskip -5pt
        \caption*{}
        %\vskip -10pt
        \caption*{(a)Blurred image\centering} 
	\label{Blurry image}%文中引用该图片代号
    \end{minipage}
    \begin{minipage}{0.16\linewidth}
		\centering
		\includegraphics[width=1\linewidth]{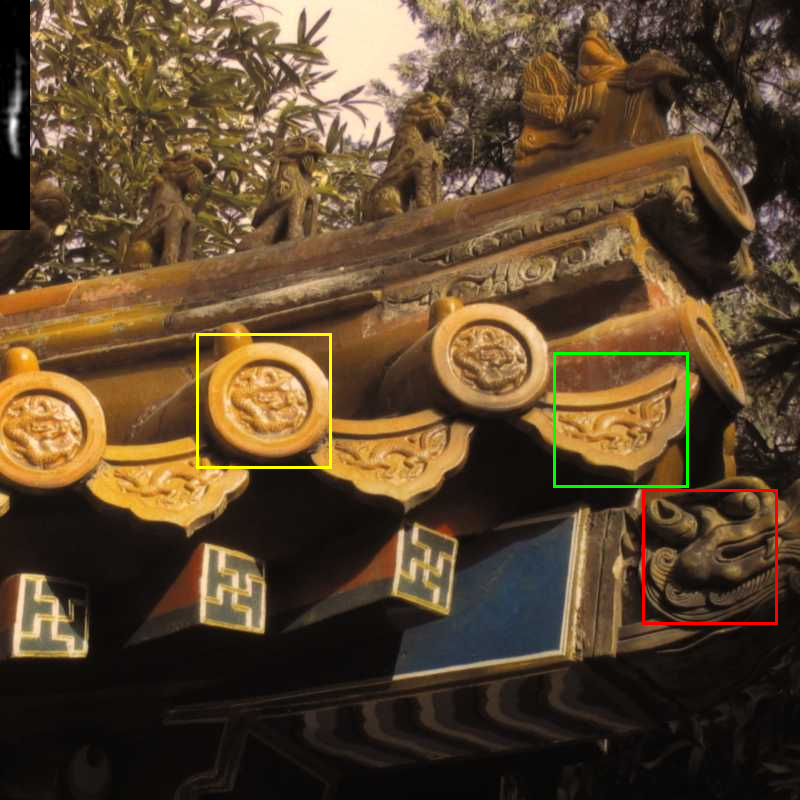}
		\vskip 2pt
		\includegraphics[width=1\linewidth]{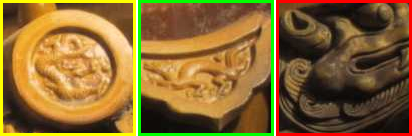}
        %\vskip -5pt
        \caption*{}
        %\vskip -10pt
        \caption*{(b)Ground-truth\centering}
        \label{Ground-truth}%文中引用该图片代号
    \end{minipage}
    \begin{minipage}{0.16\linewidth}
		\centering
		\includegraphics[width=1\linewidth]{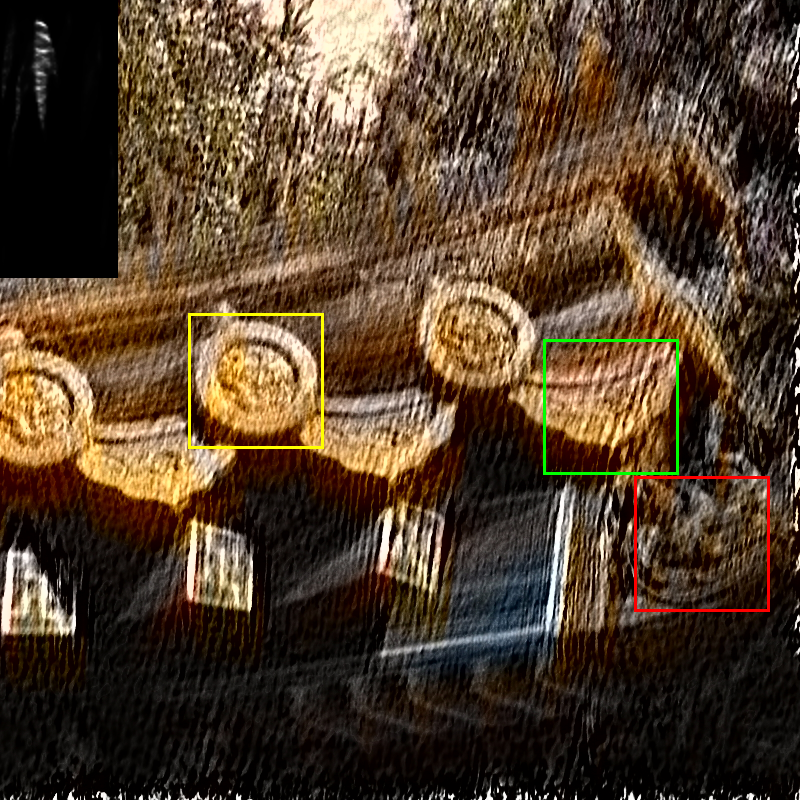}
		\vskip 2pt
		\includegraphics[width=1\linewidth]{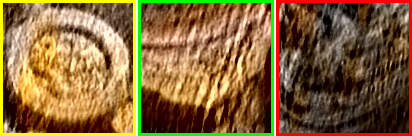}
        %\vskip -5pt
        \caption*{PSNR: 17.68}
        %\vskip -10pt
        \caption*{(c)SelfDeblur\cite{ren2020neural}\centering} %SelfDeblur
	\label{SelfDeblur}%文中引用该图片代号
    \end{minipage}
    \begin{minipage}{0.16\linewidth}
		\centering
		\includegraphics[width=1\linewidth]{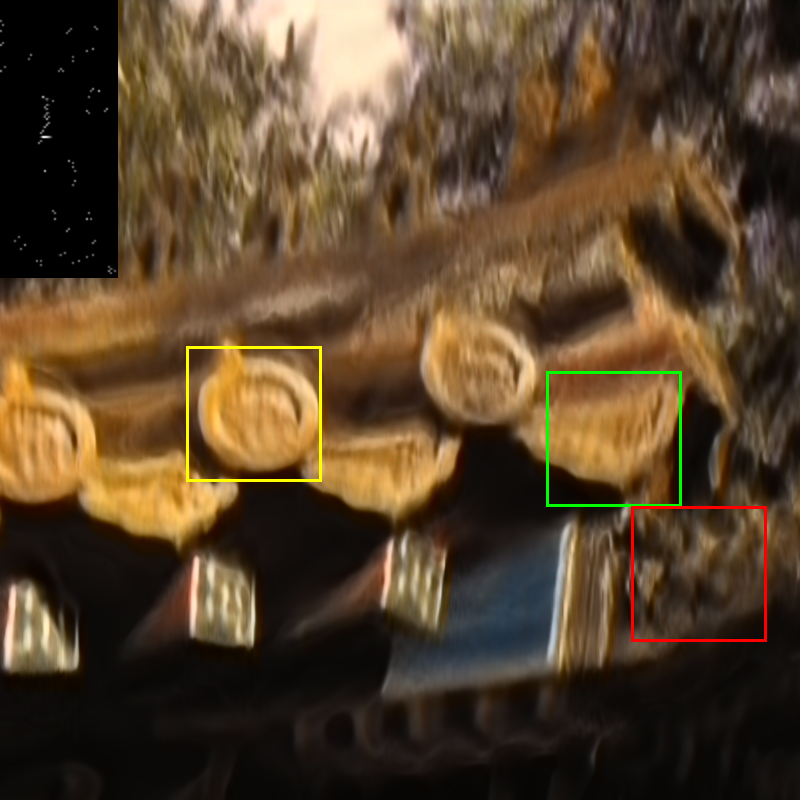}
		\vskip 2pt
		\includegraphics[width=1\linewidth]{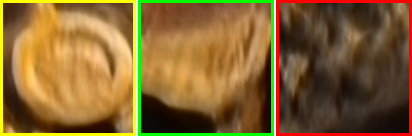}
        %\vskip -5pt
        \caption*{PSNR: 21.91}
        %\vskip -10pt
        \caption*{\leftline{(d)Fast-SelfDeblur\cite{bai2023fastselfdeblur}}} %Fast-SelfDeblur
	\label{Fast-SelfDeblur}%文中引用该图片代号
    \end{minipage}
    \begin{minipage}{0.16\linewidth}
		\centering
		\includegraphics[width=1\linewidth]{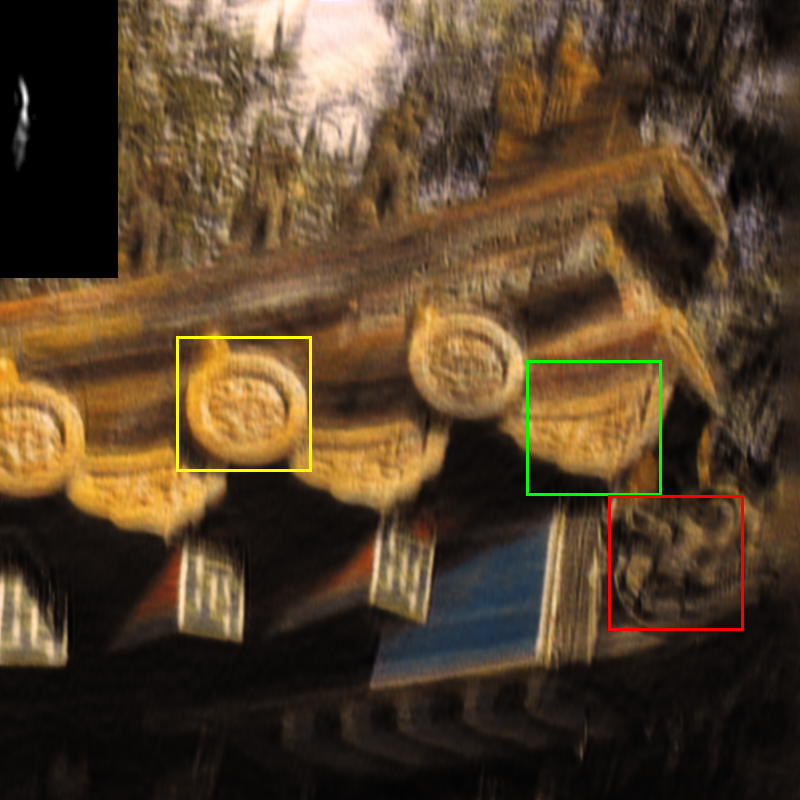}
		\vskip 2pt
		\includegraphics[width=1\linewidth]{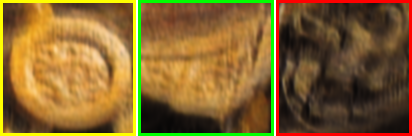}
        %\vskip -5pt
        \caption*{PSNR: 22.4}
        %\vskip -10pt
        \caption*{(e){\textbf{Ours}}}
	\label{Ours}%文中引用该图片代号
    \end{minipage}
    %\vskip -8pt
    \caption{Visual comparison on example images with 11th blur kernel from Kohler's dataset.}
    \label{fig:Kohler-visual-ker11}
\end{figure*}

%\subsection{Qualitative results - Kohler}
Many methods often cannot provide satisfactory deblurring results in the challenging scenarios involving large blur kernels. To further compare the image deblurring performance of our method with that of the other two self-supervised learning-based methods \cite{ren2020neural, bai2023fastselfdeblur}, we present visual results of example blurred images with different large blur kernels (8th-11th) from Kohler et al.'s dataset \cite{Kohler2012}, as shown in Figs.\ref{fig:Kohler-visual-ker8}-\ref{fig:Kohler-visual-ker11}. %,fig:Kohler-visual-ker9,fig:Kohler-visual-ker10,fig:Kohler-visual-ker11,  of Appendix \ref{sec:appendix_results_Kohler}
It can be observed that, the methods proposed in \cite{ren2020neural, bai2023fastselfdeblur} don't have accurate estimation of blur kernels and even fail to adequately capture the physics of the blur kernels. The results of \cite{ren2020neural} suffer from significant ringing artifacts, and blurs are not completely removed from the estimated images in \cite{bai2023fastselfdeblur}. In contrast, our method exhibits superior visual quality by effectively restoring sharp detail textures. These qualitative results intuitively reveal the effectiveness of the proposed Self-MSNet, particularly when dealing with largely blurred images.

\subsection{Ablation Study}
\label{ablationstudy}
The blur becomes smaller and more easily removed when the image is downscaled to coarser resolutions. Therefore, the incorporation of the multi-scale blur estimation can achieve better deblurring performance. An ablation study is conducted to evaluate the effectiveness of the multi-scale estimation employed in the proposed Self-MSNet.

The single-scale version (denoted as \textbf{Self-MSNet-S}) exhibits a similar architecture as Self-MSNet, except for a single input and a single output at the original image resolution. Our Self-MSNet is modified to Self-MSNet-S through the following steps: \textbf{(i)} the inputs and outputs at the other three coarser-scales are removed from the Input Converter and Output Converter part respectively, \textbf{(ii)} the SFF module is modified with only ${\rm FE}^{\rm out}$ as input, and \textbf{(iii)} the loss function is defined at the original image scale of $s=0$.

The PSNR/SSIM results of Self-MSNet-S on Lai et al.'s dataset\cite{Lai2016} and Kohler et al.'s dataset\cite{Kohler2012} are reported in Tab.\ref{tab:Lai-ave-PSNRssim} and Tab.\ref{tab:Kohler-ave-PSNRssim} respectively. As can be seen, Self-MSNet-S achieves inferior performance to Self-MSNet and these quantitative comparisons between Self-MSNet and Self-MSNet-S demonstrate the benefits of the multi-scale design. Moreover, although Self-MSNet-S operates only at the original image resolution, it achieves superior results to most of the other compared methods.

%\subsection{Experimental comparisons on Lai's dataset}
%\label{sec:appendix_results_Lai}
%==================================================================================
% [fig 4] manmade_01_kernel_04

% [fig 5] manmade_05_kernel_04

% [fig 6] natural_04_kernel_02

% [fig 7] natural_05_kernel_04

% [fig 8] people_03_kernel_02

% [fig 9] people_05_kernel_04

% [fig 10] saturated_05_kernel_03

% [fig 11] text_01_kernel_02

% [fig 12] text_04_kernel_02

%=================kohler的定性结果转去补充材料=================
%\section{Experimental comparisons on Kohler's dataset}
%\label{sec:appendix_results_Kohler}
% kohler 定性结果
%---------------kernel 8

%---------------kernel 9

%---------------kernel 10

%---------------kernel 11

\section{Conclusion}
In this paper, our Self-MSNet integrates the learning capability of deep learning-based methods and the adaptability of mathematical optimization-based models, with the aid of alternating optimization algorithm. By incorporating the multi-scale scheme into a single U-Net, a multi-scale generator network with multiple inputs and multiple outputs is presented to simultaneously estimate sharp images at four scales. The quadratic regularized least-squares model is independently solved for the blur kernel at each scale. We iteratively alternate between updating the latent image and the blur kernel estimation. This multi-scale design significantly enhances the collaboration of multi-scale image information, and proves effective in dealing with large blur kernels. Extensive performance evaluations have demonstrated the superiority of the proposed Self-MSNet over other competing methods, especially in handling largely blurred images.

\section*{Acknowledgments}
This work was supported by the Beijing Natural Science Foundation of China under Grant 4172002.

\bibliographystyle{IEEEtran}
\bibliography{main}

\begin{thebibliography}{10}
\providecommand{\url}[1]{#1}
\csname url@samestyle\endcsname
\providecommand{\newblock}{\relax}
\providecommand{\bibinfo}[2]{#2}
\providecommand{\BIBentrySTDinterwordspacing}{\spaceskip=0pt\relax}
\providecommand{\BIBentryALTinterwordstretchfactor}{4}
\providecommand{\BIBentryALTinterwordspacing}{\spaceskip=\fontdimen2\font plus
\BIBentryALTinterwordstretchfactor\fontdimen3\font minus \fontdimen4\font\relax}
\providecommand{\BIBforeignlanguage}[2]{{%
\expandafter\ifx\csname l@#1\endcsname\relax
\typeout{** WARNING: IEEEtran.bst: No hyphenation pattern has been}%
\typeout{** loaded for the language `#1'. Using the pattern for}%
\typeout{** the default language instead.}%
\else
\language=\csname l@#1\endcsname
\fi
#2}}
\providecommand{\BIBdecl}{\relax}
\BIBdecl

\bibitem{schuler2015learning}
C.~J. Schuler, M.~Hirsch, S.~Harmeling, and B.~Sch{\"o}lkopf, ``Learning to deblur,'' \emph{IEEE Trans. Pattern Anal. Mach. Intell.}, vol.~38, no.~7, pp. 1439--1451, 2015.

\bibitem{kupyn2018deblurgan}
O.~Kupyn, V.~Budzan, M.~Mykhailych, D.~Mishkin, and J.~Matas, ``Deblurgan: Blind motion deblurring using conditional adversarial networks,'' in \emph{IEEE Conf. Comput. Vis. Pattern Recog.}, 2018, pp. 8183--8192.

\bibitem{kupyn2019deblurganv2}
O.~Kupyn, T.~Martyniuk, J.~Wu, and Z.~Wang, ``Deblurgan-v2: Deblurring (orders-of-magnitude) faster and better,'' in \emph{Int. Conf. Comput. Vis.}, 2019, pp. 8878--8887.

\bibitem{cho2021rethinking}
S.-J. Cho, S.-W. Ji, J.-P. Hong, S.-W. Jung, and S.-J. Ko, ``Rethinking coarse-to-fine approach in single image deblurring,'' in \emph{Int. Conf. Comput. Vis.}, 2021, pp. 4641--4650.

\bibitem{nah2017deep}
S.~Nah, T.~H. Kim, and K.~M. Lee, ``Deep multi-scale convolutional neural network for dynamic scene deblurring,'' in \emph{IEEE Conf. Comput. Vis. Pattern Recog.}, 2017, pp. 3883--3891.

\bibitem{zhang2017learning}
K.~Zhang, W.~Zuo, S.~Gu, and L.~Zhang, ``Learning deep cnn denoiser prior for image restoration,'' in \emph{IEEE Conf. Comput. Vis. Pattern Recog.}, 2017, pp. 3929--3938.

\bibitem{ulyanov2018deep}
D.~Ulyanov, A.~Vedaldi, and V.~Lempitsky, ``Deep image prior,'' in \emph{IEEE Conf. Comput. Vis. Pattern Recog.}, 2018, pp. 9446--9454.

\bibitem{UNet2015}
O.~Ronneberger, P.~Fischer, and T.~Brox, ``U-net: Convolutional networks for biomedical image segmentation,'' in \emph{Medical Image Computing and Computer-Assisted Intervention (MICCAI)}.\hskip 1em plus 0.5em minus 0.4em\relax Cham, Switzerland: Springer, 2015, pp. 234--241.

\bibitem{ren2020neural}
D.~Ren, K.~Zhang, Q.~Wang, Q.~Hu, and W.~Zuo, ``Neural blind deconvolution using deep priors,'' in \emph{IEEE Conf. Comput. Vis. Pattern Recog.}, 2020, pp. 3341--3350.

\bibitem{bai2023fastselfdeblur}
Y.~Bai, J.~Yu, Y.~LI, and C.~Xiao, ``Deep prior-based blind image deblurring,'' \emph{Acta Electronica Sinica}, vol.~51, no.~4, pp. 1050--1067, 2023.

\bibitem{chan1998total}
T.~F. Chan and C.-K. Wong, ``Total variation blind deconvolution,'' \emph{IEEE Trans. Image Process.}, vol.~7, no.~3, pp. 370--375, 1998.

\bibitem{whyte2014IJCVshaken}
O.~Whyte, J.~Sivic, and A.~Zisserman, ``Deblurring shaken and partially saturated images,'' \emph{Int. J. Comput. Vis.}, vol. 110, pp. 185--201, 2014.

\bibitem{perrone2014total}
D.~Perrone and P.~Favaro, ``Total variation blind deconvolution: The devil is in the details,'' in \emph{IEEE Conf. Comput. Vis. Pattern Recog.}, 2014, pp. 2909--2916.

\bibitem{fergus2006removing}
R.~Fergus, B.~Singh, A.~Hertzmann, S.~T. Roweis, and W.~T. Freeman, ``Removing camera shake from a single photograph,'' \emph{ACM Trans. Graph.}, vol.~25, no.~3, pp. 787--794, 2006.

\bibitem{Levin2011CVPRefficient}
A.~Levin, Y.~Weiss, F.~Durand, and W.~Freeman, ``Efficient marginal likelihood optimization in blind deconvolution,'' in \emph{IEEE Conf. Comput. Vis. Pattern Recog.}, 2011, pp. 2657--2664.

\bibitem{shan2008highquality}
Q.~Shan, J.~Jia, and A.~Agarwala, ``High-quality motion deblurring from a single image,'' \emph{ACM Trans. Graph.}, vol.~27, no.~3, pp. 1--10, 2008.

\bibitem{levin2009understanding}
A.~Levin, Y.~Weiss, F.~Durand, and W.~T. Freeman, ``Understanding and evaluating blind deconvolution algorithms,'' in \emph{IEEE Conf. Comput. Vis. Pattern Recog.}\hskip 1em plus 0.5em minus 0.4em\relax IEEE, 2009, pp. 1964--1971.

\bibitem{Yang2019CVPRAdaptiveEdgeSelection}
L.~Yang and H.~Ji, ``A variational em framework with adaptive edge selection for blind motion deblurring,'' in \emph{IEEE Conf. Comput. Vis. Pattern Recog.}, 2019, pp. 10\,159--10\,168.

\bibitem{Krishnan2009hyperLaplacian}
D.~Krishnan and R.~Fergus, ``Fast image deconvolution using hyper-laplacian priors,'' in \emph{Adv. Neural Inform. Process. Syst.}, vol.~22, 2009, pp. 1--9.

\bibitem{Krishnan2011CVPRnormalized}
D.~Krishnan, T.~Tay, and R.~Fergus, ``Blind deconvolution using a normalized sparsity measure,'' in \emph{IEEE Conf. Comput. Vis. Pattern Recog.}, 2011, pp. 233--240.

\bibitem{xu2013unnatural}
L.~Xu, S.~Zheng, and J.~Jia, ``Unnatural $\ell_0$ sparse representation for natural image deblurring,'' in \emph{IEEE Conf. Comput. Vis. Pattern Recog.}, 2013, pp. 1107--1114.

\bibitem{pan2014CVPRtextL0}
J.~Pan, Z.~Hu, Z.~Su, and M.~Yang, ``Deblurring text images via $\ell_0$-regularized intensity and gradient prior,'' in \emph{IEEE Conf. Comput. Vis. Pattern Recog.}, 2014, pp. 2901--2908.

\bibitem{Anger2019L0}
J.~Anger, G.~Facciolo, and M.~Delbracio, ``Blind image deblurring using the $\ell_0$ gradient prior,'' \emph{Image Processing On Line}, vol.~9, pp. 124--142, 2019.

\bibitem{pan2016PAMIL0text}
J.~Pan, Z.~Hu, Z.~Su, and M.~Yang, ``$\ell_0$-regularized intensity and gradient prior for deblurring text images and beyond,'' \emph{IEEE Trans. Pattern Anal. Mach. Intell.}, vol.~39, no.~2, pp. 342--355, 2017.

\bibitem{cho2009fast}
S.~Cho and S.~Lee, ``Fast motion deblurring,'' \emph{ACM Trans. Graph.}, vol.~28, no.~5, pp. 1--8, 2009.

\bibitem{Joshi2008CVPRsharpedge}
N.~Joshi, R.~Szeliski, and D.~J. Kriegman, ``Psf estimation using sharp edge prediction,'' in \emph{IEEE Conf. Comput. Vis. Pattern Recog.}, 2008, pp. 1--8.

\bibitem{xu2010two}
L.~Xu and J.~Jia, ``Two-phase kernel estimation for robust motion deblurring,'' in \emph{Eur. Conf. Comput. Vis.}\hskip 1em plus 0.5em minus 0.4em\relax Berlin, Heidelberg: Springer, 2010, pp. 157--170.

\bibitem{money2008TVshock}
J.~H. Money and S.~H. Kang, ``Total variation minimizing blind deconvolution with shock filter reference,'' \emph{Image and Vision Computing}, vol.~26, no.~2, pp. 302--314, 2008.

\bibitem{sun2013ICCPedgeBased}
L.~Sun, S.~Cho, J.~Wang, and J.~Hays, ``Edge-based blur kernel estimation using patch priors,'' 2013, pp. 1--8.

\bibitem{lai2015CVPRnormalized}
W.-S. Lai, J.-J. Ding, Y.-Y. Lin, and Y.-Y. Chuang, ``Blur kernel estimation using normalized color-line priors,'' in \emph{IEEE Conf. Comput. Vis. Pattern Recog.}, 2015, pp. 64--72.

\bibitem{Pan2013SPICsalient}
J.~Pan, R.~Liu, Z.~Su, and X.~Gu, ``Kernel estimation from salient structure for robust motion deblurring,'' \emph{Signal Processing: Image Communication}, vol.~28, no.~9, pp. 1156--1170, 2013.

\bibitem{gong2016CVPRautomatic}
D.~Gong, M.~Tan, Y.~Zhang, A.~V.~D. Hengel, and Q.~Shi, ``Blind image deconvolution by automatic gradient activation,'' in \emph{IEEE Conf. Comput. Vis. Pattern Recog.}, 2016, pp. 1827--1836.

\bibitem{zhang2011close}
H.~Zhang, J.~Yang, Y.~Zhang, N.~M. Nasrabadi, and T.~S. Huang, ``Close the loop: Joint blind image restoration and recognition with sparse representation prior,'' in \emph{Int. Conf. Comput. Vis.}\hskip 1em plus 0.5em minus 0.4em\relax IEEE, 2011, pp. 770--777.

\bibitem{czc2017sparse}
Z.~Chang, J.~Yu, C.~Xiao, and W.~Sun, ``Single image blind deconvolution using sparse representation and structural self-similarity,'' \emph{Acta Automatica Sinica}, vol.~43, no.~11, pp. 1908--1919, 2017.

\bibitem{peng2021ksvd}
T.~Peng, J.~Yu, L.~Guo, and C.~Xiao, ``Blind image deconvolution via cross-scale dictionary learning,'' \emph{Optics and Precision Engineering}, vol.~29, no.~2, pp. 338--348, 2021.

\bibitem{michaeli2014blind}
T.~Michaeli and M.~Irani, ``Blind deblurring using internal patch recurrence,'' in \emph{Eur. Conf. Comput. Vis.}\hskip 1em plus 0.5em minus 0.4em\relax Cham, Switzerland: Springer, 2014, pp. 783--798.

\bibitem{wang2013ACCVnsp}
S.~Wang, L.~Zhang, and Y.~Liang, ``Nonlocal spectral prior model for low-level vision,'' in \emph{ACCV}, K.~M. Lee, Y.~Matsushita, J.~M. Rehg, and Z.~Hu, Eds.\hskip 1em plus 0.5em minus 0.4em\relax Berlin, Heidelberg: Springer, 2013, pp. 231--244.

\bibitem{ren2016image}
W.~Ren, X.~Cao, J.~Pan, X.~Guo, W.~Zuo, and M.~Yang, ``Image deblurring via enhanced low-rank prior,'' \emph{IEEE Trans. Image Process.}, vol.~25, no.~7, pp. 3426--3437, 2016.

\bibitem{peng2022lowrank}
T.~Peng, J.~Yu, and C.~Xiao, ``Blind image deconvolution via cross-scale low rank prior,'' \emph{Acta Automatica Sinica}, vol.~48, no.~10, pp. 2508--2525, 2022.

\bibitem{pan2016CVPRdarkchannel}
J.~Pan, D.~Sun, H.~Pfister, and M.-H. Yang, ``Blind image deblurring using dark channel prior,'' in \emph{IEEE Conf. Comput. Vis. Pattern Recog.}, 2016, pp. 1628--1636.

\bibitem{pan2018PAMIdarkchannel}
J.~Pan, D.~Sun, H.~Pfister, and M.~Yang, ``Deblurring images via dark channel prior,'' \emph{IEEE Trans. Pattern Anal. Mach. Intell.}, vol.~40, no.~10, pp. 2315--2328, 2018.

\bibitem{yan2017extremechannel}
Y.~Yan, W.~Ren, Y.~Guo, R.~Wang, and X.~Cao, ``Image deblurring via extreme channels prior,'' in \emph{IEEE Conf. Comput. Vis. Pattern Recog.}, 2017, pp. 4003--4011.

\bibitem{chen2019blind}
L.~Chen, F.~Fang, T.~Wang, and G.~Zhang, ``Blind image deblurring with local maximum gradient prior,'' in \emph{IEEE Conf. Comput. Vis. Pattern Recog.}, 2019, pp. 1742--1750.

\bibitem{Zoran2011ICCVnaturalimagepatch}
D.~Zoran and Y.~Weiss, ``From learning models of natural image patches to whole image restoration,'' in \emph{Int. Conf. Comput. Vis.}, 2011, pp. 479--486.

\bibitem{cai2012TIPframelet}
J.~Cai, H.~Ji, C.~Liu, and Z.~Shen, ``Framelet-based blind motion deblurring from a single image,'' \emph{IEEE Trans. Image Process.}, vol.~21, no.~2, pp. 562--572, 2012.

\bibitem{Kotera2013CAIPL1norm}
J.~Kotera, F.~{\v{S}}roubek, and P.~Milanfar, ``Blind deconvolution using alternating maximum a posteriori estimation with heavy-tailed priors,'' in \emph{Computer Analysis of Images and Patterns}, 2013, pp. 59--66.

\bibitem{yue2014ECCVhybrid}
T.~Yue, S.~Cho, J.~Wang, and Q.~Dai, ``Hybrid image deblurring by fusing edge and power spectrum information,'' in \emph{Eur. Conf. Comput. Vis.}\hskip 1em plus 0.5em minus 0.4em\relax Cham, Switzerland: Springer, 2014, pp. 79--93.

\bibitem{shao2015BiL0L2norm}
W.~Shao, H.~Li, and M.~Elad, ``Bi-$\ell_0$-$\ell_2$-norm regularization for blind motion deblurring,'' \emph{Journal of Visual Communication and Image Representation}, vol.~33, pp. 42--59, 2015.

\bibitem{Almeida2010TIPsemi}
M.~S.~C. Almeida and L.~B. Almeida, ``Blind and semi-blind deblurring of natural images,'' \emph{IEEE Trans. Image Process.}, vol.~19, no.~1, pp. 36--52, 2010.

\bibitem{you1999TIPanisotropicTV}
Y.~You and M.~Kaveh, ``Blind image restoration by anisotropic regularization,'' \emph{IEEE Trans. Image Process.}, vol.~8, no.~3, pp. 396--407, 1999.

\bibitem{liu2014spectral}
G.~Liu, S.~Chang, and Y.~Ma, ``Blind image deblurring using spectral properties of convolution operators,'' \emph{IEEE Trans. Image Process.}, vol.~23, no.~12, pp. 5047--5056, 2014.

\bibitem{liang2021CVPRFKP}
J.~Liang, K.~Zhang, S.~Gu, L.~V. Gool, and R.~Timofte, ``Flow-based kernel prior with application to blind super-resolution,'' in \emph{IEEE Conf. Comput. Vis. Pattern Recog.}, 2021, pp. 10\,596--10\,605.

\bibitem{fang2023CVPRflow}
Z.~Fang, F.~Wu, W.~Dong, X.~Li, J.~Wu, and G.~Shi, ``Self-supervised non-uniform kernel estimation with flow-based motion prior for blind image deblurring,'' in \emph{IEEE Conf. Comput. Vis. Pattern Recog.}, 2023, pp. 18\,105--18\,114.

\bibitem{li2023deepEM}
J.~Li, W.~Wang, Y.~Nan, and H.~Ji, ``Self-supervised blind motion deblurring with deep expectation maximization,'' in \emph{IEEE Conf. Comput. Vis. Pattern Recog.}, 2023, pp. 13\,986--13\,996.

\bibitem{gandelsman2019double}
Y.~Gandelsman, A.~Shocher, and M.~Irani, ``"double-dip": Unsupervised image decomposition via coupled deep-image-priors,'' in \emph{IEEE Conf. Comput. Vis. Pattern Recog.}, 2019, pp. 11\,026--11\,035.

\bibitem{tao2018scale}
X.~Tao, H.~Gao, X.~Shen, J.~Wang, and J.~Jia, ``Scale-recurrent network for deep image deblurring,'' in \emph{IEEE Conf. Comput. Vis. Pattern Recog.}, 2018, pp. 8174--8182.

\bibitem{gao2019CVPRnested}
H.~Gao, X.~Tao, X.~Shen, and J.~Jia, ``Dynamic scene deblurring with parameter selective sharing and nested skip connections,'' in \emph{IEEE Conf. Comput. Vis. Pattern Recog.}, 2019, pp. 3843--3851.

\bibitem{kaufman2020deblurring}
A.~Kaufman and R.~Fattal, ``Deblurring using analysis-synthesis networks pair,'' in \emph{IEEE Conf. Comput. Vis. Pattern Recog.}, 2020, pp. 5811--5820.

\bibitem{zhang2019deep}
H.~Zhang, Y.~Dai, H.~Li, and P.~Koniusz, ``Deep stacked hierarchical multi-patch network for image deblurring,'' in \emph{IEEE Conf. Comput. Vis. Pattern Recog.}, 2019, pp. 5978--5986.

\bibitem{Zamir2021CVPRMPRNet}
S.~W. Zamir, A.~Arora, S.~Khan, M.~Hayat, F.~S. Khan, M.~Yang, and L.~Shao, ``Multi-stage progressive image restoration,'' in \emph{IEEE Conf. Comput. Vis. Pattern Recog.}, 2021, pp. 14\,816--14\,826.

\bibitem{li2022ECCVdegradation}
D.~Li, Y.~Zhang, K.~C. Cheung, X.~Wang, H.~Qin, and H.~Li, ``Learning degradation representations for image deblurring,'' in \emph{Eur. Conf. Comput. Vis.}\hskip 1em plus 0.5em minus 0.4em\relax Cham, Switzerland: Springer, 2022, pp. 736--753.

\bibitem{xu2018TIPmotion}
X.~Xu, J.~Pan, Y.-J. Zhang, and M.~Yang, ``Motion blur kernel estimation via deep learning,'' \emph{IEEE Trans. Image Process.}, vol.~27, no.~1, pp. 194--205, 2018.

\bibitem{Ramakrishnan2017ICCVWGFMD}
S.~Ramakrishnan, S.~Pachori, A.~Gangopadhyay, and S.~Raman, ``Deep generative filter for motion deblurring,'' 2017, pp. 2993--3000.

\bibitem{yuan2020efficient}
Y.~Yuan, W.~Su, and D.~Ma, ``Efficient dynamic scene deblurring using spatially variant deconvolution network with optical flow guided training,'' in \emph{IEEE Conf. Comput. Vis. Pattern Recog.}, 2020, pp. 3555--3564.

\bibitem{Purohit2020regionadaptive}
K.~Purohit and A.~N. Rajagopalan, ``Region-adaptive dense network for efficient motion deblurring,'' \emph{AAAI}, vol.~34, no.~7, pp. 11\,882--11\,889, 2020.

\bibitem{sun2015CVPRlearning}
J.~Sun, W.~Cao, Z.~Xu, and J.~Ponce, ``Learning a convolutional neural network for non-uniform motion blur removal,'' in \emph{IEEE Conf. Comput. Vis. Pattern Recog.}, 2015, pp. 769--777.

\bibitem{yan2016blind}
R.~Yan and L.~Shao, ``Blind image blur estimation via deep learning,'' \emph{IEEE Trans. Image Process.}, vol.~25, no.~4, pp. 1910--1921, 2016.

\bibitem{gong2017CVPRfromMBMF}
D.~Gong, J.~Yang, L.~Liu, Y.~Zhang, I.~Reid, C.~Shen, A.~V.~D. Hengel, and Q.~Shi, ``From motion blur to motion flow: a deep learning solution for removing heterogeneous motion blur,'' in \emph{IEEE Conf. Comput. Vis. Pattern Recog.}, 2017, pp. 3806--3815.

\bibitem{zuo2016TIPtriple}
W.~Zuo, D.~Ren, D.~Zhang, S.~Gu, and L.~Zhang, ``Learning iteration-wise generalized shrinkage-thresholding operators for blind deconvolution,'' \emph{IEEE Trans. Image Process.}, vol.~25, no.~4, pp. 1751--1764, 2016.

\bibitem{li2019ICASSPunrolling}
Y.~Li, M.~Tofighi, V.~Monga, and Y.~C. Eldar, ``An algorithm unrolling approach to deep image deblurring,'' in \emph{ICASSP}, 2019, pp. 7675--7679.

\bibitem{Pan2017ICCVdataFitting}
J.~Pan, J.~Dong, Y.-W. Tai, Z.~Su, and M.~Yang, ``Learning discriminative data fitting functions for blind image deblurring,'' in \emph{Int. Conf. Comput. Vis.}, 2017, pp. 1077--1085.

\bibitem{Lu2019CVPRgan}
B.~Lu, J.-C. Chen, and R.~Chellappa, ``Unsupervised domain-specific deblurring via disentangled representations,'' in \emph{IEEE Conf. Comput. Vis. Pattern Recog.}, 2019, pp. 10\,217--10\,226.

\bibitem{xia2019NIPStraining}
Z.~Xia and A.~Chakrabarti, ``Training image estimators without image ground truth,'' in \emph{Adv. Neural Inform. Process. Syst.}, H.~Wallach, H.~Larochelle, A.~Beygelzimer, F.~d\textquotesingle Alch\'{e}-Buc, E.~Fox, and R.~Garnett, Eds., vol.~32.\hskip 1em plus 0.5em minus 0.4em\relax Curran Associates, Inc., 2019, pp. 1--11.

\bibitem{zhang2020CVPRrealistic}
K.~Zhang, W.~Luo, Y.~Zhong, L.~Ma, B.~Stenger, W.~Liu, and H.~Li, ``Deblurring by realistic blurring,'' in \emph{IEEE Conf. Comput. Vis. Pattern Recog.}, 2020, pp. 2734--2743.

\bibitem{Chen2023TCSVTensemble}
M.~Chen, Y.~Quan, Y.~Xu, and H.~Ji, ``Self-supervised blind image deconvolution via deep generative ensemble learning,'' \emph{IEEE Trans. Circuit Syst. Video Technol.}, vol.~33, no.~2, pp. 634--647, 2023.

\bibitem{kaiming2015initialization}
K.~He, X.~Zhang, S.~Ren, and J.~Sun, ``Delving deep into rectifiers: surpassing human-level performance on imagenet classification,'' in \emph{Int. Conf. Comput. Vis.}, 2015, pp. 1026--1034.

\bibitem{kingma2014adam}
D.~P. Kingma and J.~Ba, ``Adam: A method for stochastic optimization,'' \emph{CoRR}, vol. abs/1412.6980, 2014.

\bibitem{Lai2016}
W.-S. Lai, J.-B. Huang, Z.~Hu, N.~Ahuja, and M.~Yang, ``A comparative study for single image blind deblurring,'' in \emph{IEEE Conf. Comput. Vis. Pattern Recog.}\hskip 1em plus 0.5em minus 0.4em\relax IEEE, 2016, pp. 1701--1709.

\bibitem{Kohler2012}
R.~K{\"o}hler, M.~Hirsch, B.~Mohler, B.~Sch{\"o}lkopf, and S.~Harmeling, ``Recording and playback of camera shake: Benchmarking blind deconvolution with a real-world database,'' in \emph{Eur. Conf. Comput. Vis.}\hskip 1em plus 0.5em minus 0.4em\relax Berlin, Heidelberg: Springer, 2012, pp. 27--40.

\bibitem{whyte2010nonUniform}
O.~Whyte, J.~Sivic, A.~Zisserman, and J.~Ponce, ``Non-uniform deblurring for shaken images,'' in \emph{IEEE Conf. Comput. Vis. Pattern Recog.}, 2010, pp. 491--498.

\bibitem{krishnan2011normalized}
D.~Krishnan, T.~Tay, and R.~Fergus, ``Blind deconvolution using a normalized sparsity measure,'' in \emph{IEEE Conf. Comput. Vis. Pattern Recog.}, 2011, pp. 233--240.

\bibitem{levin2011understanding}
A.~Levin, Y.~Weiss, F.~Durand, and W.~T. Freeman, ``Understanding blind deconvolution algorithms,'' \emph{IEEE Trans. Pattern Anal. Mach. Intell.}, vol.~33, no.~12, pp. 2354--2367, 2011.

\bibitem{zhang2013multiImage}
H.~Zhang, D.~Wipf, and Y.~Zhang, ``Multi-image blind deblurring using a coupled adaptive sparse prior,'' in \emph{IEEE Conf. Comput. Vis. Pattern Recog.}, 2013, pp. 1051--1058.

\bibitem{Zhong2013handling}
L.~Zhong, S.~Cho, D.~Metaxas, S.~Paris, and J.~Wang, ``Handling noise in single image deblurring using directional filters,'' in \emph{IEEE Conf. Comput. Vis. Pattern Recog.}, 2013, pp. 612--619.

\bibitem{RDS2014}
\BIBentryALTinterwordspacing
``Robust deblurring software,'' Jan. 2014. [Online]. Available: \url{http://www.cse.cuhk.edu.hk/~leojia/deblurring.htm}
\BIBentrySTDinterwordspacing

\bibitem{bai2019TIPgraph}
Y.~Bai, G.~Cheung, X.~Liu, and W.~Gao, ``Graph-based blind image deblurring from a single photograph,'' \emph{IEEE Trans. Image Process.}, vol.~28, no.~3, pp. 1404--1418, 2019.

\bibitem{chen2020enhanced}
L.~Chen, F.~Fang, S.~Lei, F.~Li, and G.~Zhang, ``Enhanced sparse model for blind deblurring,'' in \emph{Eur. Conf. Comput. Vis.}\hskip 1em plus 0.5em minus 0.4em\relax Cham, Switzerland: Springer, 2020, pp. 631--646.

\bibitem{chen2020OID}
L.~Chen, F.~Fang, J.~Zhang, J.~Liu, and G.~Zhang, ``Oid: Outlier identifying and discarding in blind image deblurring,'' in \emph{Eur. Conf. Comput. Vis.}\hskip 1em plus 0.5em minus 0.4em\relax Cham, Switzerland: Springer, 2020, pp. 598--613.

\bibitem{shao2020gradient}
S.~Wenze, L.~Yunzhi, L.~Yuanyuan, W.~Liqian, G.~qi, B.~Bingkun, and L.~Haibo, ``Gradient-based discriminative modeling for blind image deblurring,'' \emph{Neurocomputing}, vol. 413, pp. 305--327, 2020.

\bibitem{wen2021TCSVT}
F.~Wen, R.~Ying, Y.~Liu, P.~Liu, and T.-K. Truong, ``A simple local minimal intensity prior and an improved algorithm for blind image deblurring,'' \emph{IEEE Trans. Circuit Syst. Video Technol.}, vol.~31, no.~8, pp. 2923--2937, 2021.

\bibitem{shao2023Revisiting}
W.~Shao, ``Revisiting the regularizers in blind image deblurring with a new one,'' \emph{IEEE Trans. Image Process.}, vol.~32, pp. 3994--4009, 2023.

\bibitem{whyte2011shaken}
O.~Whyte, J.~Sivic, and A.~Zisserman, ``Deblurring shaken and partially saturated images,'' 2011, pp. 745--752.

\bibitem{Hirsch2011fastremoval}
M.~Hirsch, C.~J. Schuler, S.~Harmeling, and B.~Sch{\"o}lkopf, ``Fast removal of non-uniform camera shake,'' in \emph{Int. Conf. Comput. Vis.}, 2011, pp. 463--470.

\bibitem{Goldstein2012ECCVspectralirregularities}
A.~Goldstein and R.~Fattal, ``Blur-kernel estimation from spectral irregularities,'' in \emph{Eur. Conf. Comput. Vis.}\hskip 1em plus 0.5em minus 0.4em\relax Berlin, Heidelberg: Springer, 2012, pp. 622--635.

\end{thebibliography}

\appendices
\section{Derivation for Blur Kernel Estimation Subproblem}
\label{sec:appendix_blur_kernel_derivation}
As mentioned in Sec.\ref{sec:blur_kernel_estimation_subproblem} of our paper, the blur kernel estimation subproblem at each scale can be formulated as
\begin{equation}
    \begin{aligned}
        {{\bm{h}}_k} =
        &\mathop{\arg}\min_{\bm{h}} \Vert  \nabla \bm{y} - \bm{h} \ast \nabla \bm{x}_{k-1} \Vert^2_{\mathrm F} + \lambda_h \Vert \bm{h} \Vert^2_{\mathrm F} + \\
        &\, \gamma \bigg{\lVert} {\left( x_0, y_0 \right)^{\rm{T}}} - {\left(  \frac{ {\mathbbm{1}}_{n}^{\rm{T}} {\bm{h}}^{\rm{T}} {{\bm t}_m} }{ {\mathbbm{1}}_{m}^{\rm{T}} {\bm{h}} {{\mathbbm{1}}_{n}} }, \frac{ {\mathbbm{1}}_{m}^{\rm{T}} {\bm{h}} {{\bm t}_n} }{ {\mathbbm{1}}_{m}^{\rm{T}} {\bm{h}} {{\mathbbm{1}}_{n}} } \right)^{\rm{T}}} \bigg{\rVert}^2_2
    \end{aligned}
    \label{eq:sup_1}
\end{equation}

Equation (\ref{eq:sup_1}) is a quadratic regularized least-squares problem with respect to $\bm{h}$, which admits a global minimum solution. To simplify the calculation of the derivative, we first reduce it to
\begin{equation}
    \begin{aligned}
        {{\bm{h}}_k} =  &\mathop{\arg}\min_{\bm{h}} \Vert  \nabla \bm{y} - \bm{h} \ast \nabla \bm{x}_{k-1} \Vert^2_{\mathrm F} + \lambda_h \Vert \bm{h} \Vert^2_{\mathrm F} + \\
        &\, \gamma \left( \left( {\bm t}_{m} - x_0 \right)^{\rm{T}} {\bm h} {\mathbbm{1}}_{n} \right)^2 + \gamma \left( {\mathbbm{1}}_{m}^{\rm{T}} {\bm h} \left( {\bm t}_{n} - y_0 \right) \right)^2
    \end{aligned}
    \label{eq:sup_2}
\end{equation}
We set the derivative of the objective function with respect to $\bm{h}$ in Eq.(\ref{eq:sup_1}) to zero, that is,
\begin{equation}
    \begin{aligned}
        \frac{\partial \, \bm{h}_{k}} {\partial \, {\bm{h}}} = 
        &-2 \, {\widehat{ \partial_r \bm{x}_{k-1}}} \ast \left( \partial_r \bm{y}- \bm{h}_{k} \ast \partial_r \bm{x}_{k-1} \right) \\
        &- 2 \, {\widehat{\partial_c \bm{x}_{k-1}}} \ast \left( \partial_c \bm{y} - \bm{h}_{k} \ast \partial_c \bm{x}_{k-1} \right) + 2 \, \lambda_h \bm{h}_{k} \\
        &+ 2 \gamma \left( {\bm t}_{m} - x_0 \right) {\mathbbm{1}}_{n}^{\rm{T}} \left( {\bm t}_{m} - x_0 \right)^{\rm{T}} {\bm h}_{k} {\mathbbm{1}}_{n} \\
        &+ 2 \gamma {\mathbbm{1}}_{m} \left( {\bm t}_{n} - y_0 \right)^{\rm{T}} {\mathbbm{1}}_{m}^{\rm{T}} {\bm h}_{k} \left( {\bm t}_{n} - y_0 \right) \\
        &= 0
    \end{aligned}
    \label{eq:sup_3}
\end{equation}
That can be rearranged as
\begin{equation}
\label{eq:sup_4}
    \begin{aligned}
        & \Bigl[ \left( {\widehat{\partial_r \bm{x}_{k-1}}} \right) \ast \left( \partial_r \bm{x}_{k-1} \right) + \left( {\widehat{\partial_c \bm{x}_{k-1}}} \right) \ast \left( \partial_c \bm{x}_{k-1} \right) + \lambda_h \Bigr] \ast \bm{h}_{k} \\
        & + \gamma \left( {\bm t}_{m} - x_0 \right) {\mathbbm{1}}_{n}^{\rm{T}} \left( {\bm t}_{m} - x_0 \right)^{\rm{T}} {\bm h}_{k} {\mathbbm{1}}_{n} \\
        & + \gamma {\mathbbm{1}}_{m} \left( {\bm t}_{n} - y_0 \right)^{\rm{T}} {\mathbbm{1}}_{m}^{\rm{T}} {\bm h}_{k} \left( {\bm t}_{n} - y_0 \right) \\
        &=  \left( {\widehat{\partial_r \bm{x}_{k-1}}} \right) \ast (\partial_r \bm{y})  + \left( {\widehat{\partial_c \bm{x}_{k-1}}} \right) \ast \left( \partial_c \bm{y} \right) 
    \end{aligned}
\end{equation}
where $\,\, \widehat{} \,\,$ indicates the reverse operation. The \textit{Convolution Theorem} states that the Fourier transform of the convolution of two images in the spatial domain is equal to the product in the frequency domain of their Fourier transforms. By performing Fourier transform on both sides of Eq.(\ref{eq:sup_4}), we can obtain
\begin{equation}
    \begin{aligned}
        &{\bm{\mathit \Psi}} \odot \mathcal{F} \left( \bm{h}_{k} \right) + \gamma {\mathcal{F}} \left( \left( {\bm t}_{m} - x_0 \right) {\mathbbm{1}}_{n}^{\rm{T}} \right) \left( {\bm t}_{m} - x_0 \right)^{\rm{T}} {\bm h}_{k} {\mathbbm{1}}_{n}\\
        &+ \gamma {\mathcal{F}} \left( {\mathbbm{1}}_{m} \left( {\bm t}_{n} - y_0 \right)^{\rm{T}} \right) {\mathbbm{1}}_{m}^{\rm{T}} {\bm h}_{k} \left( {\bm t}_{n} - y_0 \right) = {\bm{\mathit \Gamma}}
    \end{aligned}
    \label{eq:sup_5}
\end{equation}
where $\odot$ represents the element-wise product operation, and
\begin{equation}
\begin{aligned}
    {\bm{\mathit \Psi}} = \,\, &\lambda_{h} + \overline{\mathcal{F} \left( \partial_r {\bm x}_{k-1} \right)} \odot \mathcal{F} \left( \partial_r {\bm x}_{k-1} \right)  \\
    &\quad\, + \overline{\mathcal{F} \left( \partial_c {\bm x}_{k-1} \right)} \odot \mathcal{F} \left( \partial_c {\bm x}_{k-1} \right), \\
    {\bm{\mathit \Gamma}} = \,\, &\overline{\mathcal{F} \left(\partial_r {\bm x}_{k-1} \right)} \odot \mathcal{F} \left( \partial_r {\bm y} \right) + \overline{\mathcal{F} \left( \partial_c {\bm x}_{k-1} \right)} \odot \mathcal{F} \left( \partial_c {\bm y} \right),
    \notag
\end{aligned}
\end{equation}
with $\mathcal{F} \left( \cdot \right)$ and $\overline{\mathcal{F} \left( \cdot \right)}$ denoting Fourier transform and complex conjugate of Fourier transform, respectively. The expression in Eq.(\ref{eq:sup_5}) can be reformulated as follows
\begin{equation}
    \begin{aligned}
        &\mathcal{F} \left( {\bm h}_{k} \right) + \gamma \frac{ \mathcal{F} \left( {{ \left( {{\bm t}_m} - {x_0} \right)} {\mathbbm 1}_{n}^{\rm{T}}} \right)} {\bm{\mathit \Psi}} {{\left( {{\bm t}_m} - {x_0} \right)}^{\rm{T}} {\bm h}_{k} {\mathbbm 1}_{n}} \\
        &+ \gamma \frac{\mathcal{F} \left( {{\mathbbm1}_{m} { \left({{\bm t}_n} - {y_0} \right)}^{\rm{T}}} \right)} {\bm{\mathit \Psi}} { {\mathbbm 1}_{m}^{\rm{T}} {\bm h}_{k} \left( {{\bm t}_n} - {y_0} \right)} = \frac{\bm{\mathit \Gamma}} {\bm{\mathit \Psi}}
    \end{aligned}
    \label{eq:sup_6}
\end{equation}
where the fraction represents element-wise division. To obtain the solution in the spatial domain, we perform the inverse Fourier transform on Eq.(\ref{eq:sup_6}) resulting in
\begin{equation}
    \begin{aligned}
        &{\bm h}_{k} + \gamma \mathcal{F}^{-1} \Bigl[ \frac{ \mathcal{F} \left( {{ \left( {{\bm t}_m} - {x_0} \right)} {\mathbbm 1}_{n}^{\rm{T}}} \right)} {\bm{\mathit \Psi}} \Bigr] {{\left( {{\bm t}_m} - {x_0} \right)}^{\rm{T}} {\bm h}_{k} {\mathbbm 1}_{n}}\\
        &+ \gamma \mathcal{F}^{-1} \Bigl[ \frac{\mathcal{F} \left( {{\mathbbm1}_{m} { \left({{\bm t}_n} - {y_0} \right)}^{\rm{T}}} \right)} {\bm{\mathit \Psi}} \Bigr] { {\mathbbm 1}_{m}^{\rm{T}} {\bm h}_{k} \left( {{\bm t}_n} - {y_0} \right)}\\
        &= \mathcal{F}^{-1} \left( \frac{\bm{\mathit \Gamma}} {\bm{\mathit \Psi}} \right)
    \end{aligned}
    \label{eq:sup_7}
\end{equation}
where $\mathcal{F}^{-1} \left( \cdot \right)$ is inverse Fourier transform. And Eq.(\ref{eq:sup_7}) can be rewritten in the following matrix-vector form
\begin{equation}
    {\bm h}_{k} 
    + \gamma {\bm F} {{ \left( {{\bm t}_m} - {x_0} \right)}^{\rm{T}} {\bm h}_{k} {\mathbbm 1}_{n}}
    + \gamma {\bm K} {\mathbbm 1}_{m}^{\rm{T}} {\bm h}_{k} \left( {{\bm t}_n} - {y_0} \right)
    = {\bm Z}
    \label{eq:sup_8}
\end{equation}
with these matrices denoted as
\begin{equation}
    \begin{aligned}
        &{\bm{F}} = {\mathcal F}^{-1} \Bigl[ \frac{ \mathcal{F} \left({{\left( {\bm t}_m - {x_0} \right)}  {\mathbbm 1}_{n}^{\rm{T}}} \right)} {\bm{\mathit \Psi}} \Bigr], \\
        &{\bm{K}} = {\mathcal F}^{-1} \Bigl[ \frac{\mathcal{F} \left( {{\mathbbm 1}_m {\left( {\bm t}_n - {y_0} \right)} ^{\rm{T}}} \right)} {\bm{\mathit \Psi}} \Bigr], \\
        &{\bm Z} = {\mathcal F}^{-1} \left( \frac{\bm{\mathit \Gamma}} {\bm{\mathit \Psi}} \right).
    \end{aligned}
    \notag
\end{equation}
In this case, the minimization in Eq.(\ref{eq:sup_2}) converts to solving the linear system of equations as follows
\begin{equation}
    {\bm A} {\rm{vec}} \left( {{\bm h}_{k}}  \right) = {\rm{vec}} \left( {\bm Z} \right)
    \label{eq:sup_9}
\end{equation}
where ${\rm{vec}} \left( \cdot \right)$ represents the operation of stacking the elements of a matrix column-wise, and the coefficient matrix
\begin{equation}
    {\bm A} = \gamma \left( {\rm{vec}} \left( {\bm F} \right) {\rm{vec}} \left( {\bm U} \right)^{\rm T}
    + {\rm{vec}} \left( {\bm K} \right) {\rm{vec}} \left( {\bm V} \right)^{\rm T} \right) + {\bm I}
    \notag
\end{equation}
with ${\bm U} = \left( {\bm t}_{m} - {x_0} \right) {\mathbbm 1}_{n}^{\rm T}, {\bm V} = {\mathbbm 1}_{m} \left( {\bm t}_{n} - {y_0} \right)^{\rm T}$ and ${\bm I}$ denoting an identity matrix of size $mn \times mn$.

And then, we can obtain a direct, exact and analytic solution for the blur kernel ${\bm h}_{k}$ by solving Eq.(\ref{eq:sup_9}).

\end{document}